\def\year{2021}\relax
\documentclass[letterpaper]{article} 
\usepackage{aaai21}  
\usepackage{times}  
\usepackage{helvet} 
\usepackage{courier}  
\usepackage[hyphens]{url}  
\usepackage{graphicx} 
\usepackage{natbib}  
\usepackage{caption} 
\title{Analysing the Noise Model Error for Realistic Noisy Label Data}
\author{
   Michael A. Hedderich, Dawei Zhu, Dietrich Klakow\\
}
\affiliations{
   Saarland University, Saarland Informatics Campus, Germany\\
   \texttt{\{mhedderich,dzhu,dietrich.klakow\}@lsv.uni-saarland.de}
}

\graphicspath{{figures/}}
\usepackage{subfigure}
\usepackage{booktabs}
\usepackage{amssymb}
\usepackage{amsmath}

\newcommand{\sumik}{\sum_{i=1}^{k}}
\newcommand{\Mtil}{\widetilde{M}}
\newcommand{\Ex}{\mathbb{E}}
\newcommand{\Var}{\mathrm{Var}}
\newcommand{\Cov}{\mathrm{Cov}}

\begin{document}

\maketitle

\begin{abstract}
Distant and weak supervision allow to obtain large amounts of labeled training data quickly and cheaply, but these automatic annotations tend to contain a high amount of errors. A popular technique to overcome the negative effects of these noisy labels is noise modelling where the underlying noise process is modelled. In this work, we study the quality of these estimated noise models from the theoretical side by deriving the expected error of the noise model. Apart from evaluating the theoretical results on commonly used synthetic noise, we also publish NoisyNER, a new noisy label dataset from the NLP domain that was obtained through a realistic distant supervision technique. It provides seven sets of labels with differing noise patterns to evaluate different noise levels on the same instances. Parallel, clean labels are available making it possible to study scenarios where a small amount of gold-standard data can be leveraged. Our theoretical results and the corresponding experiments give insights into the factors that influence the noise model estimation like the noise distribution and the sampling technique.
\end{abstract}
\section{Introduction}
\label{label1}
One of the factors in the success of deep neural networks is the availability of large, labeled datasets. Where such labeled data is not available, distant and weak supervision have become popular. Related but different to semi-supervised learning, in distant supervision, the unlabeled data is automatically annotated by a separate process using e.g. rules and heuristics created by expert \cite{distant/DataProgramming} or exploiting the context of images \cite{distant/LimitsWeaklyInstagram}. For tasks like information extraction from text \cite{distant/mintz2009distantRE}, this has become one of the dominant techniques to overcome the lack of labeled data.

\begin{figure}
    \centering
    \includegraphics[width=0.65\columnwidth]{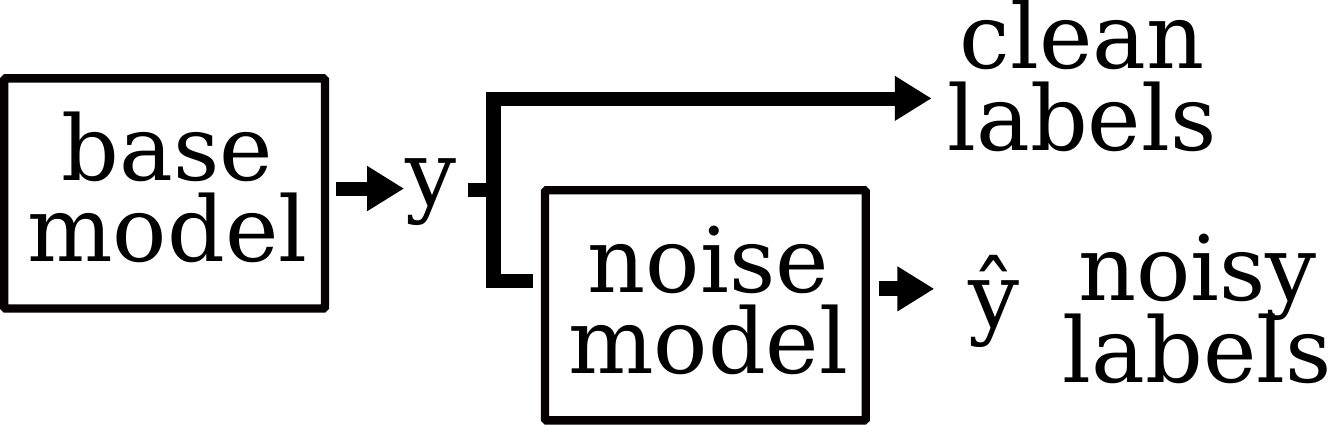}
    \caption{Visualization of the general noise model architecture. The \textit{base model} works directly on the clean data and predicts the clean label $y$. For noisily-labeled data, a \textit{noise model} is added after the base model's predictions.}
    \label{fig:general_noise_model}
\end{figure}

While distant supervision allows generating labels in a cheap and fast way, these labels tend to contain more errors than gold standard ones and training with this additional data might even deteriorate the performance of a classifier (see e.g. \citet{noise/Fang16POS}). Effectively leveraging this noisily-labeled data for machine learning algorithms has, therefore, become a very active field of research. One of the major approaches is the explicit modeling of the noise. This general concept is task-independent and can be added to existing deep learning architectures. It is visualized in Figure \ref{fig:general_noise_model}. The \textit{base model} is the model that was originally developed for a specific classification task. It is directly used during testing and when dealing with other clean data. When working with noisily-labeled data during training, a \textit{noise model} is added after the base model's output. The noise model is an estimate of the underlying noise process. The training process of the base model can benefit from it as the noise model can be seen as changing the distribution of the labels from the clean to the noisy. This will be properly defined below.

Many works on noise modeling assume that no manually annotated, clean data is available. The recent trend in few-shot learning and specific works like \cite{data/Lauscher2020ZeroHero, noise/hedderich2020african} have shown, however, that it is both realistic and beneficial to assume a small amount of manually labeled instances. This motivates us to study in this work scenarios where a small amount of clean, gold-standard data, as well as a large amount of noisily labeled data, are available. 

In this paper, we focus on the quality of the noise model. The better such a model is estimated, the better it can model the relationship between clean and noisy labels. We are interested in the factors on which the quality of the noise model depends. We show, both theoretically and empirically, the influence of the clean data and the noise properties. We also propose to adapt the sampling technique to obtain more accurate estimates. Apart from helping to improve the understanding of these noise models in general, we hope that the insights can also be useful guidance for practitioners. 

In the noisy labels literature, theoretical insights are often evaluated only on simulated noise processes, e.g. on MNIST with added single-label-flip noise \cite{noise/Reed, noise/Bekker16, noise/Goldberger16, noise/patrini2017losscorrection, noise/Coteaching}. This synthetic noise has the advantage that it can be controlled completely and allows to rigorously and continuously evaluate aspects like the noise level. However, certain assumptions about the noise have to be taken. And these are usually the same assumptions that are chosen for the noise model itself. It might, therefore, not be too surprising that such a model is suited for such a noise. Recently, efforts have been taken to also evaluate on more realistic scenarios, mostly in the vision domain, e.g. the Clothing1M dataset by \citet{noise/Xiao2015}. We want to add to this by making available a noisy label dataset from the natural language processing (NLP) domain based on an existing named entity recognition (NER) corpus. It provides parallel clean and noisy labels for the full data allowing to evaluate different scenarios of resource availability of both clean and noisy data. This new dataset also contains properties that can make learning with noisy labels more challenging such as skewed label distributions and a noise level higher than the true label probability in some settings. In contrast to existing work, we provide seven different sets of noisy labels, each obtained by a realistic noise process via different heuristics in the distant supervision. This makes it possible to experiment with different noise levels for the same instances. The dataset along with the code for the experiments is made publicly available\footnote{\url{https://github.com/uds-lsv/noise-estimation}}.

\vspace{0.3cm}
\textbf{Our key contributions}: 
\begin{itemize}
    \item A derivation of the expected error of the noise model estimated from pairs of clean and noisy labels with empirical verification of the derived results on both simulated and realistic noisy labels.
    \item A set of experiments analyzing how the noise model estimation influences the test performance of the base model.
    \item NoisyNER, an NLP dataset with noisy labels obtained through non-synthetic, realistic distant supervision that also provides different levels of noise and parallel clean labels.
\end{itemize}

\section{Background}
\label{noisy-labels-theory}

We are given a dataset $D$ consisting of instances $(x,\hat{y})$ where $\hat{y}$ is a noisy label that can differ from the unknown, clean/true label $y$. Both clean and noisy label have one of $k$ possible label values/classes. We assume that the change from the clean to the noisy label was performed by a probabilistic noise process described as $p(\hat{y}=j|y=i)$. This describes the probability of a true label $y$ being changed from value $i$ to the noisy label $\hat{y}$ with label value $j$. With probability $p(\hat{y}=j|y=j)$ the label value is kept unchanged. This is a common approach to describe noisy label settings. Under this process, a uniform noise \cite{noise/Larsen98} with noise level $\epsilon$ is obtained with
\begin{align}
    p_{\text{uni}}(\hat{y}=j|y=i) = \left\{\begin{array}{lr}
        1-\epsilon, & \text{for } i = j\\
        \frac{\epsilon}{k-1}, & \text{for } i \neq j
        \end{array} \right. \label{eq:uniform} \; ,
\end{align}
\vspace{0.5cm}

and a single label-flip noise \cite{noise/Reed} via
\begin{align}
    p_{\text{flip}}(\hat{y}=j|y=i) = \left\{\begin{array}{lr}
        1-\epsilon, & \text{for } i = j \\
        \epsilon, & \text{for one } i \neq j\\
        0, & else
        \end{array} \right. \label{eq:labelflip} \;.
\end{align}

In this work, we will use the more general form that just requires that a noise process can be described with a valid noise transition probability $p(\hat{y}=j|y=i)$ \cite{noise/Bekker16}, i.e.
\begin{align} \label{eq:requirements_M}
    \sum_{j=1}^{k} p(\hat{y}=j|y=i) = 1 \text{ and }  p(\hat{y}=j|y=i) \geq 0 \; \forall i,j.
\end{align}
This allows to model more complex noise processes where a true label can be confused with multiple other labels at different noise rates/probabilities. We call this multi-flip noise. This definition generalizes to multi-class classification what in the binary case \citet{noise/Natarajan13} and \citet{noise/Scott13} describe as class-conditional and asymmetric noise. It is called Markov label corruption by \citet{noise/Rooyen17}.

\begin{figure}
    \centering
    \subfigure[uniform]{
        \includegraphics[height=2.3cm]{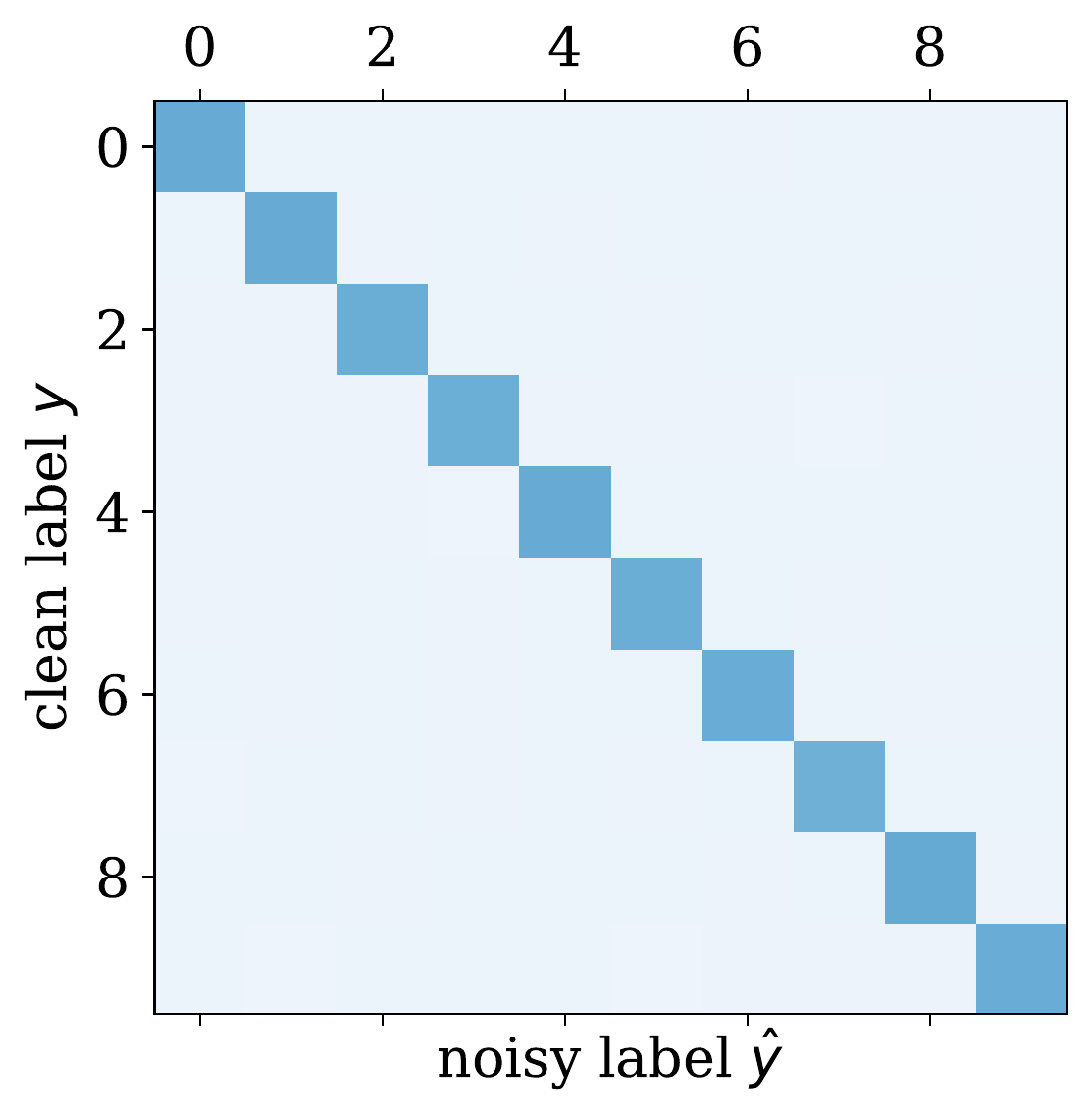} 
    }
    \subfigure[single-flip]{
        \includegraphics[height=2.3cm]{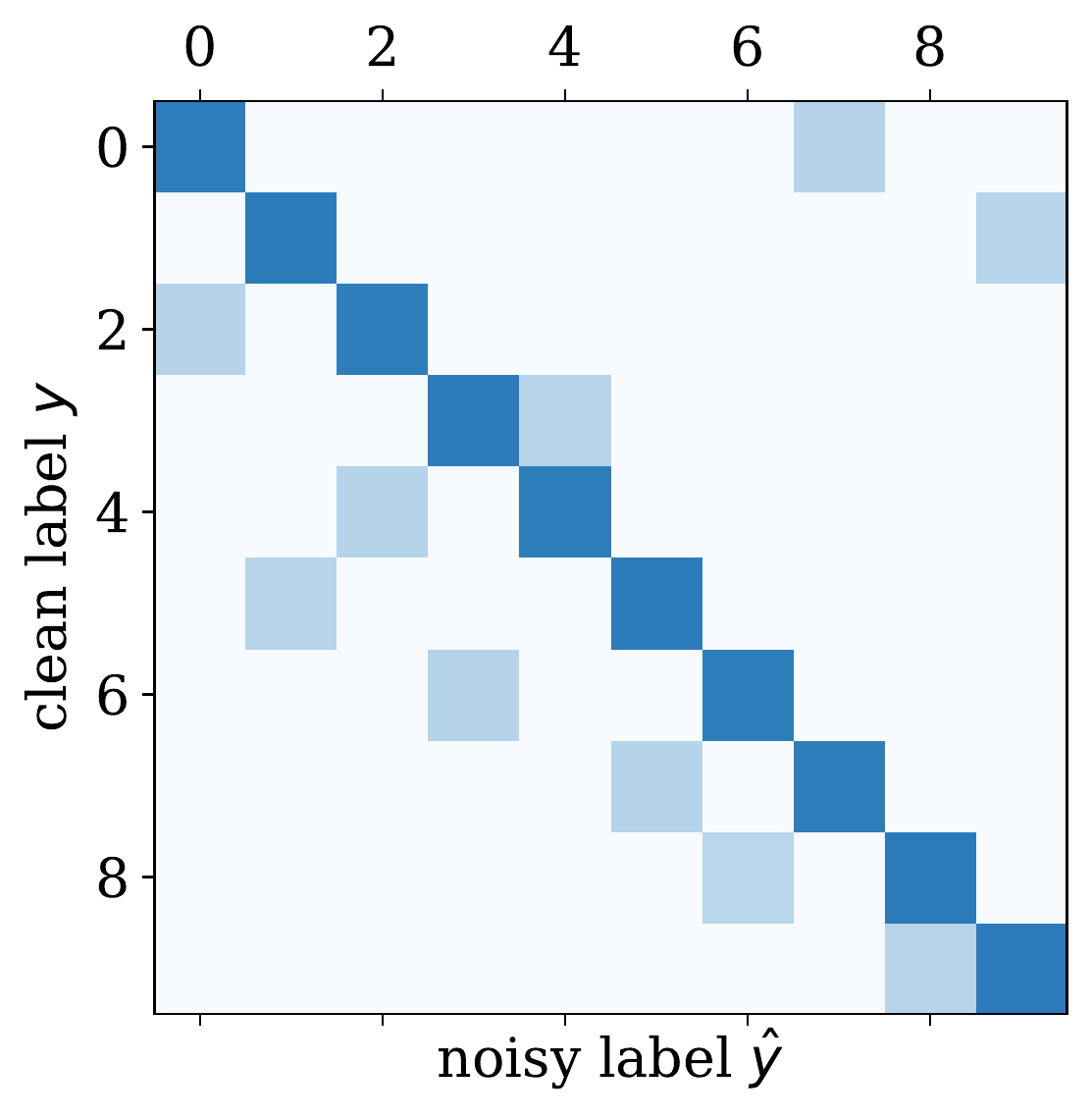} 
    }
    \subfigure[multi-flip]{
        \includegraphics[height=2.3cm]{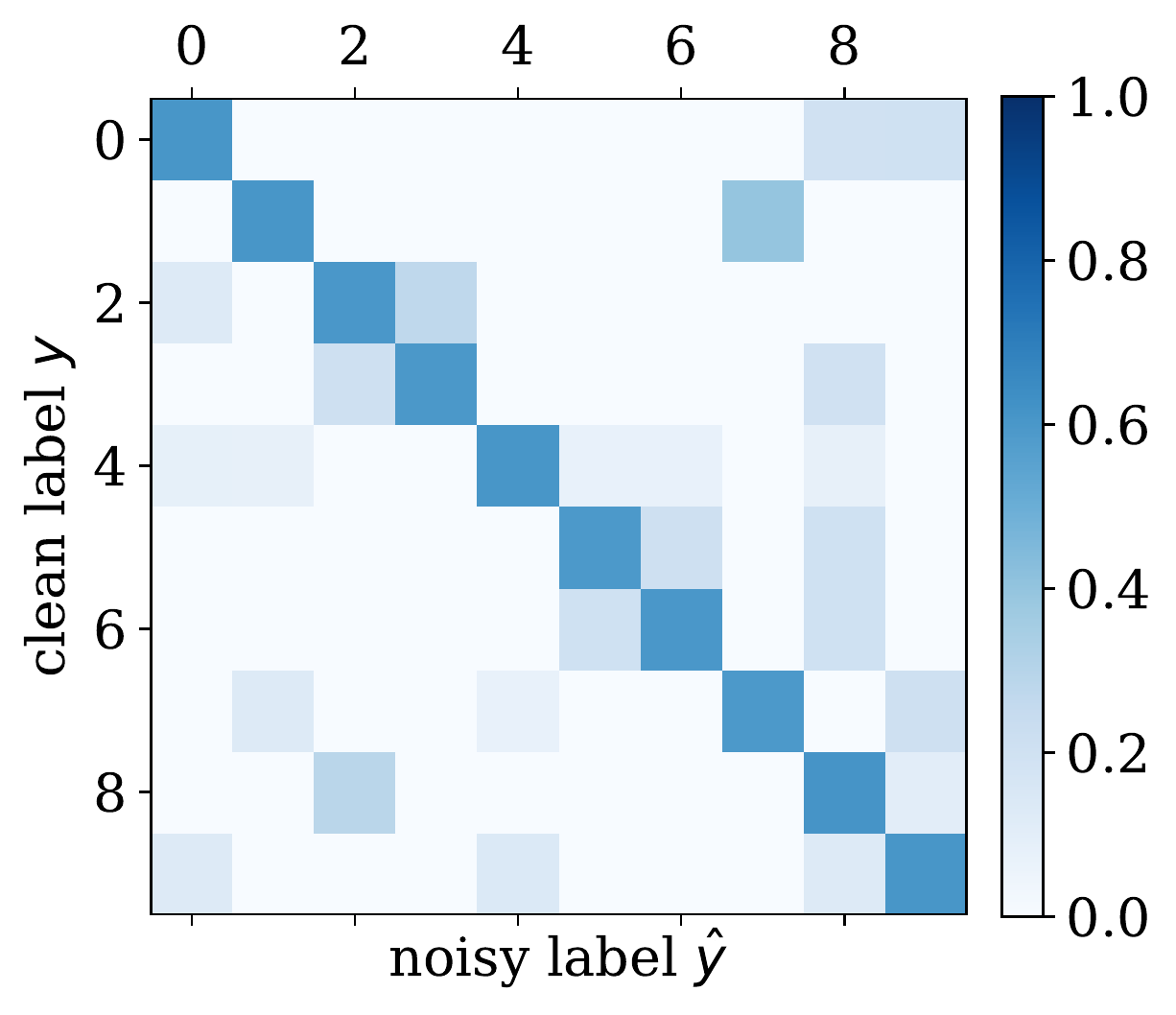}
    }
    \caption{Different noise processes visualized as noise matrices $M$: uniform noise ($\epsilon = 0.5$), single-flip noise  ($\epsilon = 0.3$) and multi-flip noise ($\epsilon = 0.4$). \label{fig:noise_types}}
\end{figure}

The noise process can also be described as a matrix $M \in \mathbb{R}^{k \times k}$ where 
\begin{align} \label{eq:mij_p}
    M_{i,j} = p(\hat{y}=j|y=i) \;.
\end{align}
The matrix $M$ is called confusion or noise transition matrix. See Figure \ref{fig:noise_types} for example matrices with uniform, single-flip and a more complex multi-flip noise.

For a multi-class classification setting, this noise process results in the following relation: 
\begin{align} \label{eq:pyhat_givenx}
    p(\hat{y}=j|x) = \sum_{i=1}^{k} p(\hat{y}=j|y=i) p(y=i|x) \;.
\end{align}
This is used to adapt the predictions from the clean label distribution $p(y=i|x)$ (learned by the base model) to the noisy label distribution $p(\hat{y}=j|x)$ for the noisily labeled data via the noise process or model $p(\hat{y}=j|y=i)$.

Several ways have been proposed to estimate the noise model when only noisy data is available. This includes the use of expert insight \cite{noise/Mnih2012}, EM-algorithm \cite{noise/Bekker16, noise/Paul19, noise/chen2019EM}, backpropagation with regularizers \cite{noise/Sukhbaatar14, noise/Luo17} and the estimates of a pretrained neural network \cite{ noise/Goldberger16, noise/patrini2017losscorrection, noise/Dgani2018, noise/Wang19}. For settings where a small portion of clean data can be used, noise model estimation methods have been proposed by \citet{noise/Xiao2015, noise/Fang16POS, noise/Hedderich18, noise/Hendrycks2018, noise/Lange19} and \citet{noise/hedderich2020african}.

Out of the different, similar approaches to estimate a noise model given a small amount of clean labels, we will follow the specific definitions of \citet{noise/Hedderich18} which are based on \cite{noise/Goldberger16} and also used in \cite{noise/Lange19}. The authors assume that a small dataset $D_C$ with clean labels exists, i.e. $(x,y) \in D_C$  is known. The size of this clean dataset is much smaller than that of the noisy dataset $D$. They then relabel the instances in $D_C$ with the same mechanism that was used for $D$ (e.g. distant supervision) to obtain $\hat{y}$. This results in a set $S_{NC}$ of pairs $(y,\hat{y})$ with $|S_{NC}| = |D_C|$. The noise matrix $M$ is then estimated as $\widetilde{M}$ using the transitions from $y$ to $\hat{y}$ (or equivalently the confusion between $y$ and $\hat{y}$):
\begin{align} \label{eq:m_estimate}
    \widetilde{M}_{ij} = \frac{m_{ij}}{n_i} = \frac{\sum\limits_{(y,\hat{y}) \in S_{NC}} 1_{\{y=i, \hat{y}=j\}}}{\sum\limits_{(y,\hat{y}) \in S_{NC}} 1_{\{y=i\}}} \;,
\end{align}
where $m_{ij}$ is the number of times that the label was observed to change from $i$ to $j$ due to the noise process and $n_i$ is the number of instances in $D_C$ with label $y=i$. $\widetilde{M}$ is the estimated model of the noise process. This noise model is then integrated into the training process using Equation \ref{eq:mij_p} and \ref{eq:pyhat_givenx} and as visualized in Figure \ref{fig:general_noise_model}. 

\section{Expected Error of the Noise Model}
\label{sec:new-theory}

The noise model obtained in Equation \ref{eq:m_estimate} is an approximation of the underlying true noise process estimated on a small number of instances $D_C$. In this section, we derive a formula for the expected error of the estimated noise model $\widetilde{M}$.  This gives us insights into the factors that influence the noise model's quality as well as their effect.

\textbf{Assumptions} In the following proofs, we assume that $M$ describes a noise process following Equations \ref{eq:requirements_M} and \ref{eq:mij_p}. $\widetilde{M}_{ij}$ is estimated using Equation \ref{eq:m_estimate}.

We study two sampling techniques on how to obtain the set of clean and noisy label pairs $|S_{NC}|$. Commonly, a fixed number of unlabeled instances $n$ is obtained and then manually annotated with gold labels. The value of $n_i$ then follows the distribution of classes in the data. We call this \textbf{Variable Sampling} as the value of $n_i$ varies.

In contrast to that, for \textbf{Fixed Sampling}, for each label value $i$, we sample $n_i$ instances with $y=i$. This could be conducted e.g. by asking annotators to provide a specific number of labeled instances per class. In this case, $n_i$ is fixed. For readability, we write $\Ex$ for $\Ex_{\sim S_{NC}}$ and analogously for $\Var$ and $\Cov$. We assume that the instances are sampled independently.

As quality metric for evaluating the noise model, we use squared error which is in this matrix case the square of the Frobenius norm

\begin{align}
    SE = ||M - \Mtil||_F^2 = \sum_{i=1}^{k} \sum_{j=1}^{k} (M_{ij} - \Mtil_{ij})^2 \;.
\end{align}

\textbf{Theorem 1}

The expected squared error of the noise model is 
\begin{align*}
    \Ex[SE] = &\sumik \sum_{j=1}^{k} \Var[\Mtil_{ij}] \;.
\end{align*}

\textit{Sketch of the proof:} We show that $\widetilde{M}$ is an unbiased estimator, i.e. $\Ex[\Mtil_{ij}] = M_{ij}$. From that, Theorem 1 can be followed. For space reasons, the full proofs are given in the Appendix.  

\textbf{Theorem 2a} Assuming \textit{Variable Sampling}, it holds $\Var[\Mtil_{ij}] = M_{ij}(1-M_{ij})\sum\limits_{n_i=1}^nP(n_i)\frac{1}{n_i}$ where $P(n_i)$ is the probability of sampling $n_i$ instances with label $y=i$ from the data.

\textbf{Theorem 2b} Assuming \textit{Fixed Sampling}, it holds $\Var[\Mtil_{ij}] = \frac{M_{ij}(1-M_{ij})}{n_i}$.

\textit{Sketch of the proof:} The proofs for both variants of the theorem work on the main insight that given $n_i$, the value of $m_{ij}$ follows a multinomial distribution defined by $M_{ij}$.

Combining Theorem 1 and 2, we obtain a closed-form solution for the expected error of the estimated noise model for both Fixed and Variable Sampling. From this, we can see that
\begin{itemize}
    \item the error changes with the amount of sampled instances by factor $\frac{1}{n_i}$.
    \item the error depends on the noise distribution as well as the level of noise $M_{ij}$. In the single-flip scenario, it reaches its maximum when the noise is as dominant as the true label value.
    \item Fixed Sampling obtains lower error than Variable Sampling in most cases.
\end{itemize}
These results are visualized and experimentally verified below.

\section{Data with Synthetic Noise}

Experiments with synthetic or simulated noise allow fine-grained control of the noise level and type of noise. An existing dataset is taken and the labels are assumed to be all correct and clean. Then, to obtain a noisy label dataset, for each instance, the label is flipped according to the noise process. \citet{noise/Reed} and \citet{noise/Goldberger16} use the MNIST dataset \cite{data/MNIST} and apply single-flip noise (Equation \ref{eq:labelflip}) to obtain the noisy labels. We follow their approach and label-flip pattern. Additionally, we also generate noisy labels with uniform noise (Equation \ref{eq:uniform}) and a more complex, multi-flip noise where one label can be changed into one of several incorrect labels (Equation in the Appendix). All three noise types are visualized in Figure \ref{fig:noise_types}. We see the multi-flip noise as the most realistic of these synthetic noises, as it resembles most the two realistic datasets presented in the next section.

\section{Data with Realistic Noise}

While evaluating on synthetically generated noise is popular and allows for an easy evaluation in a controlled environment, it is limited by the assumptions on the noise. Certain assumptions are taken when building a model of the noise and the same assumptions are used to generate the noisy labels.

In real-world scenarios, some of these assumptions might not apply. Inspecting realistic noise matrices (Figures \ref{fig:noisematrix_clothing1m} and \ref{fig:noisematrix_ner}), it is already quite obvious that these do not resemble the popular uniform or single-flip noise. We think it is therefore important to also evaluate on more realistic data that does not rely on the noise being simulated. Nevertheless, having parallel clean and noisy labels for the same instances is very useful as it allows e.g. to compute the upper bounds of training on clean data compared to training with noisy labels. In this specific work, it is required to obtain an approximation of the true noise pattern and to flexibly vary the number of clean labels. The Clothing1M dataset by \citet{noise/Xiao2015} and our newly proposed NoisyNER dataset offer these possibilities.   

\subsection{Clothing1M}

\begin{figure}
    \centering
    \includegraphics[width=0.73\columnwidth]{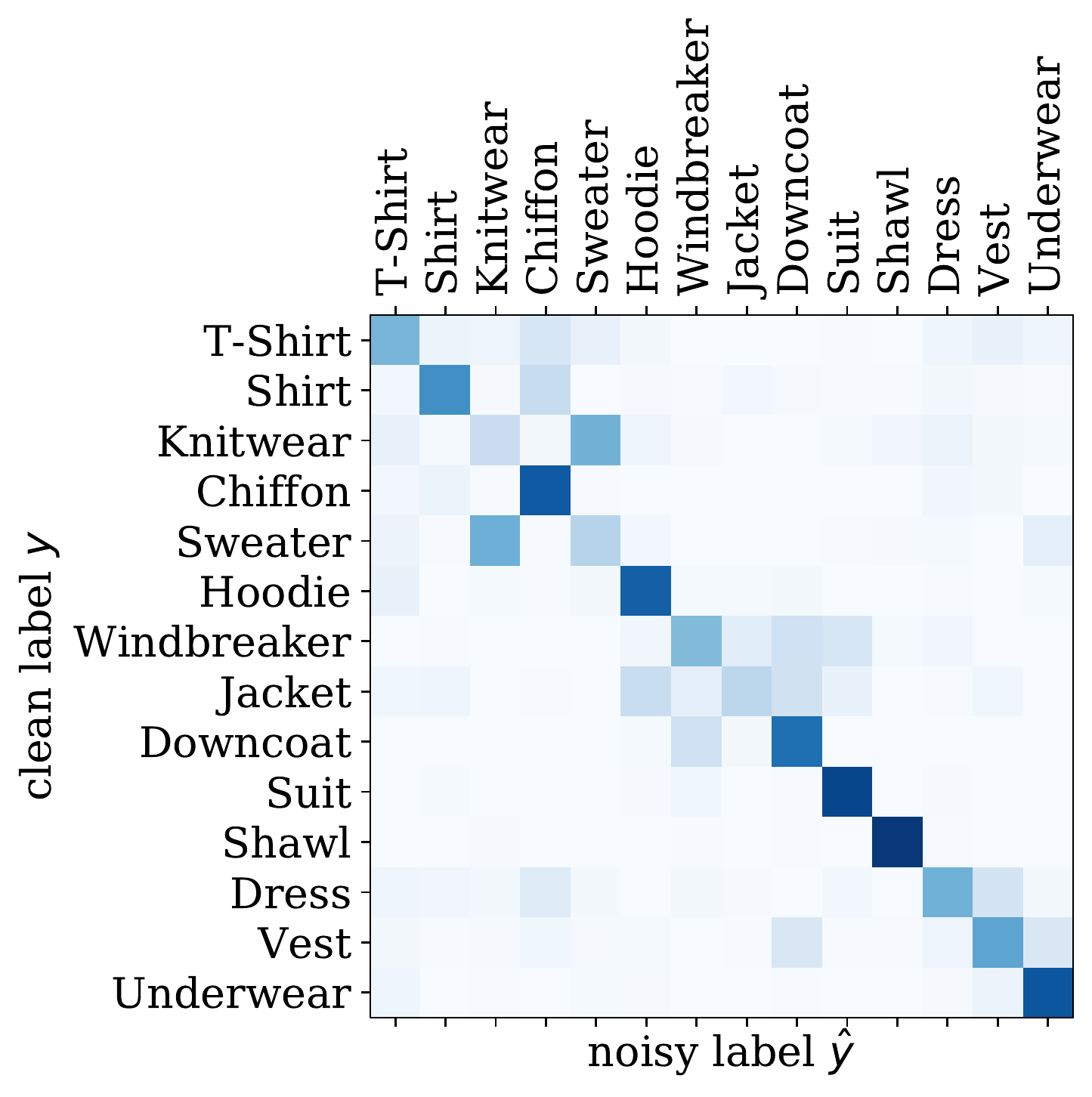}
    \caption{Noise matrix for Clothing1M computed over all pairs of clean and noisy labels. \label{fig:noisematrix_clothing1m}}
\end{figure}

The Clothing1M dataset is part of a classification task to label clothing items present in an image. The noisy labels were obtained through a distant supervision process that used the text context of the images appearing on a shopping website. For 37k images, both clean and noisy labels are available. The percentage of correct labels in the noisy data is 38\% and a visualization of the noise is given in Figure \ref{fig:noisematrix_clothing1m}. One can see that the noise distributes neither uniformly nor is there a single label flip. Rather a label tends to be confused with several other related labels, e.g. a "Jacket" with a "Hoodie" and a "Downcoat".

\subsection{NoisyNER}
\label{sec: NoisyNER}

In this work, we propose another noisy label dataset. It is from the text classification domain with word-level labels for named entity recognition (NER). The labels are persons, locations and organizations. The language is Estonian, a typical low-resource language with a demand for natural language processing techniques. The text and the clean labels were collected by \citet{data/EstonianNERDataset} through expert annotations \cite{data/EstonianNER}. The noisy labels are obtained through a distant supervision/automatic annotation approach. Using a knowledge base for the distant supervision was first introduced by \citet{distant/mintz2009distantRE} and is still commonly used, e.g. by \citet{distant/lison2020NER}. In our case, lists of named entities were extracted from Wikidata and matched against the words in the text via the ANEA tool \cite{distant/hedderich2021anea}. If a word/string appears in a list of entities, it is labeled as the corresponding entity. This allows to quickly annotate large text corpora without manually labeling each word. However, these labels are not error-free. Reasons include non-complete lists of entities, grammatical complexities of Estonian that do not allow simple string matching or entity lists in conflict with each other (e.g. "Tartu" is both the name of an Estonian city and a person name). Heuristic functions allow to leverage insights from experts efficiently \cite{distant/ratner20snorkel} and they can also be applied to correct some of these error sources, e.g. by normalizing (lemmatizing) the grammatical form of a word or by excluding certain high false-positive words. Details are given in the Appendix.

\begin{figure}
    \centering
    \includegraphics[width=0.36\columnwidth]{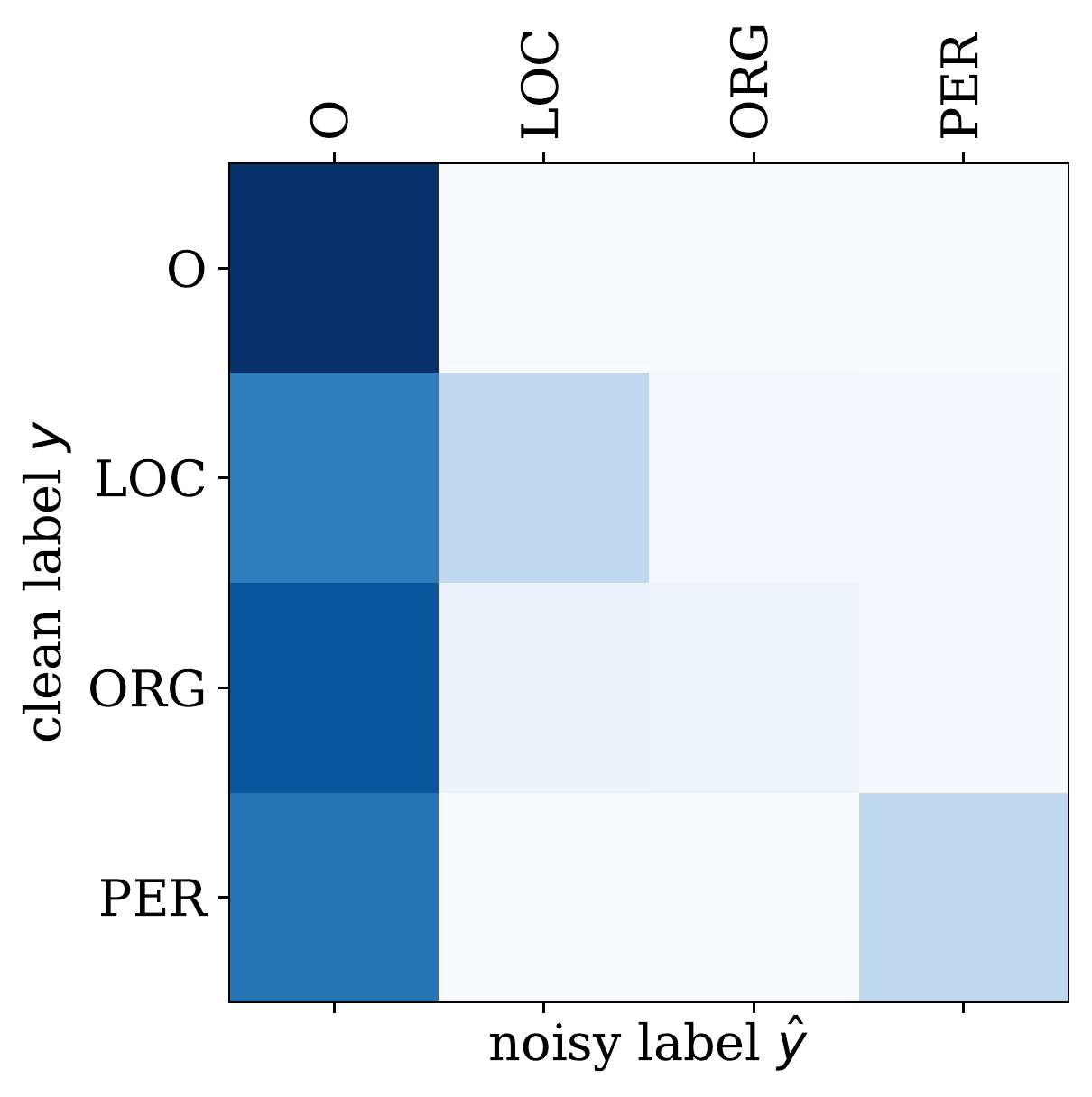}
    \includegraphics[width=0.36\columnwidth]{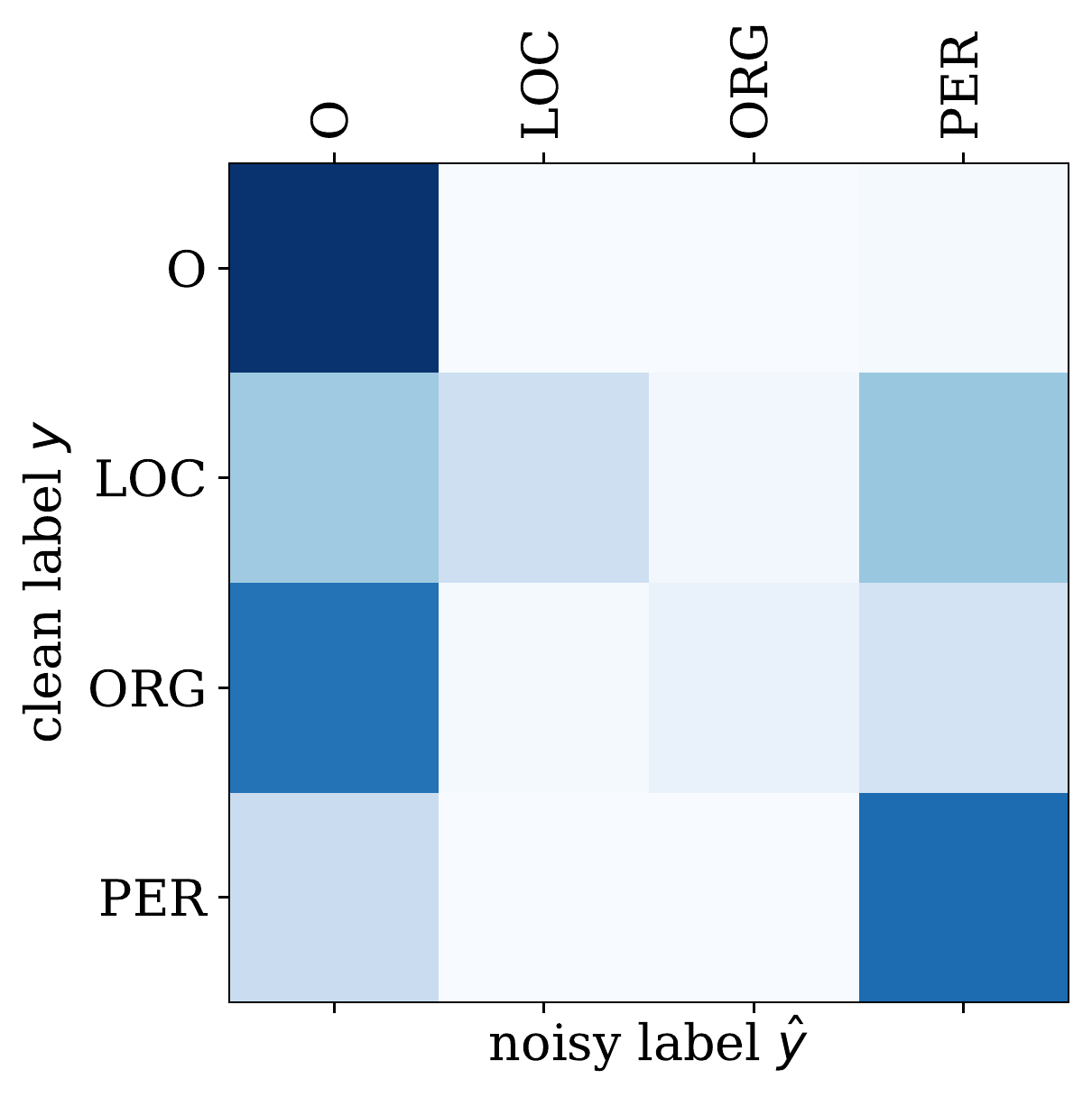}\\
    \includegraphics[width=0.36\columnwidth]{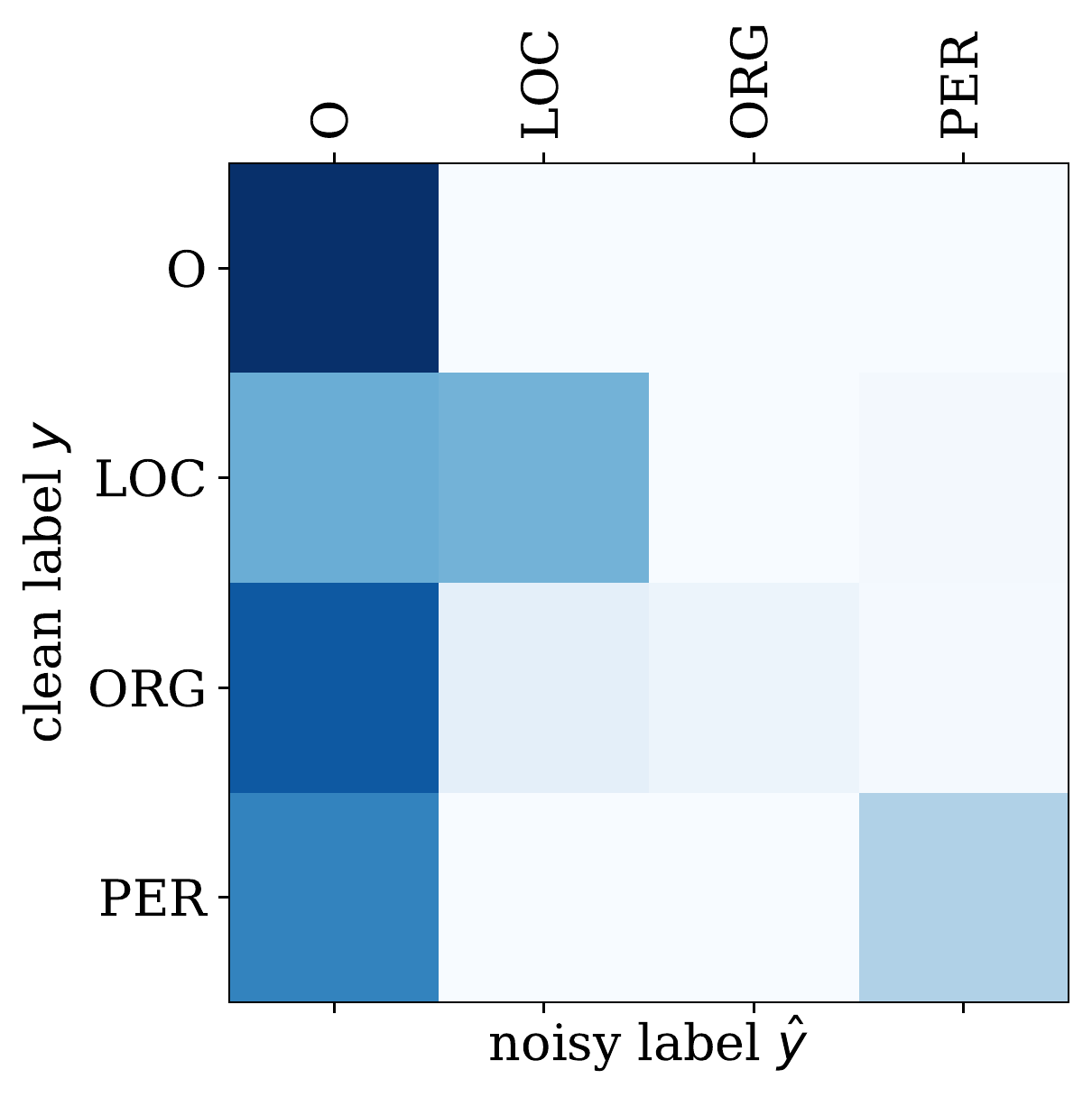}
    \includegraphics[width=0.36\columnwidth]{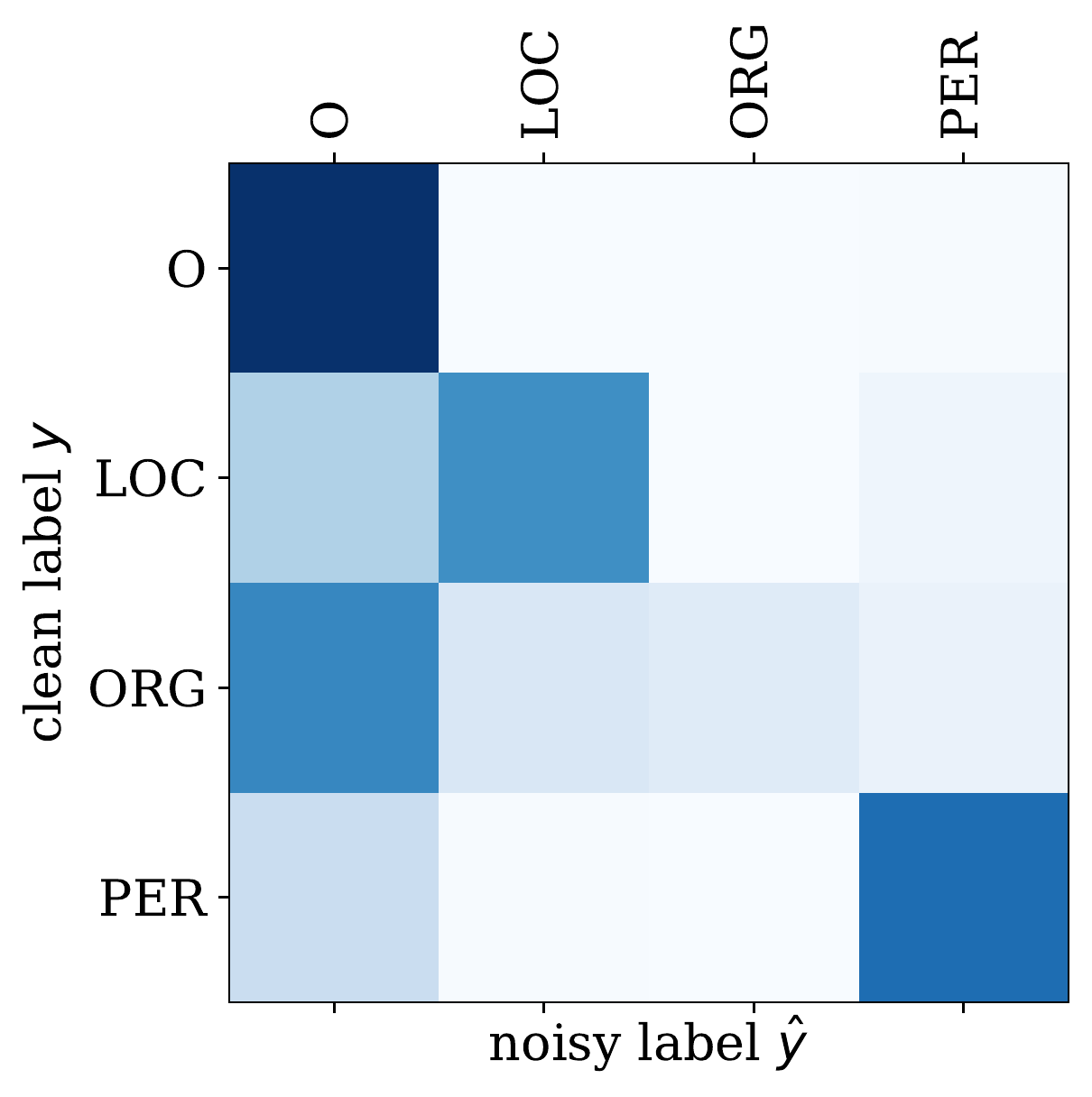}
    \caption{Noise matrices for NoisyNER (label sets 1, 3, 4 and 7) computed over all pairs of clean and noisy labels.}
    \label{fig:noisematrix_ner}
\end{figure}

In contrast to Clothing1M, we provide seven sets of labels that differ in the noise process and therefore in the amount of noise and the noise pattern. Four of these are visualized in Figure \ref{fig:noisematrix_ner} with the rest in the Appendix. Table \ref{tab:errors_ner} lists an overview of the percentage of correct labels. The different labels are obtained by using varying amounts of heuristics during the automatic annotation process. This reflects different levels of manual effort that one might be able to spend on the creation of distantly supervised data. Having these multiple sets of labels for the same instances allows to directly evaluate for different noise levels while excluding side effects that differing features might have. 

\noindent We want to highlight several properties of the dataset:

\begin{itemize}
    \item  Clean labels are available for all instances. This allows studying different splits of clean and noisy labels as well as computing upper bounds of performance on only clean data. 
    \item The distribution of the labels is skewed. Out of the ca. 217k instances, ca. 8k are persons, 6k are locations and 6k are organizations. All other instances are labeled as non-entity O. Such a skewed label distribution is typical for a named entity recognition dataset.
    \item In past works, experiments are often only performed until the probability of a noisy class reaches the probability of the true class, i.e. it is assumed that $p(\hat{y}=j|y=i) < p(\hat{y}=i|y=i) \; \forall  j \neq i$. This assumption does not hold for several of our label sets which can make learning on the data more challenging. 
    \item While not studied in this work, the labels in the dataset also contain sequential dependencies. A clean or noisy label can span over several words/instances, e.g. for the mention of a person with a first and last name. These sequential dependencies could be leveraged in future work.
\end{itemize}

\begin{table}[]
    \vskip 0.15in
    \begin{center}
    \begin{small}
    \begin{sc}
    \centering
    \begin{tabular}{c|c|c|c|c|c|c|c}
        \toprule
        Label Set & 1 & 2 & 3 & 4 & 5 & 6 & 7\\
        \midrule
        Precision & 67 & 73 & 37 & 75 & 48 & 53 & 59 \\
        Recall &  18 & 27 & 31 & 27 & 41 & 41 & 49 \\
        F1 & 28 & 39 & 34 & 40 & 44 & 46 & 54 \\
        \bottomrule
    \end{tabular}
    \end{sc}
    \end{small}
    \end{center}
        \caption{Percentages of correct labels in the NoisyNER dataset for the seven different label sets. For each subsequent label set, more manual effort in the labeling heuristics was invested. As the label distribution is highly skewed, precision, recall and F1 score are reported. Following the standard approach for named entity recognition \cite{data/CoNLL03}, the micro-average is computed excluding the non-entity label.}
    \label{tab:errors_ner}
\end{table}
\section{Analysis of the Noise Model Error}

In this section, the theoretically expected squared error between the noise model estimate and the true noise matrix is compared to the empirically measured one. We vary the two parameters found in Theorem 2: the amount of sampled data $n_i$/$n$ and the amount of noise $M_{ij}$. For Clothing1M only the data size can be varied while for NoisyNER the variation of the sample size can be compared across different noise levels and noise distributions.

\subsection{Experimental Setup}
\label{sec:experimental_setup}

From the full dataset, a small subset $D_C$ and corresponding $S_{NC}$ is sampled uniformly at random either using the Fixed or Variable Sampling approach. The noise model $\Mtil$ is estimated on this sample and compared to the true noise process $M$. The process is repeated 500 times and the average empirical squared error is reported as well as its standard deviation on the error bars.

For the synthetic noisy labels, the true noise process is known by construction. For the realistic datasets, the true noise process is unknown. Instead, the noise matrix $M$ is computed over the whole data as an approximation. For the distribution $P(y)$, that is part of the Variable Sampling formula, we assume a uniform distribution for MNIST and a multinomial distribution for Clothing1M and NoisyNER with the parameters of the distribution estimated over the whole dataset. In praxis, one might also rely on expert knowledge for this distribution (e.g. from \citet{ner/Augenstein17} for NER).

\subsection{Results \& Analysis}
\label{sec:results}

\begin{figure}
\centering
\includegraphics[width=0.8\columnwidth]{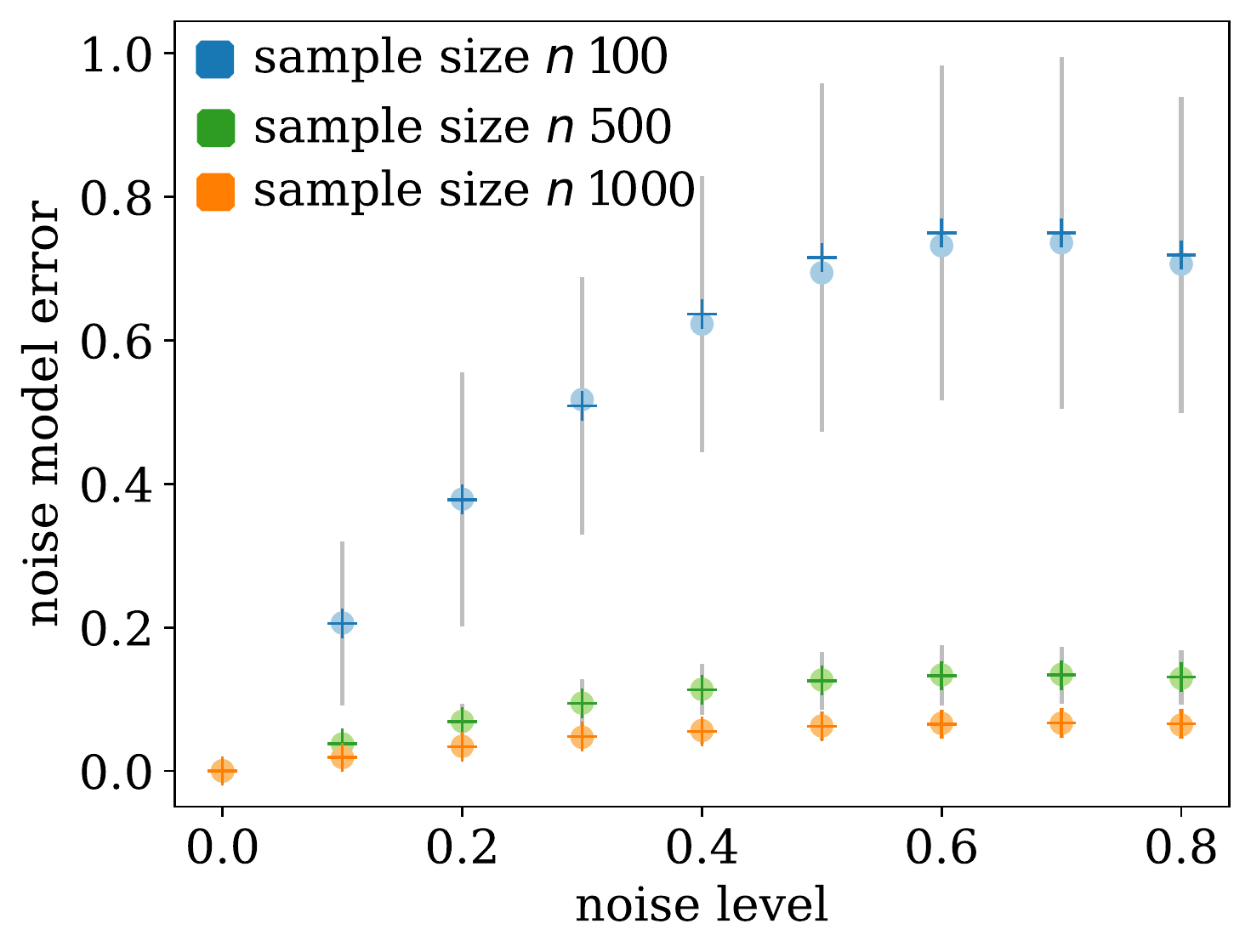}
\caption{Comparison between the theoretically expected mean error (circle marker) and the empirically measured mean error (cross) of the noise model on the MNIST dataset with \textit{multi-flip noise} and with \textit{Variable Sampling}. The x-axis varies the \textit{noise level}. Plots with all noise types, for Fixed Sampling and for sample size on the x-axis are given in the Appendix.}
\label{fig:exp_mnist}
\end{figure}

On the synthetic data (Figure \ref{fig:exp_mnist}), the theoretically expected error of the noise model follows closely the empirical measurements. This holds across different noise types, sample sizes and noise levels. Only for a very small set of clean labels in combination with a high noise level, there is a slight deviation.  

As stated above, we can see the influence of the sample size and noise process. The error changes with the amount of sampled instances by factor $\frac{1}{n_i}$ and it depends on the noise distribution as well as the level of noise $M_{ij}$. For the evaluated scenarios, due to the additional dependency on the clean label distribution $P(y)$, the Variable Sampling technique has a higher expected error than the Fixed Sampling approach, especially for settings with large noise. From the empirical experiments, one can see that the variance of the noise model error mostly depends on the sample size.

\begin{figure}
    \centering
    \subfigure[Clothing1M]{
        \includegraphics[ width=0.8\columnwidth]{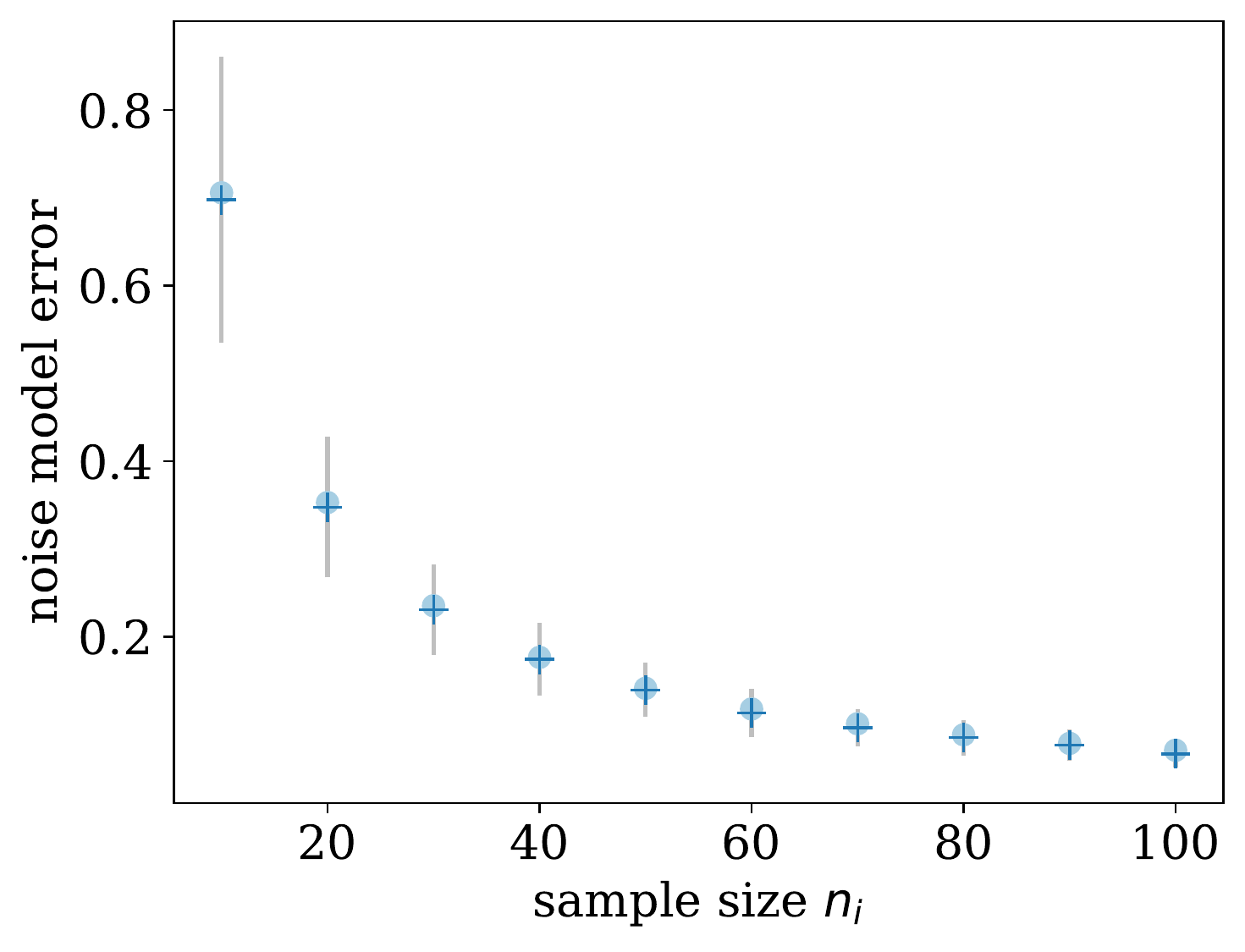}
    }
    \subfigure[NoisyNER]{
        \includegraphics[ width=0.8\columnwidth]{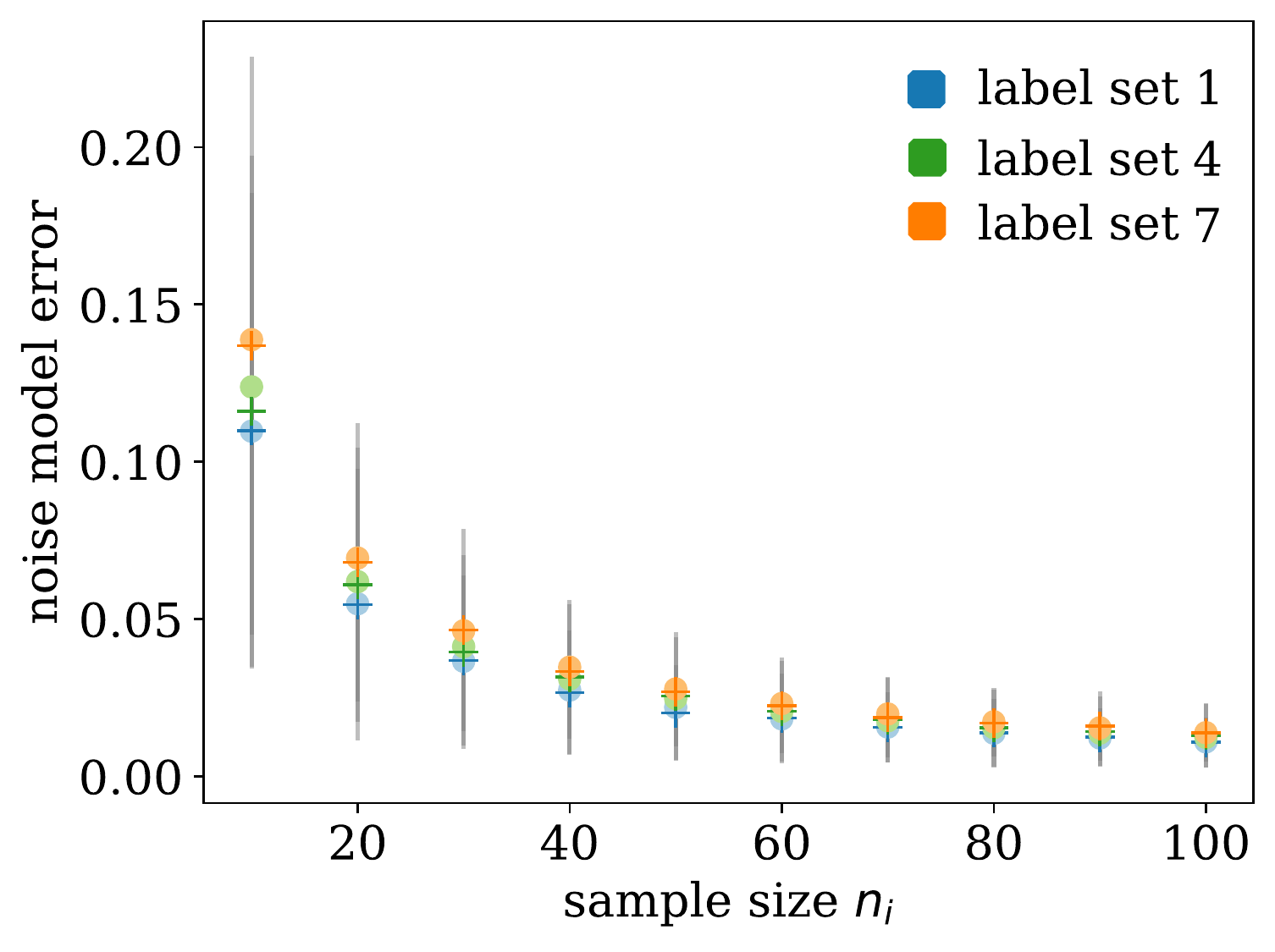}
    }
    \caption{Comparison between the theoretically expected mean error (circle marker) and the empirically measured mean error (cross) of the noise model with \textit{Fixed Sampling}. The x-axis varies the \textit{sample size}. Variable Sampling and the remaining label sets can be found in the Appendix. \label{fig:exp_cloth-ner}}
\end{figure}

The theoretical and experimental results also match on the realistic noisily labeled datasets Clothing1M and NoisyNER (Figure \ref{fig:exp_cloth-ner}). Again, only for the very low sample size, one can observe a deviation. It is interesting to note that for NoisyNER the estimation error of the noise model is higher for the data with overall lower noise level (measured in F1 score in Table \ref{tab:errors_ner}). This is due to how the noise distribution changes in the realistic setting. The difference between Fixed Sampling and Variable Sampling is most noticeable for NoisyNER increasing the error by a factor of around 3 (cf. plots in the Appendix). This suggests that in practice, especially for such skewed distributions, a sampling technique is beneficial which focuses on each label separately. 

\section{Analysing the Base Model Performance}

In the previous sections, we studied the factors on which the quality of the noise model's estimate depends. The noise model is part of a larger classifier and it is combined with the actual base model that performs the task-specific classification (cf. Figure \ref{fig:general_noise_model}). In this section, we evaluate in different experiments the base model on clean test data and analyze the effects of the noise modelling and the noise model estimation on the base model and its task performance.

As in the experimental setup detailed in the previous section, a clean subset $D_C$ of the data is sampled and the noise model estimated on it. The base model is then trained directly on the clean data $D_C$. The noise model is added to the base model for training on the noisy dataset $D$ (cf. Figure \ref{fig:general_noise_model}). For evaluation, a test set is held-out from the data. We follow the training procedure of \citet{noise/Hedderich18} and the training details are given in the Appendix. Due to the longer runtimes, the experiments are repeated 50 times for Clothing1M and NoisyNER. The error bars show the empirical standard deviation.

\subsection{Effect of Noise Handling}
\begin{figure}
    \centering
    \includegraphics[width=0.8\columnwidth]{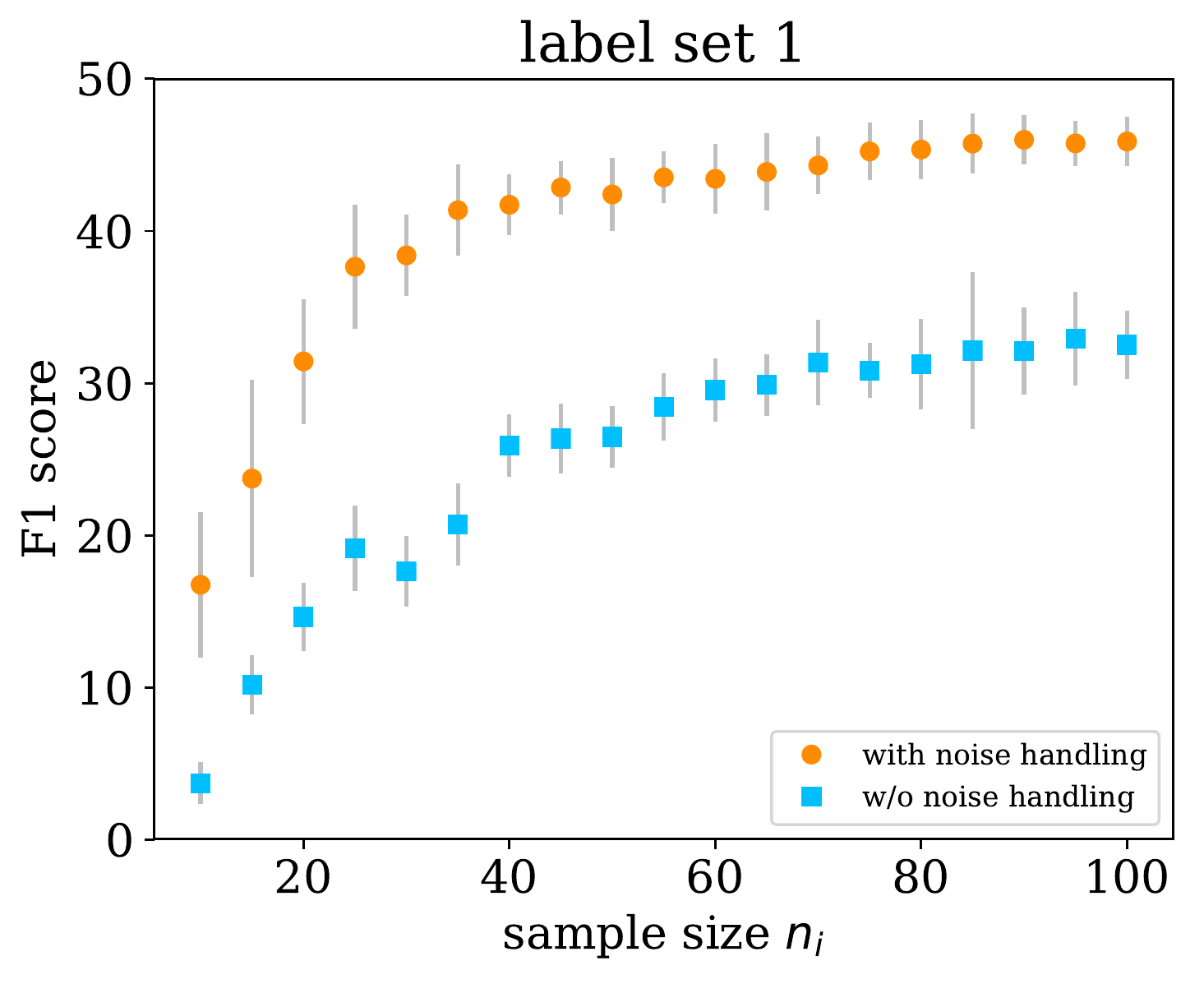}
    \caption{Mean test performance (F1 score) of the base model on NoisyNER label set 1 on clean and noisy data with and without noise handling. Using Fixed Sampling. Further plots in the Appendix.\label{fig:exp_ns_f1}}
\end{figure}

Figure \ref{fig:exp_ns_f1} shows the test performance of the base model for increasing size of $D_C$. It compares training the base model directly on the clean and the noisy label data to handling the noisy labels via a noise model. The latter improves the results in most cases. This confirms past findings that noise handling is an important technique to leverage distantly supervised training data. Comparing the different noise levels on NoisyNER, one can see that larger noise levels also result in larger improvements via noise handling. Only in a few cases with a very small amount of clean training samples does noise handling deteriorate the results, possibly due to a bad estimation of the noise model (see Appendix).

\subsection{Relationship between Noise Estimation and Base Model Performance}
There are several factors on which the base model's performance depends. These could include the amount of clean and noisy training data, the noise distribution and the quality of the estimated noise model. Here, we show experimentally on Clothing1M and NoisyNER that the noise model estimation error directly influences the performance of the base model.

\begin{figure}
    \centering
    \includegraphics[width=0.71\columnwidth]{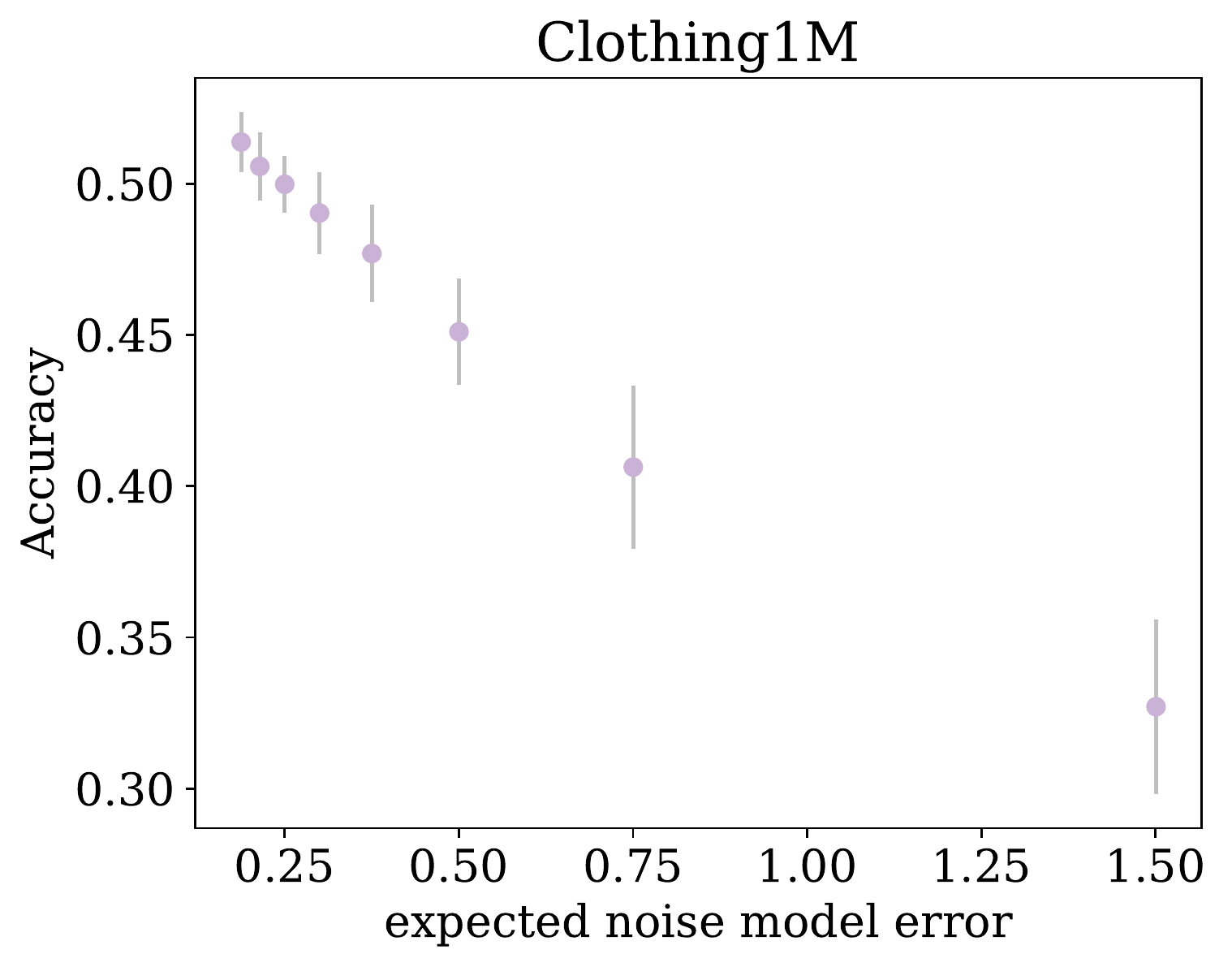}\\
    \vspace{0.1cm}
    \includegraphics[width=0.71\columnwidth]{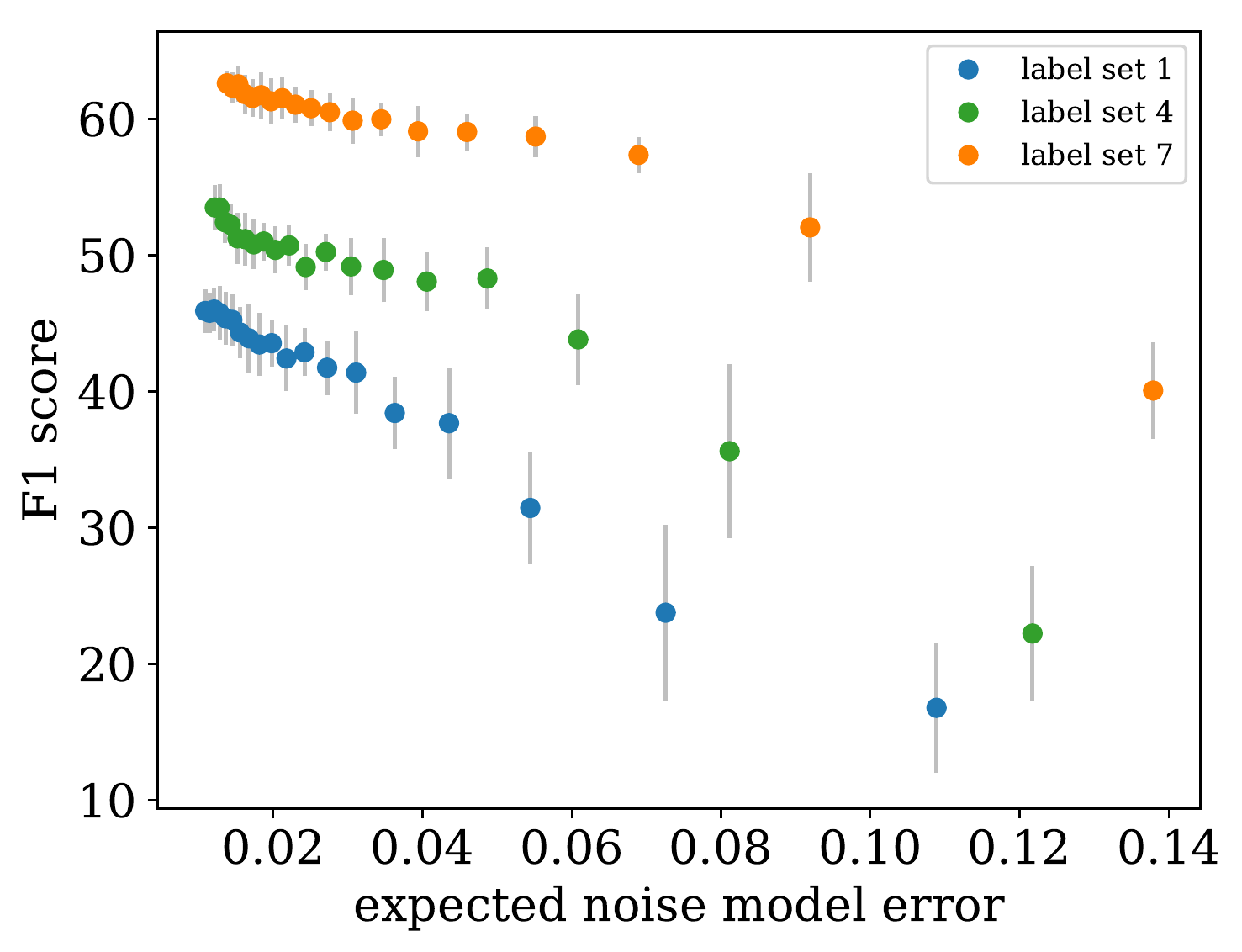}
    \caption{Relationship between the theoretically expected noise model error and the test performance (Accuracy/F1 score) of the base model for Clothing1M and NoisyNER (label set 1, 4 and 7) with Fixed Sampling. Each point corresponds to one sample size $n_i$ (cf. Figure \ref{fig:exp_ns_f1}). Additional plots in the Appendix. \label{fig:exp_estonian_ns_se_f1}}
\end{figure}

In Figure \ref{fig:exp_estonian_ns_se_f1}, the expected noise model error is plotted against the test performance of the base model for Fixed Sampling. They show a clear negative correlation. The influence of the noise distribution is visible in the different slopes in the plots for the different noisy label sets of NoisyNER. For all settings, the Pearson Correlation between the mean test performance and the expected noise model error is at least $-0.96$.

Increasing the number of clean labels is not only beneficial to the estimation of the noise model but also to the base model itself. To remove this effect, the experiment is repeated with a fixed amount of training instances for the base model ($n_i = 50$ for NoisyNER and $n_i=25$ for Clothing1M) and a varying amount of clean labels for the noise model estimation (plots in the Appendix). The same linear relationship can be seen. The Pearson Correlation is again at least $-0.96$ for all settings.

\begin{figure*}[t]
    \centering
    \subfigure[NoisyNER - label set 1]{
        \includegraphics[width=0.65\columnwidth]{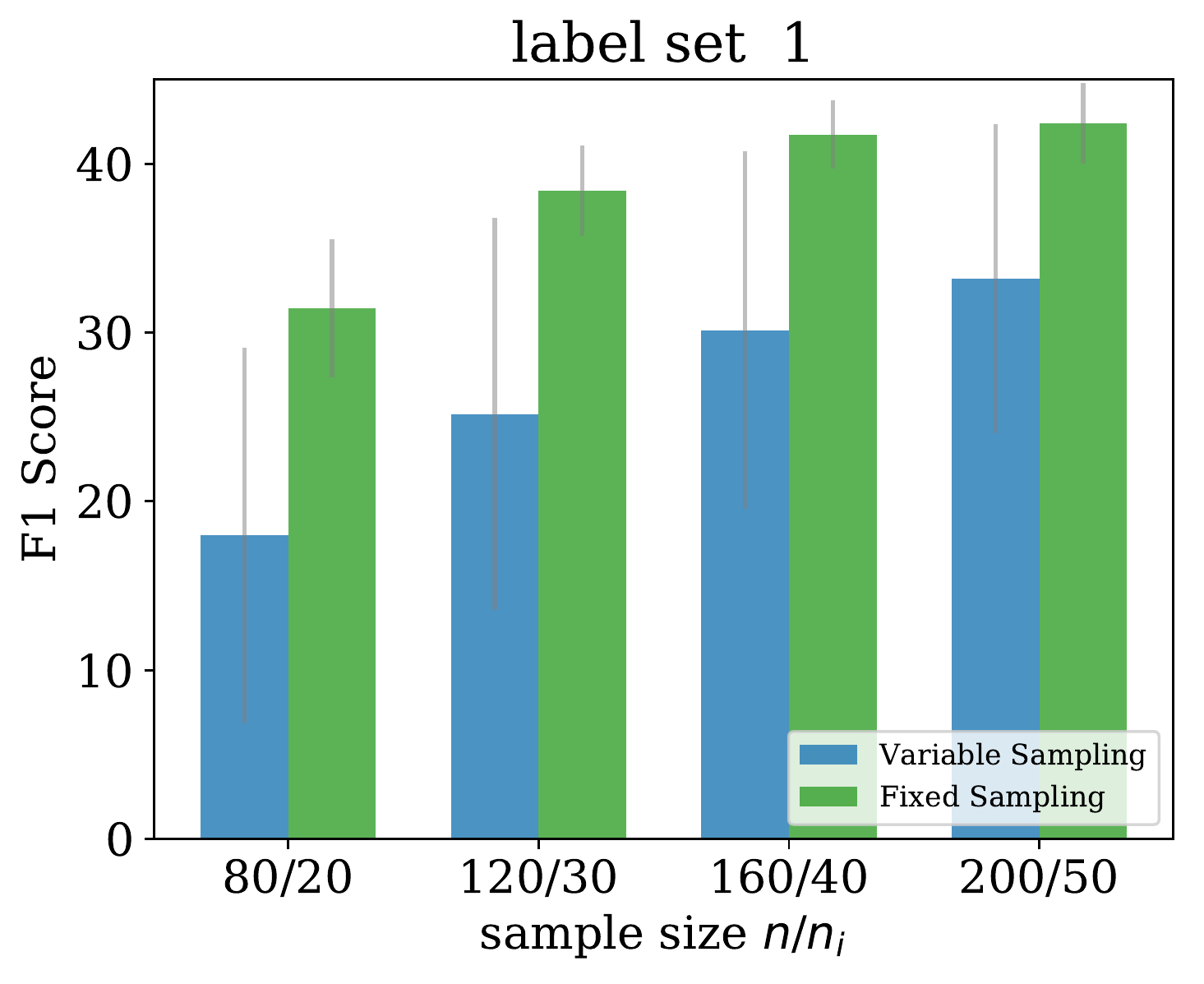}
    }
    \subfigure[NoisyNER - label set 4]{
        \includegraphics[width=0.65\columnwidth]{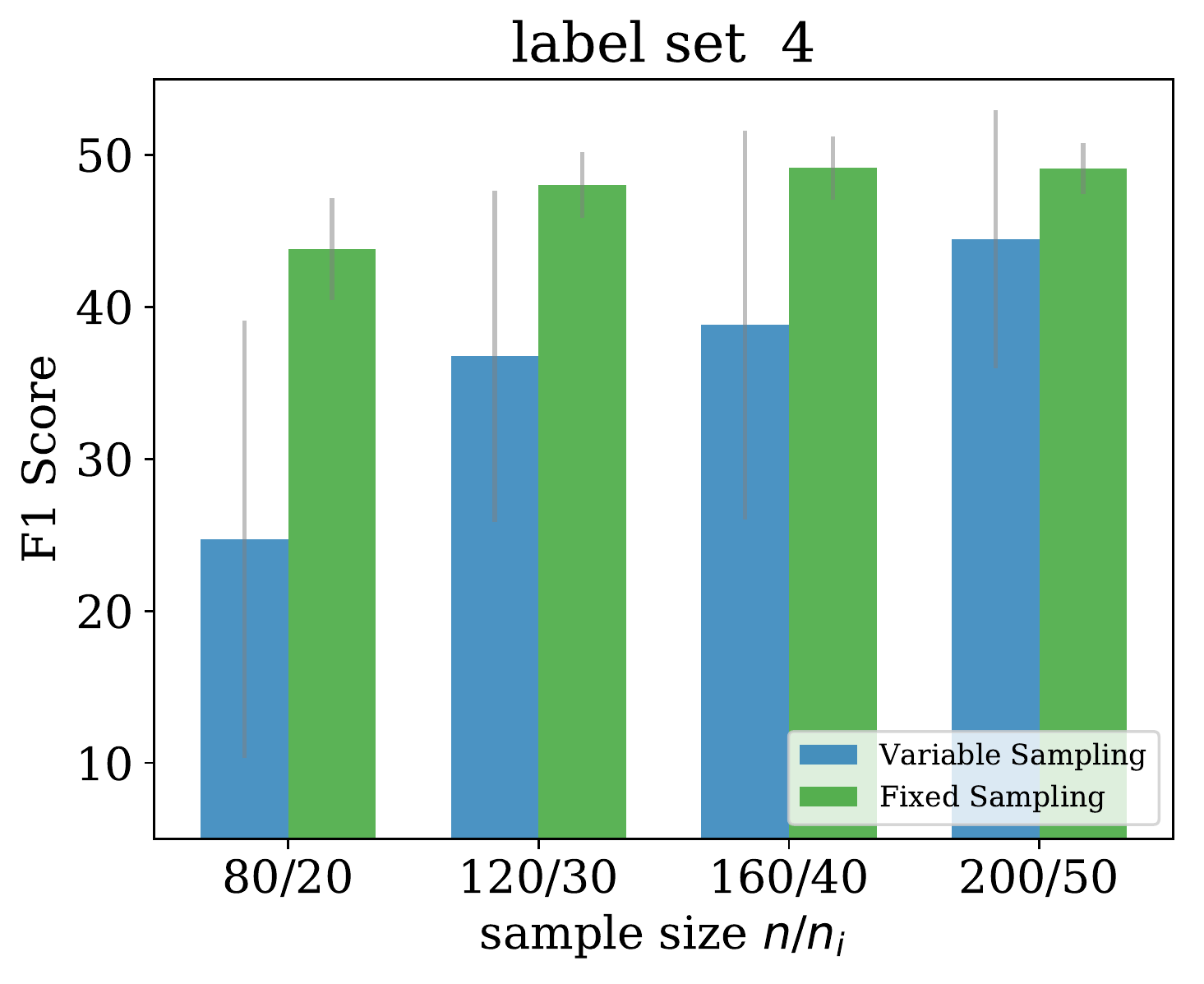}
    }
    \subfigure[NoisyNER - label set 7]{
        \includegraphics[width=0.65\columnwidth]{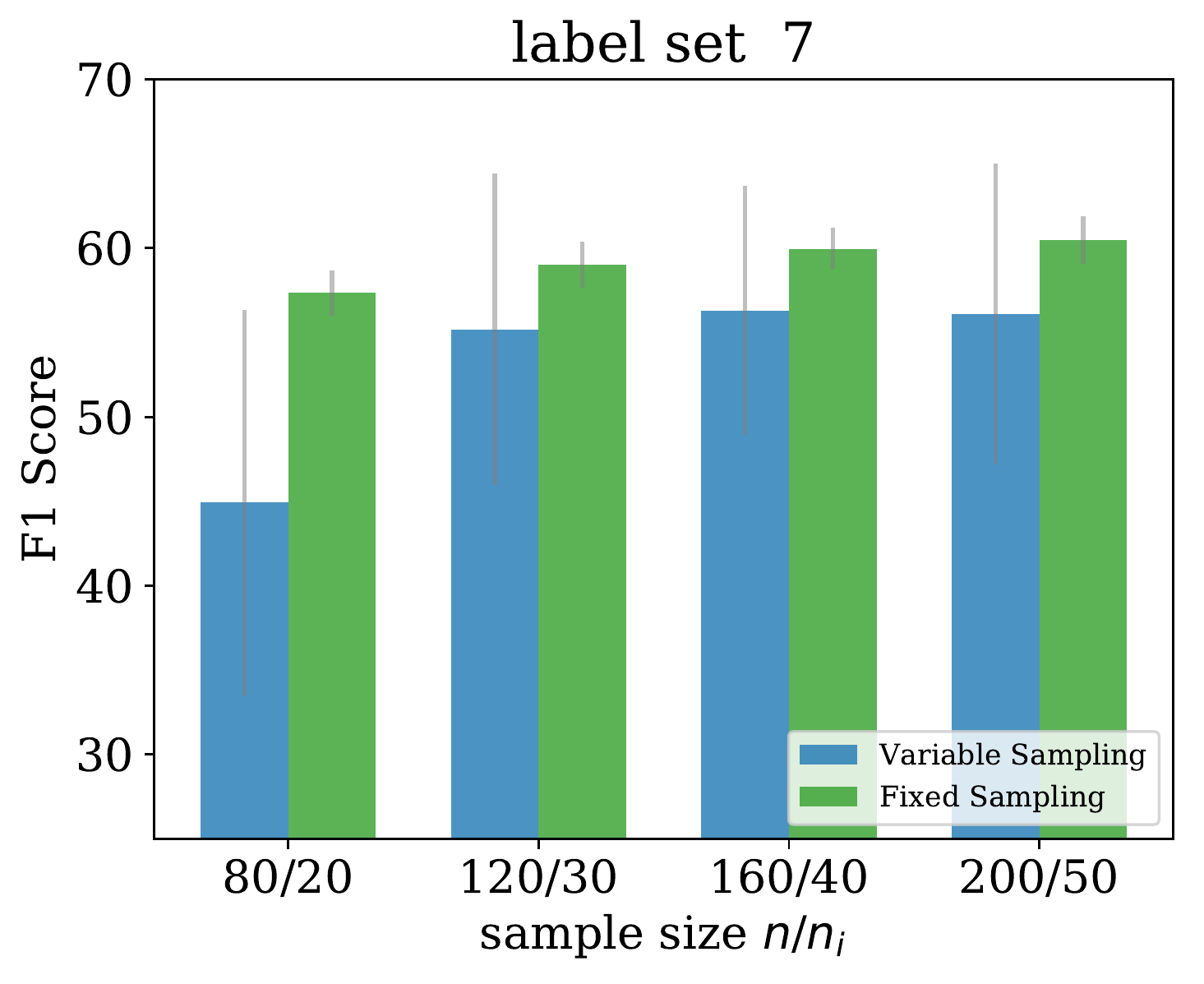}
    }
    \caption{Comparing Variable and Fixed Sampling by mean test performance (F1 score) of the base model on NoisyNER label set 1, 4 and 7. The two approaches were given the same amount of clean data with $n_i = n / k$ for Fixed Sampling. Further plots in the Appendix.}
    \label{fig:var-vs-fixed}
\end{figure*}

\subsection{Effect of Sampling during Estimation}

Above, we saw both from the theoretical analysis as well as in the experiments that Variable versus Fixed Sampling has a strong influence on the noise model estimation error. Figure \ref{fig:var-vs-fixed} shows that this effect also transfers to the performance of the base model on the test set. Fixed Sampling consistently outperforms Variable Sampling on the average performance across different noise levels. It also reduces the variance of the reached test performance, another important issue models trained on noisy labels suffer from. The same holds again when a fixed amount of training instances for the base model is used to remove the effect of just having more clean data.

As far as we are aware, Variable Sampling is more common in the literature, suggesting that Fixed Sampling could be a useful and model-independent way to improve noise handling.

\section{Other Related Work}

Apart from the publications mentioned in the previous sections, we want to highlight the following related work. The survey by \citet{noise/survey/Frenay2014} gives a comprehensive overview of the literature on learning with noisily labeled data. More recently, \citet{noise/survey/Algan19} and \citet{noise/hedderich2020survey}  surveyed deep learning techniques for noisy labels in the image classification domain and the NLP domain respectively. 

Other aspects of learning with noisy labels have been studied from a theoretical viewpoint. For the binary classification setting, \citet{noise/Liu} prove an upper bound for the noise rate. \citet{noise/Natarajan13} and \citet{noise/Patrini16} derive upper bounds on the risk when learning in a similar noise scenario. Assuming a known and correct noise model, \citet{noise/Rooyen17} derive upper and lower bounds on learning with noisy labeled data. The existence of anchor points is an important assumption for several recent works on estimating a noise matrix, i.a. \cite{noise/liu2016reweighting, noise/patrini2017losscorrection, noise/xia2019anchor}. \citet{noise/yao2020dualT} propose to learn an intermediate class to reduce the estimation error of the noise matrix. \citet{noise/Chen19} propose a formula to calculate the test accuracy assuming, however, that the test set is noisy as well. 

On the empirical side, \citet{noise/Veit17} model the relationship $p(y|\hat{y})$, i.e. the opposite of the noise models in this work. They also use pairs of clean and corresponding noisy labels. \citet{ner/Rahimi19} include the concept of noise transition matrices in a Bayesian framework to model several sources of noisy supervision. In their few-shot evaluation, they also leverage pairs of clean and noisy labels. \citet{noise/Ortego19} study the behaviour of the loss for different noise types. Some recent works have taken instance-dependent (instead of only label-dependent) noise into consideration both from a theoretic and an empiric viewpoint \cite{noise/Luo17, noise/Menon18, noise/Cannings2018, noise/Cheng20BoundedInstance, noise/Xia20PartDependent}. \citet{noise/Luo17} successfully modeled instance-dependent noise for an information extraction task, making instance-dependence an interesting future work for the NoisyNER dataset.

In the image classification domain, further noisy label datasets exist. The distantly supervised WebVision \cite{data/Webvision} was obtained by searching for images on Google and Flickr where the labels (and related words) are used as search keywords. It contains 2.4 million images but it lacks clean labels for the training data. The Food101N dataset \cite{data/FoodN} was created similarly, focusing, like Clothing1M, on a specific image domain. The authors provide ca. 300k images, ca. 50k of which have additional, human-annotated labels. The noise rate is lower with around 20\%. The older Tiny Images dataset \cite{data/TinyImages} consists of 80 million images at a very low resolution with 3526 training images having clean labels. Recently,  \citet{data/jiang2020controlledNoisyLabels} proposed artificially adding different amounts of incorrectly labeled images to existing datasets for evaluation. For the task of text sentiment analysis, \citet{noise/Wang19} proposed a dataset that obtained sentence level labels by exploiting document level ratings. Clean sentence level labels exist for two of their studied text domains.

\section{Conclusion}
In this work, we analyzed the factors that influence the estimation of the noise model in noisy label modelling for distant supervision. We derived the expected error of a noise model estimated from pairs of clean and noisy labels highlighting factors like noise distribution and sampling technique. Extensive experiments on synthetic and realistic noise showed that it matches the empirical error. We also analyzed how the noise model estimation affects the test performance of the base model. Our experiments show the strong influence of the noise model estimation and how theoretical insights on e.g. the sampling technique can be used to improve the task performance. A major contribution of this work is our newly created NoisyNER dataset, a named entity recognition dataset consisting of seven sets of labels with differing noise levels and patterns. It allows a simple and extensive evaluation of noisy label approaches in a realistic setting and for different levels of resource availability. With its interesting properties, we hope that it is useful for the community to develop noise-handling methods that are applicable to real scenarios.

\section{Acknowledgments}
We would like to thank the reviewers as well as Gabriele Bettgenhäuser and Alexander Blatt for their extensive and helpful feedback. This work is funded by the Deutsche Forschungsgemeinschaft (DFG, German Research Foundation) – Project-ID 232722074 – SFB 1102 and the EU Horizon 2020 project ROXANNE under grant number 833635.

\section{Ethics Statement}
Distant supervision and handling of noisy labels reduce the need for costly, labeled supervision. While this could lower the entrance bar for nefarious uses of machine learning, it also makes it possible to train and use machine learning approaches in low-resource scenarios such as under-resourced languages or applications developed by individuals or small organizations. It can, therefore, be a step towards the democratization of AI.

\bibliography{noise_est.bib}

\clearpage

\onecolumn

\renewcommand{\thesection}{\Alph{section}}

\newcommand{\sumjk}{\sum_{j=1;j \neq i}^{k}}
\newcommand{\Mtilp}{\Mtil_{ip}}
\newcommand{\Mtilq}{\Mtil_{iq}}
\newcommand{\SumSNC}{\sum\limits_{(y,\hat{y}) \in S_{NC}}}
\newcommand{\SumPQN}{\sum_{s_p,s_q,n_i=1}^n}
\newcommand{\SumNProb}{\sum_{n_i=1}^n P(N_i=n_i) \frac{1}{n_i^2}}

\graphicspath{{figures/}}

\textbf{Appendix to ``Analysing the Noise Model Error for Realistic Noisy Label Data''}

\section{Proof of Lemma 1}

We show that $\Ex[\Mtil_{ij}]$ is an unbiased estimator of $M_{ij}$.  Assuming a noise process following Equation 3 and 4 with $p(\hat{y}=j|y=i) = M_{ij}$ per definition. Let $(y,\hat{y}) \in S_{NC}$ be sampled independently at random with fixed $n = |S_{NC}|$ (this holds both for \textit{Fixed} and \textit{Variable Sampling}). For readability, we write $\Ex$ for $\Ex_{\sim S_{NC}}$ and analogously for $\Var$ and $\Cov$. Let $N_i$ be the random variable describing how often $y=i$, i.e. $N_i = \sum_{(y,\hat{y})\in S_{NC}} 1_{y=i}$. Let $S$ be the random variable describing how often $y=i$ and $\hat{y}=j$ on the same instance, i.e. $S = \sum_{(y,\hat{y}) \in S_{NC}} 1_{\{y=i, \hat{y}=j\}}$. Note that, given $N_i=n_i$ and independent sampling, $S$ is multinomially distributed with probabilities $M_i$ and $n_i$ trials. $\Mtil$ is defined by Equation 6. For $n_i = 0$, we define $\Mtil_{ij} = 0$.

\begin{align*}
    \Ex[\Mtil_{ij}] &= \Ex[\frac{\sum\limits_{(y,\hat{y}) \in S_{NC}} 1_{\{y=i, \hat{y}=j\}}}{\sum\limits_{y                 \in S_{NC}} 1_{\{y=i\}}}] \\
                &= \sum\limits_{s,n_i=1}^n P(S=s, N_i=n_i)\frac{s}{n_i} \\
                &= \sum\limits_{s,n_i=1}^n P(S=s|N_i=n_i)P(N_i=n_i)\frac{s}{n_i} \\
                &= \sum\limits_{n_i=1}^n P(N_i=n_i) \frac{1}{n_i} \sum\limits_{s=1}^{n_i} P(S=s|N_i=n_i)s\\
                &= \sum\limits_{n_i=1}^n P(N_i=n_i) \frac{1}{n_i} \Ex[S|N_i=n_i] \\
                &= \sum\limits_{n_i=1}^n P(N_i=n_i) \frac{1}{n_i} n_i M_{ij} \\
                &= M_{ij} \\
\end{align*}

\section{Proof of Theorem 1}
Let the assumptions be as in Lemma 1. Let $SE$ be defined as in Formula 7. Then using Lemma 1 it follows

\begin{align*}
\Ex[SE] &= \Ex\left[\sum_{i=1}^{k} \sum_{j=1}^{k} (M_{ij} - \Mtil_{ij})^2\right] \\
        &= \sum_{i=1}^{k} \sum_{j=1}^{k} \Ex[(\Ex[\Mtil_{ij}] - \Mtil_{ij})^2 ] \\
        &= \sum_{i=1}^{k} \sum_{j=1}^{k}  \Var[\Mtil_{ij}]
\end{align*}

\section{Proof of Theorem 2}

We will first show Theorem 2a. Let the assumptions be as in Theorem 1. Then

\begin{align*}
    \Var[\Mtil_{ij}] &= \Ex[\Mtil_{ij}^{2}]-\Ex[\Mtil_{ij}]^{2}\\
                    &= \Ex[\Mtil_{ij}^{2}] - M_{ij}^{2} \\
                    &= \sum\limits_{s,n_i=1}^n P(S=s, N_i=n_i)\frac{s^2}{n_{i}^{2}} - M_{ij}^{2} \\
                    &= \sum\limits_{s,n_i=1}^n P(S=s|N_i=n_i)P(N_i=n_i)\frac{s^2}{n_{i}^{2}} - M_{ij}^{2} \\
                    &= \sum\limits_{n_i=1}^n P(N_i=n_i) \frac{1}{n_{i}^{2}} \sum\limits_{s=1}^{n_i} P(S=s|N_i=n_i)s^2 -  M_{ij}^{2}\\
                    &= \sum\limits_{n_i=1}^n P(N_i=n_i) \frac{1}{n_{i}^{2}} \sum\limits_{s=1}^{n_i} \Ex[S^2|N_i=n_i] -  M_{ij}^{2}\\
                    &= \sum \limits_{n_i=1}^n P(N_i=n_i) \frac{1}{n_i^{2}}(n_i^{2}M_{ij}^2+n_i M_{ij}(1-M_{ij}))- M_{ij}^{2} \\
                    &= M_{ij}(1-M_{ij})\sum\limits_{n_i}P(N_i=n_i)\frac{1}{n_i} \\
\end{align*}

where we use $\Ex[S^2] = \Var[S] + \Ex[S]^2$ with $\Var[S] = n_i M_{ij} (1 - M_{ij})$ and $\Ex[S]^2 = n_{i}^2 M_{ij}^2$ as $S$ is multinomial distributed. 

For Theorem 2b, assuming Fixed Sampling with $n_i = n'_i \; \forall i$, it follows that $P(N_i = n'_i) = 1$ and 0 otherwise. The variance then simplifies to 

\begin{align*}
 \Var[\Mtil_{ij}] = \frac{M_{ij}(1-M_{ij})}{n'_i}
\end{align*}

\section{Multi-Flip Noise}
Multi-flip noise allows noise patterns where a true label can be corrupted into several other labels with different probability. For MNIST, the following noise pattern is used given noise level $\epsilon$. It is inspired by the similarity of digits in the dataset.

\begin{align*}
&M_{\text{multi}} = \\
&\begin{pmatrix}
        1 - \epsilon & 0 & 0 & 0 & 0 & 0 & 0 & 0 & \epsilon / 2 & \epsilon / 2 \\
        0 & 1 - \epsilon & 0 & 0 & 0 & 0 & 0 & \epsilon & 0 & 0 \\
        \epsilon / 3 & 0 & 1 - \epsilon & 2 * \epsilon / 3 & 0 & 0 & 0 & 0 & 0 & 0 \\
        0 & 0 & \epsilon / 2 & 1 - \epsilon & 0 & 0 & 0 & 0 & \epsilon / 2 & 0 \\
        \epsilon / 5 & \epsilon / 5 & 0 & 0 & 1 - \epsilon & \epsilon / 5 & \epsilon / 5 & 0 & \epsilon / 5 & 0 \\
        0 & 0 & 0 & 0 & 0 & 1 - \epsilon & \epsilon / 2 & 0 & \epsilon / 2 & 0 \\
        0 & 0 & 0 & 0 & 0 & \epsilon / 2 & 1 - \epsilon & 0 & \epsilon / 2 & 0 \\
        0 & 2 * \epsilon / 6 & 0 & 0 & \epsilon / 6 & 0 & 0 & 1 - \epsilon & 0 & 3 * \epsilon / 6 \\
        0 & 0 & 3 * \epsilon / 4 & 0 & 0 & 0 & 0 & 0 & 1 - \epsilon & \epsilon / 4 \\
        \epsilon / 3 & 0 & 0 & 0 & \epsilon / 3 & 0 & 0 & 0 & \epsilon / 3 & 1 - \epsilon
\end{pmatrix}
\end{align*}

\section{NoisyNER Dataset}
\sloppy
NoisyNER was created by automatically annotating the existing dataset by Laur (2013) using a distant supervision approach. The original dataset is licensed under the CC-BY-NC license. We make our seven label sets publicly available under CC-BY 4.0.

Lists of entity names were retrieved from Wikidata (dump 2019-10-14). For the label PER(son), all entities were used that had the property "instance of Q5" where Q5 is the Wikidata internal identifier. For LOC(ation), the identifiers were Q6256, Q515 and Q5107. For ORG(anization) the identifiers were Q43229, Q4830453, Q163740, Q79913, Q484652, Q245065, Q1156831, Q157031, Q783794, Q6881511, Q21980538, Q4287745, Q2659904, Q4438121 and Q178790. If a word or several consecutive words from the text matched an entity name, the corresponding entity type was assigned to the word(s). For conflict cases where multiple entities matched a token, the entity that resulted in the longest sequence of matched tokens was used. The original tokenization of the dataset was used. The automatic annotation was performed via the ANEA tool \cite{distant/hedderich2021anea}.

Manual heuristics were obtained by inspecting the automatic annotation and providing corrections for common mistakes. 

\begin{itemize}
    \item \textbf{Label Set 1}: No heuristics.
    \item \textbf{Label Set 2}: Applying Estonian lemmatization to normalize the words.
    \item \textbf{Label Set 3}: Splitting person entity names in the list, i.e. both first and last names can be matched separately. Person names must have a minimum length of 4. Also, lemmatization. 
    \item \textbf{Label Set 4}: If entity names from two different lists match the same word, location entities are preferred. Also, lemmatization.
    \item \textbf{Label Set 5}: Locations preferred, lemmatization, splitting names with minimum length 4.
    \item \textbf{Label Set 6}: Removing the entity names "kohta", "teine", "naine" and "mees" from the list of person names (high false-positive rate). Also, all of label set 5.
    \item \textbf{Label Set 7}: Using alternative, alias names for organizations. Using additionally the identifiers Q82794, Q3957, Q7930989, Q5119  and Q11881845 for locations and Q1572070 and Q7278 for organizations. Also, all of label set 6.
\end{itemize}

\begin{figure}
    \centering
    \includegraphics[width=0.2\textwidth]{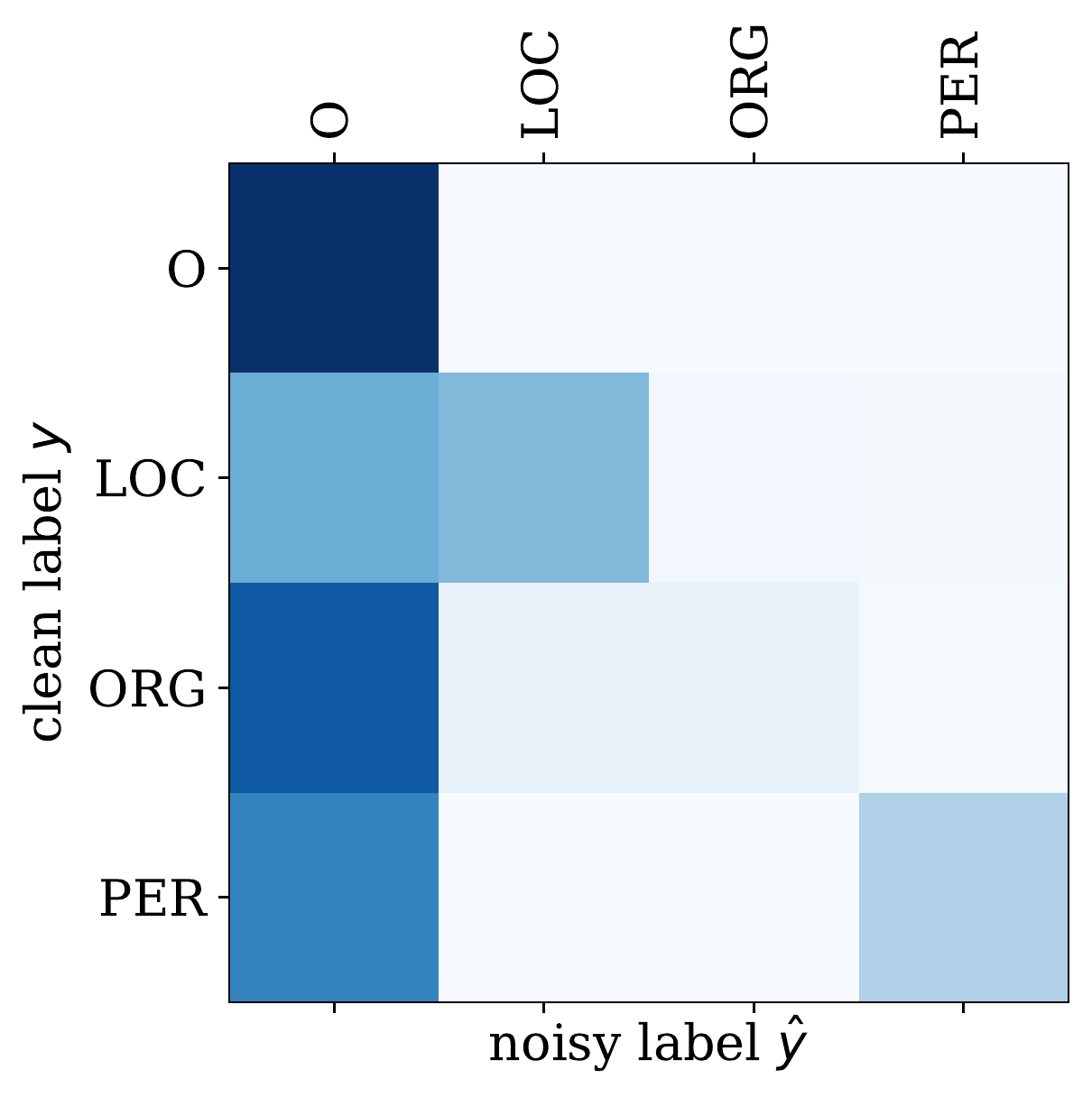}
    \includegraphics[width=0.2\textwidth]{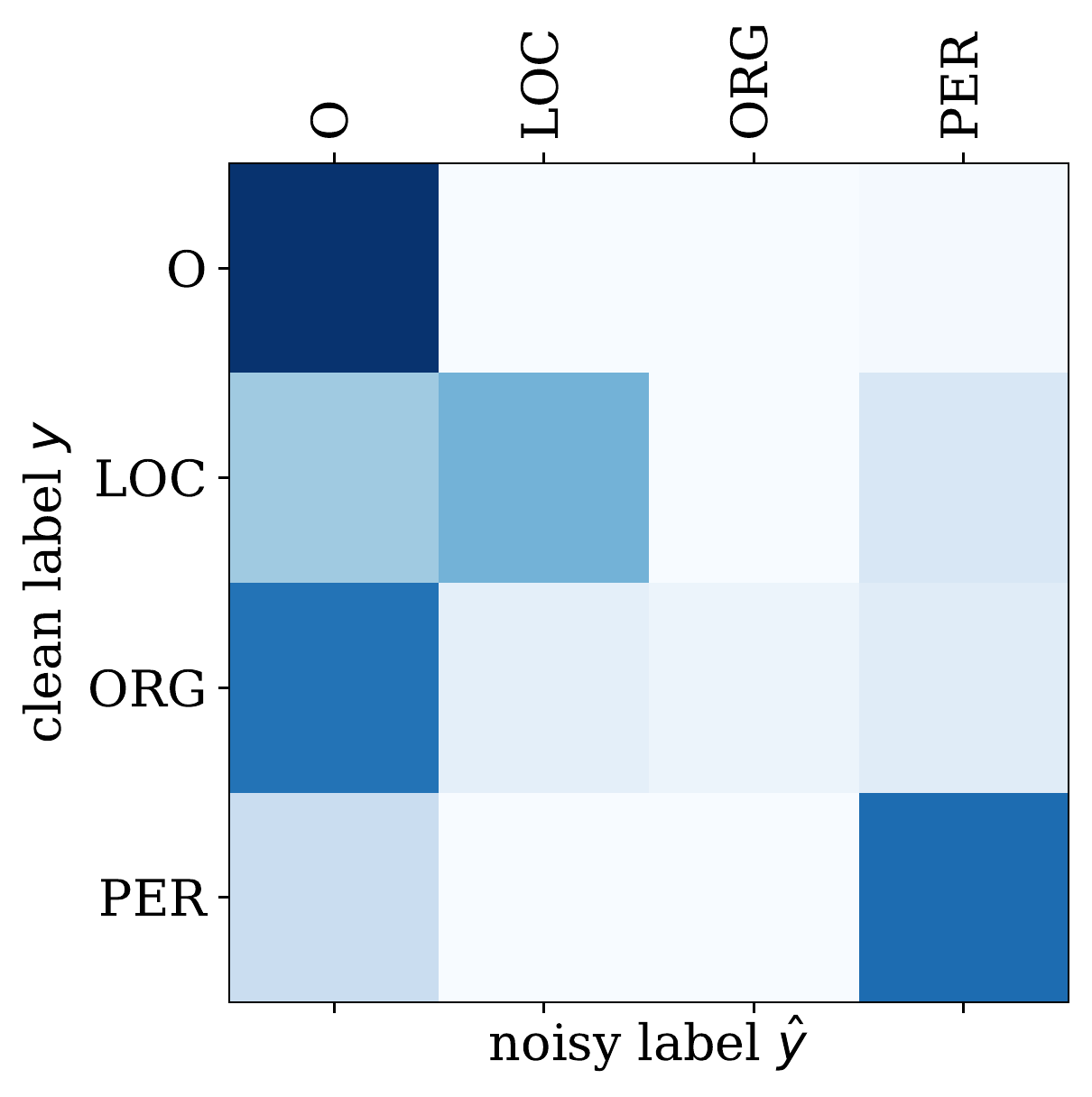}
    \includegraphics[width=0.2\textwidth]{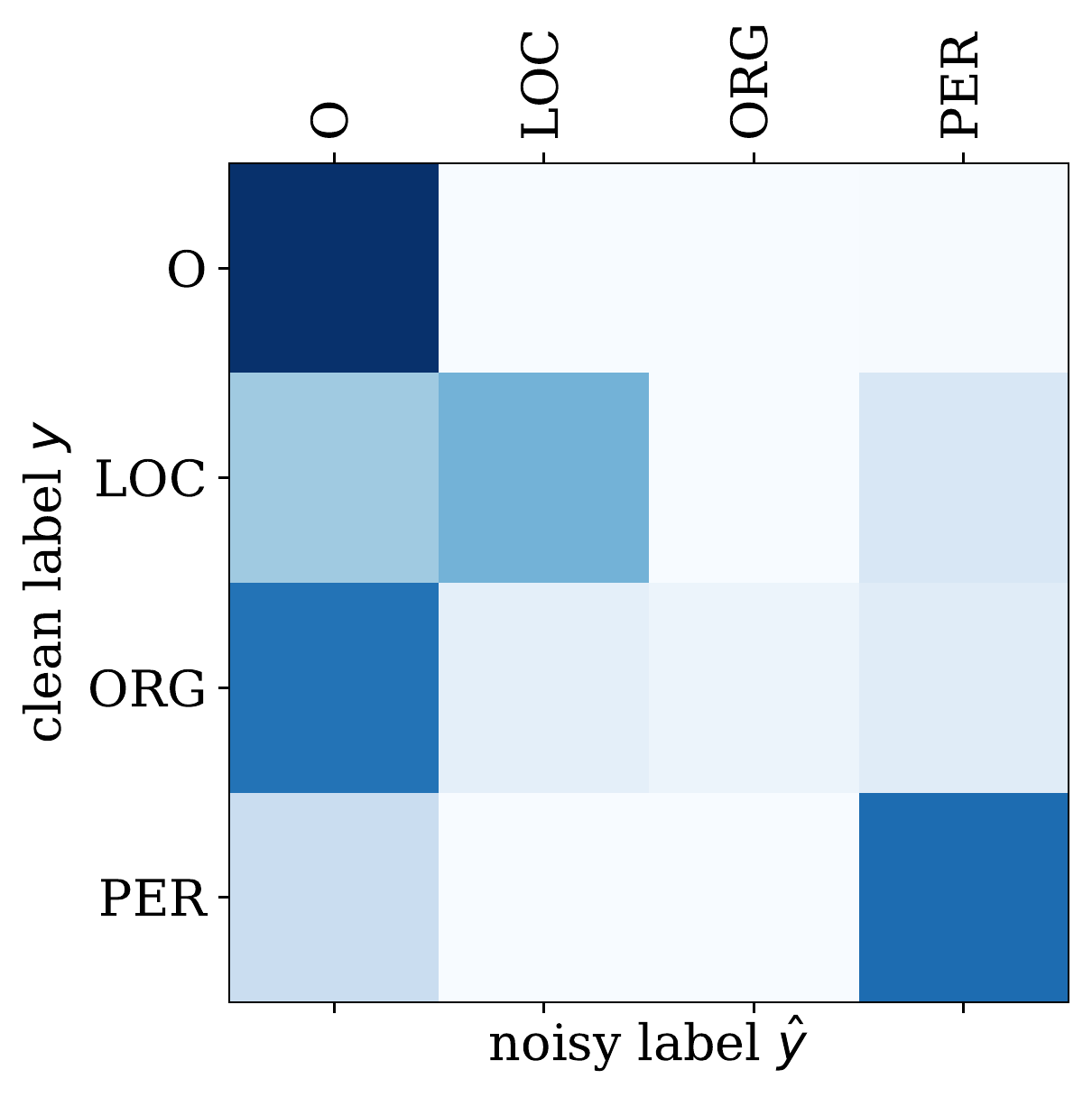}
    \caption{Noise matrices for NoisyNER (label sets 2, 5 and 6) computed over all pairs of clean and noisy labels.}
\end{figure}

\section{Experimental Setup for ``Analysing the Base Model Performance''}

In these experiments, we explore the relationship between the estimation of the noise model and the test performance of the base model on clean data. The experiments are performed on the NoisyNER dataset and the Clothing1M dataset.

\subsection{NoisyNER}
We split the NoisyNER data into an 80/10/10 train/dev/test split. Following the standard approach for named entity recognition (Tjong Kim Sang \& De Meulder, 2003), we evaluate with the micro-average F1 score excluding the non-entity label. Each experiment is repeated 50 times to report mean and standard deviation. For each run of the experiment, a small subset $D_C$ is sampled uniformly at random from the full dataset. Either Fixed or Variable Sampling are used. The noise model $\Mtil$ is estimated on this sample.

As base model, we use the architecture for named entity recognition proposed by Hedderich \& Klakow (2018). It consists of a Bi-LSTM model (state size 300 for each direction) with an additional fully connected layer (size 100) and a softmax classification layer. The context size is 3 on each side. The tokens are embedded using fixed FastText embeddings (Grave et al., 2018). The weights of the noise matrix $\Mtil$ are fixed during training. The model is optimized using Adam (Kingma \& Ba, 2014) with a learning rate of 0.001. Training is performed for 80 epochs. We test using the learnt weights of the epoch that performed best according to the clean development set. In each epoch, the model is alternately trained on the clean and the noisy instances (the latter with the noise model). Following the findings of the original authors, the model is trained with a noisy subset of the full noisy data. This noisy subset is sampled uniformly at random for each epoch. We set the size to 15 times $|D_C|$. 

In the experiments where the number of clean instances is fixed for the base model (to separate the effect on noise estimation from the effect of the base model just having access to more clean data), the base model is trained on 50 instances per class for Fixed Sampling and 50 instances in total for Variable Sampling. The other hyperparameters remain the same in both cases.

\subsection{Clothing1M}
For the Clothing1M dataset, we extract the 33,747 images from the original dataset where both clean and noisy labels are available. Among these images, we use a 90/10 split for training/test sets. Again we take a small fraction out from the training set as our $D_C$ and in training, the noise matrix is estimated by $S_{NC}$. \\
As base model, we employ a pre-trained ResNet18 (He et al, 2016) classifier obtained from the Pytorch library (Paszke et al, 2019). Specifically, we first switch the CNN header (i.e. the final linear classifier) to have the correct output dimension (14 in our case). Then we only train the new header and freeze all other layers. We fine-tune the network for 10 epochs. In training, an Adam optimizer with a learning rate of 0.005 is used for training. Again the noisy subset used in training is sampled uniformly at random for each epoch, with a size of 10 times the $|D_C|$.

In the experiments where the amount of clean labels is fixed for the base model (to separate the effect on noise estimation from the effect of the base model just having access to more clean data), the base model is trained on 25 clean instances per class for Fixed Sampling and 250 instances in total for Variable Sampling. The other hyperparameters remain the same in both settings.

\subsection{Additional Plots}

\begin{figure}[h]
    \centering
    \includegraphics[trim=0cm 1cm 0cm 0, height=3.2cm]{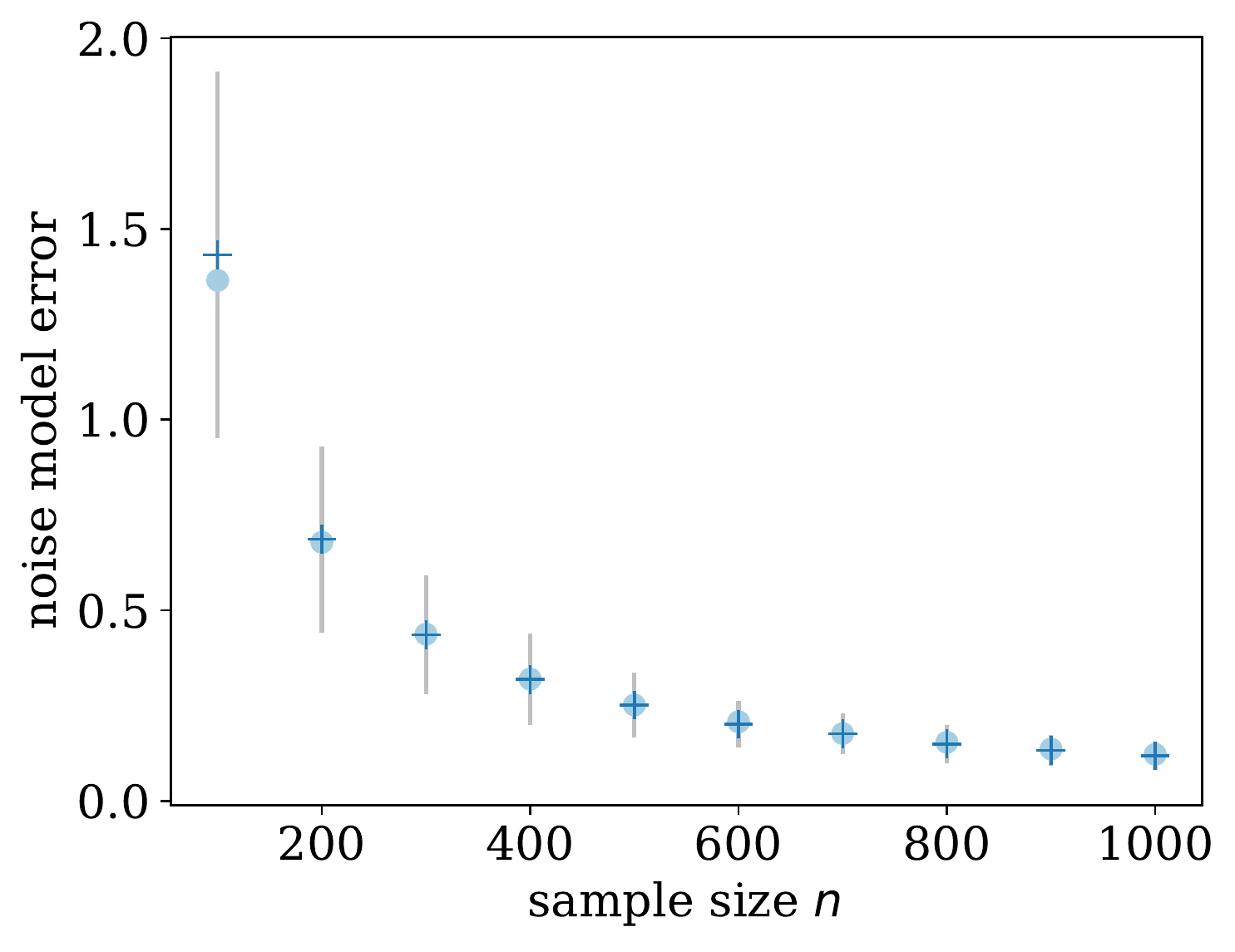}
    \includegraphics[trim=0cm 1cm 0cm 0, height=3.2cm]{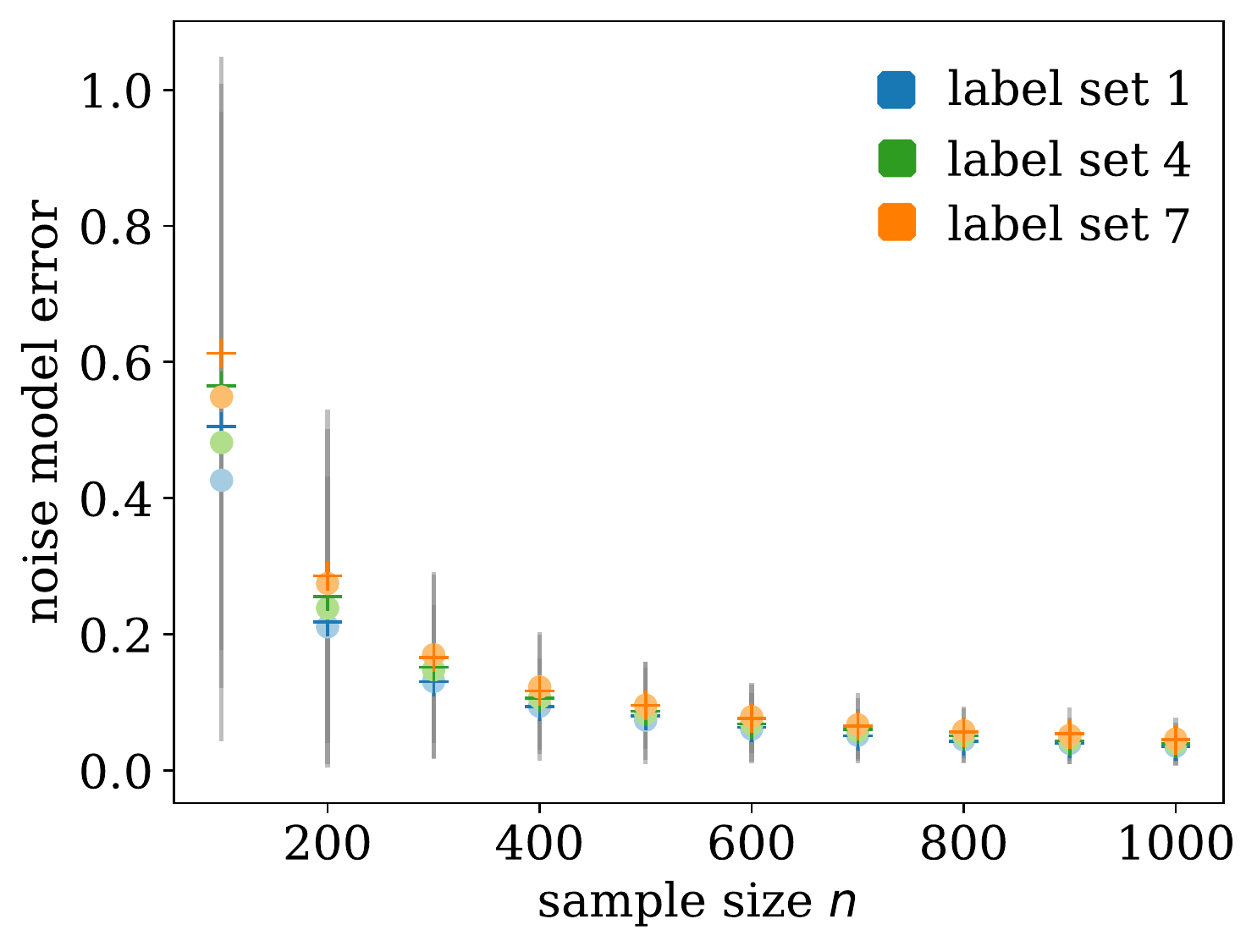}
    \caption{Comparison between the theoretically expected mean error (circle marker) and the empirically measured mean error (cross) of the noise model for Clothing1M and NoisyNER (label sets 1, 4 and 7) with \textbf{Variable Sampling}. Error bars show the empirical standard deviation.} 
\end{figure}

\begin{figure}[h]
    \centering
    \includegraphics[height=3.2cm]{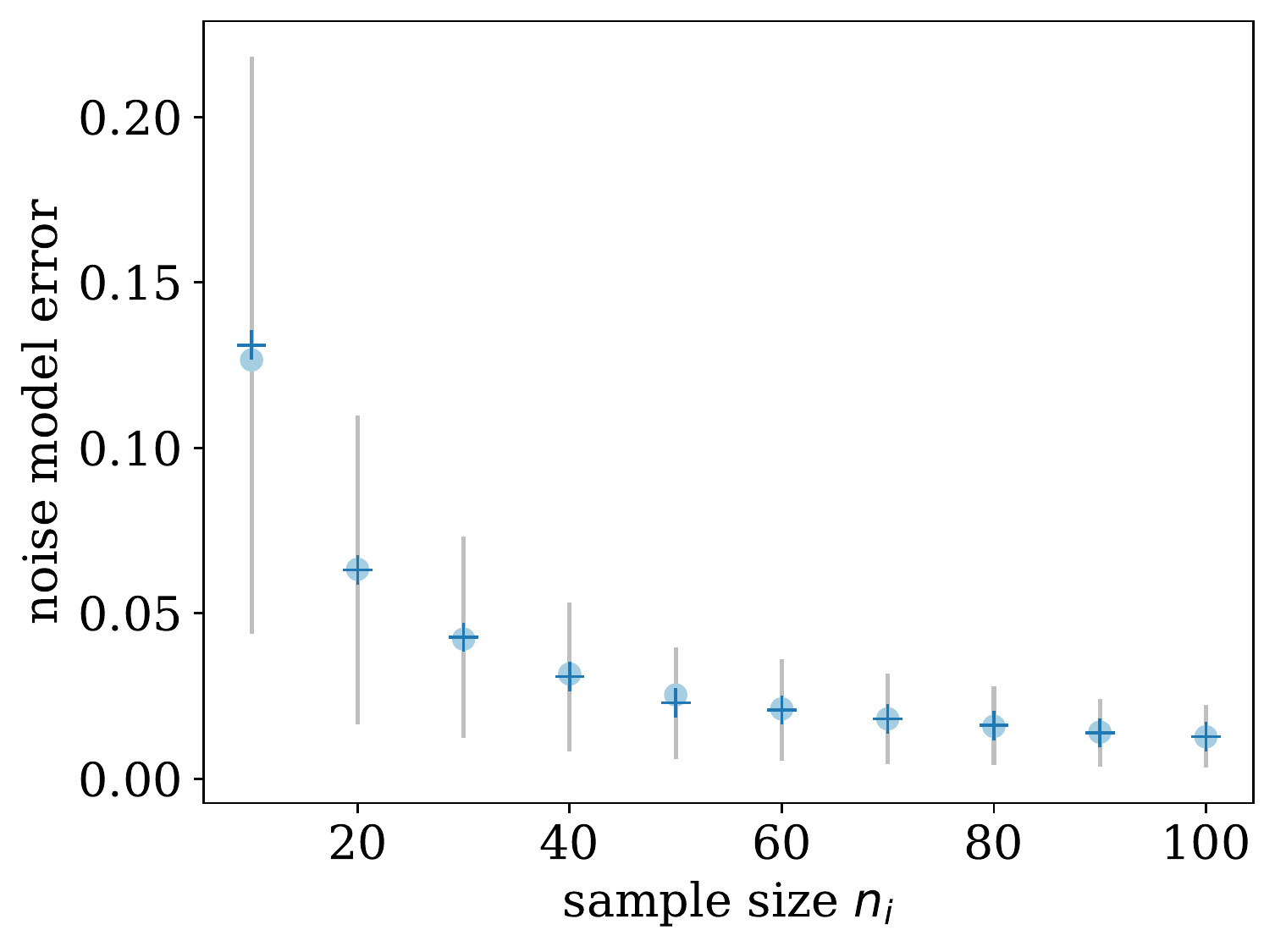}
    \includegraphics[height=3.2cm]{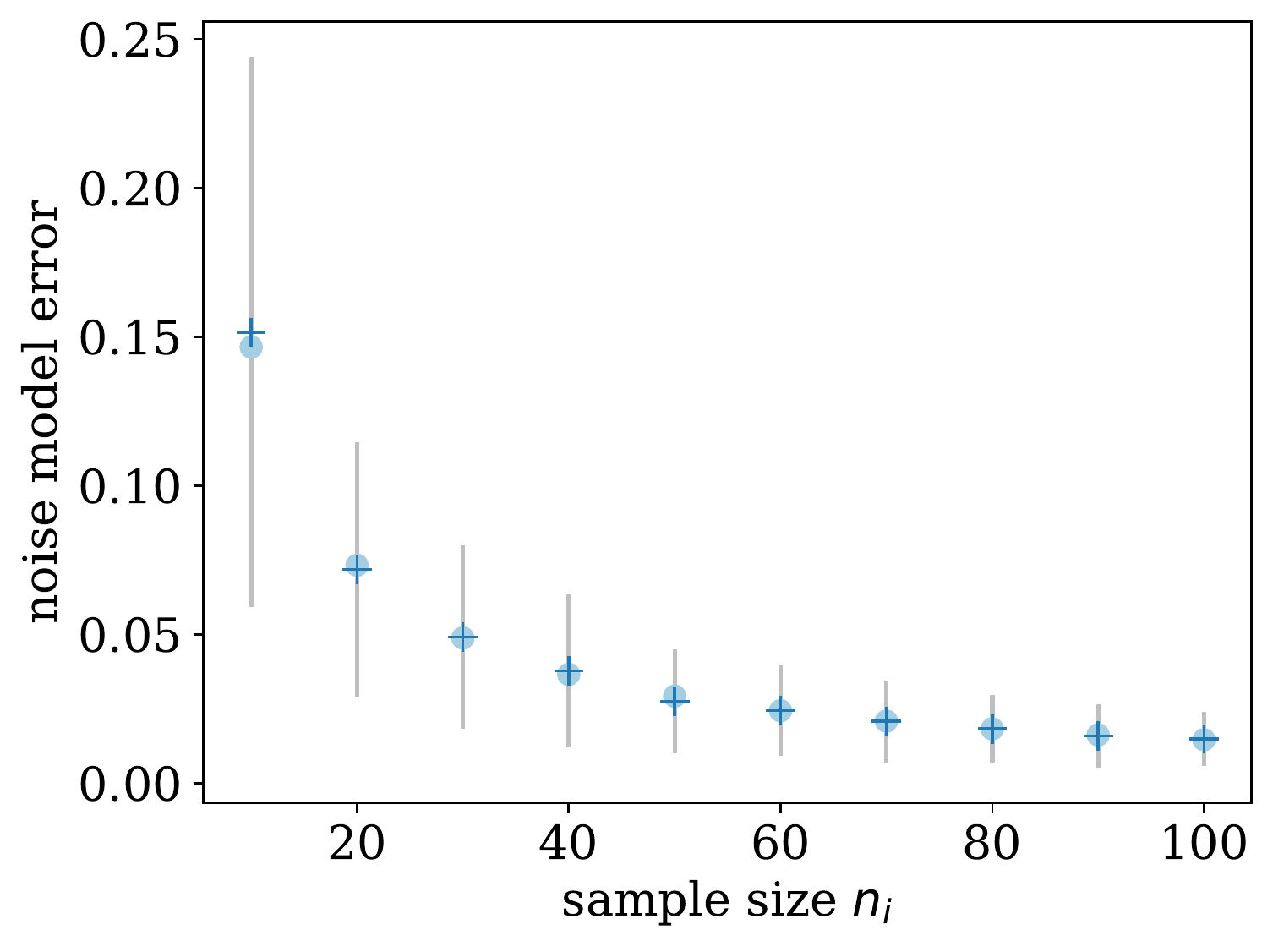}
    \includegraphics[height=3.2cm]{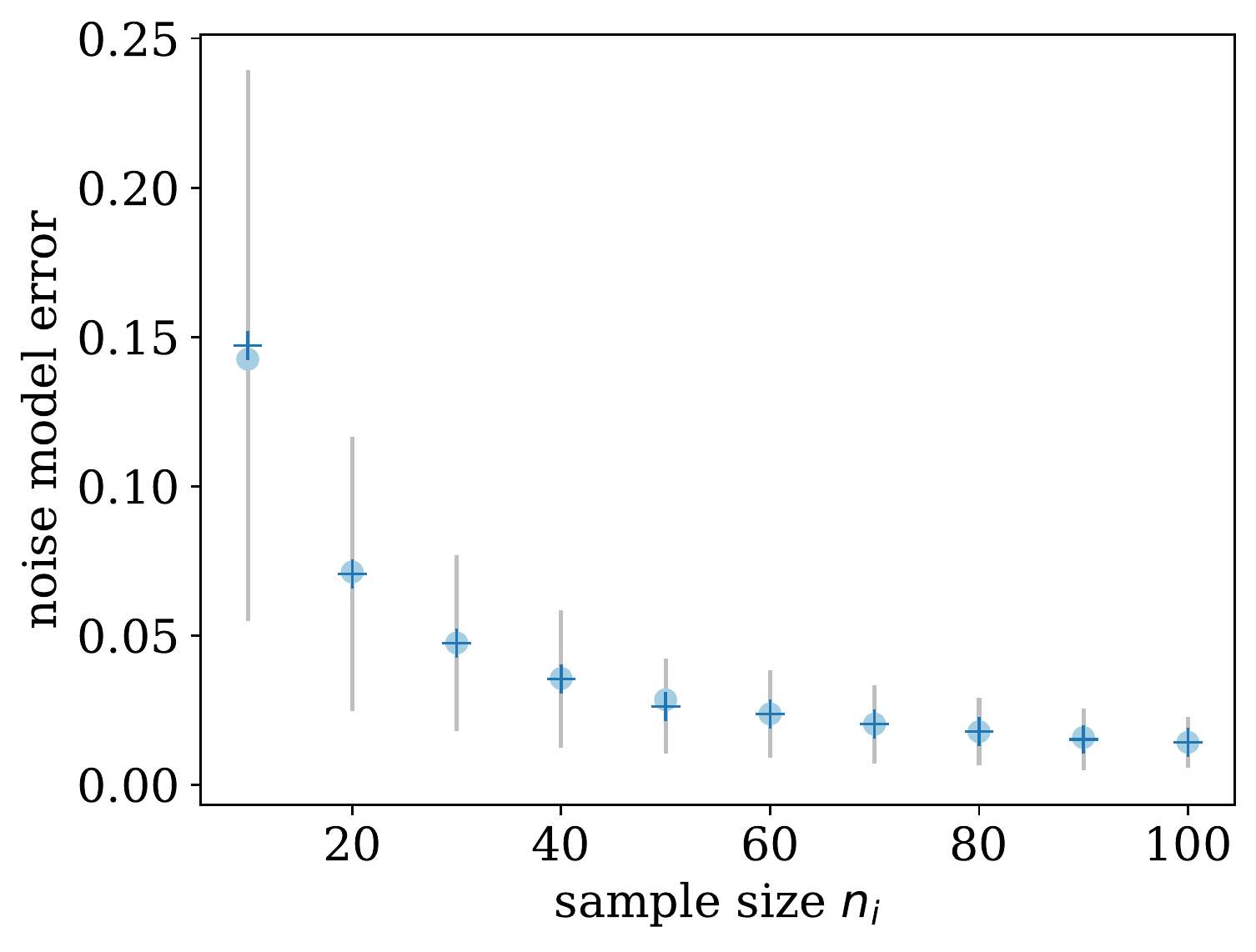}
    \includegraphics[height=3.2cm]{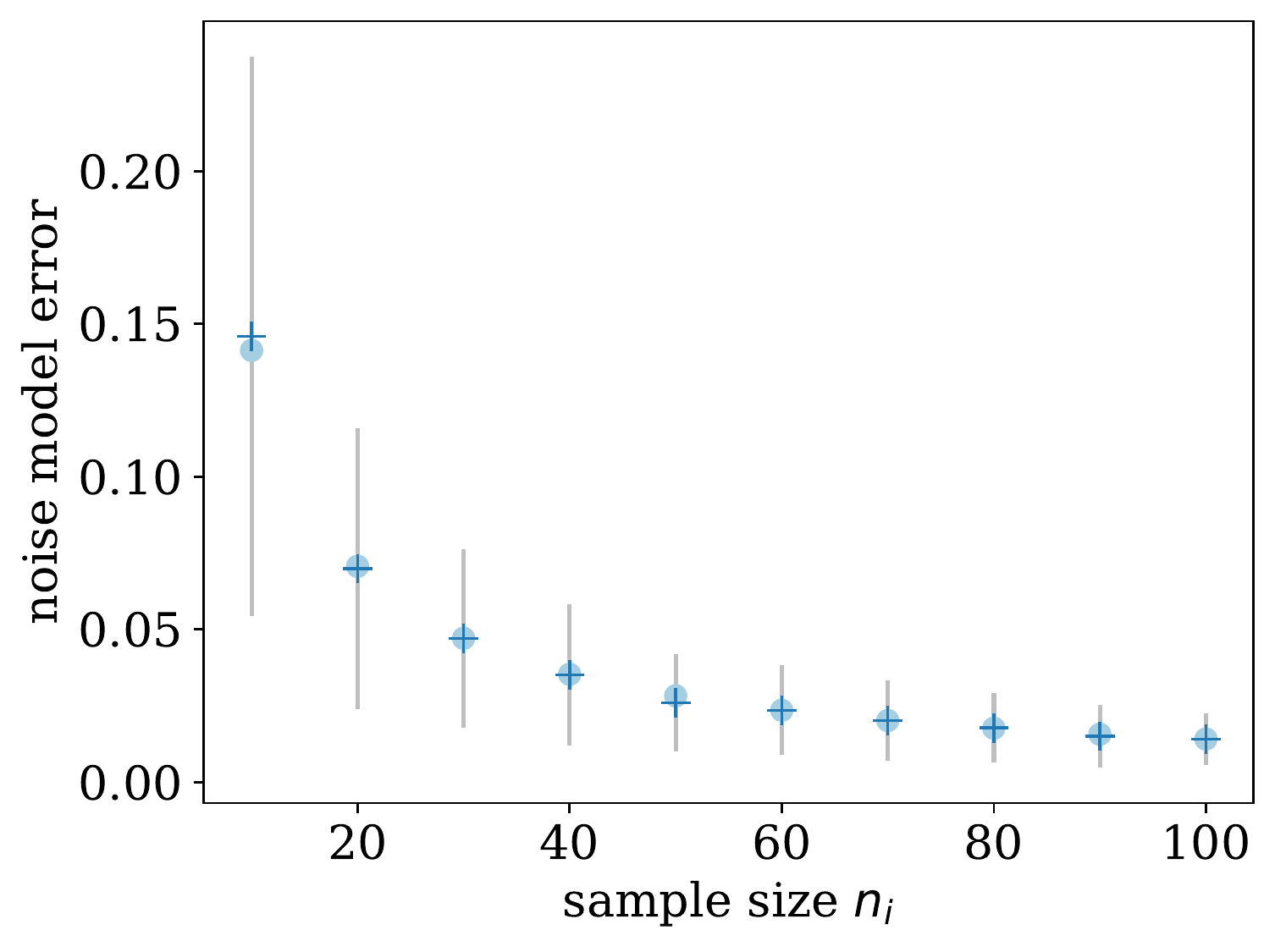}
    \caption{Comparison between the theoretically expected mean error (circle marker) and the empirically measured mean error (cross) of the noise model on the NoisyNER dataset for the noisy label sets 2, 3, 5 and 6 for \textbf{Fixed Sampling}. The grey error bars show the empirical standard deviation. }
\end{figure}

\begin{figure}[h]
    \centering
    \includegraphics[height=3.2cm]{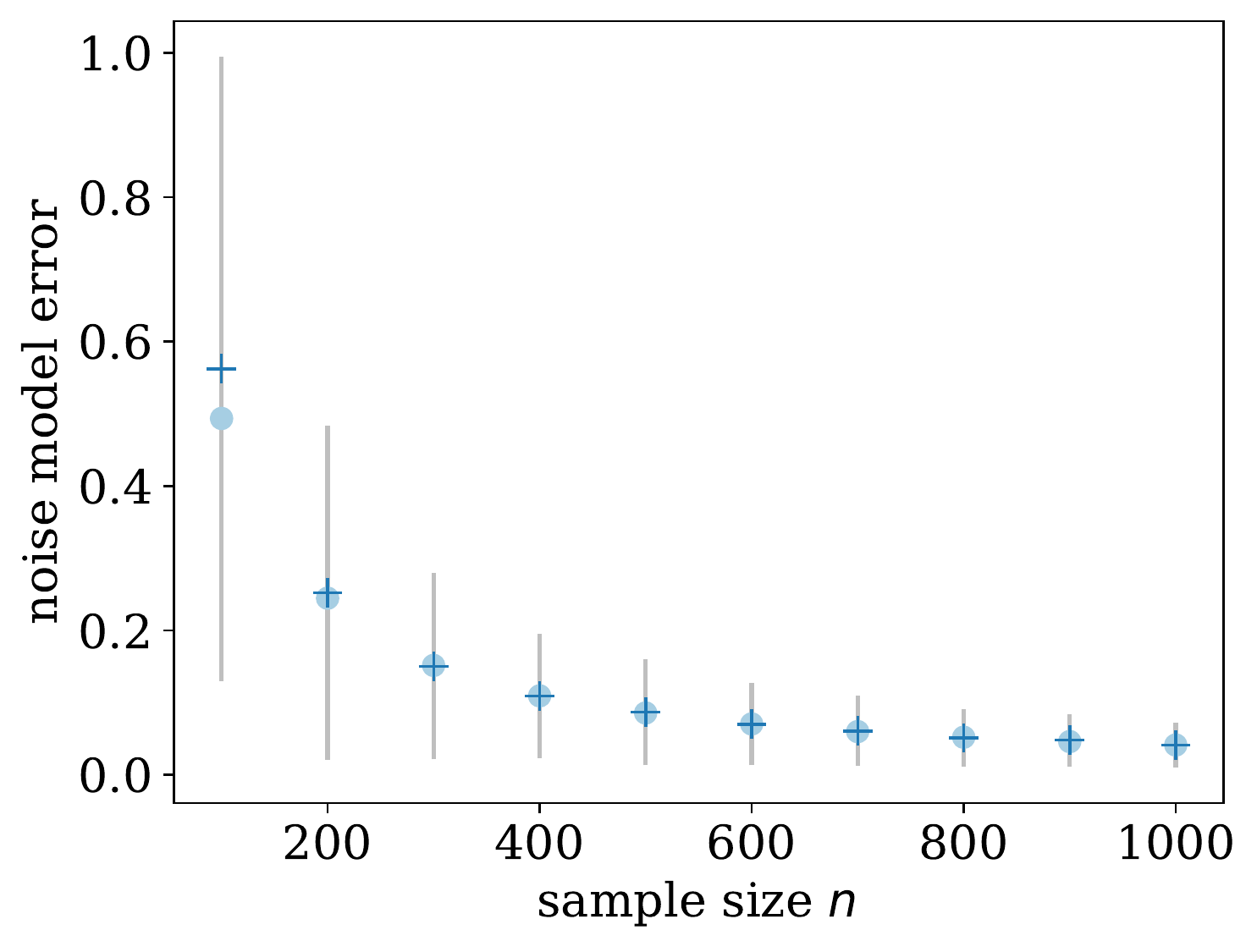}
    \includegraphics[height=3.2cm]{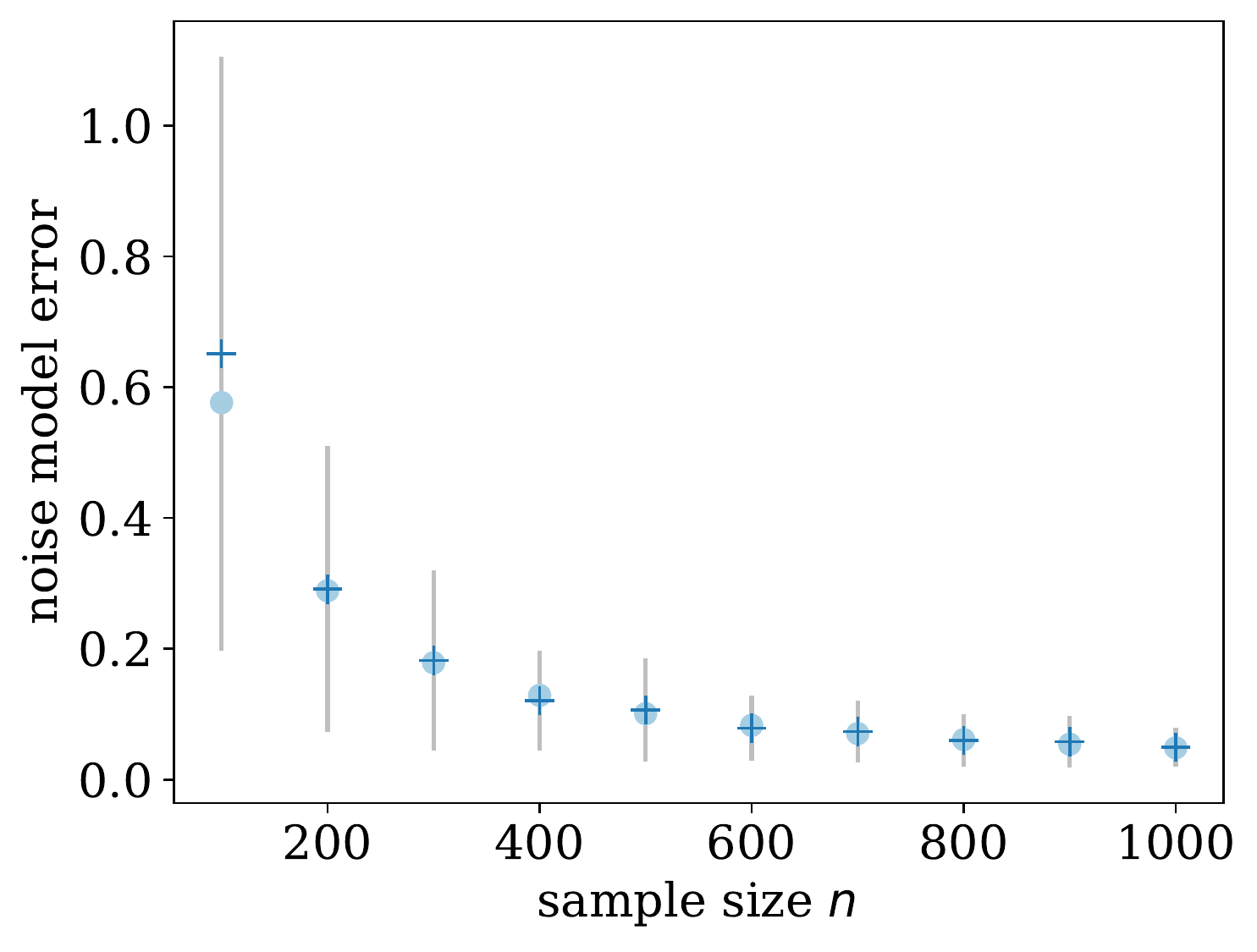}
    \includegraphics[height=3.2cm]{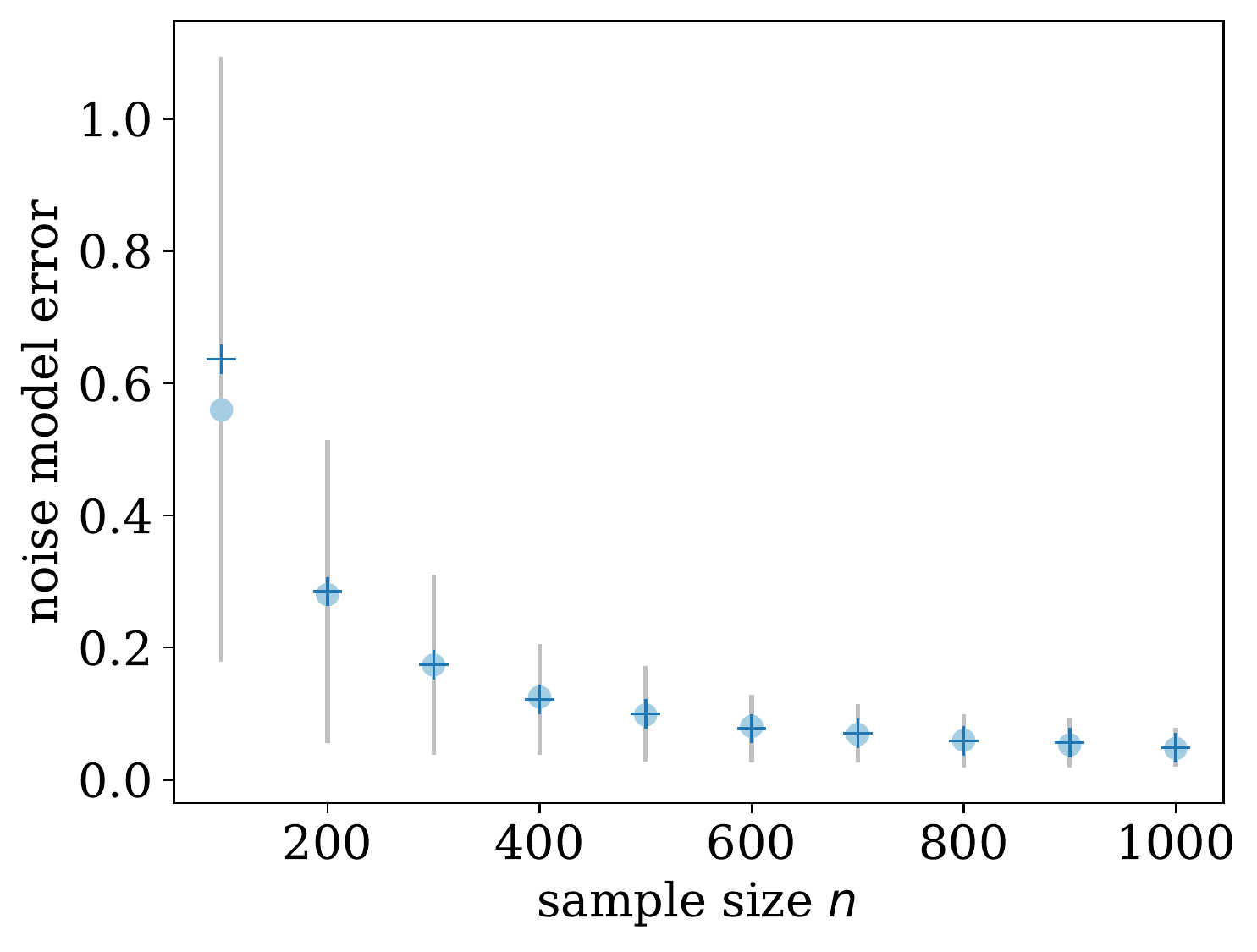}
    \includegraphics[height=3.2cm]{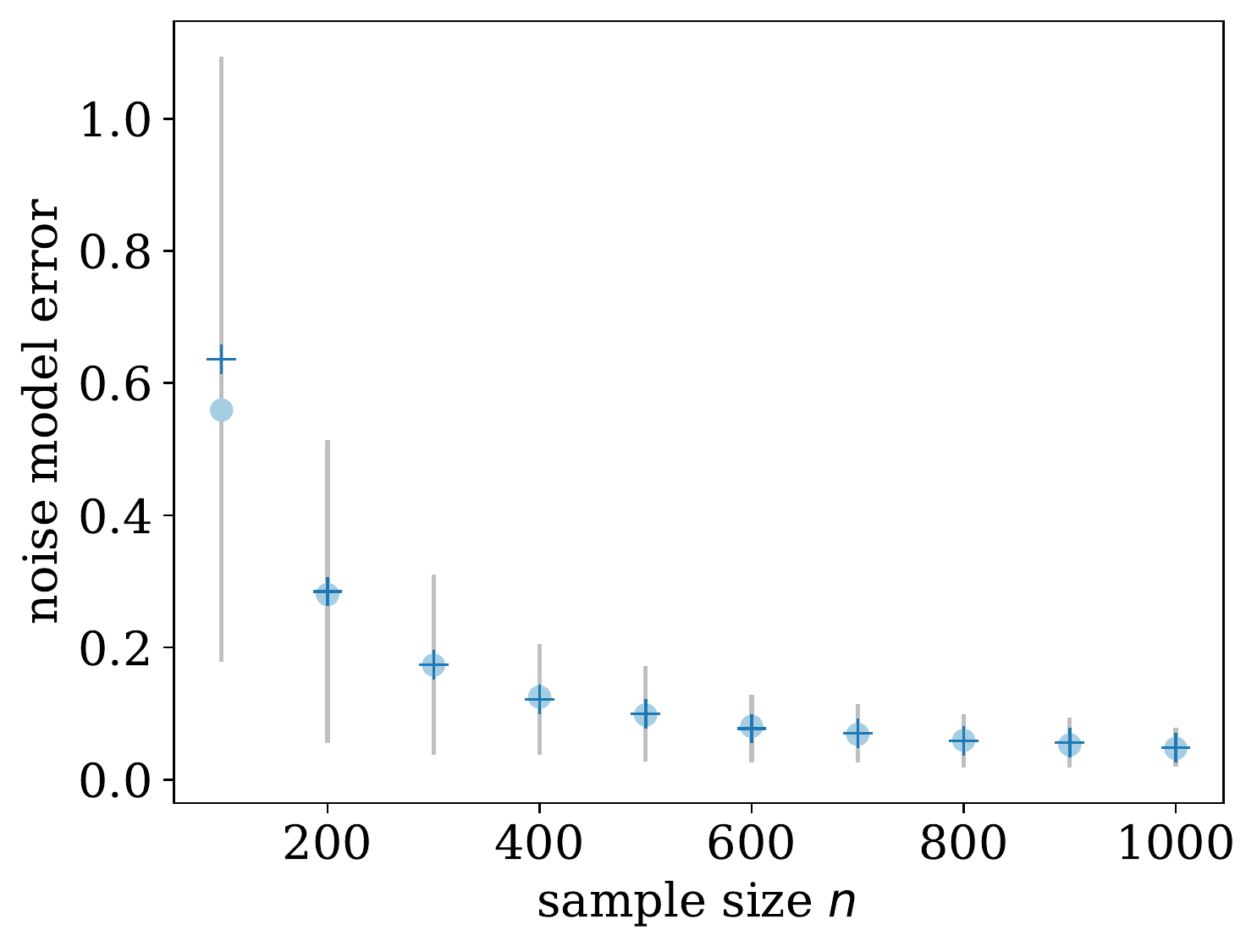}
    \caption{Comparison between the theoretically expected mean error (circle marker) and the empirically measured mean error (cross) of the noise model on the NoisyNER dataset for the noisy label sets 2, 3, 5 and 6 for \textbf{Variable Sampling}. The grey error bars show the empirical standard deviation. }
\end{figure}

\begin{figure}
\centering
\setlength{\tabcolsep}{0pt}
\renewcommand{\arraystretch}{0.7}
\begin{tabular}{p{0.4cm}cccc}
 & \hspace{0.45cm} Uniform & \hspace{0.45cm} Single-Flip & \hspace{0.45cm} Multi-Flip & \\
 \rotatebox{90}{\hspace{-1.5cm}Fixed Sampling}
 & \includegraphics[width=0.265\textwidth]{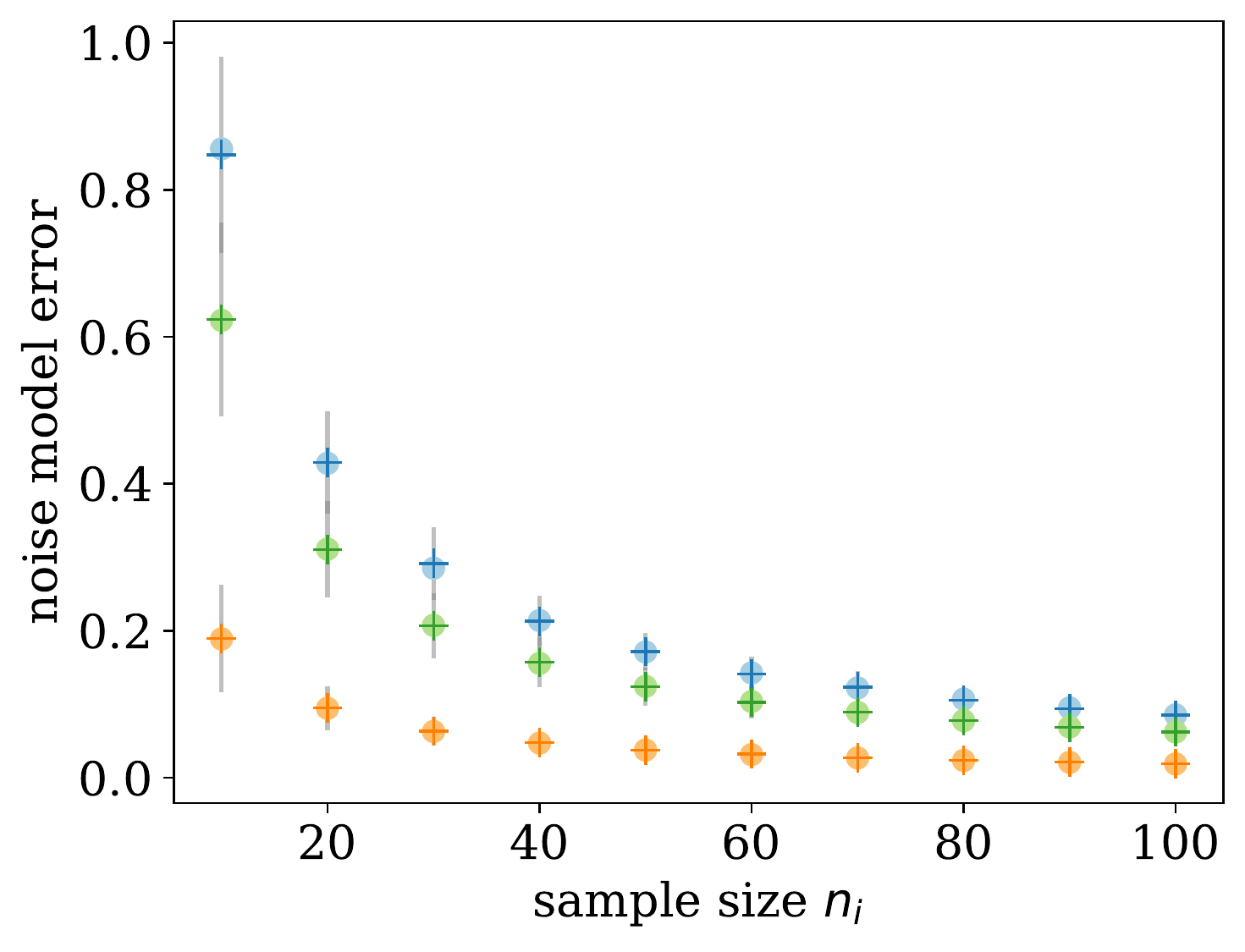}
 & \includegraphics[width=0.265\textwidth]{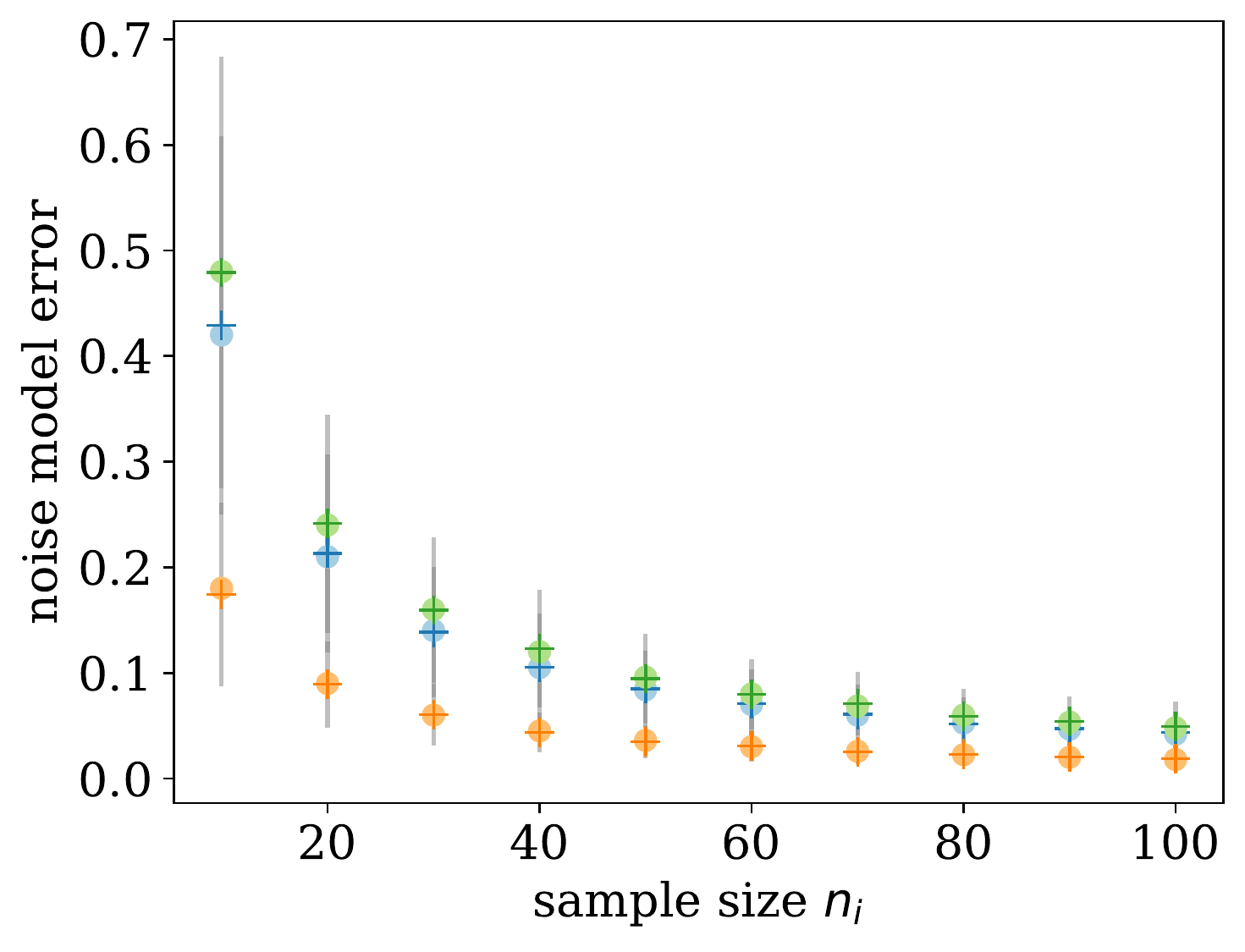}
 & \includegraphics[width=0.265\textwidth]{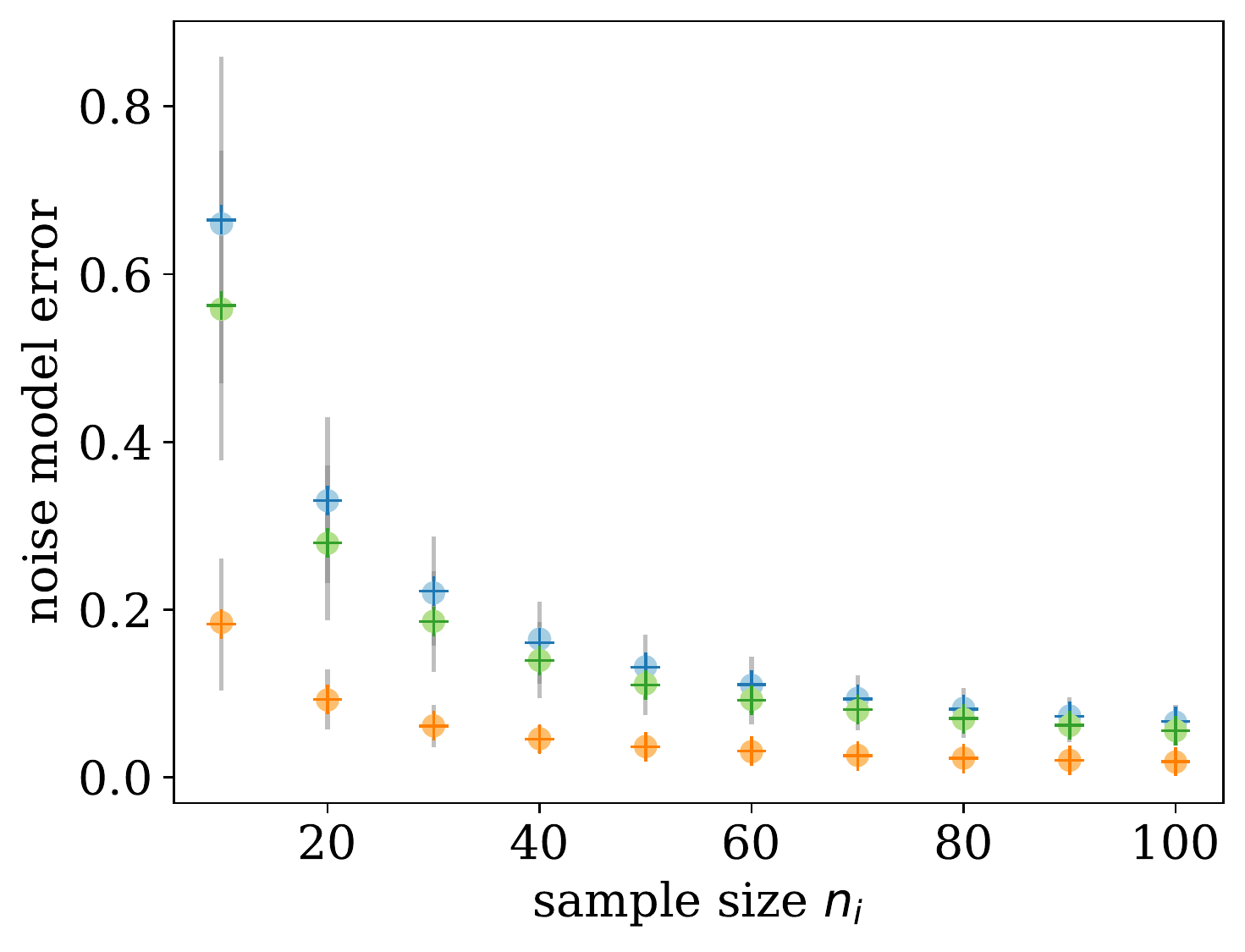}
 & \raisebox{1.6cm}{\includegraphics[width=0.15\textwidth]{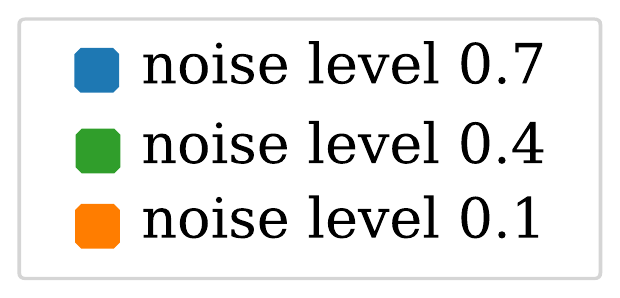}}\\
 & \includegraphics[width=0.265\textwidth]{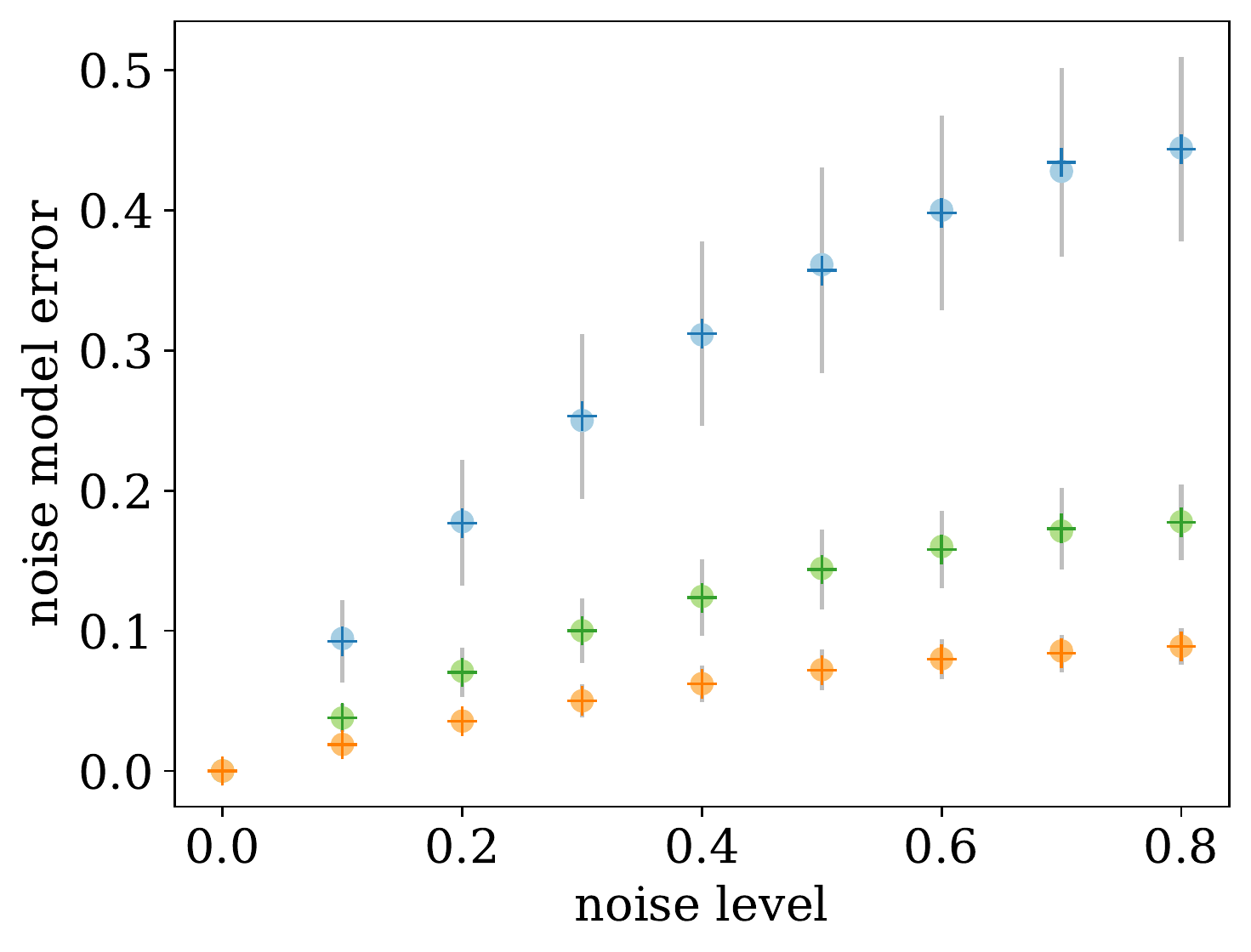}
 & \includegraphics[width=0.265\textwidth]{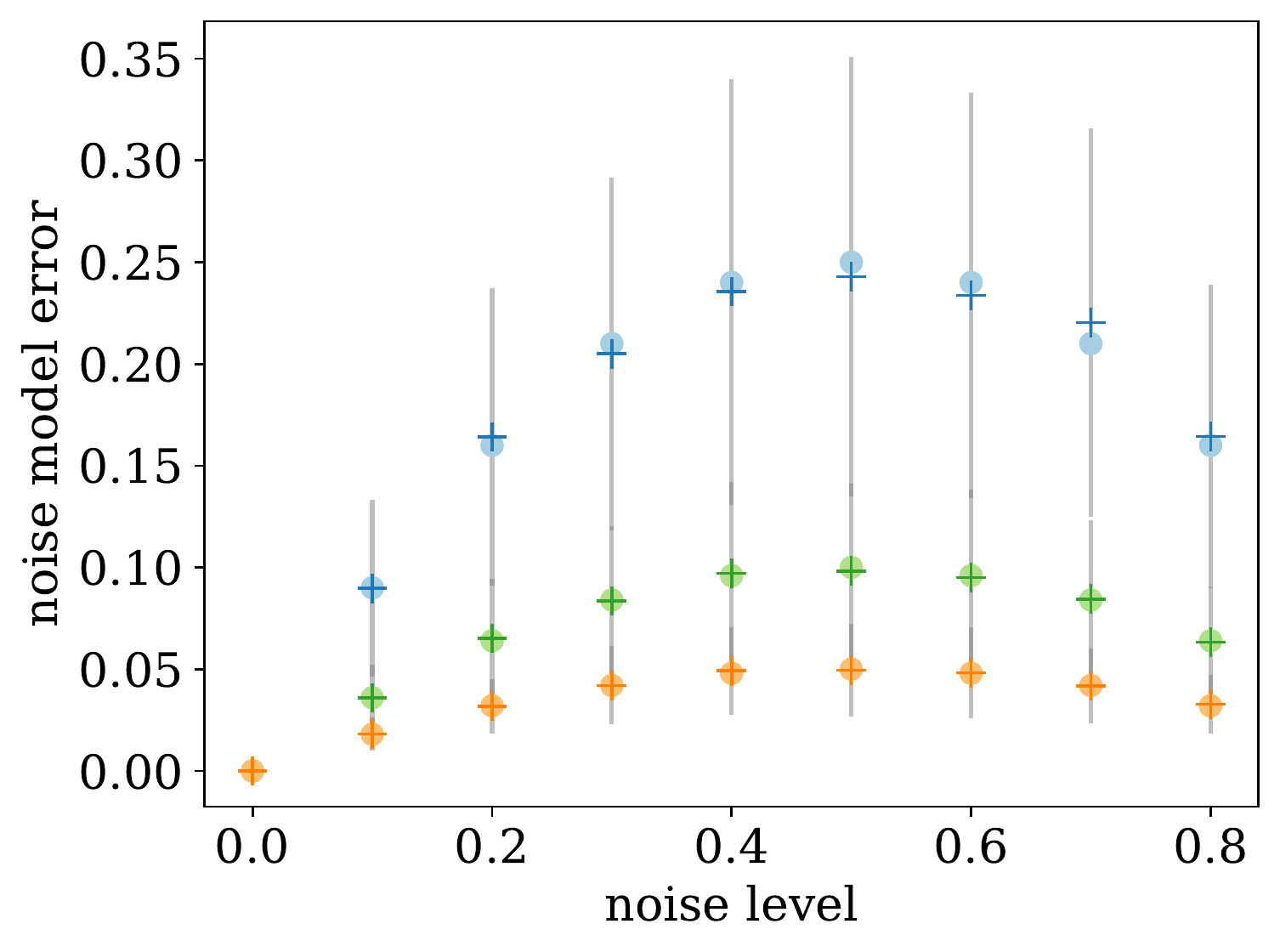}
 & \includegraphics[width=0.265\textwidth]{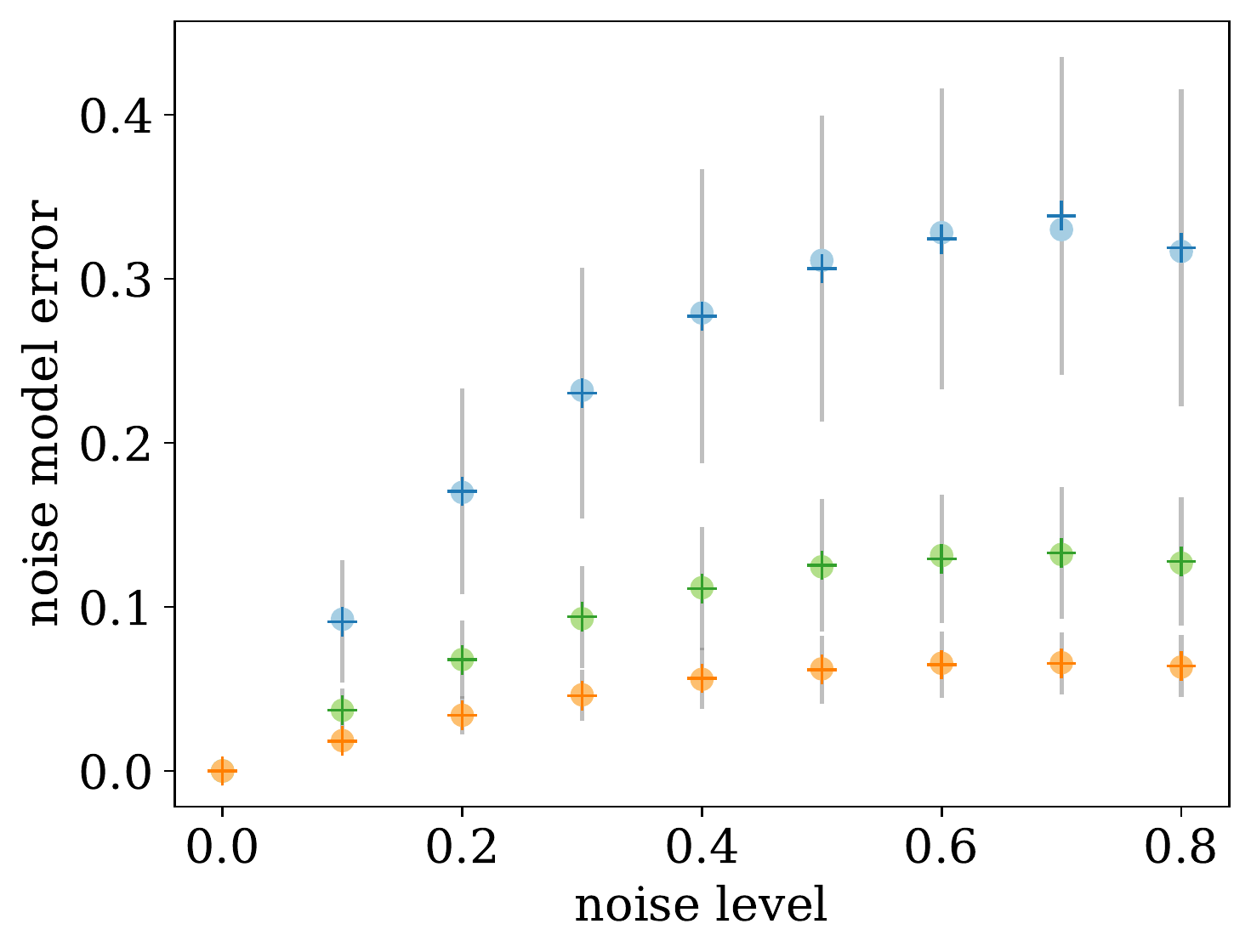}
 & \raisebox{1.65cm}{\hspace{0.05cm} \includegraphics[width=0.15\textwidth]{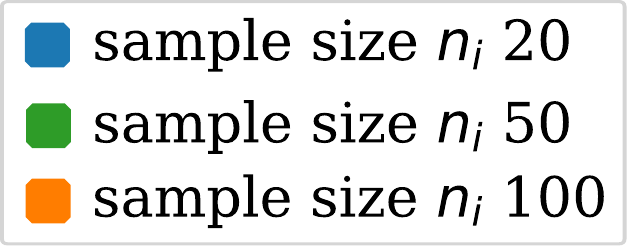}}\\ \midrule
  \rotatebox{90}{\hspace{-1.6cm}Variable Sampling}
 & \includegraphics[width=0.265\textwidth]{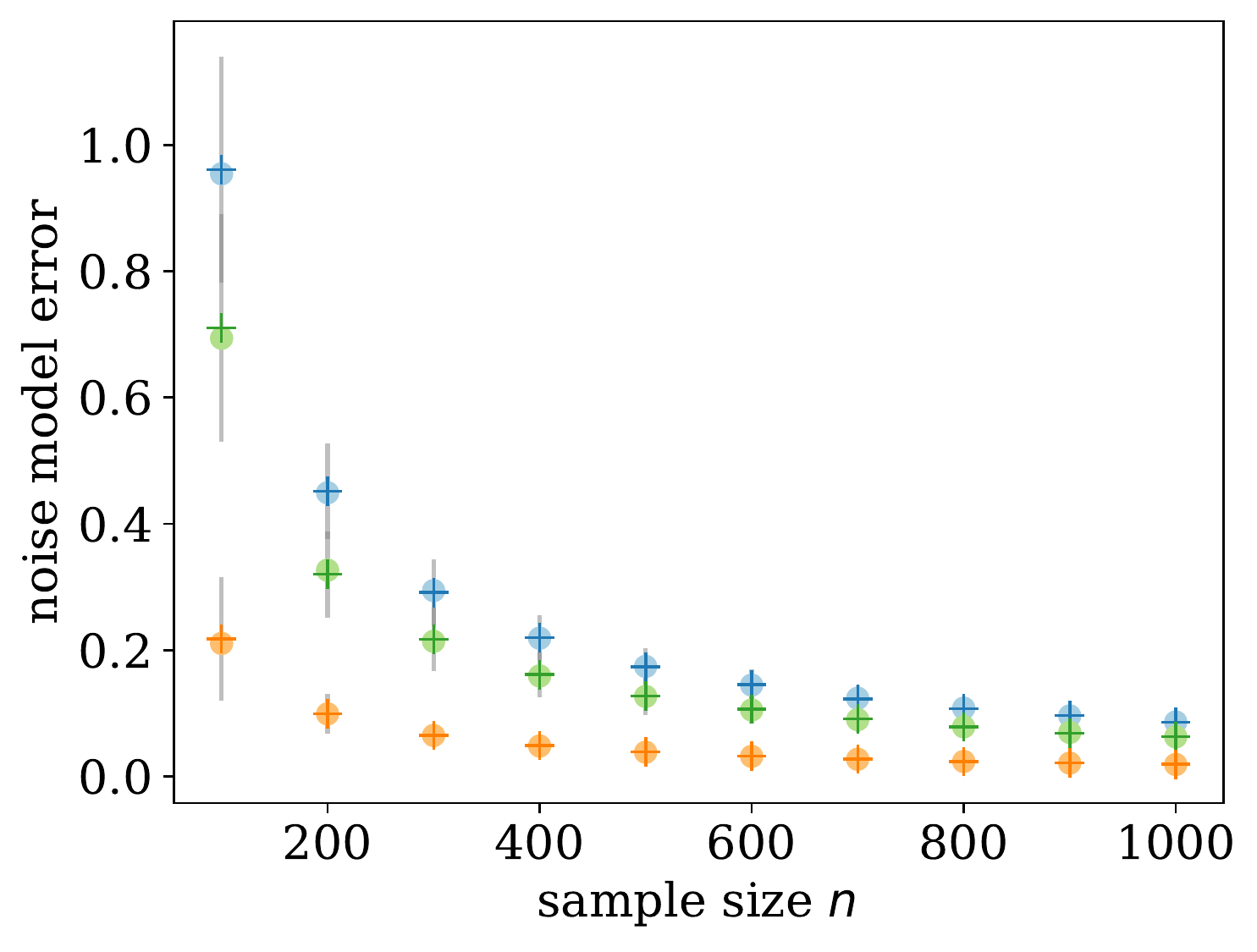}
 & \includegraphics[width=0.265\textwidth]{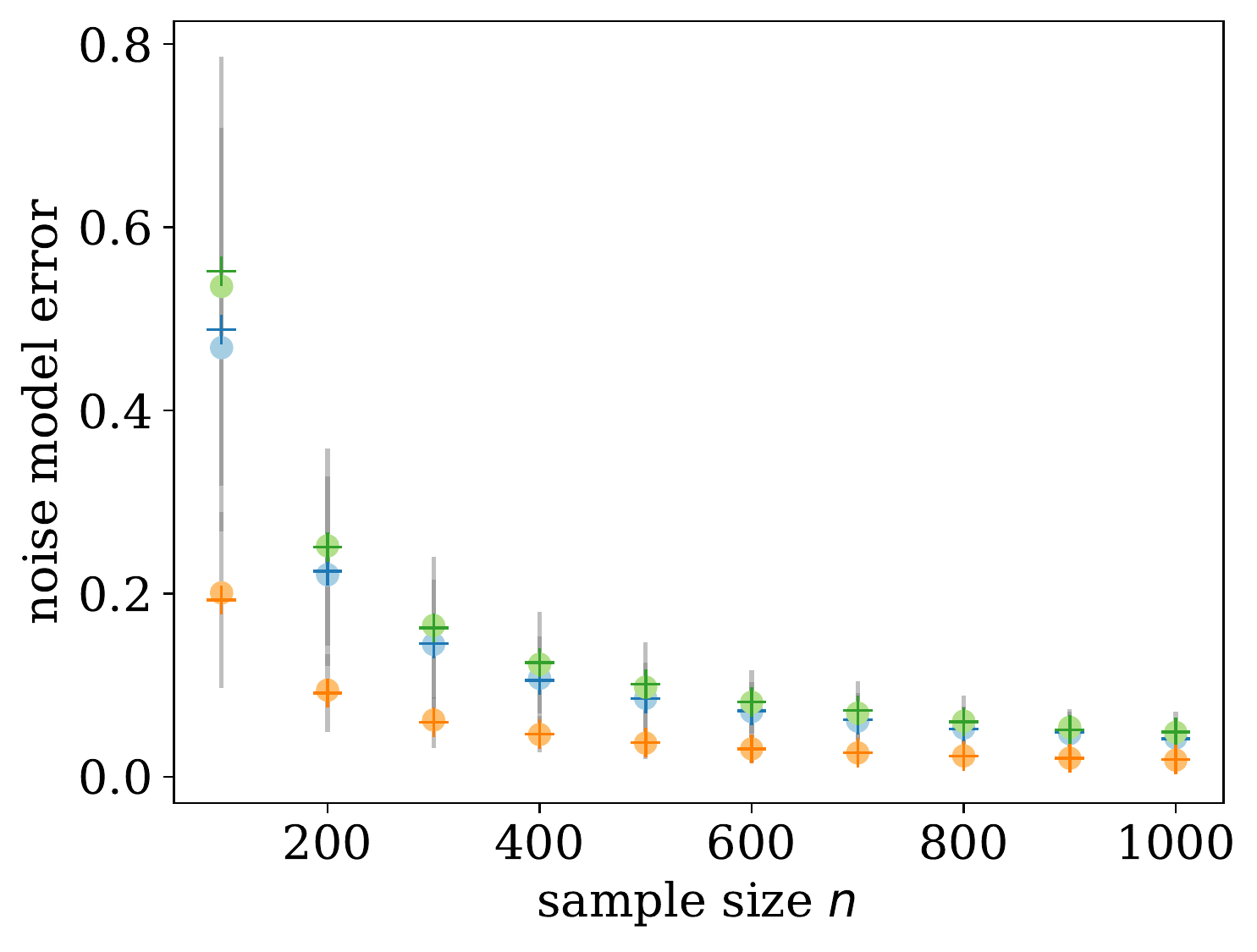}
 & \includegraphics[width=0.265\textwidth]{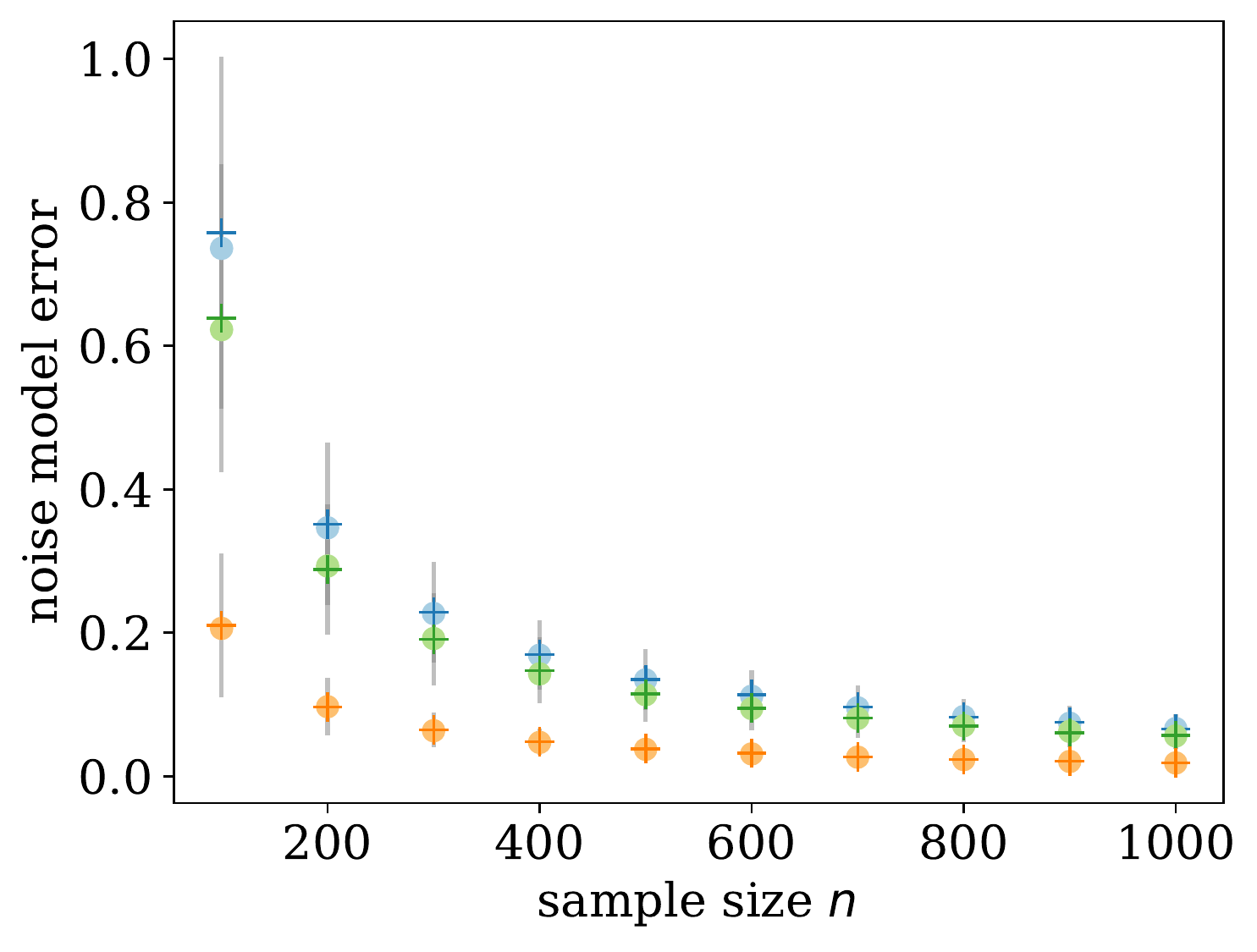} 
 & \raisebox{1.6cm}{\includegraphics[width=0.15\textwidth]{mnist_noise-level_legend.pdf}}\\
 & \includegraphics[trim=0cm 1cm 0cm 0, width=0.265\textwidth]{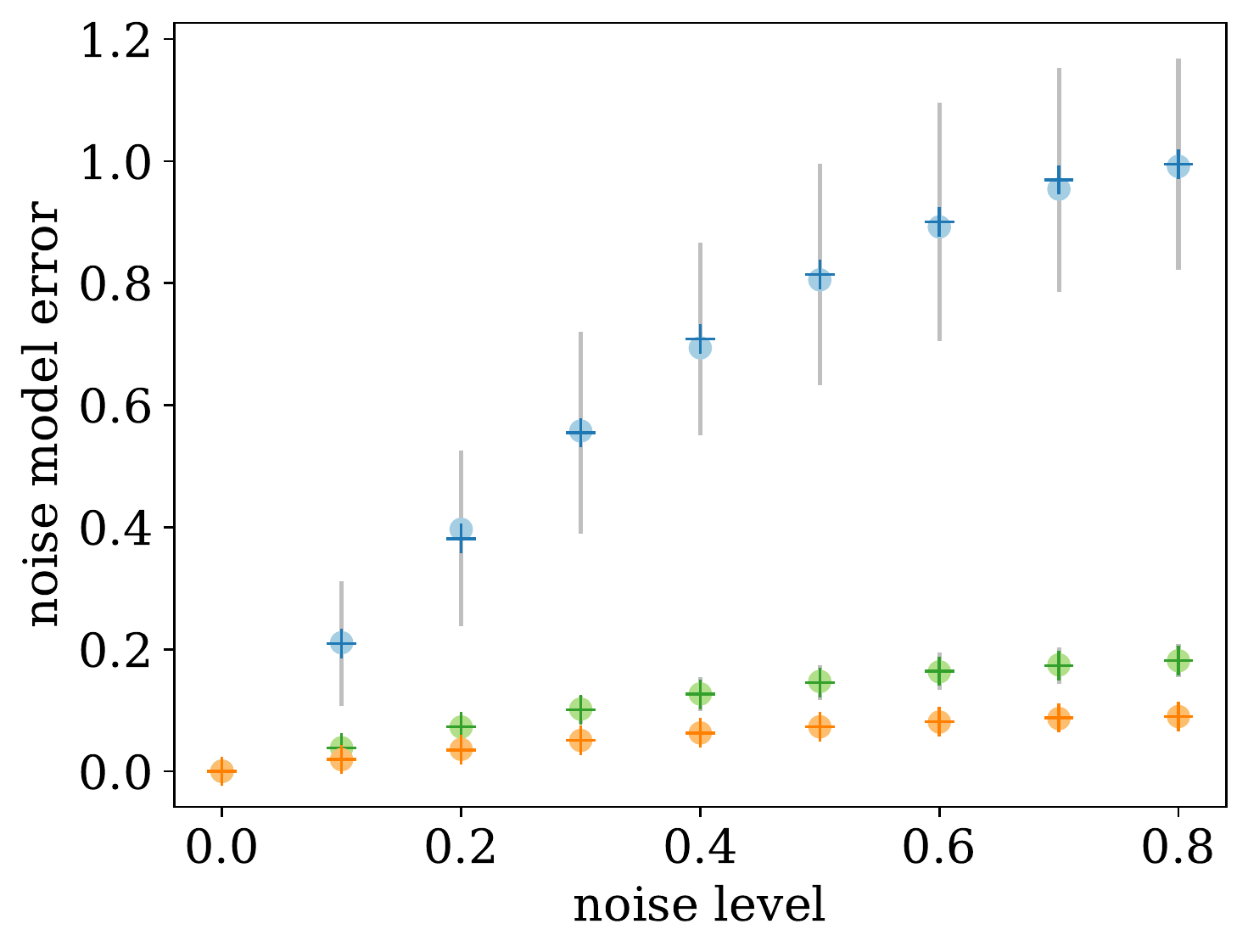}
 & \includegraphics[trim=0cm 1cm 0cm 0, width=0.265\textwidth]{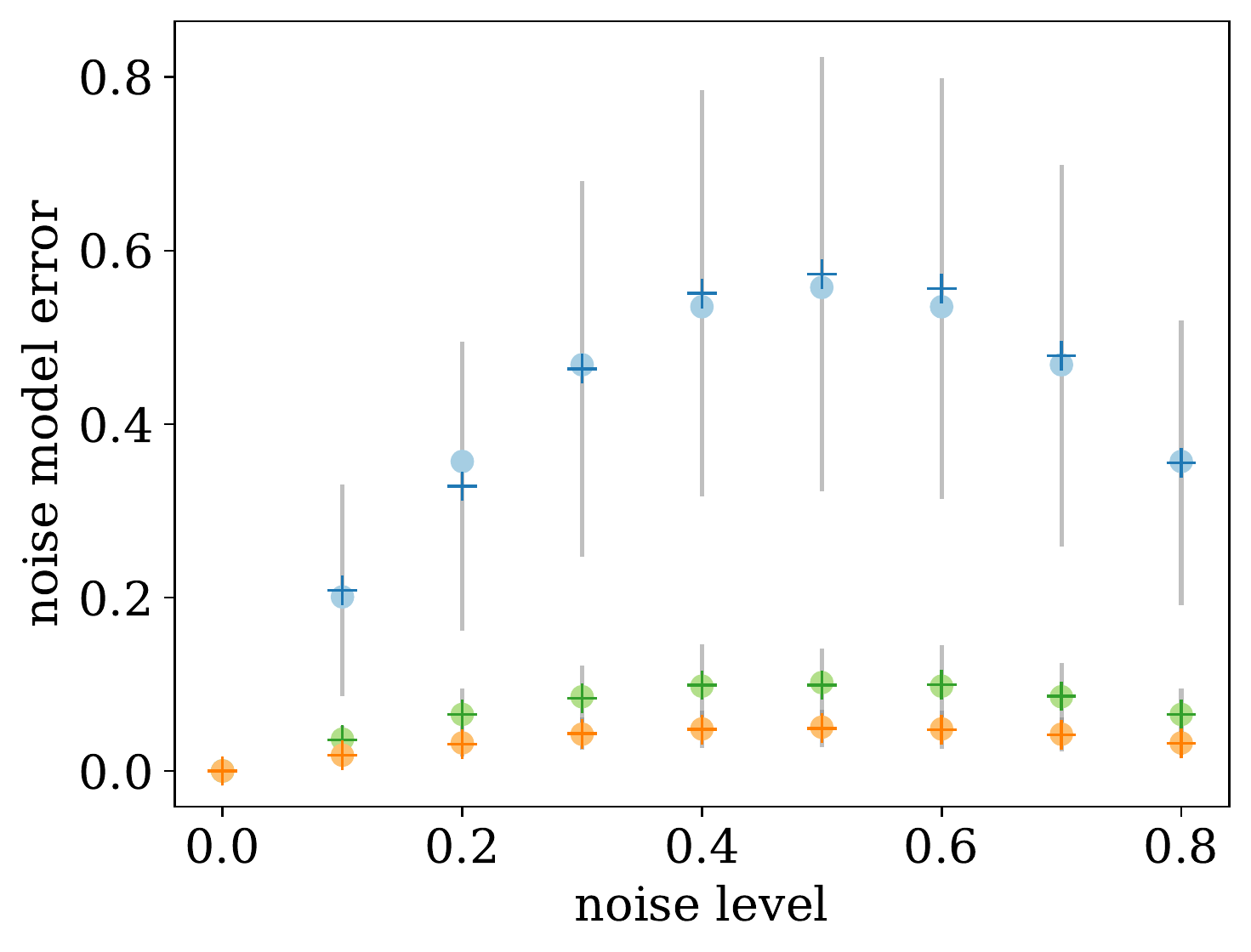}
 & \includegraphics[trim=0cm 1cm 0cm 0, width=0.265\textwidth]{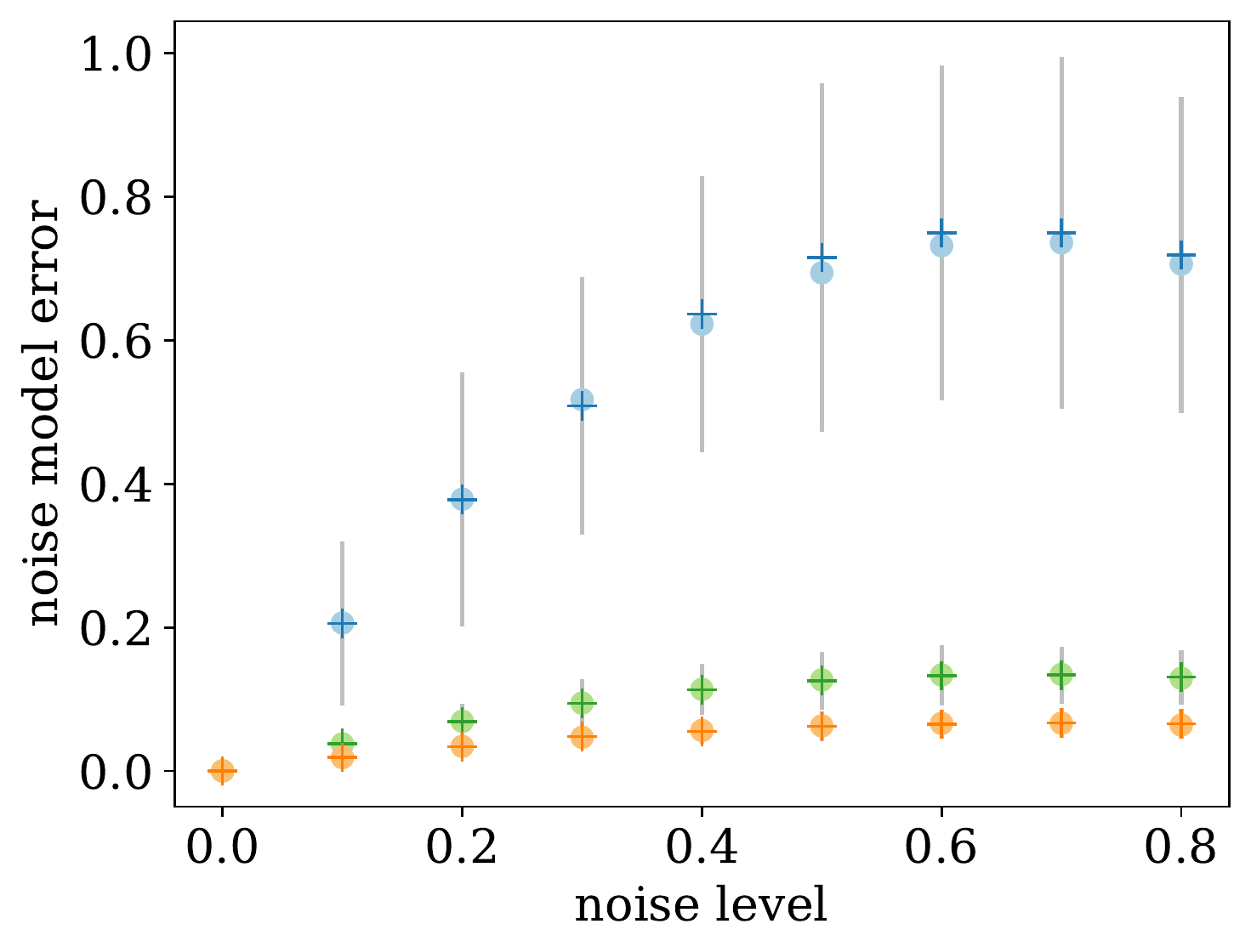}
 & \raisebox{1.5cm}{\hspace{0.05cm} \includegraphics[width=0.15\textwidth]{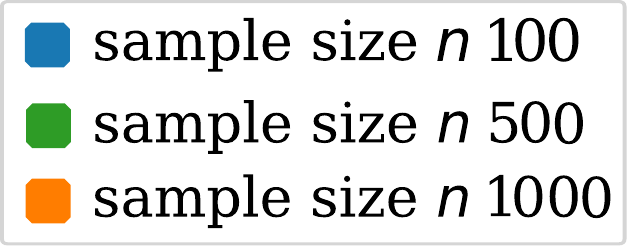}}
\end{tabular}
\caption{Comparison between the theoretically expected mean error (circle marker) and the empirically measured mean error (cross) of the noise model on the MNIST dataset. In the columns, the three different synthetic noise types "uniform", "single-flip" and "multi-flip" are given. Upper part uses Fixed Sampling, lower part uses Variable Sampling. On the x-axis, either the sample size ($n_i$ or $n$ respectively) or the noise level $\epsilon$ is varied. Error bars show the empirical standard deviation.}
\end{figure}

\begin{figure}
    \centering
    \includegraphics[height=3.5cm]{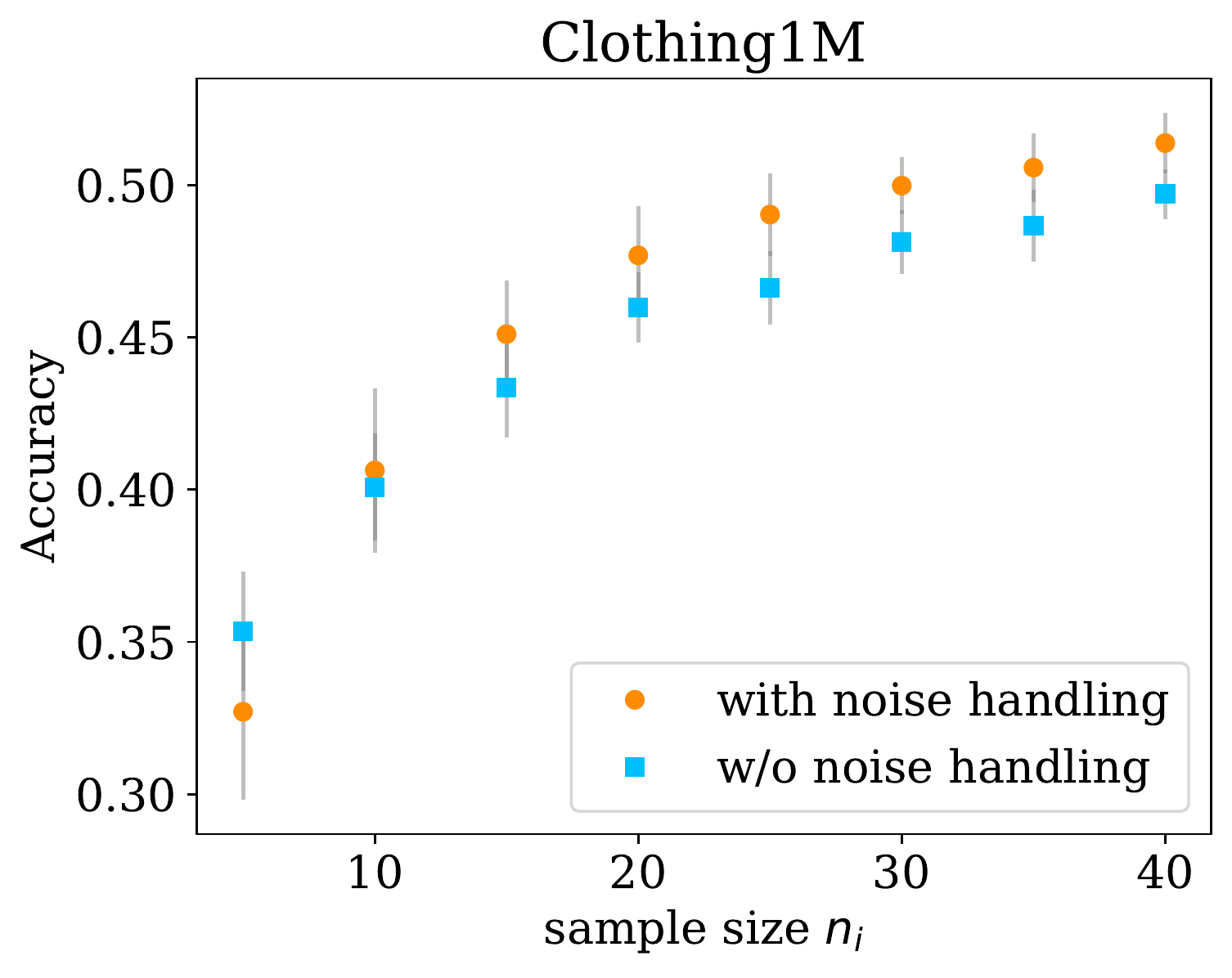}
    \includegraphics[height=3.5cm]{w_wo_noise-handling/estonian/ni/w_and_wo_noise_handling_ls_ni_1.pdf}
    \includegraphics[height=3.5cm]{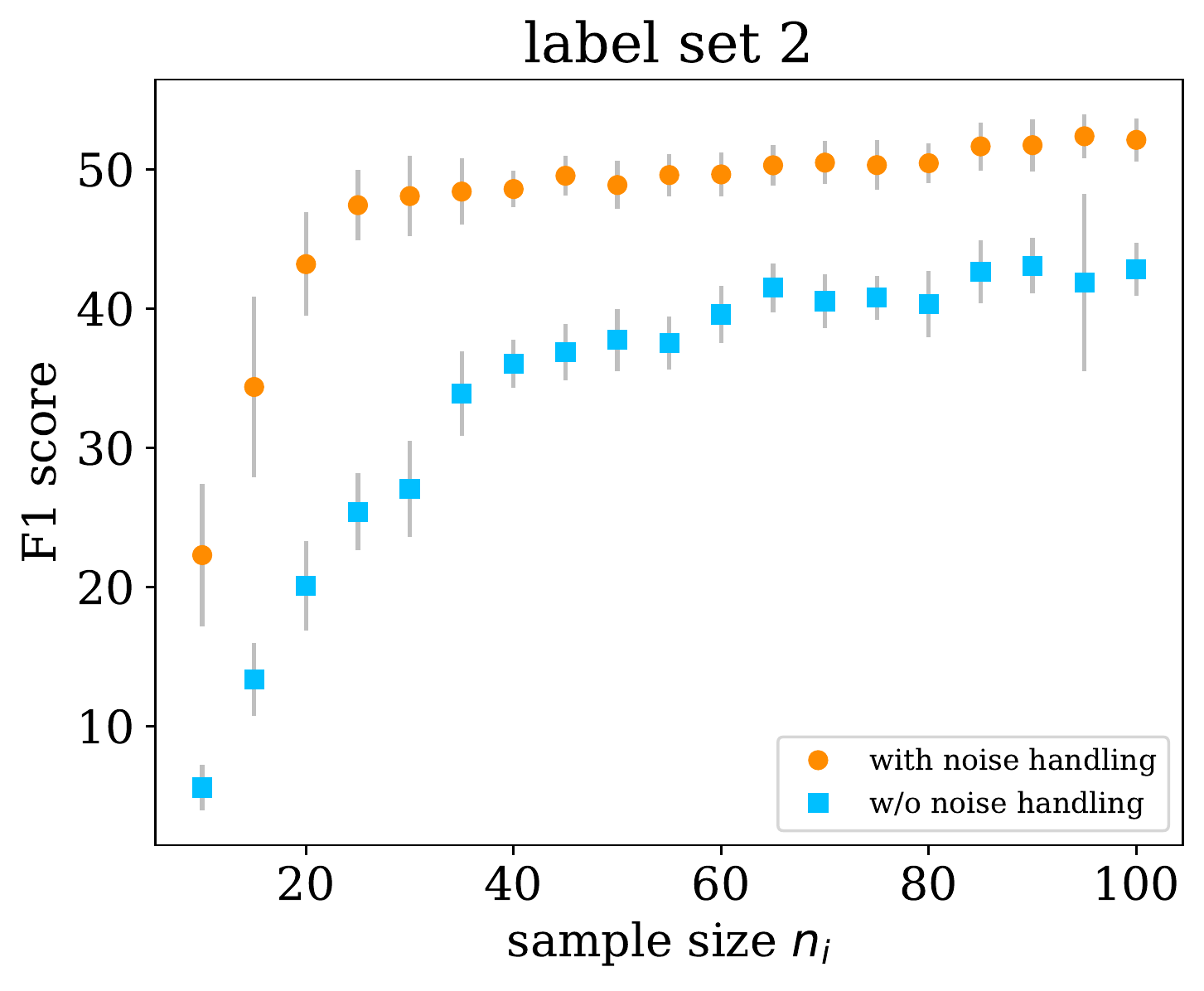}
    \includegraphics[height=3.5cm]{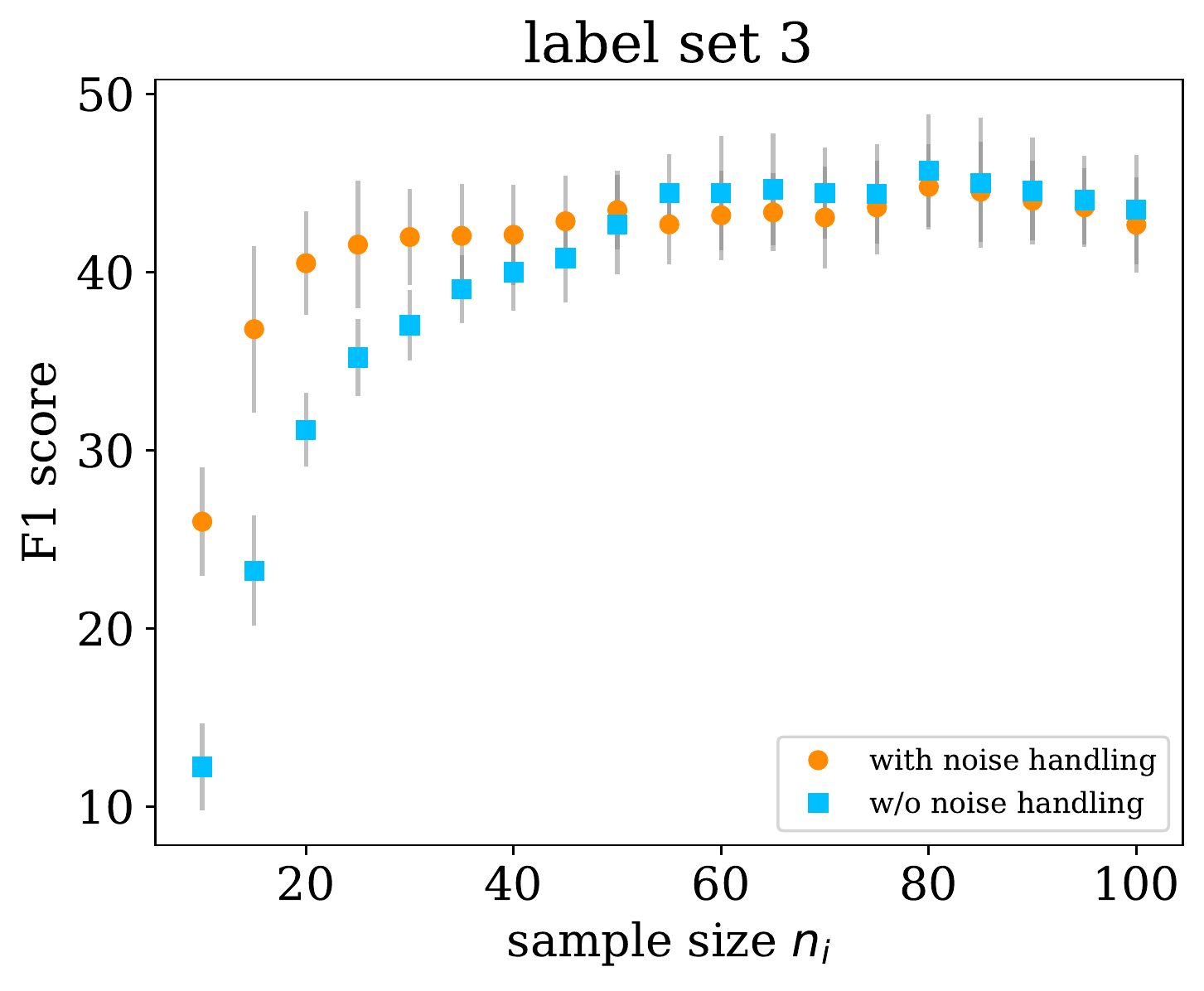}
    \includegraphics[height=3.5cm]{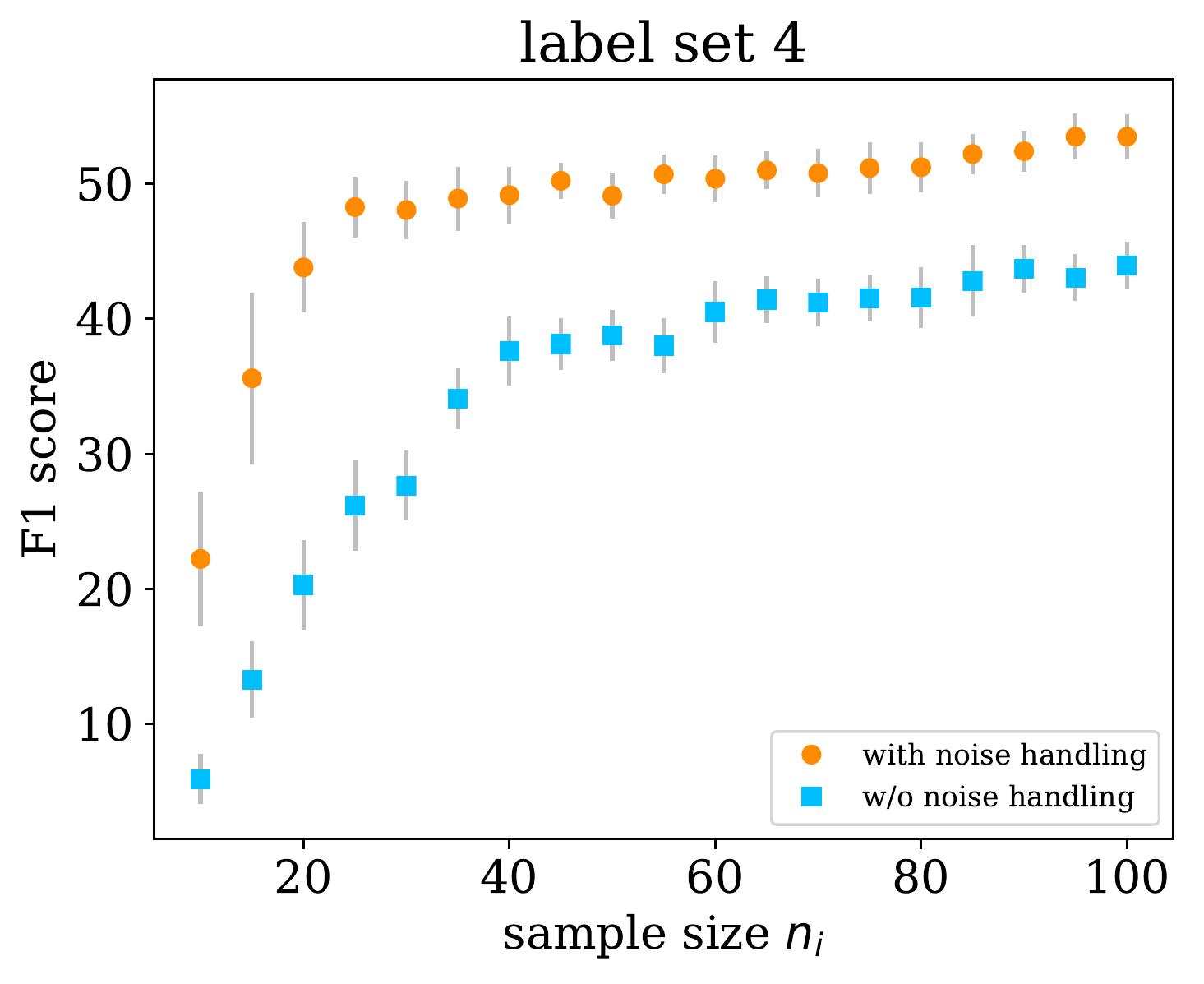}
    \includegraphics[height=3.5cm]{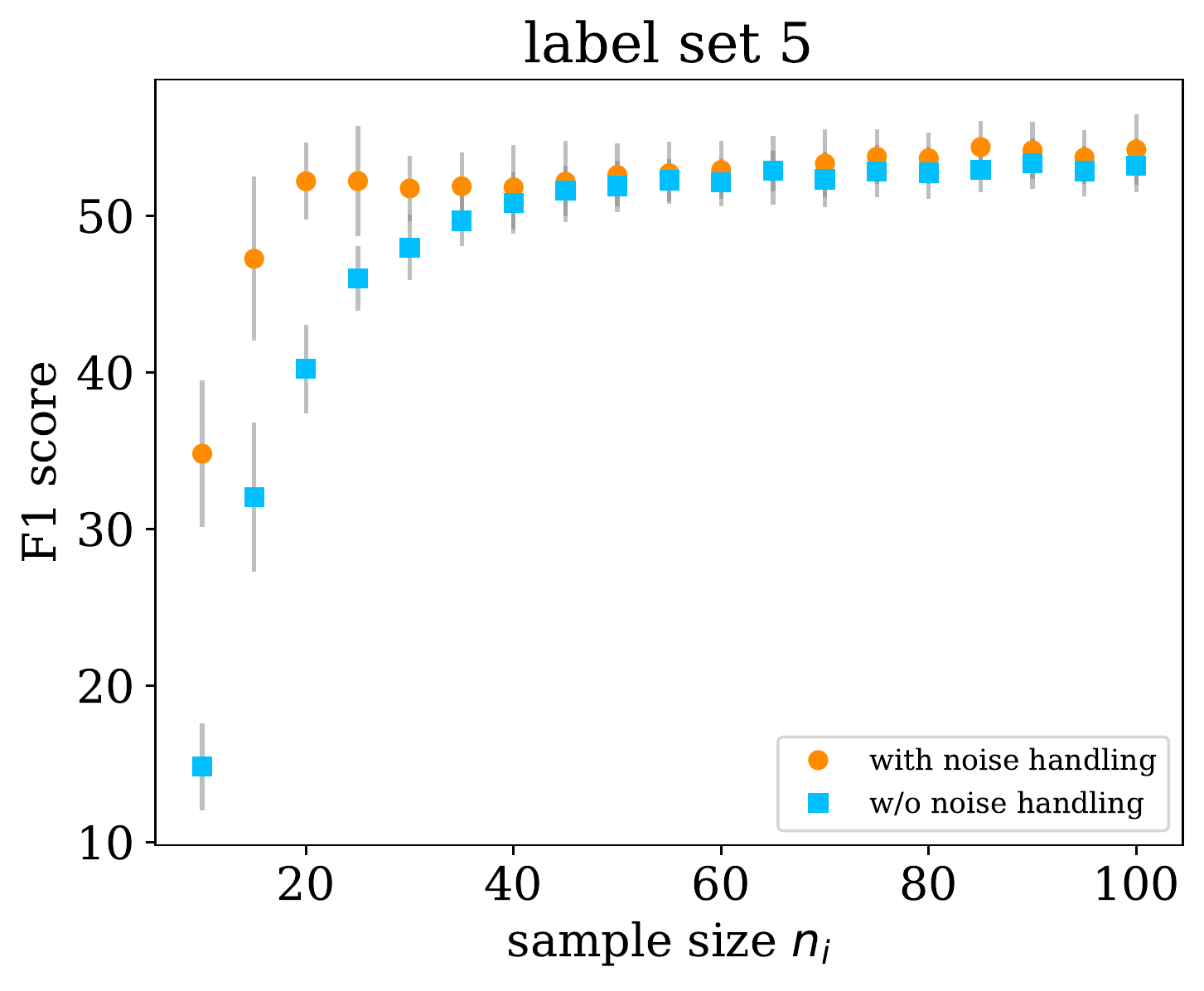}
    \includegraphics[height=3.5cm]{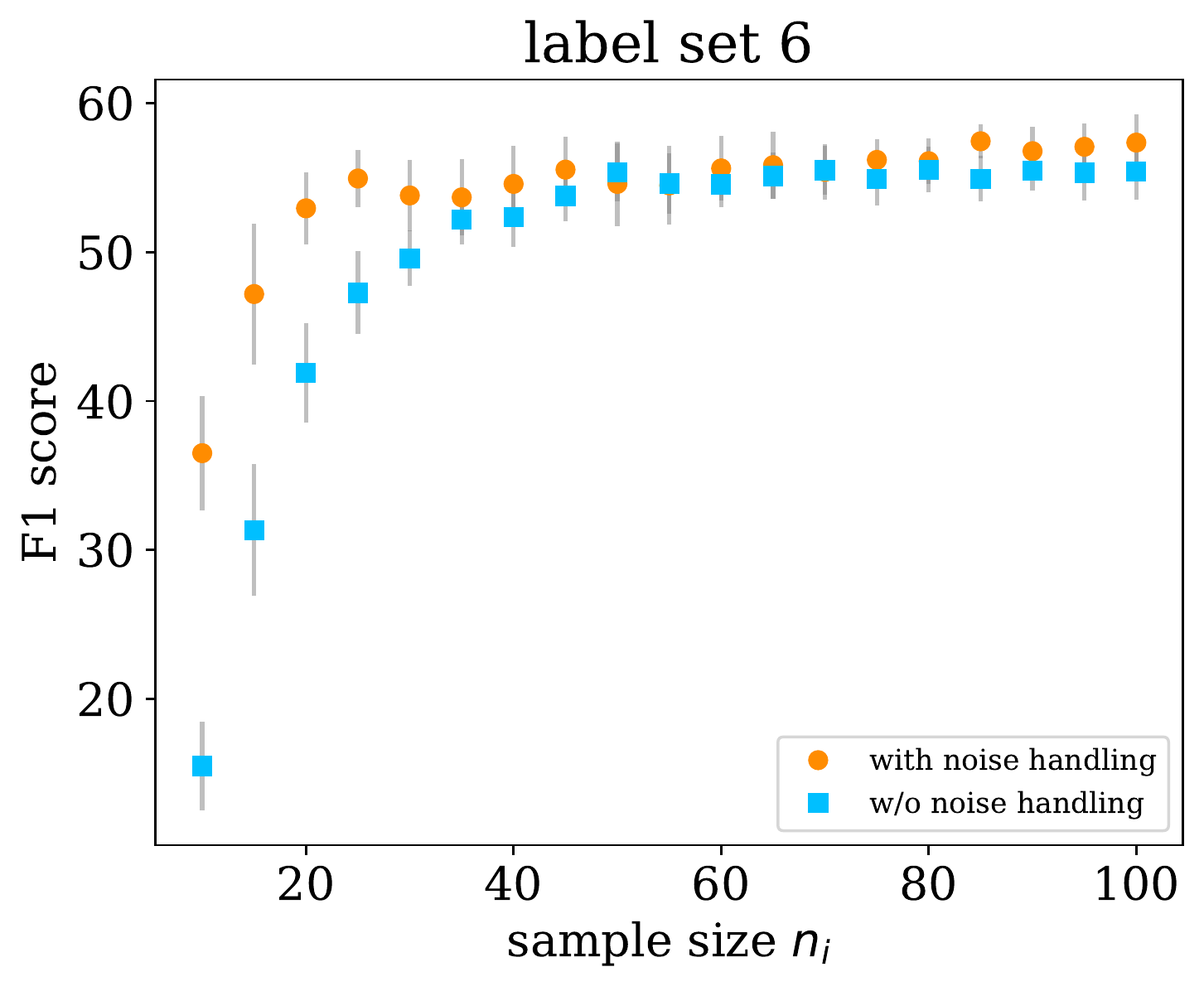}
    \includegraphics[height=3.5cm]{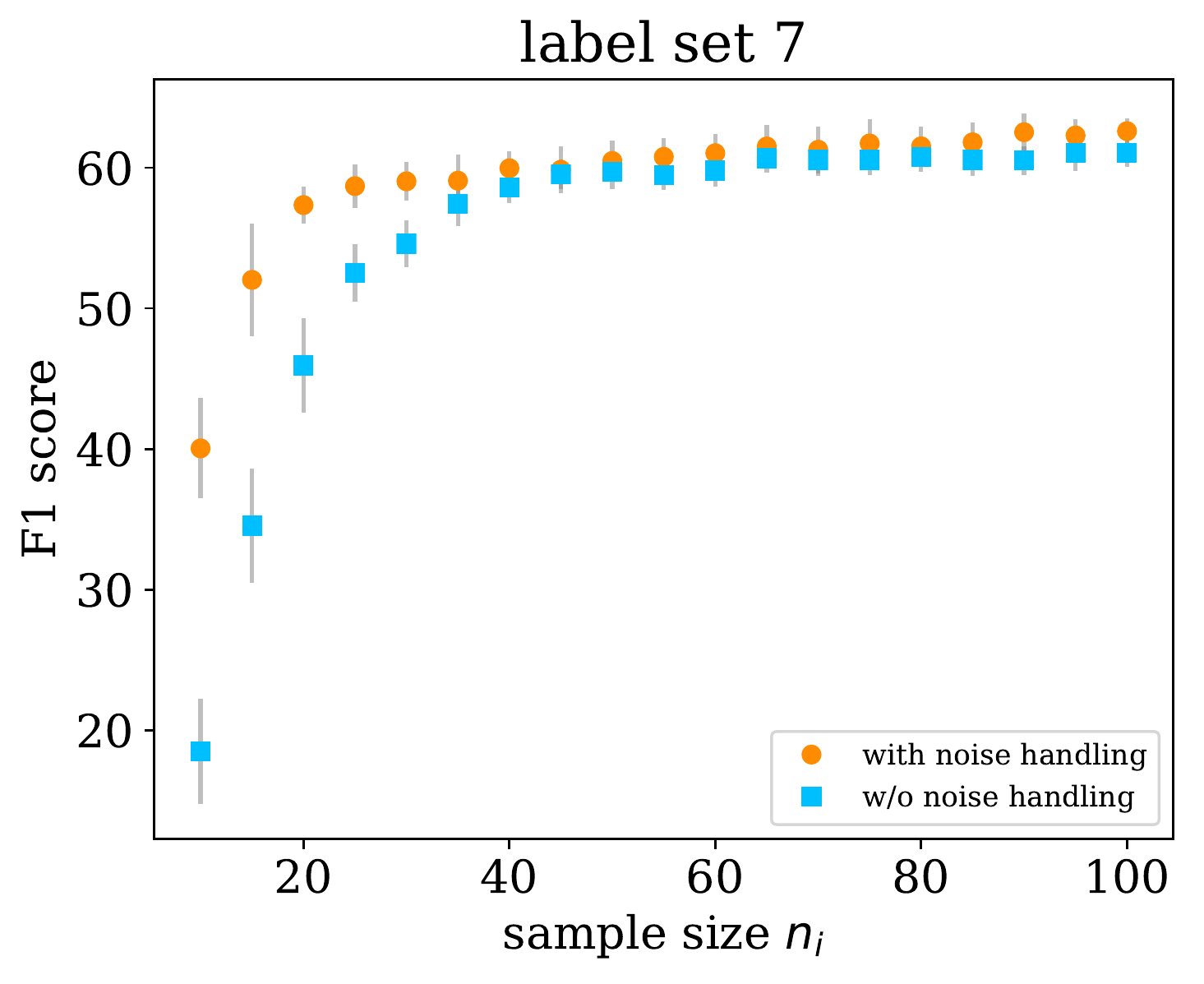}
    \caption{Mean test performance (Accuracy/F1 score) of the base model on Clothing1M and NoisyNER on clean and noisy data with and without noise handling. Using \textbf{increasing size of $\mathbf{|D_C|}$} and  \textbf{Fixed Sampling}. Error bars show the empirical standard deviation.}
\end{figure}

\begin{figure}
    \centering
    \includegraphics[height=3.5cm]{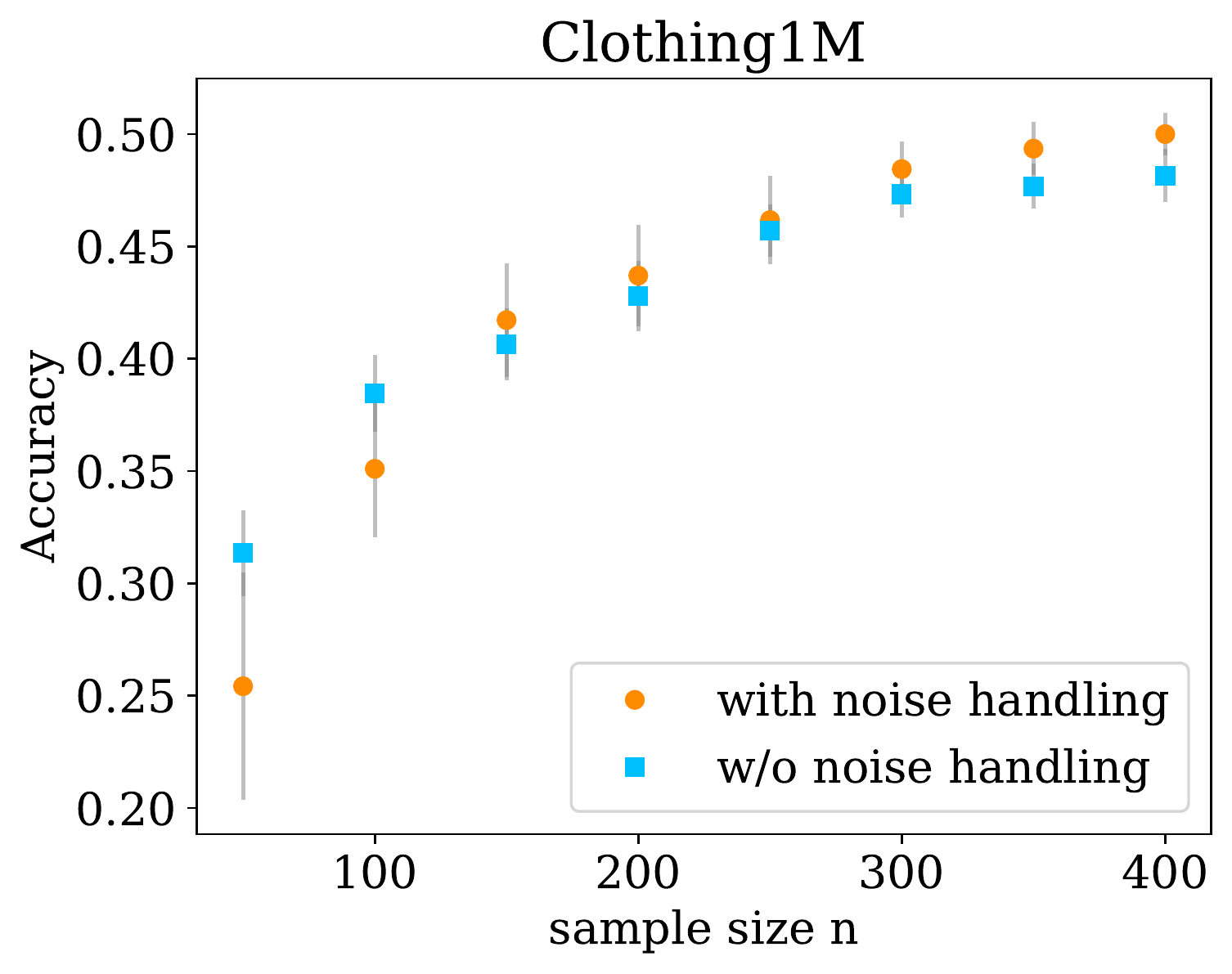}
    \includegraphics[height=3.5cm]{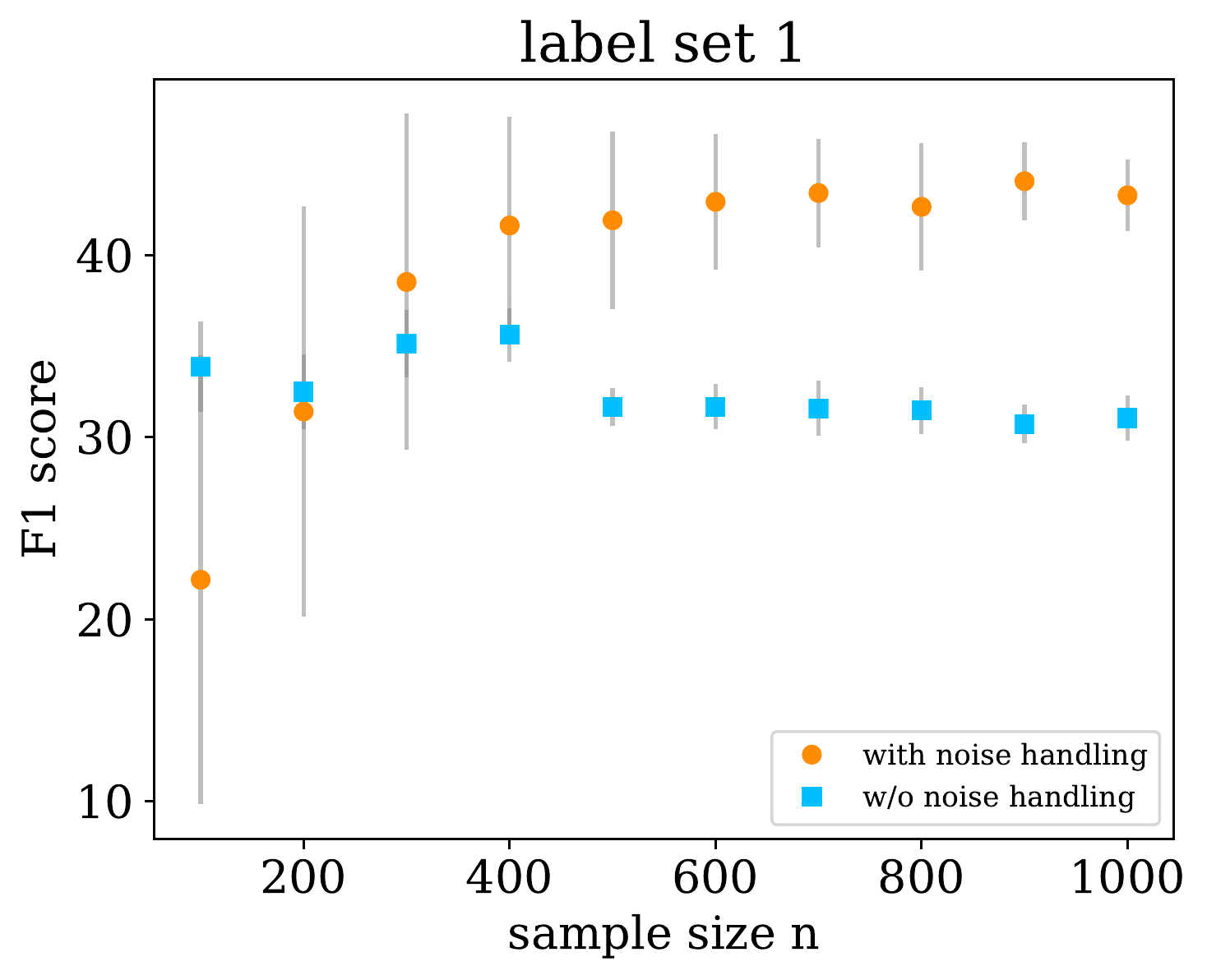}
    \includegraphics[height=3.5cm]{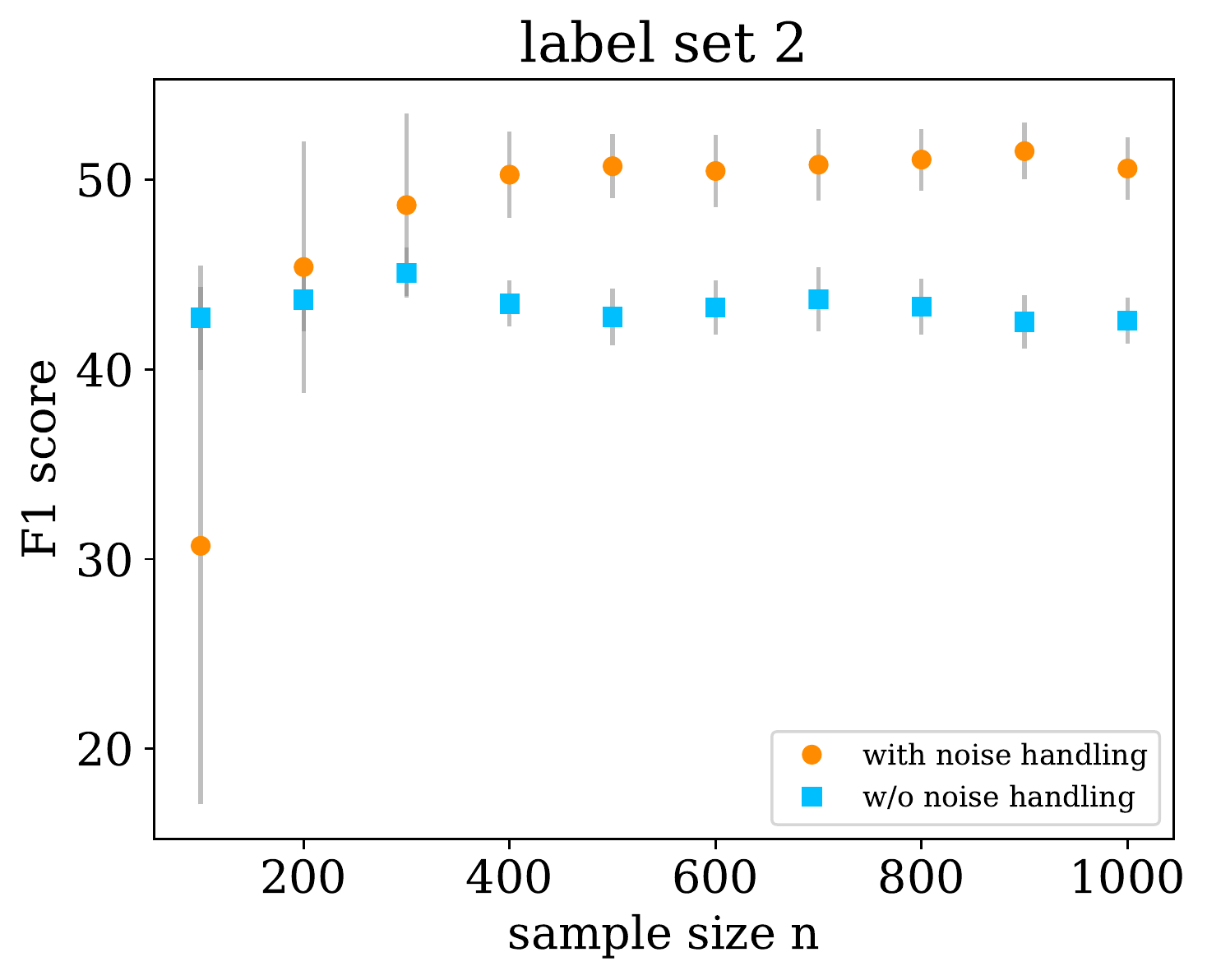}
    \includegraphics[height=3.5cm]{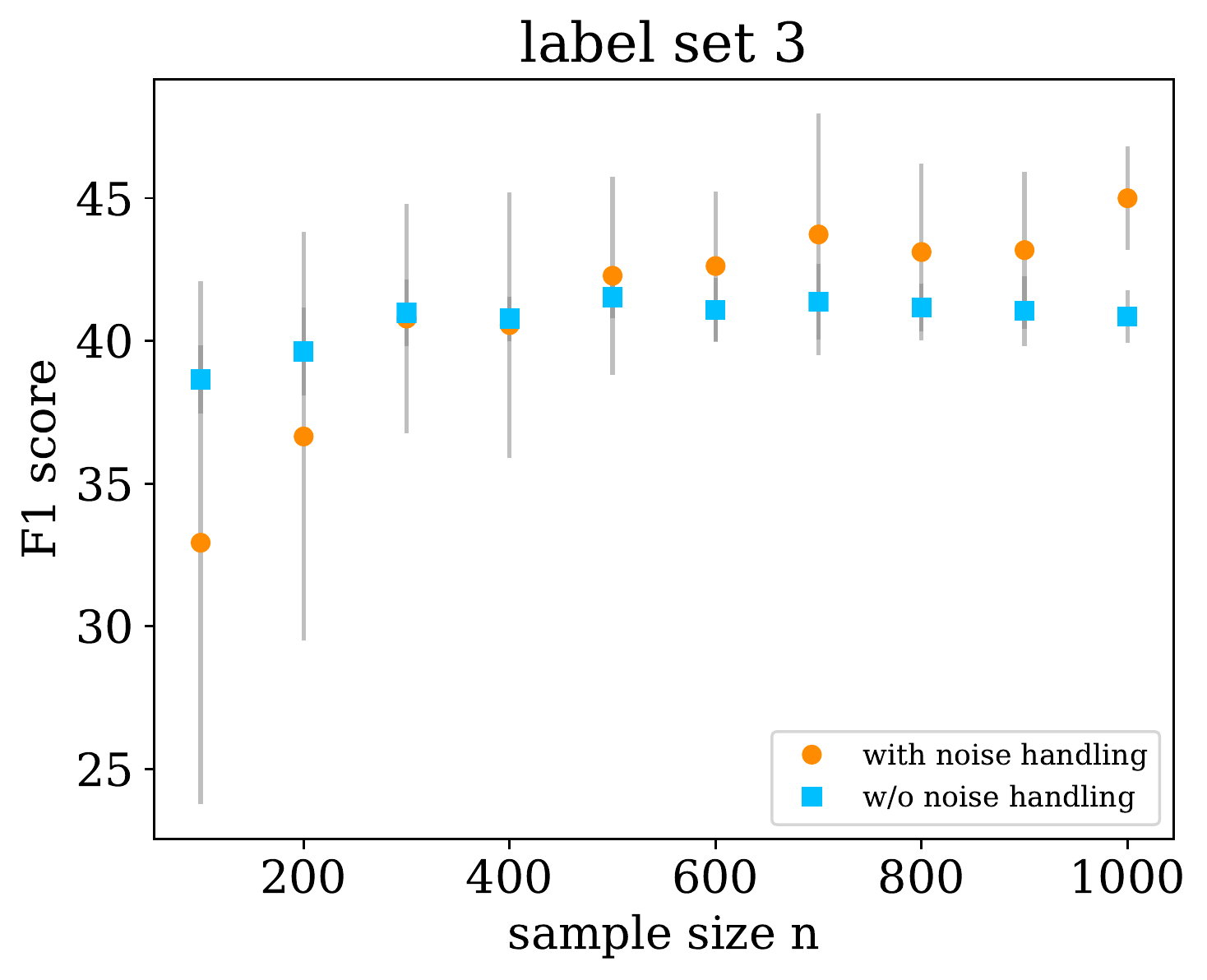}\\
    \includegraphics[height=3.5cm]{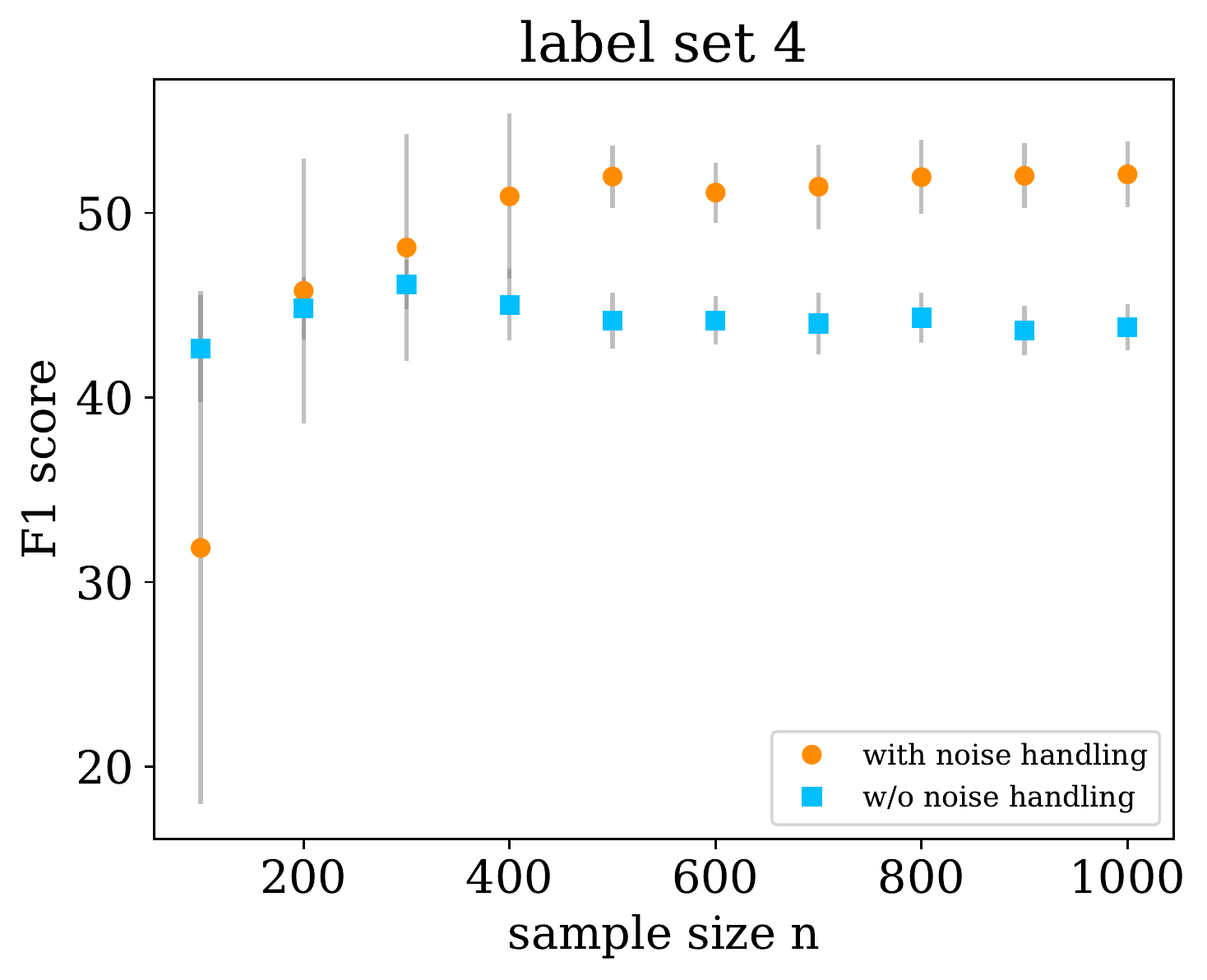}
    \includegraphics[height=3.5cm]{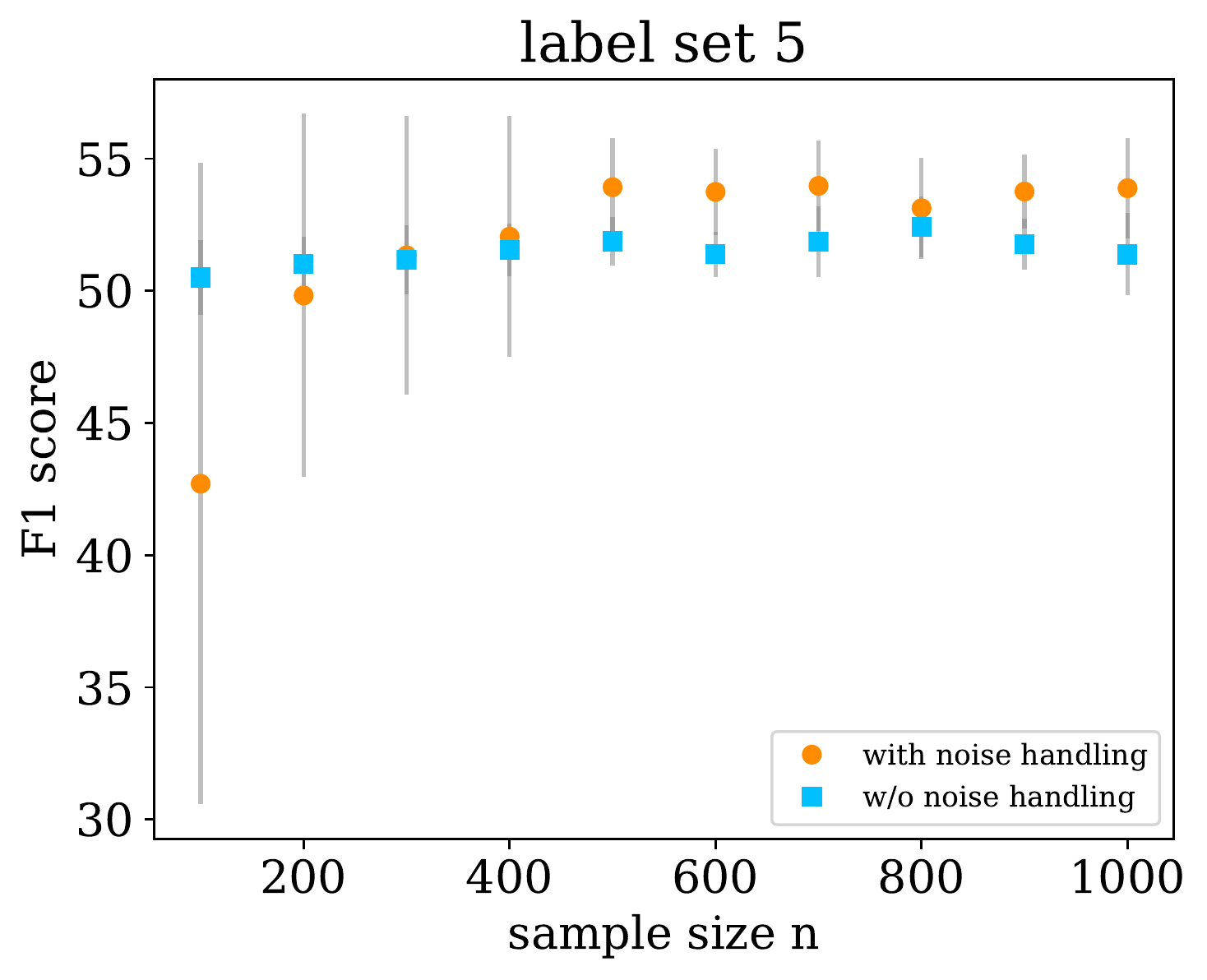}
    \includegraphics[height=3.5cm]{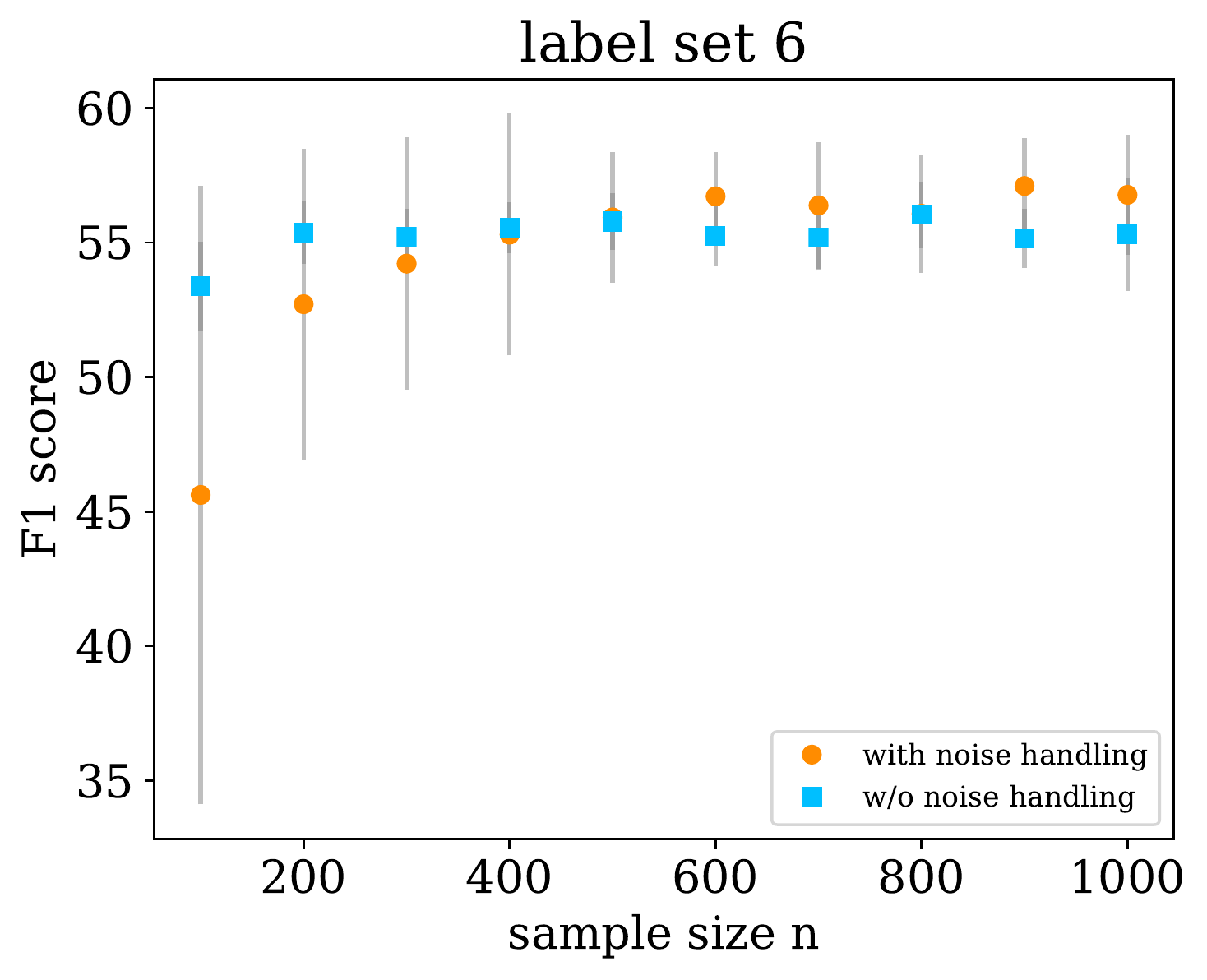}
    \includegraphics[height=3.5cm]{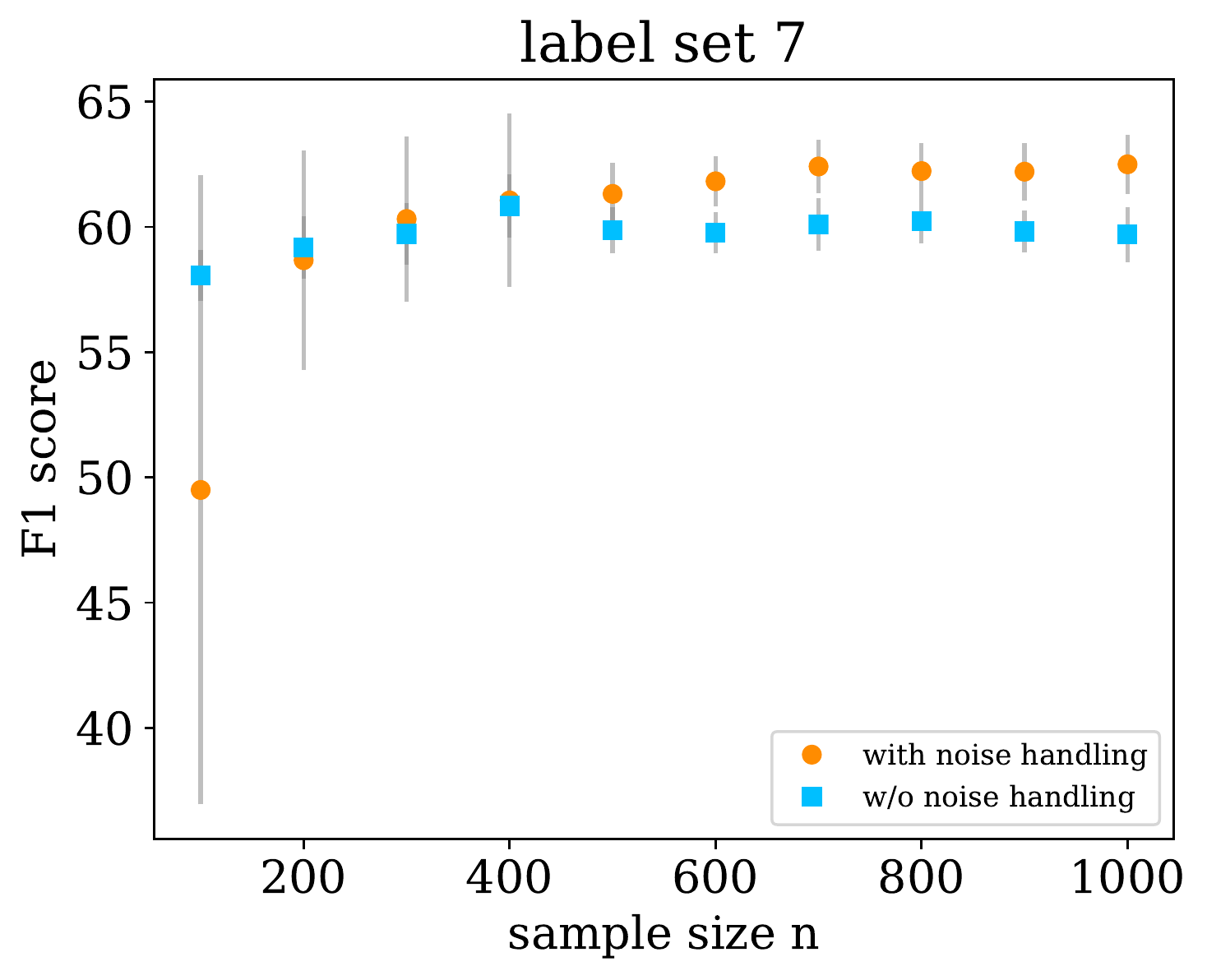}
    \caption{Mean test performance (Accuracy/F1 score) of the base model on Clothing1M and NoisyNER on clean and noisy data with and without noise handling. Using \textbf{increasing size of $\mathbf{|D_C|}$} and  \textbf{Variable Sampling}. Error bars show the empirical standard deviation.}
\end{figure}

\begin{figure}
    \centering
    \includegraphics[height=3.5cm]{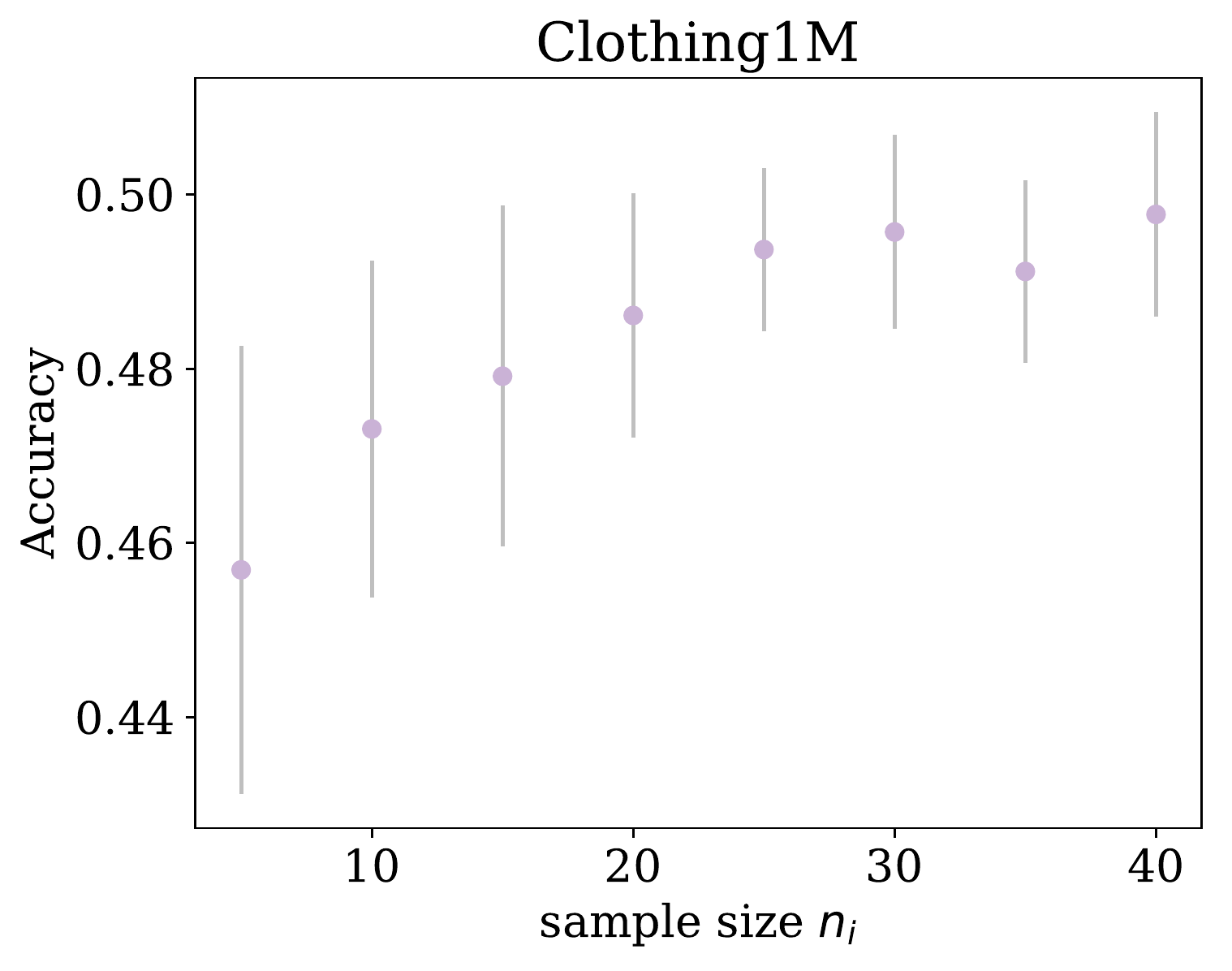}
    \includegraphics[height=3.5cm]{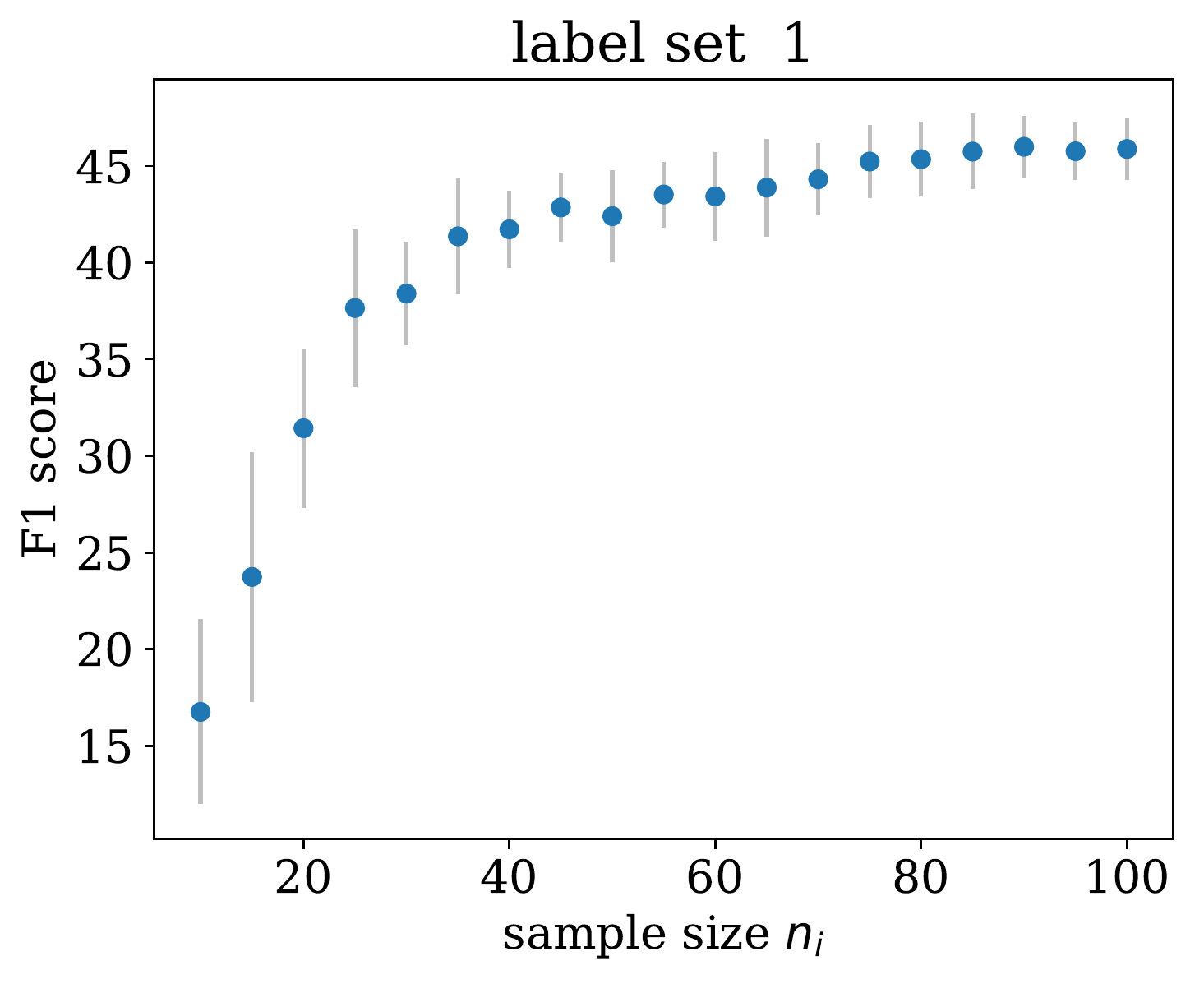}
    \includegraphics[height=3.5cm]{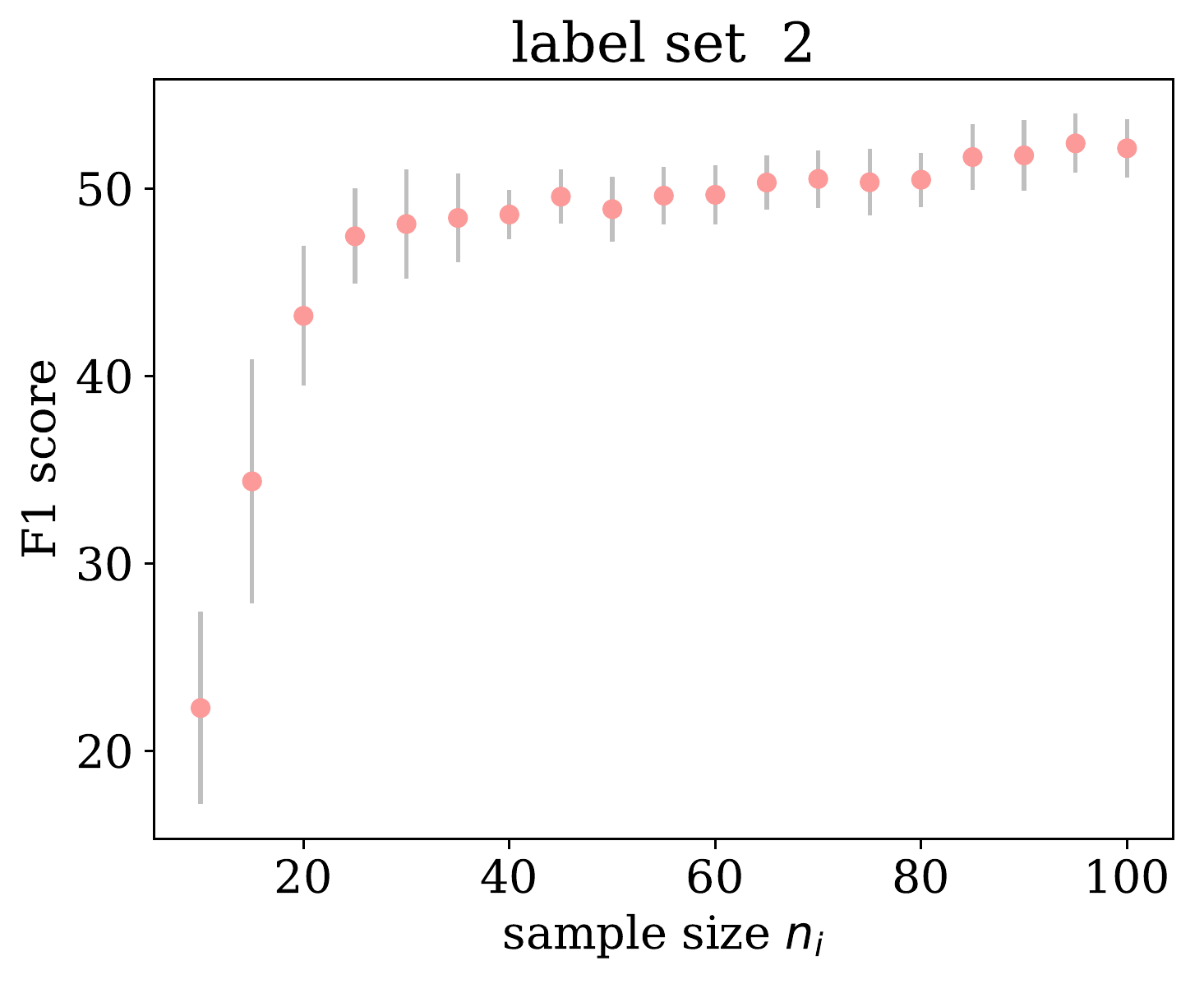}
    \includegraphics[height=3.5cm]{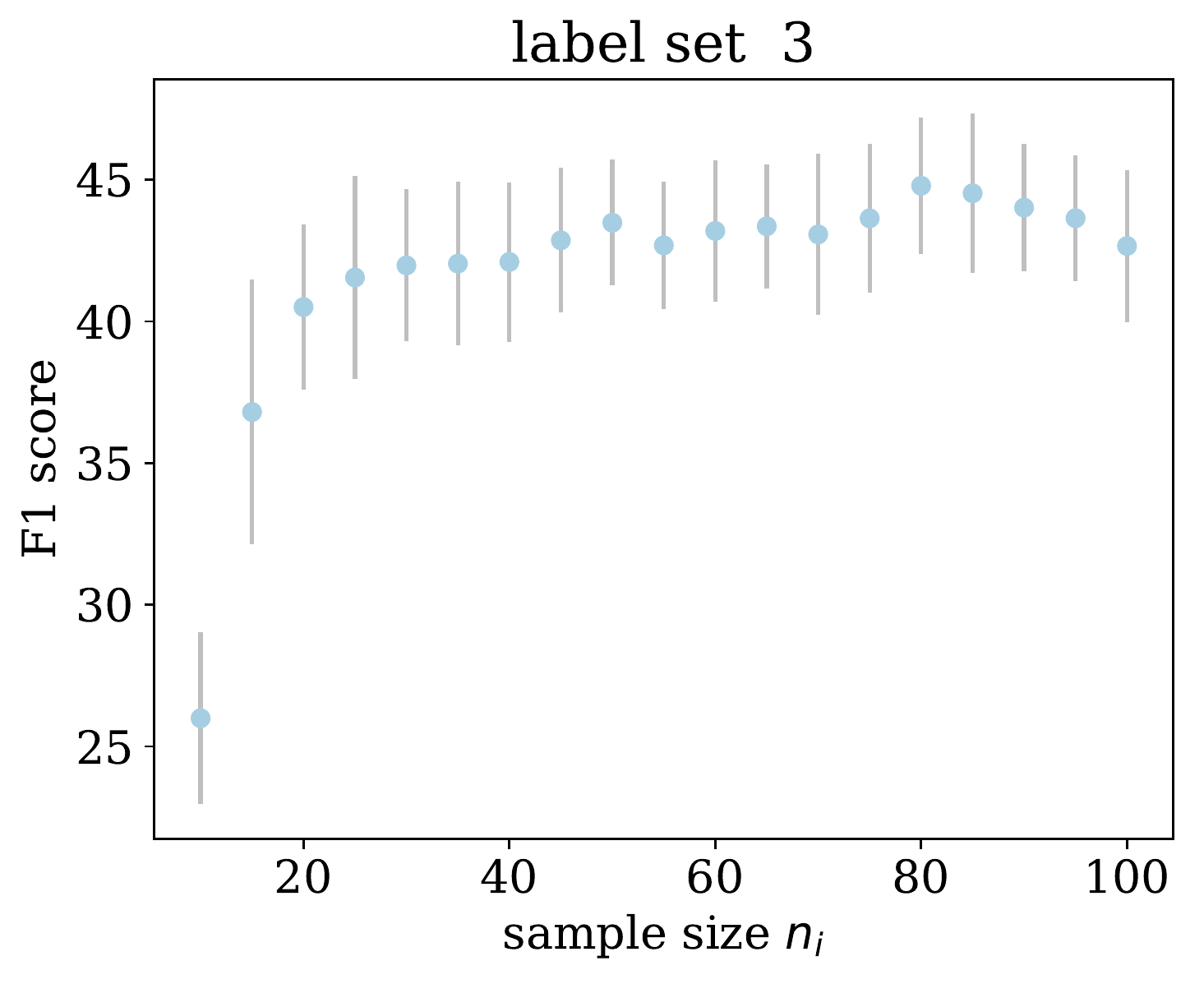}
    \includegraphics[height=3.5cm]{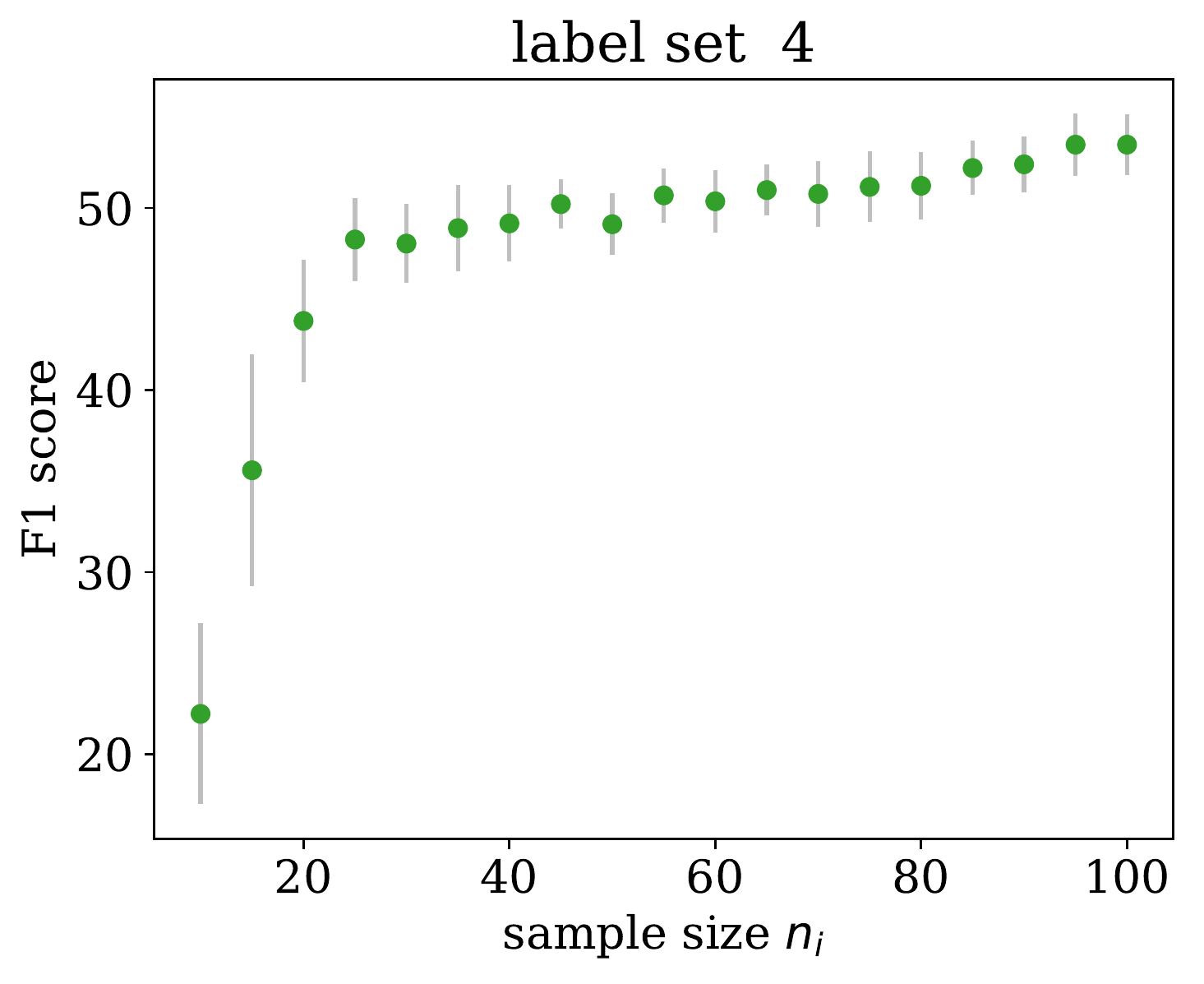}
    \includegraphics[height=3.5cm]{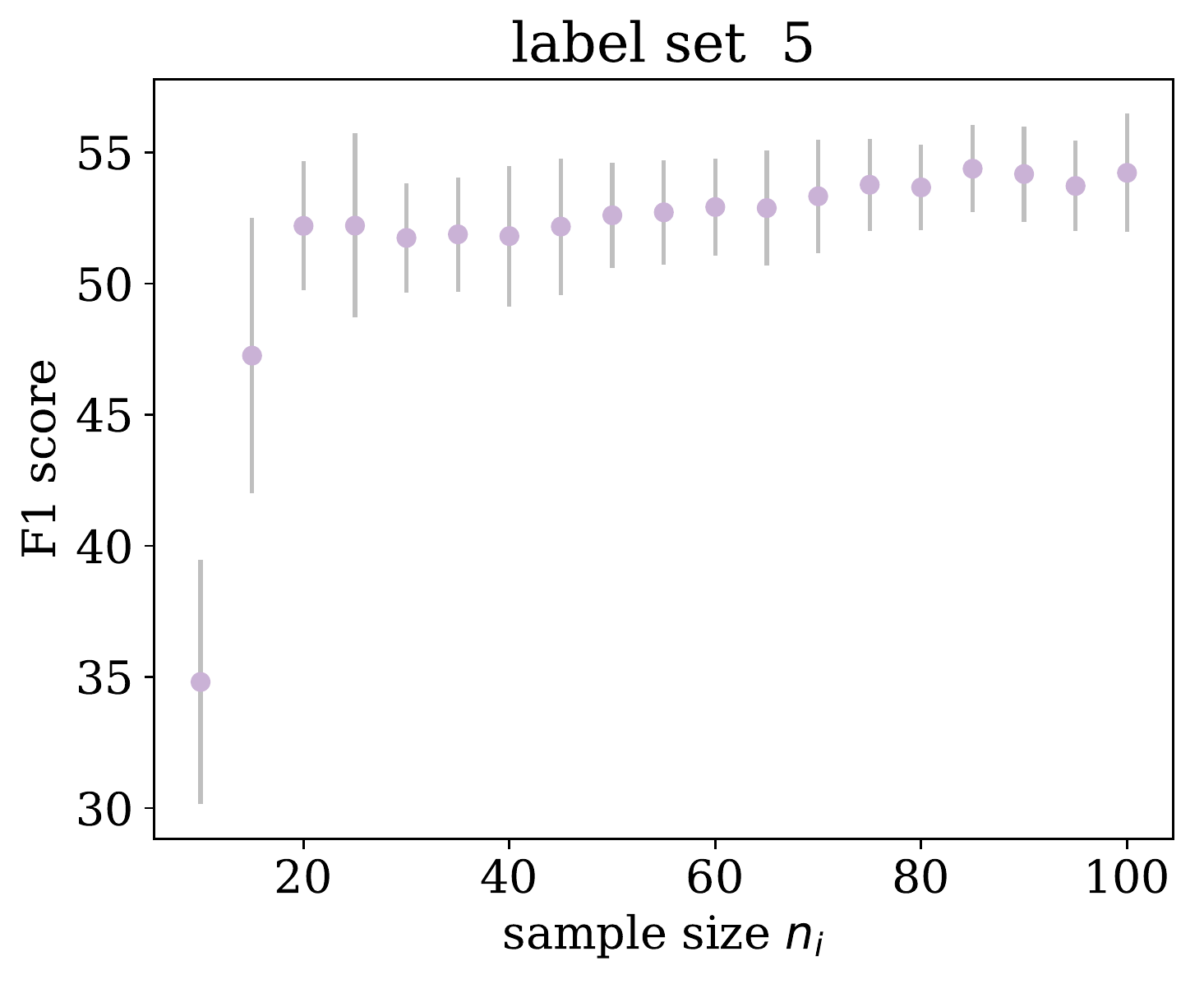}
    \includegraphics[height=3.5cm]{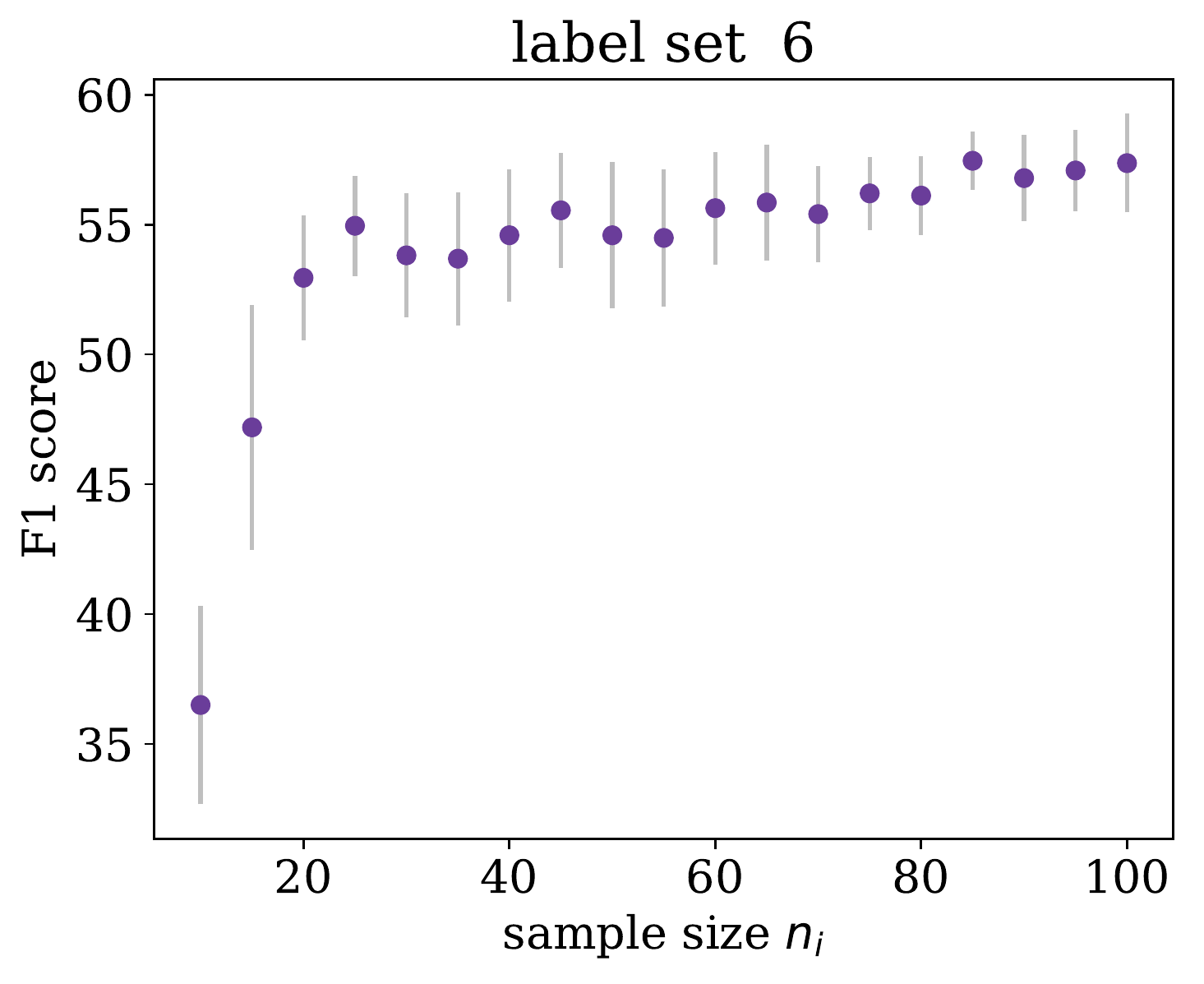}
    \includegraphics[height=3.5cm]{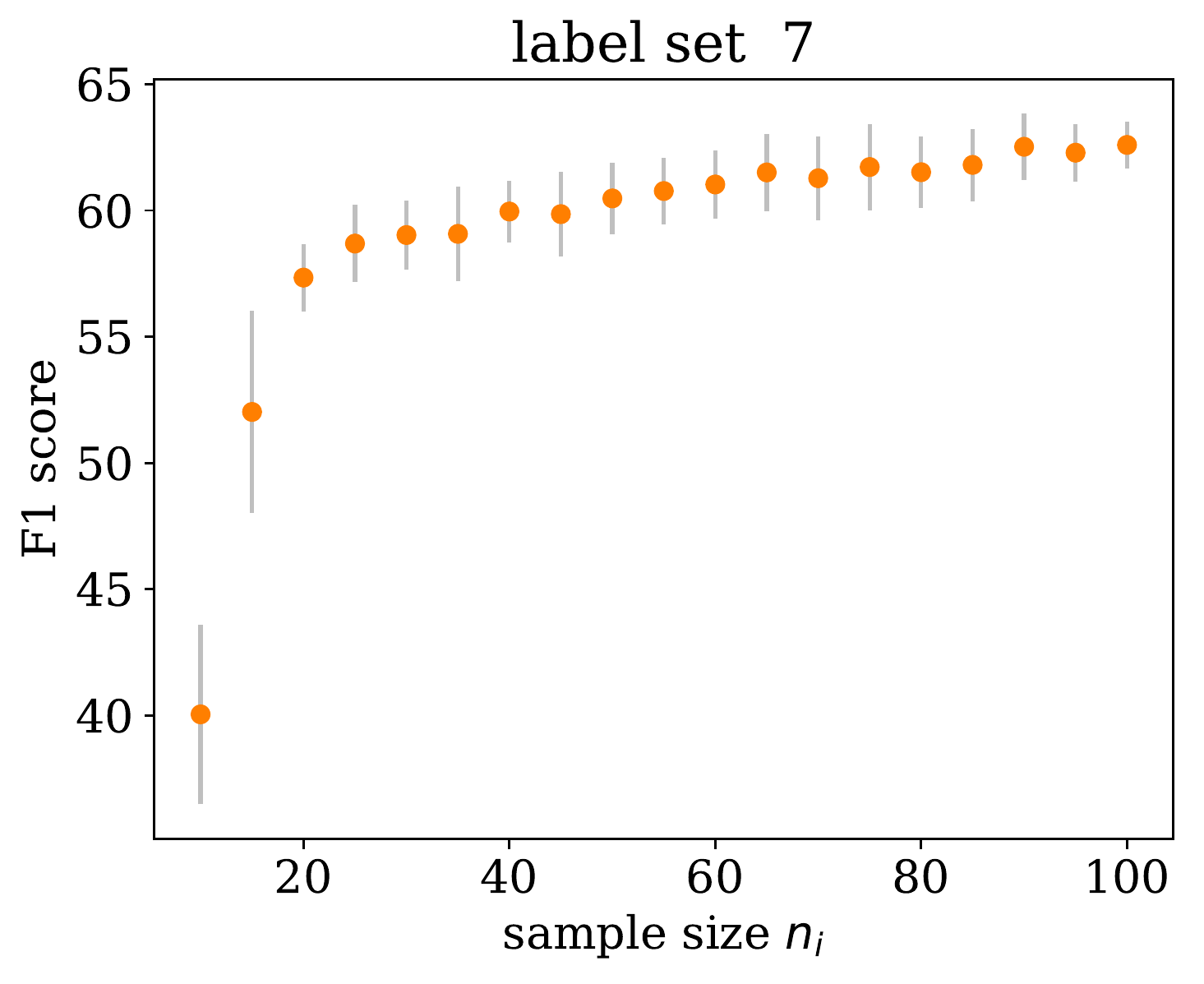}
    \caption{Mean test performance (Accuracy/F1 score) of the base model on Clothing1M and NoisyNER \textbf{with increasing $\mathbf{|D_C|}$} for the base model and varying for the noise model estimation and with \textbf{Fixed Sampling}. Grey error bars show the empirical standard deviation.}
\end{figure}

\begin{figure}
    \centering
    \includegraphics[height=3.5cm]{acc_f1_exp/clothing1m/se_acc_scale_fixC_mfixed/clothing1M_ns_acc_0.pdf}
    \includegraphics[height=3.5cm]{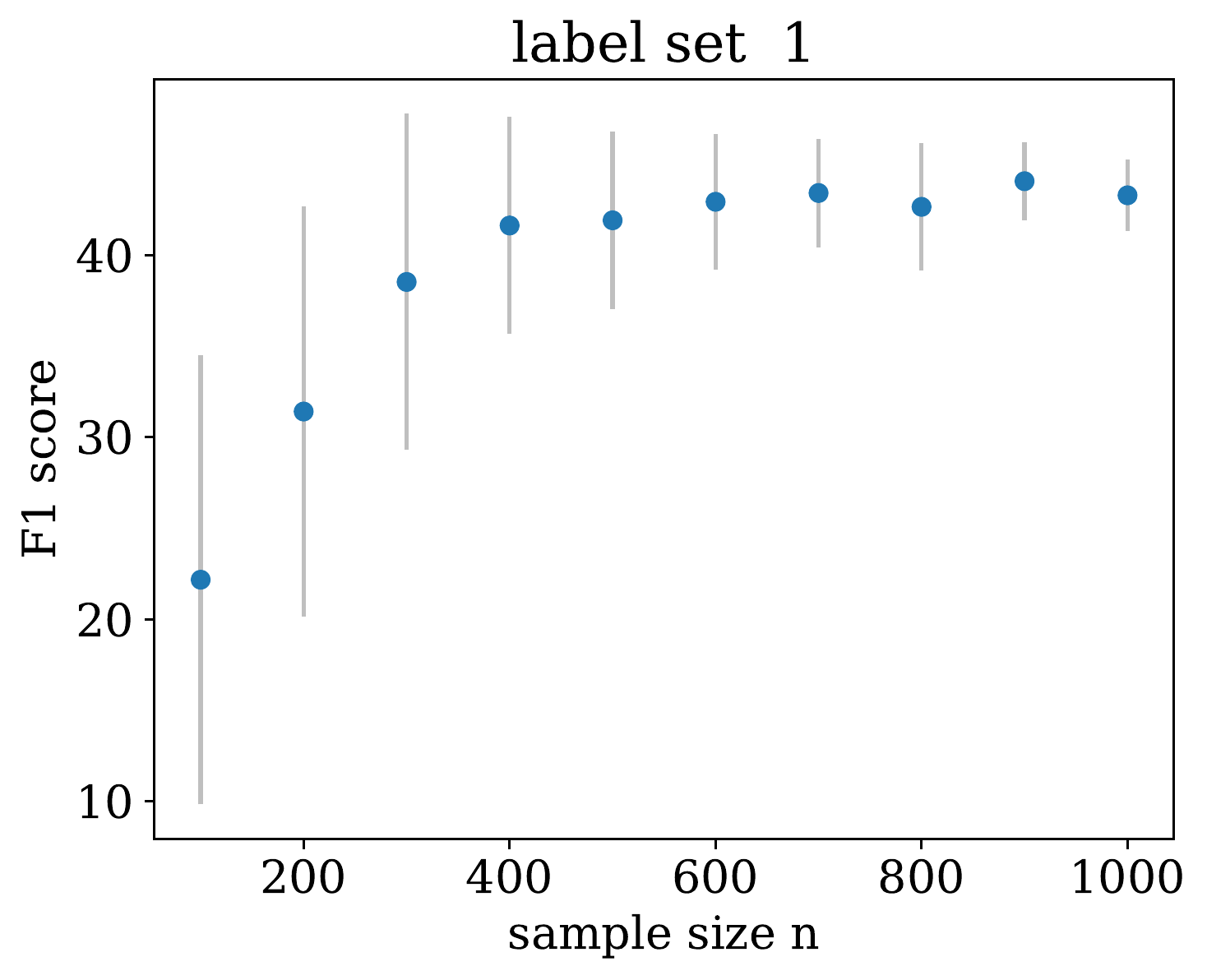}
    \includegraphics[height=3.5cm]{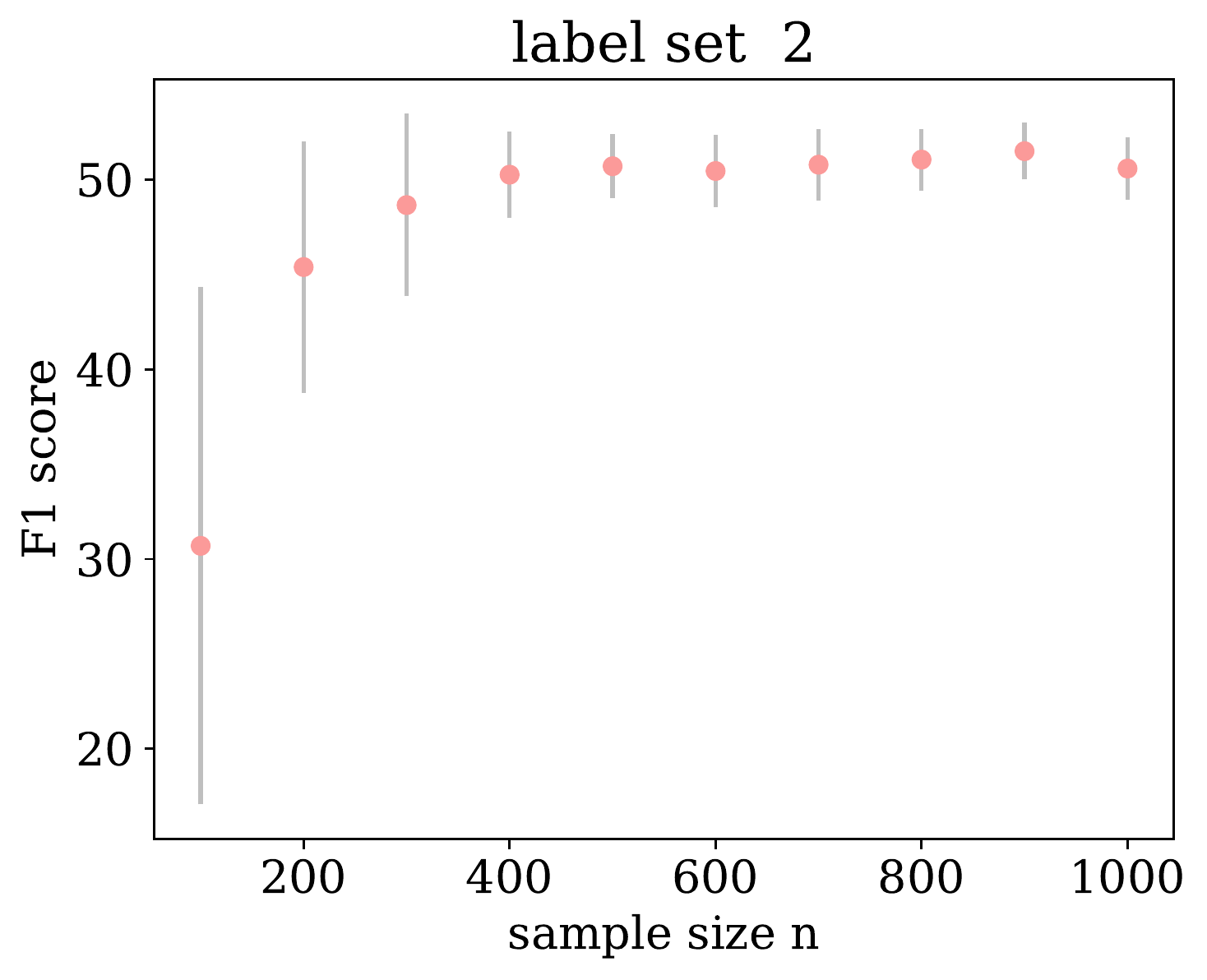}
    \includegraphics[height=3.5cm]{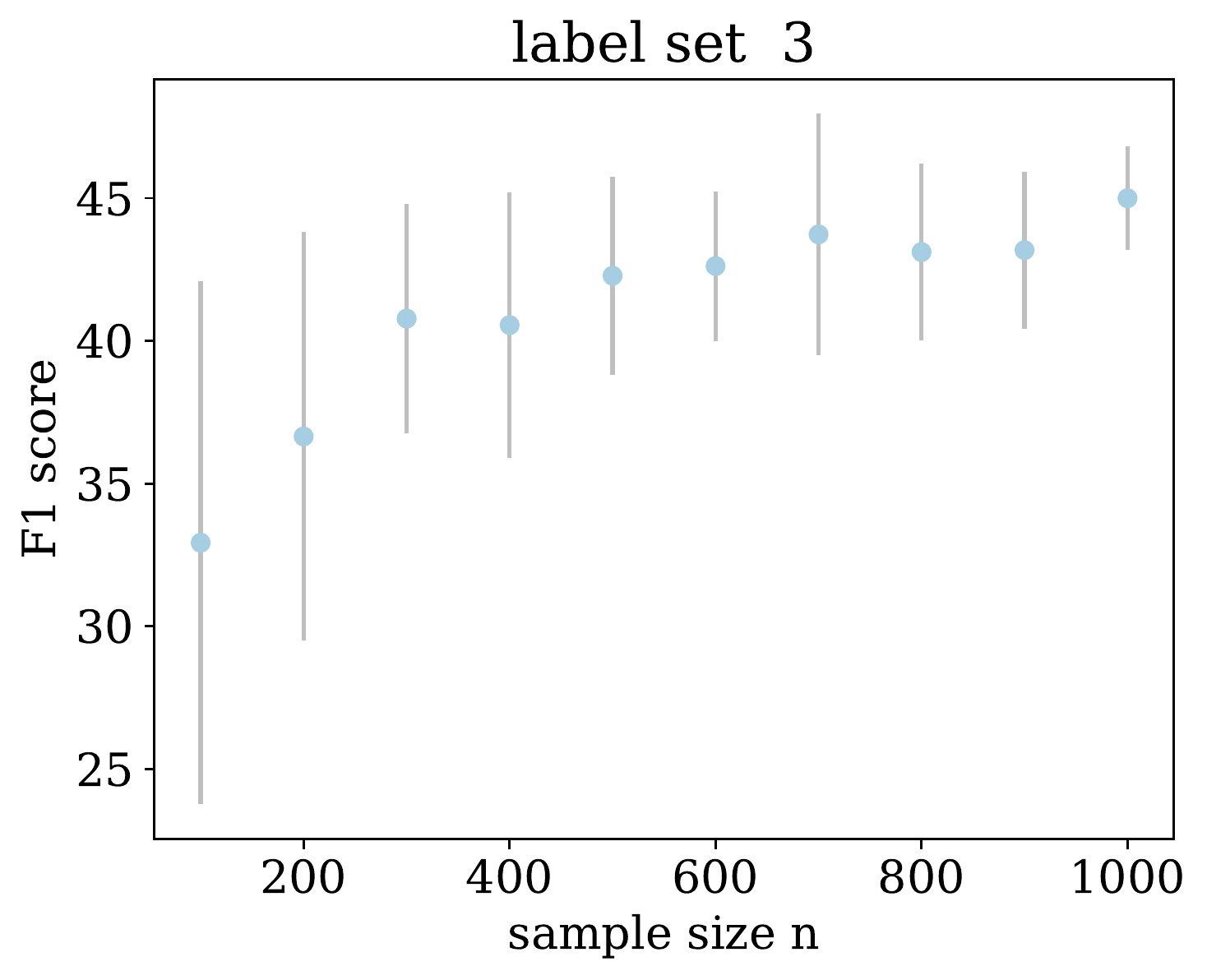}
    \includegraphics[height=3.5cm]{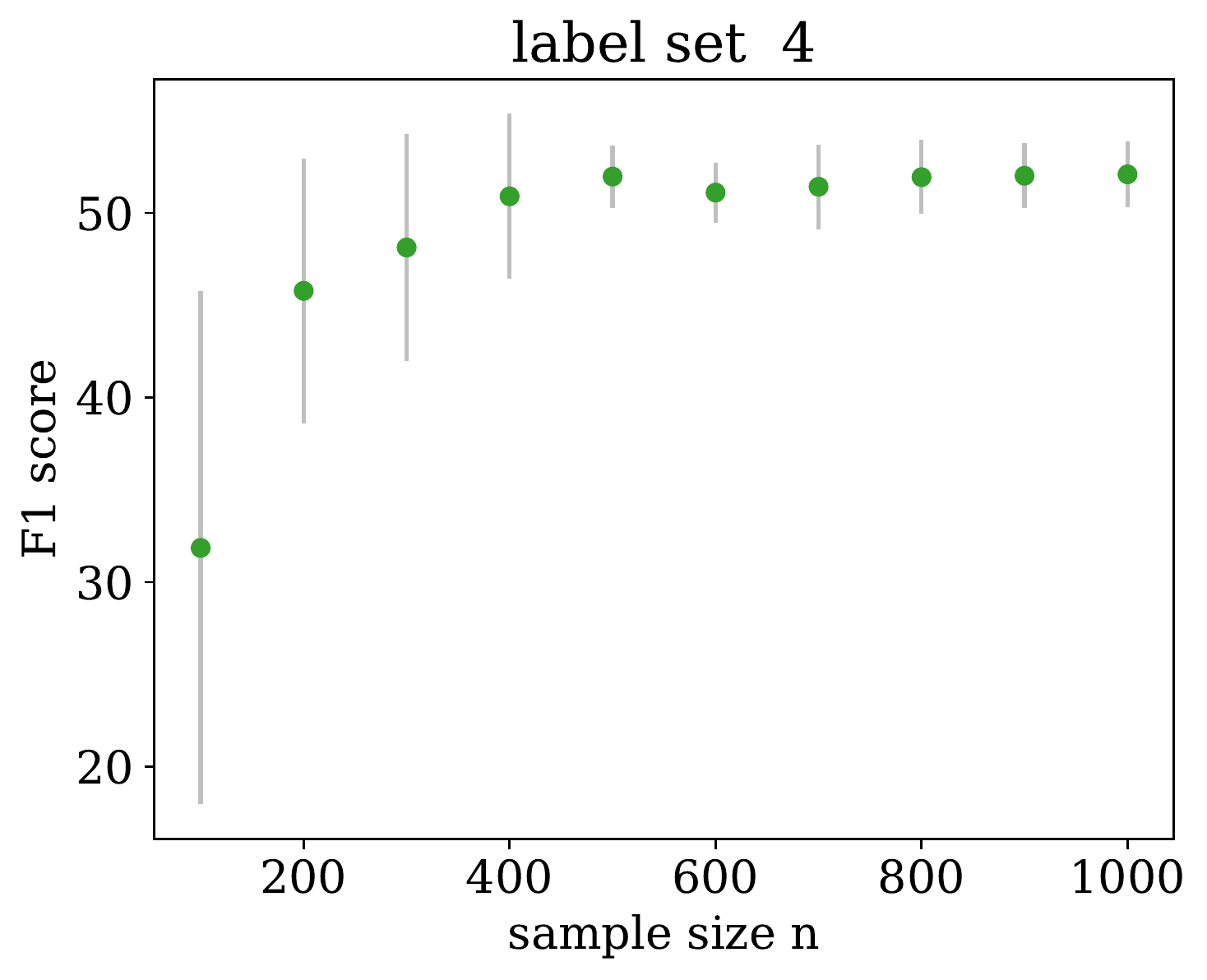}
    \includegraphics[height=3.5cm]{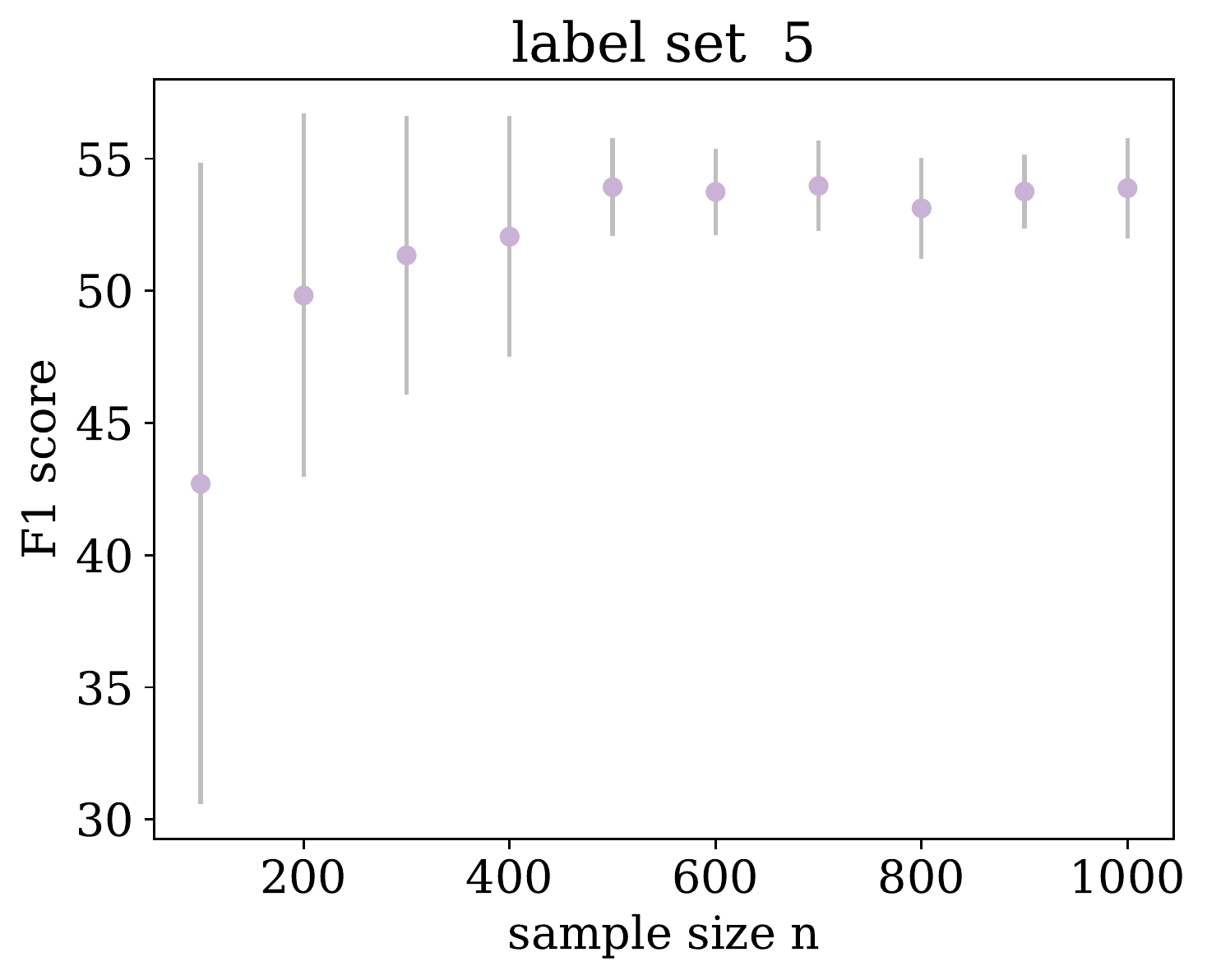}
    \includegraphics[height=3.5cm]{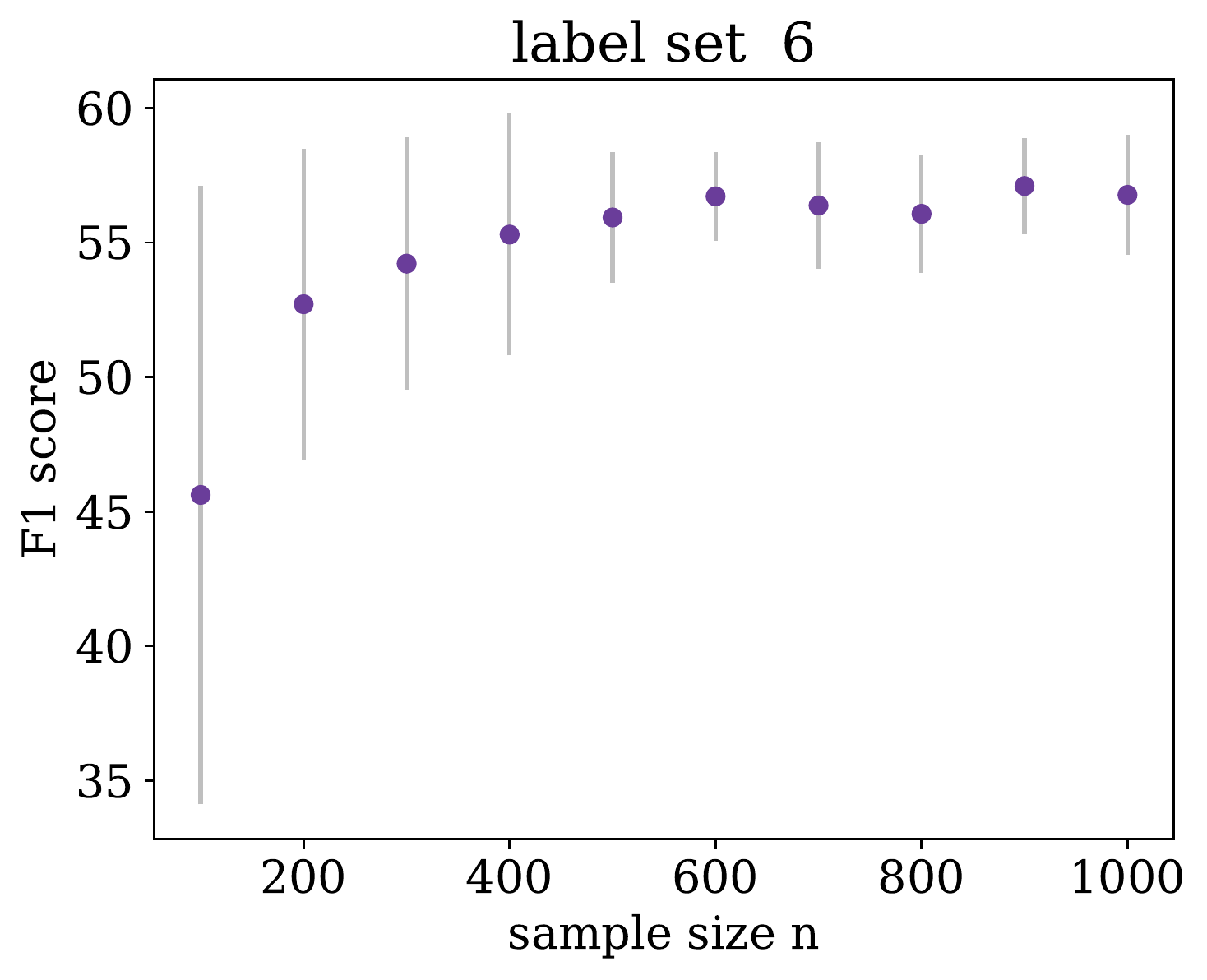}
    \includegraphics[height=3.5cm]{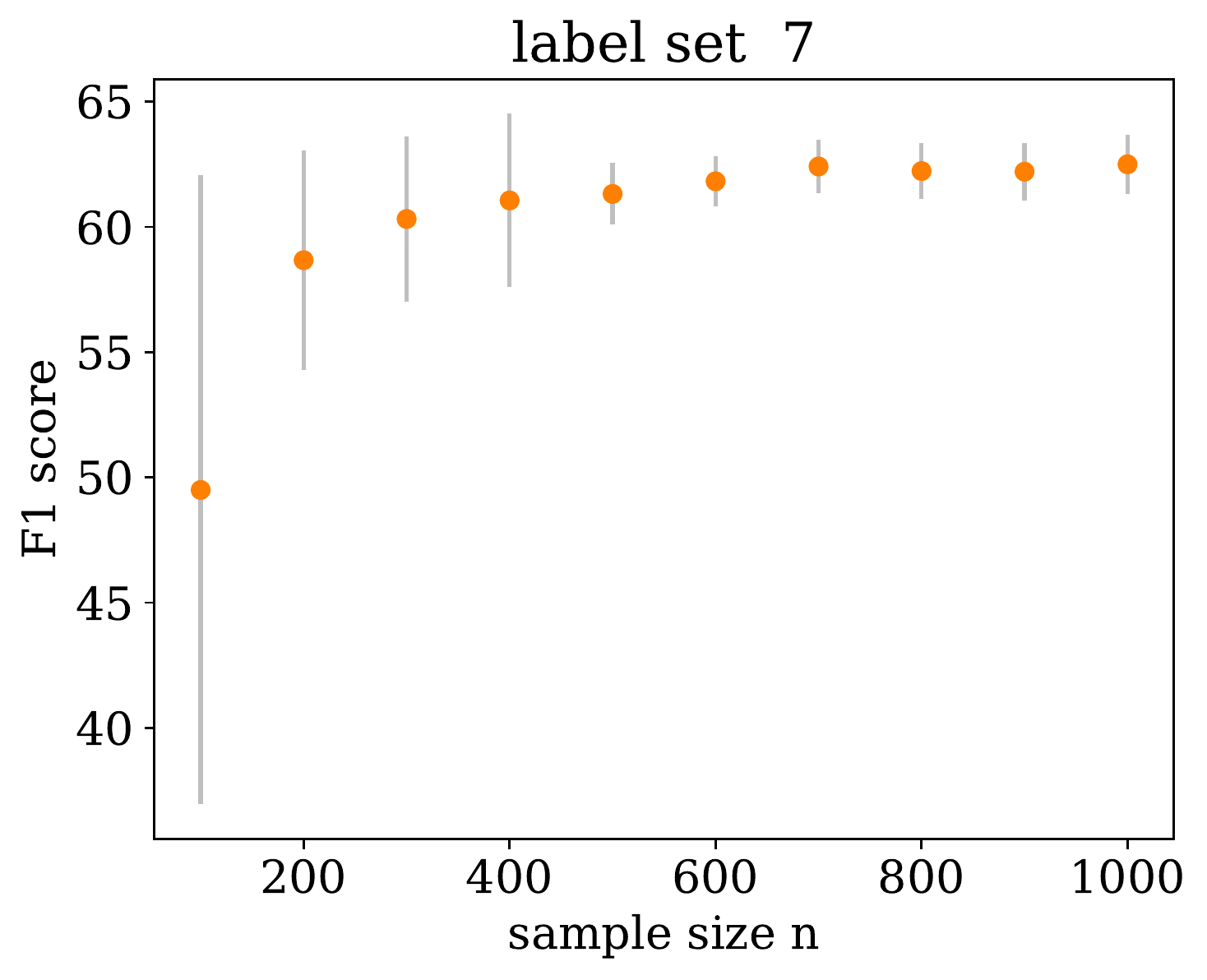}
    \caption{Mean test performance (Accuracy/F1 score) of the base model on Clothing1M and NoisyNER \textbf{with increasing $\mathbf{|D_C|}$} for the base model and varying for the noise model estimation and with \textbf{Variable Sampling}. Grey error bars show the empirical standard deviation.}
\end{figure}

\begin{figure}
    \centering
    \includegraphics[height=3.5cm]{acc_f1_exp/clothing1m/se_acc_scale_fixC_mfixed/clothing1M_ns_acc_0.pdf}
    \includegraphics[height=3.5cm]{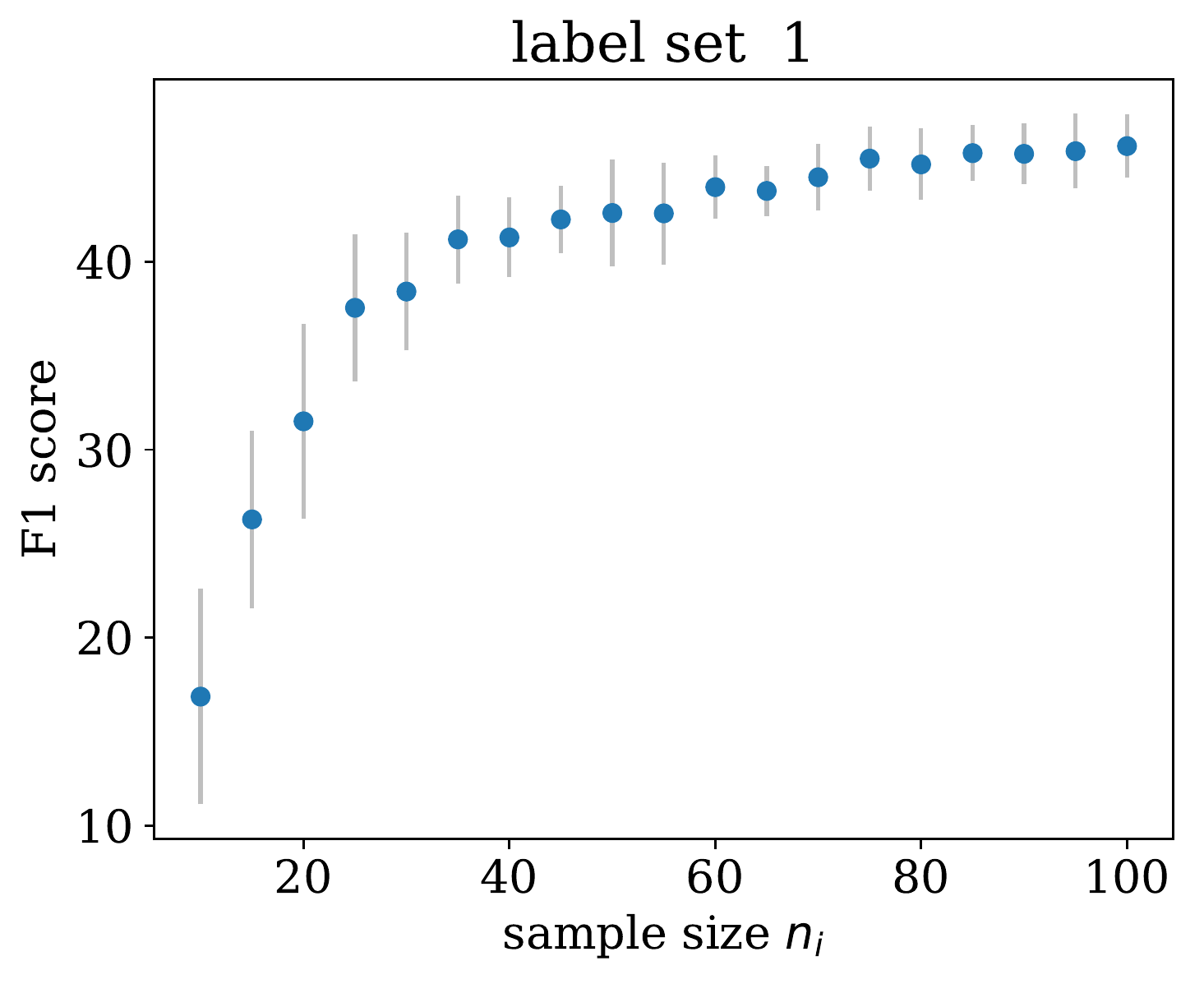}
    \includegraphics[height=3.5cm]{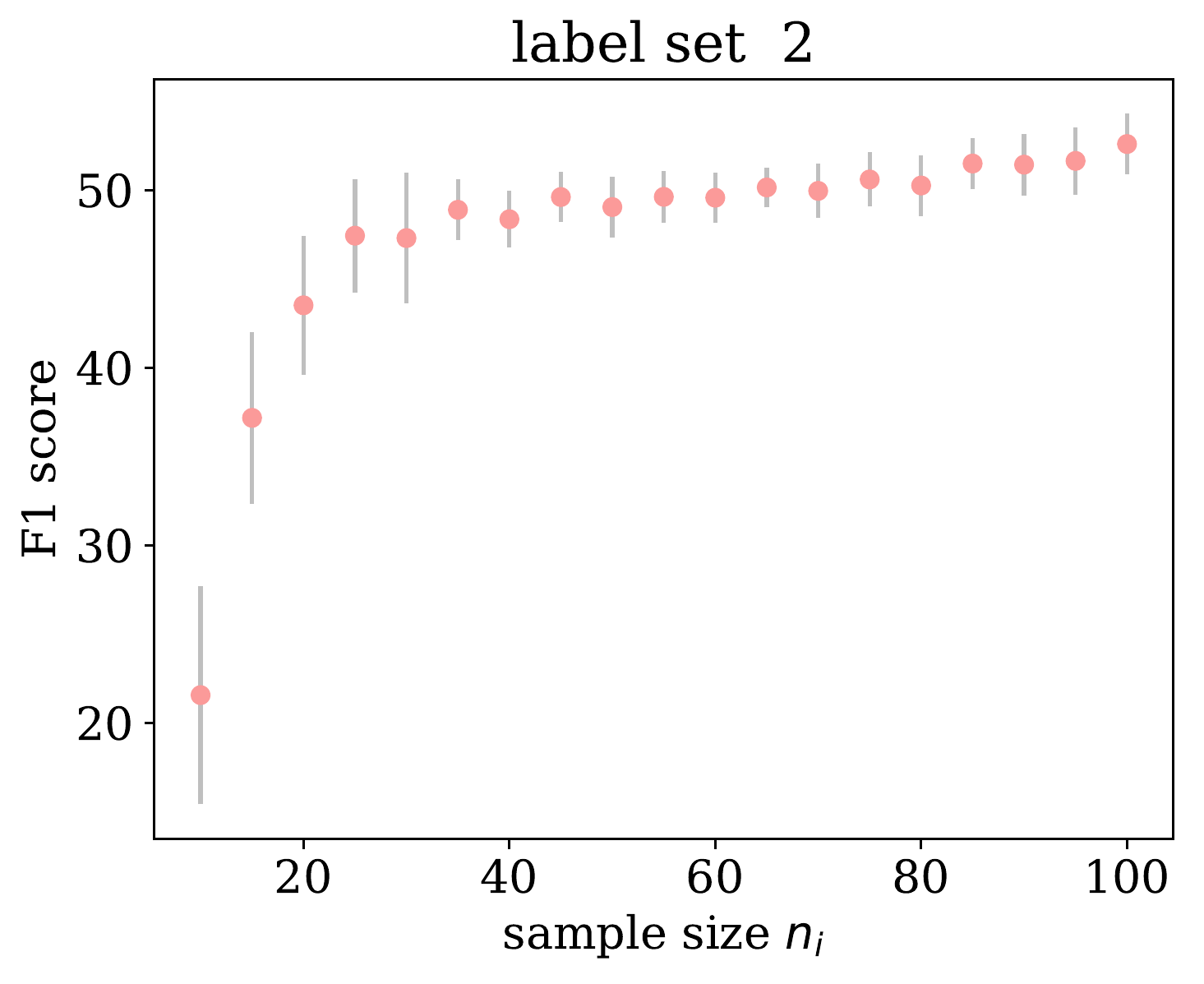}
    \includegraphics[height=3.5cm]{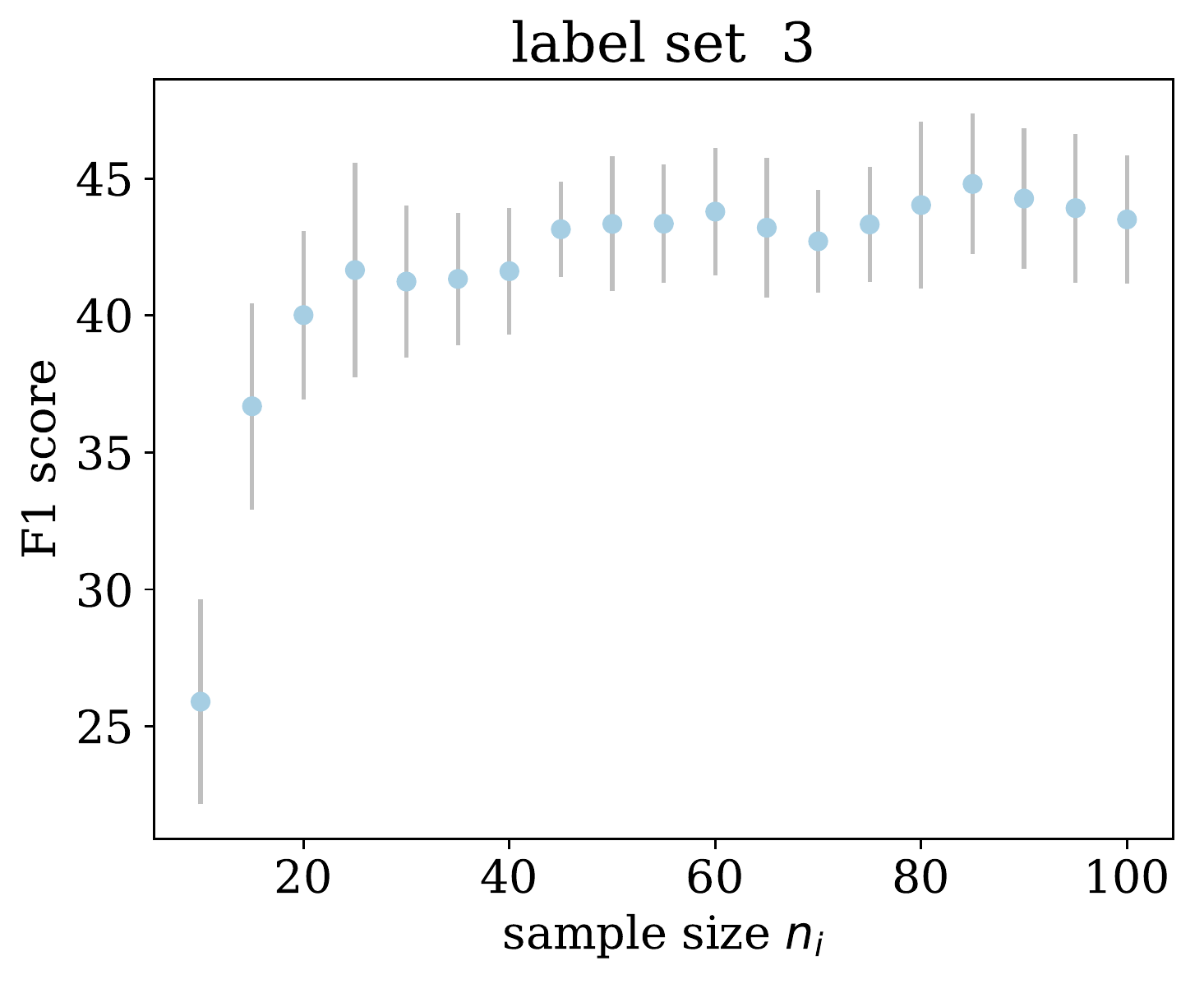}
    \includegraphics[height=3.5cm]{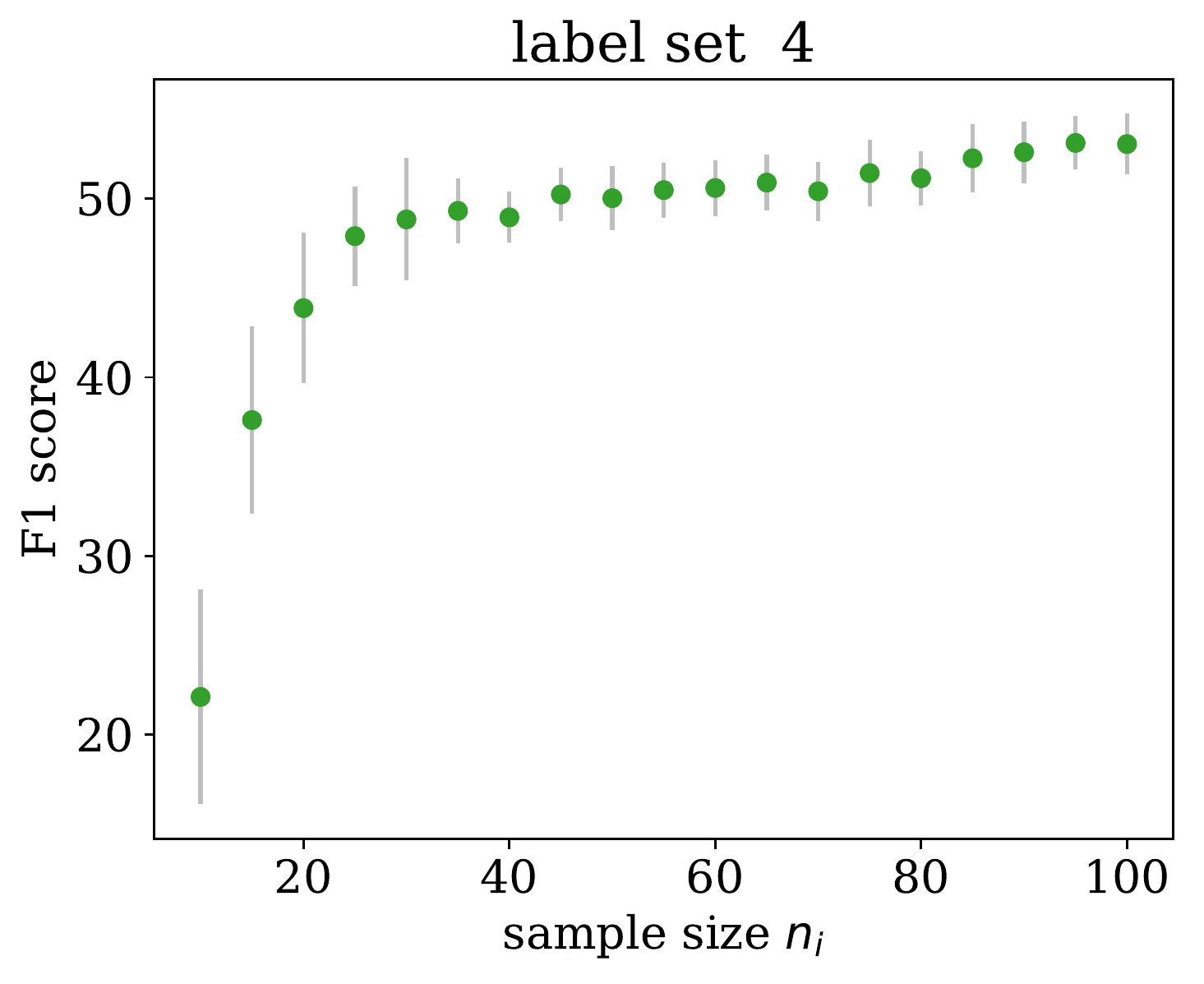}
    \includegraphics[height=3.5cm]{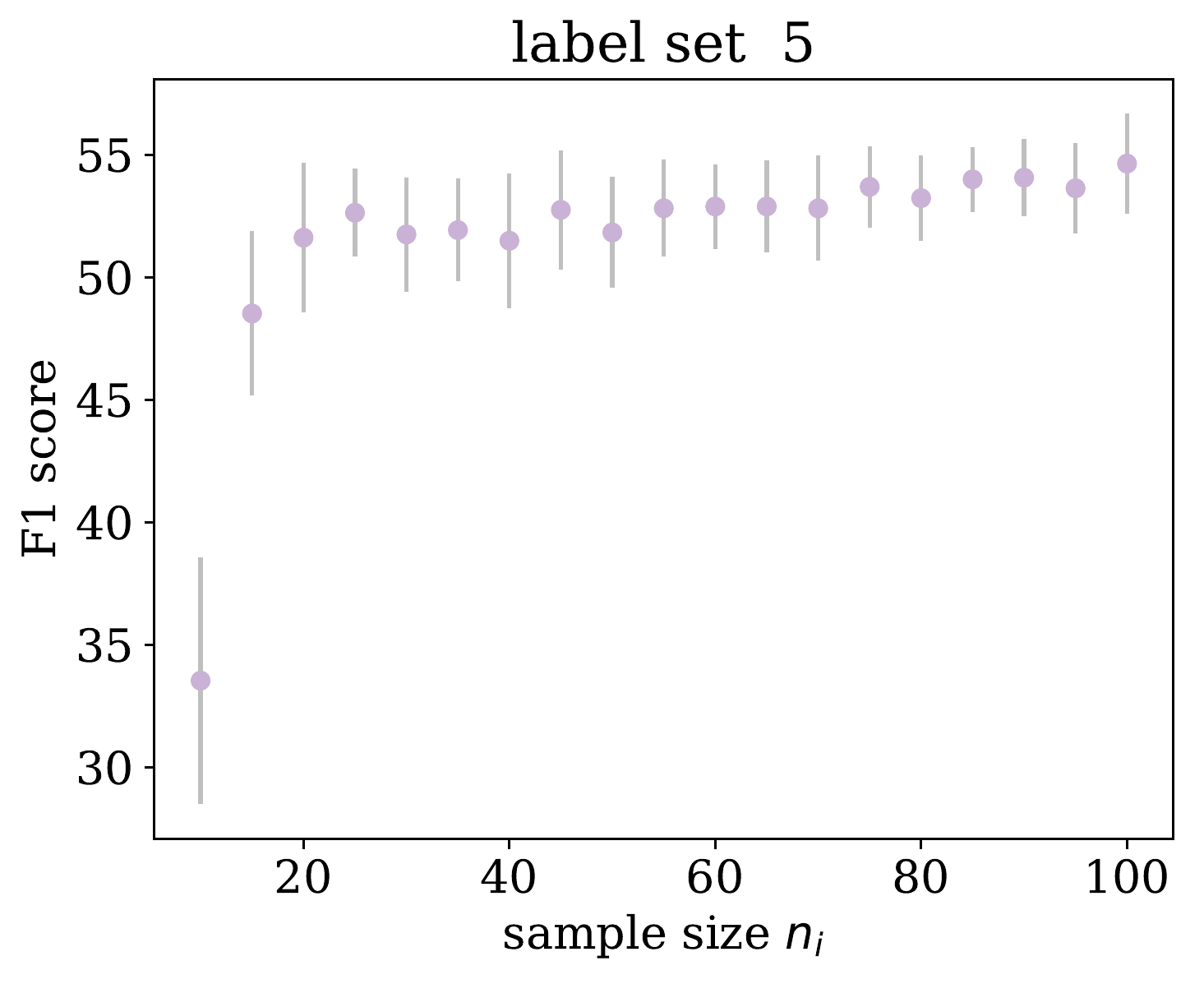}
    \includegraphics[height=3.5cm]{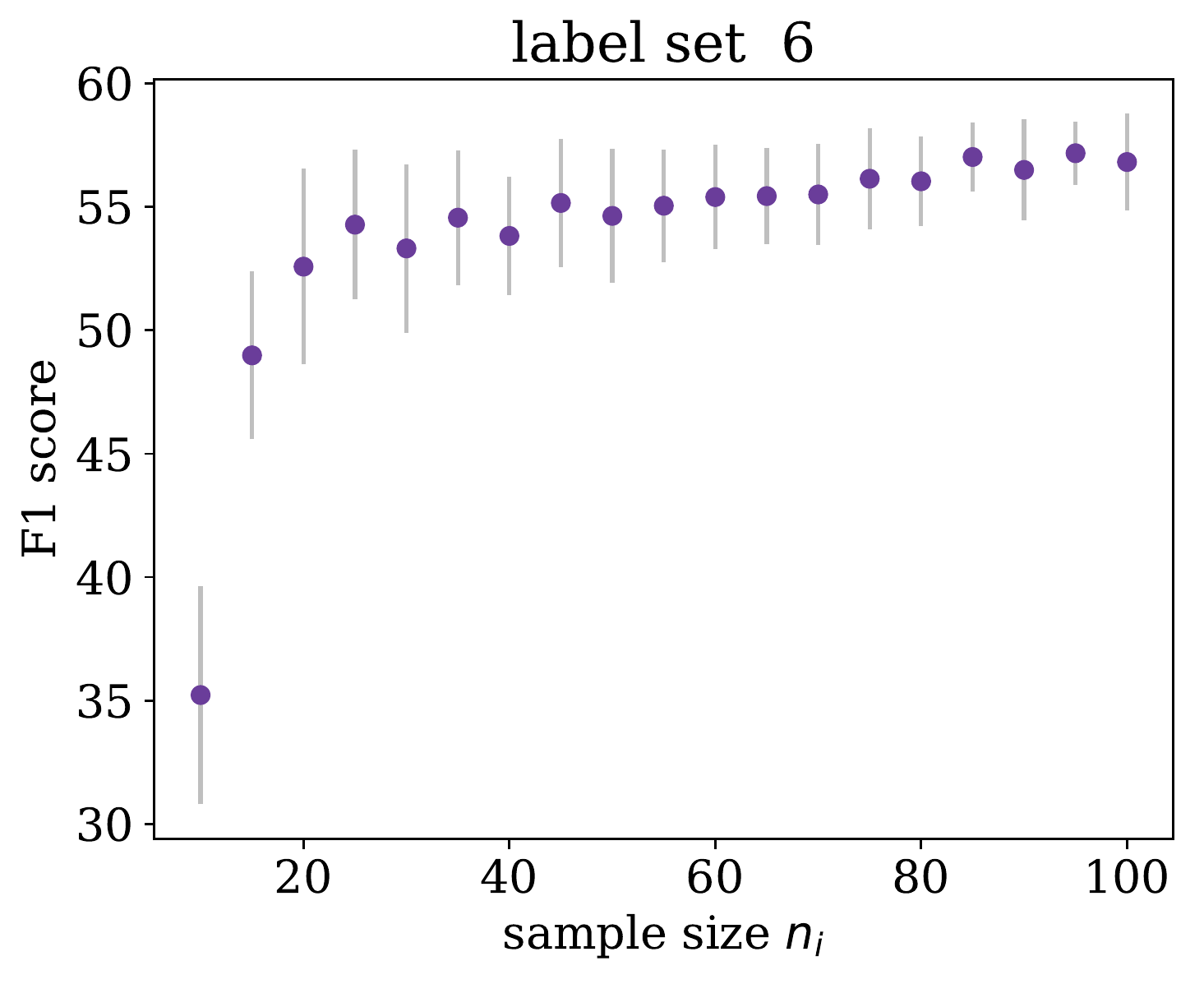}
    \includegraphics[height=3.5cm]{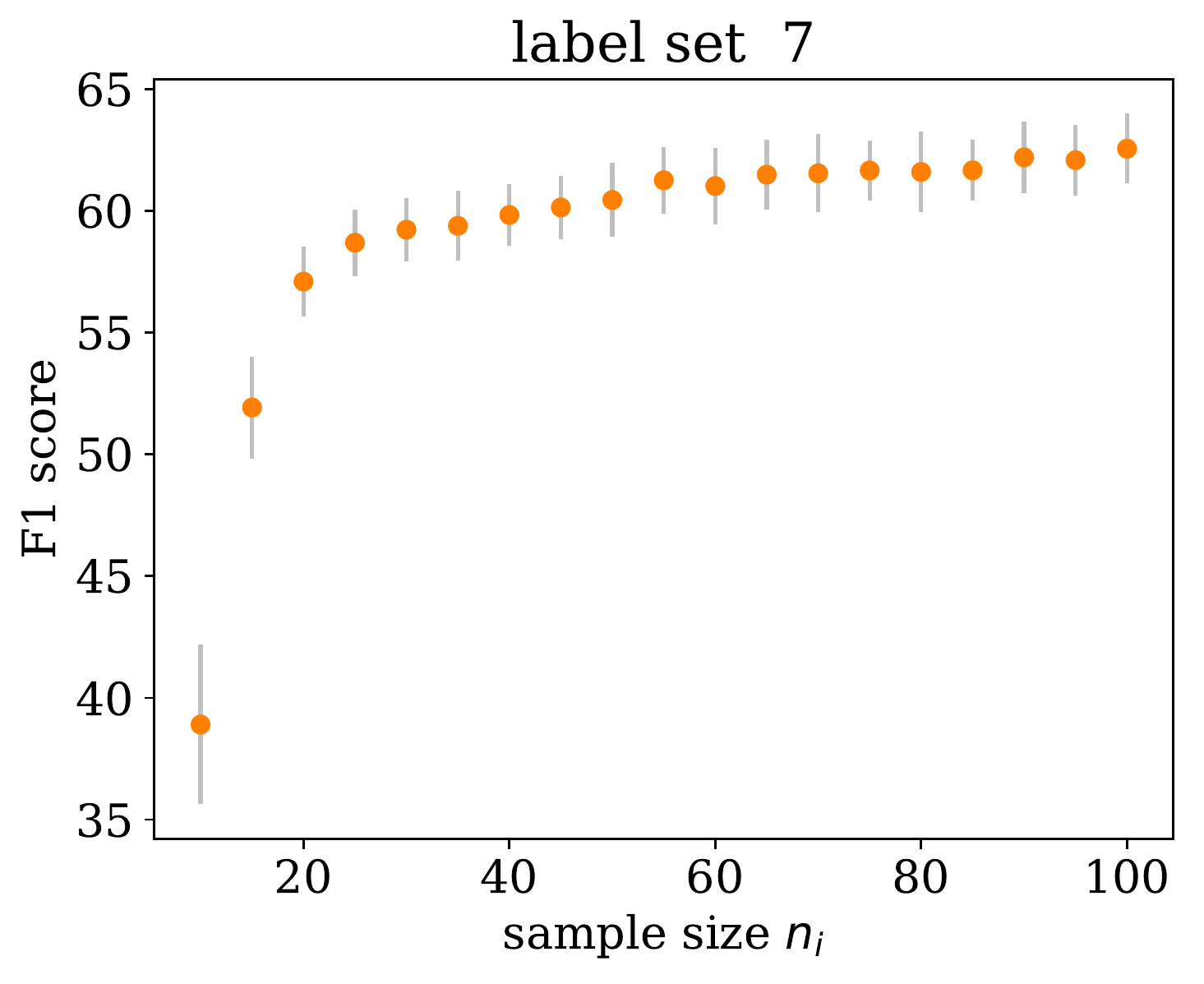}
    \caption{Mean test performance (Accuracy/F1 score) of the base model on Clothing1M and NoisyNER \textbf{with $\mathbf{|D_C|}$ fixed} for the base model and varying for the noise model estimation and with \textbf{Fixed Sampling}. Grey error bars show the empirical standard deviation.}
\end{figure}

\begin{figure}
    \centering
    \includegraphics[height=3.5cm]{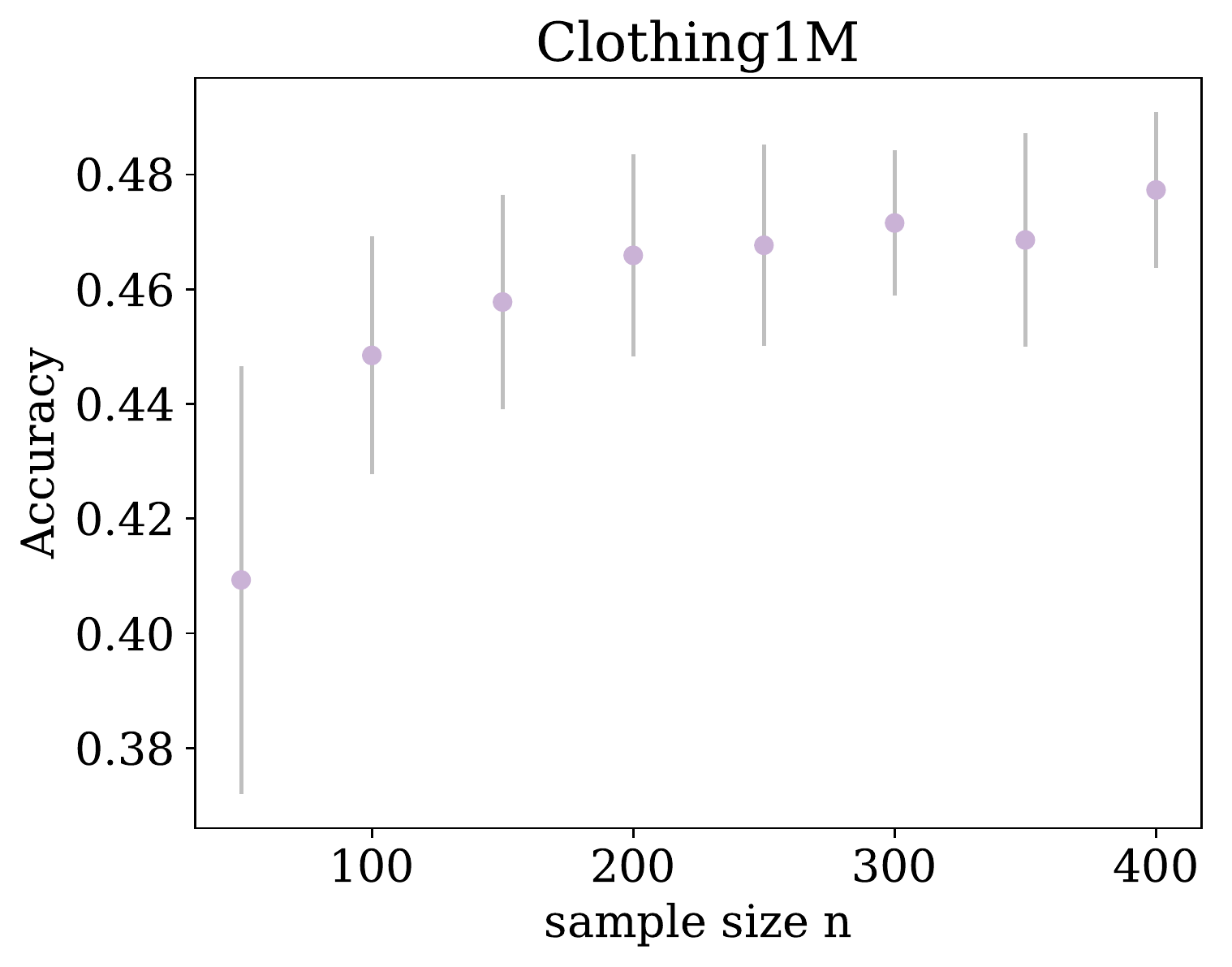}
    \includegraphics[height=3.5cm]{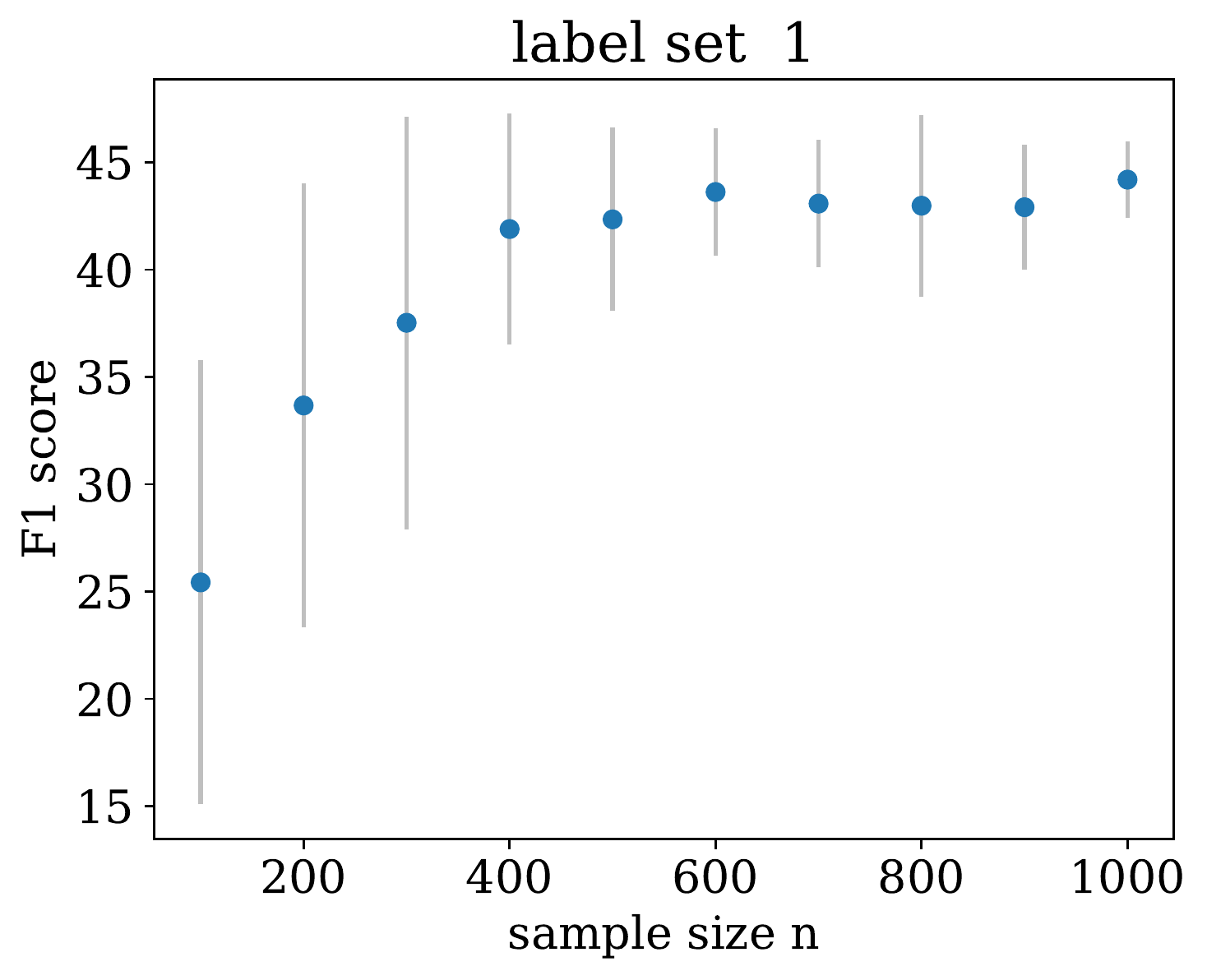}
    \includegraphics[height=3.5cm]{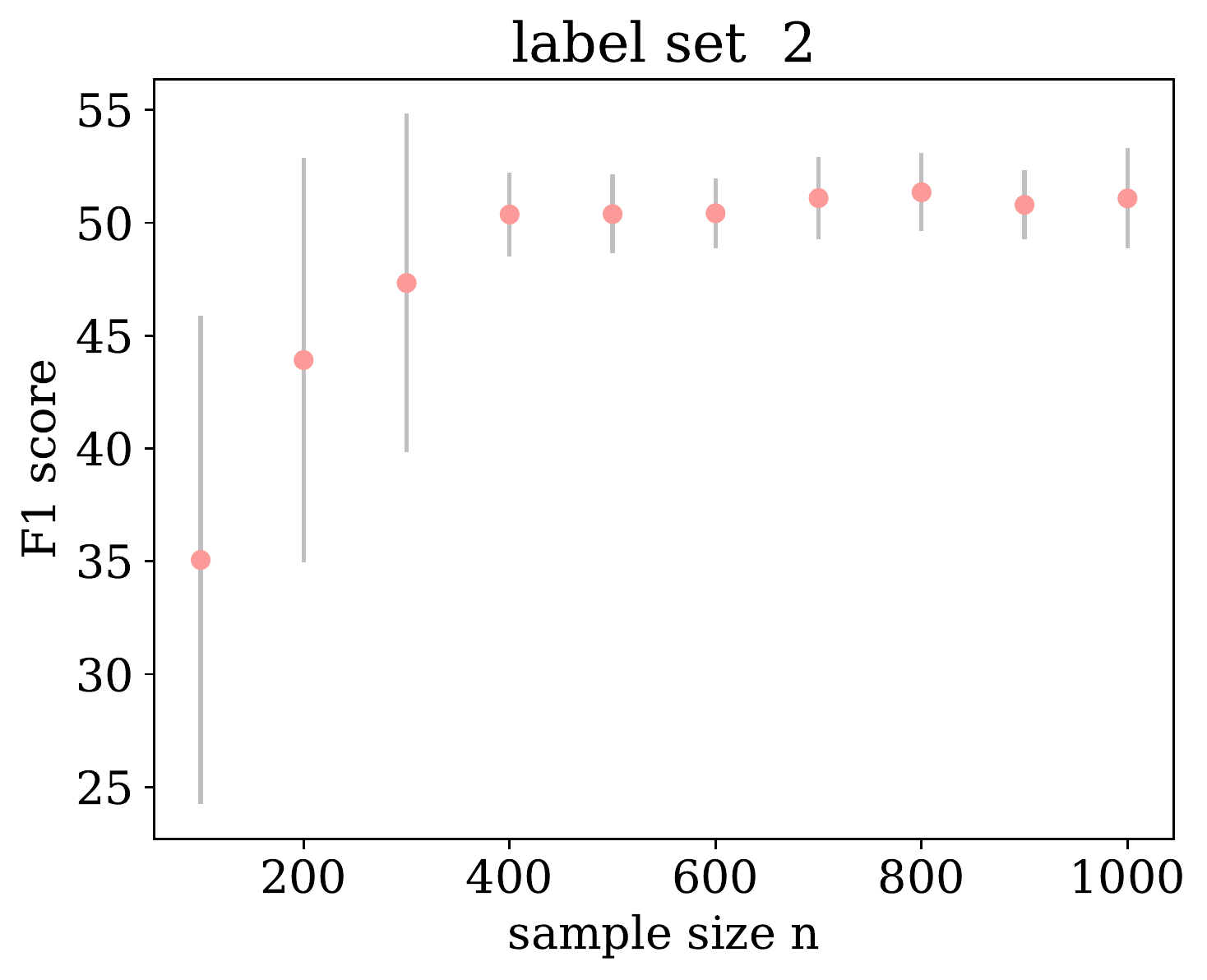}
    \includegraphics[height=3.5cm]{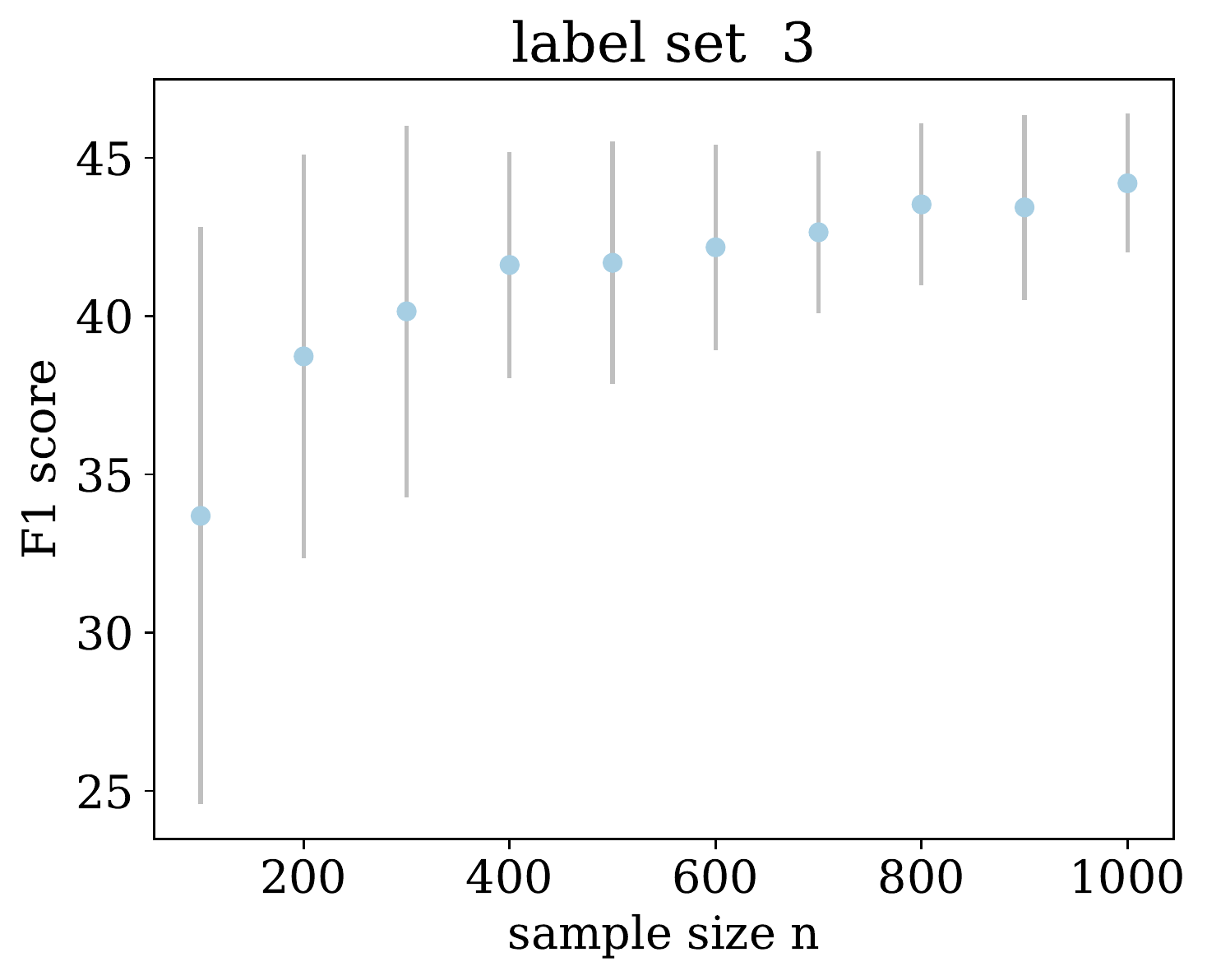}
    \includegraphics[height=3.5cm]{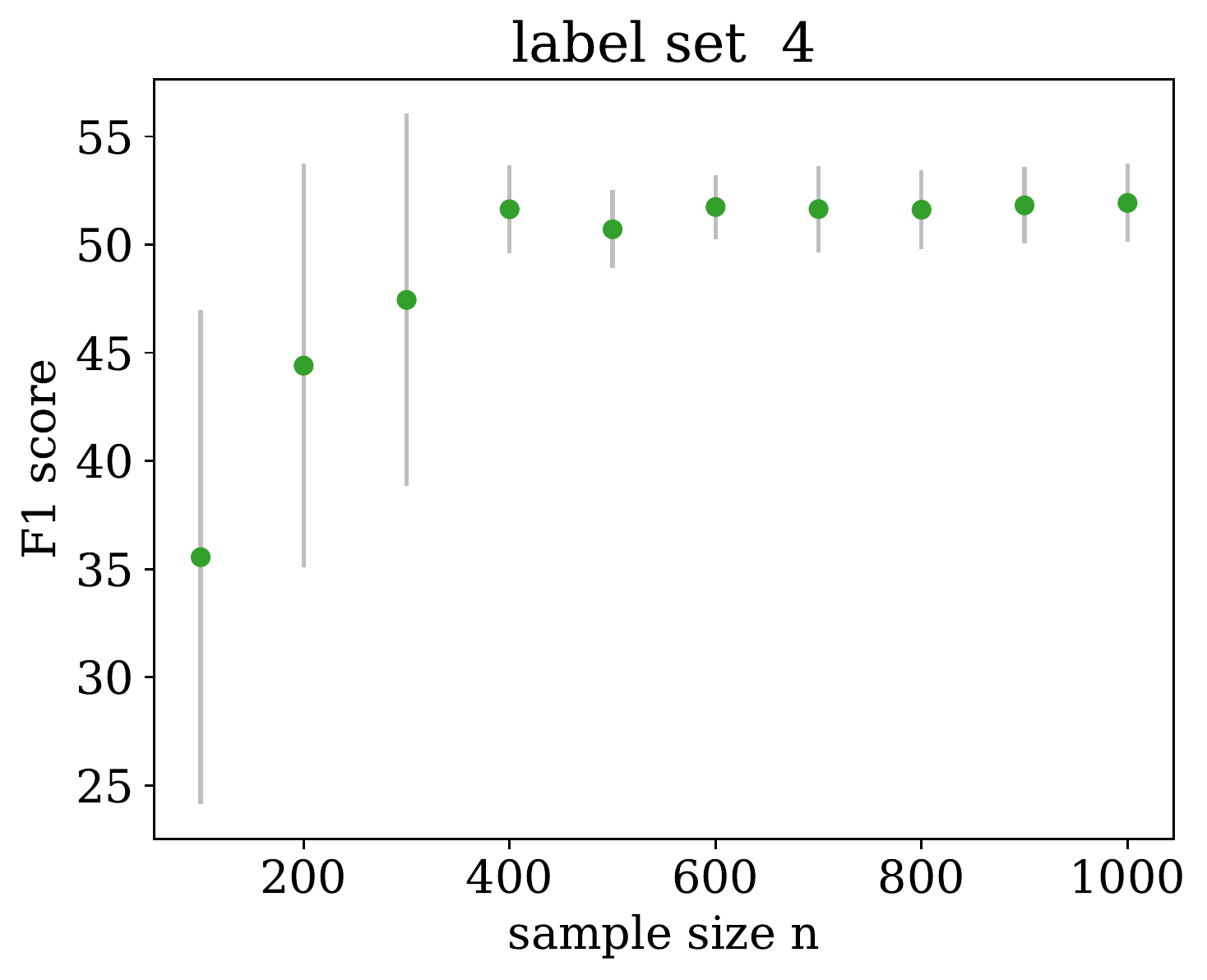}
    \includegraphics[height=3.5cm]{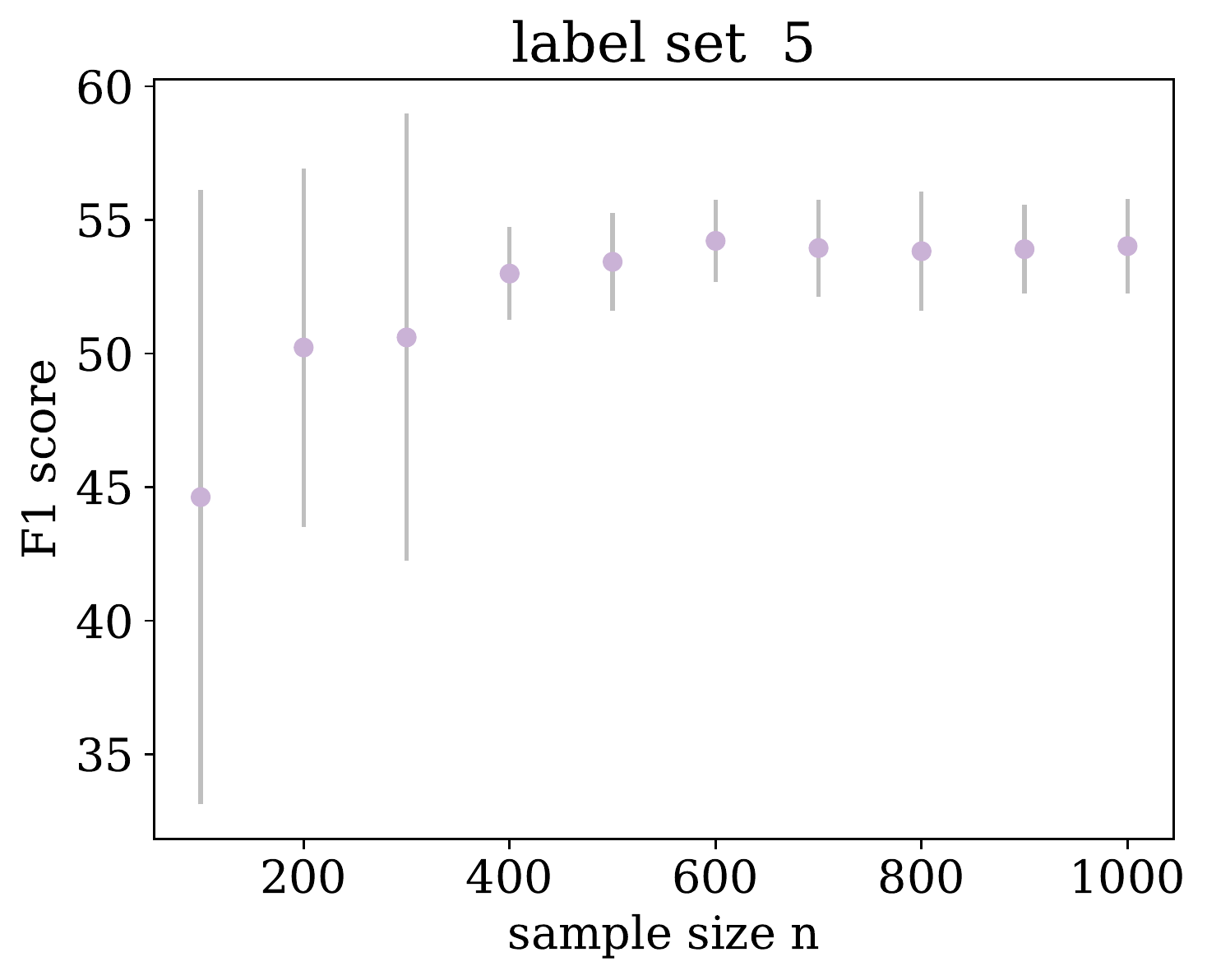}
    \includegraphics[height=3.5cm]{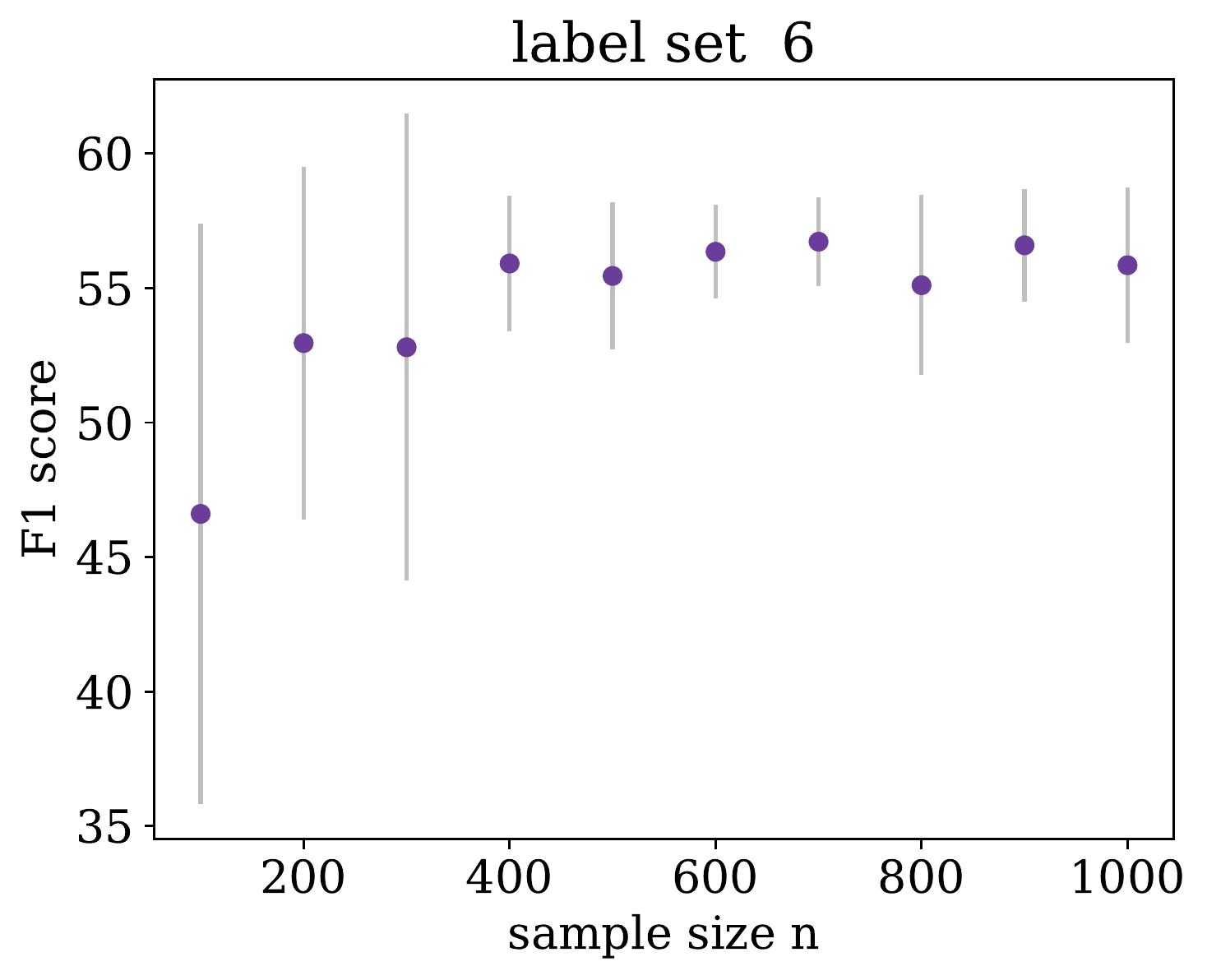}
    \includegraphics[height=3.5cm]{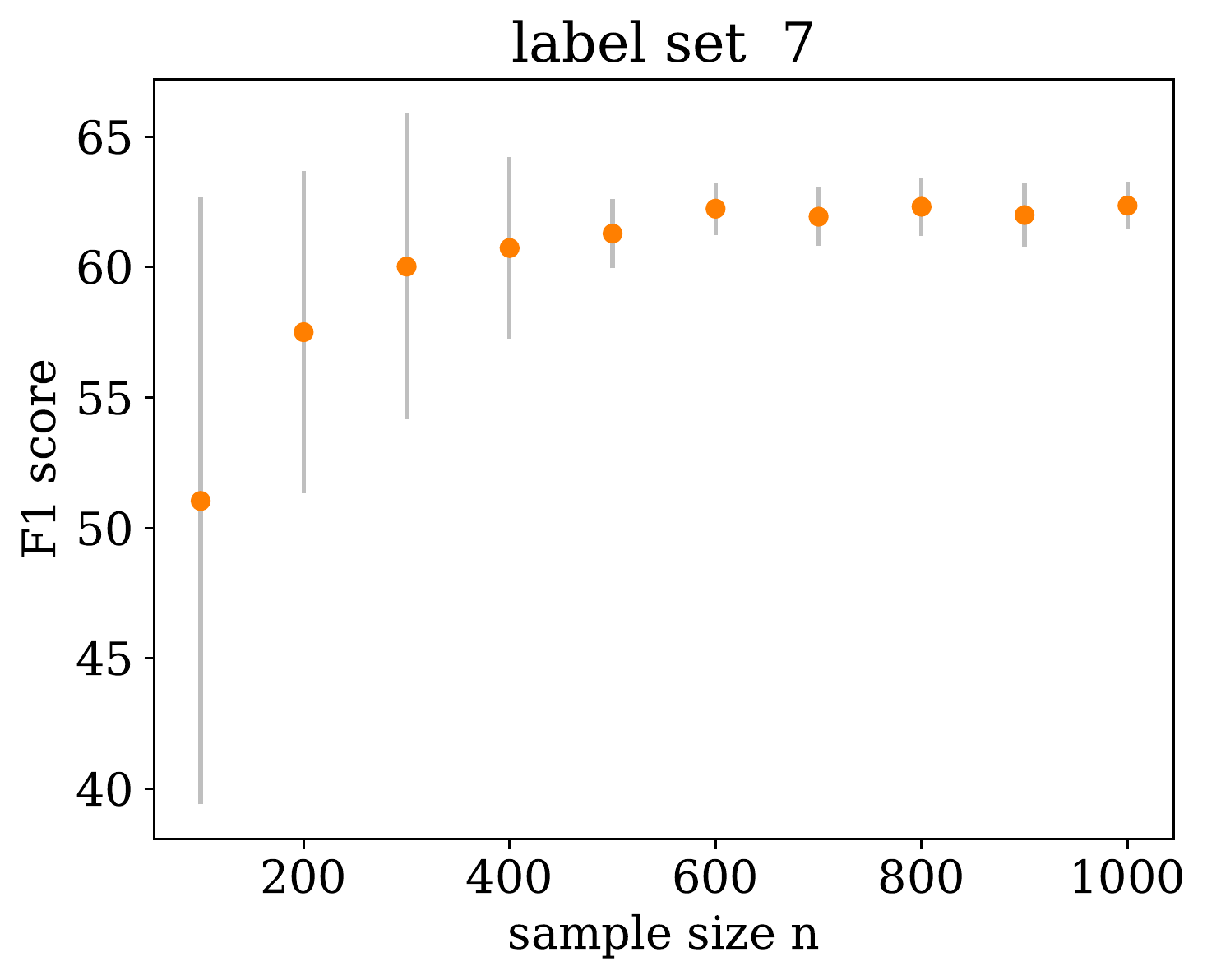}
    \caption{Mean test performance (Accuracy/F1 score) of the base model on Clothing1M and NoisyNER \textbf{with $\mathbf{|D_C|}$ fixed} for the base model and varying for the noise model estimation and with \textbf{Variable Sampling}. Grey error bars show the empirical standard deviation.}
\end{figure}

\begin{figure*}
    \centering
    \includegraphics[height=3.35cm]{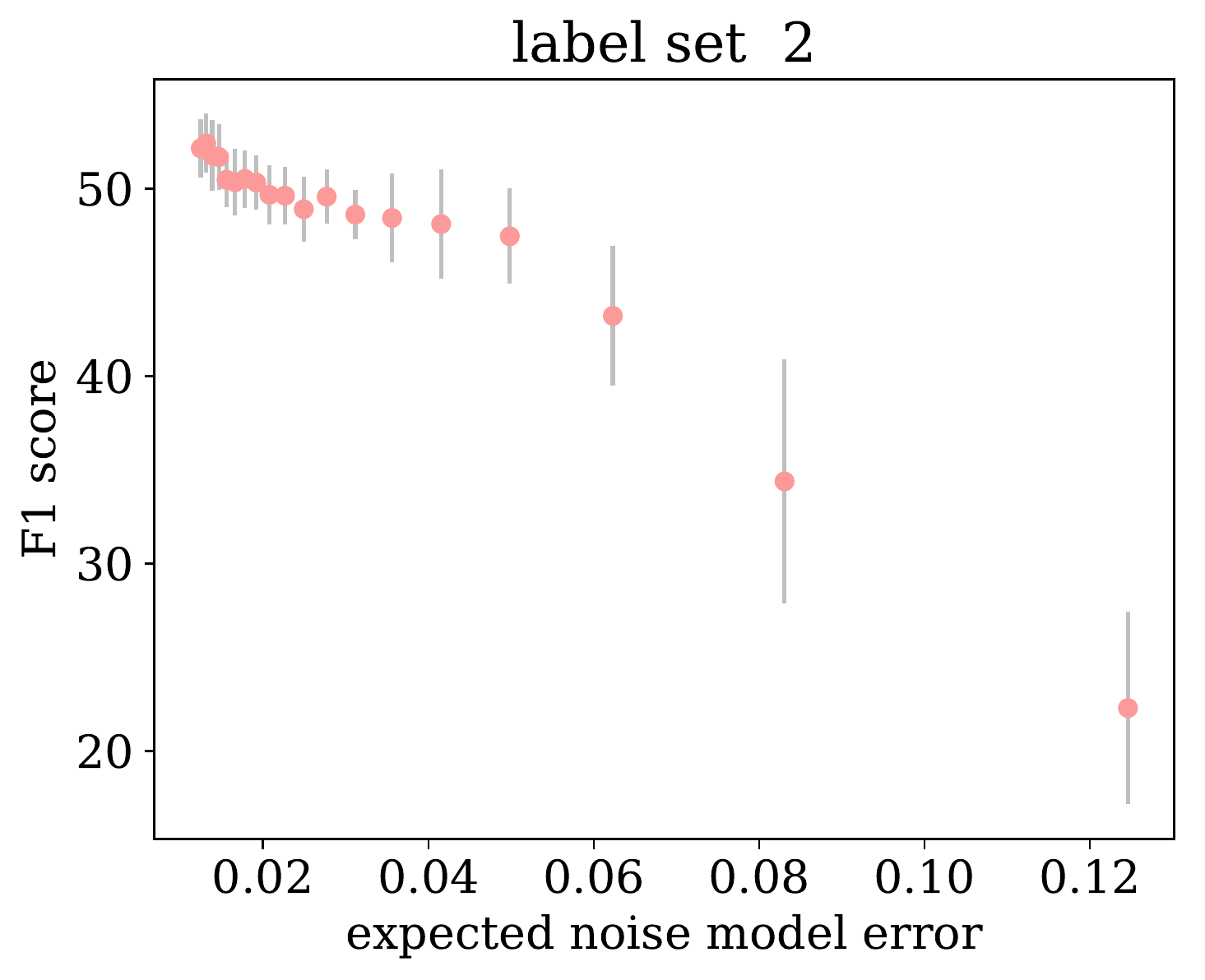}
    \includegraphics[height=3.35cm]{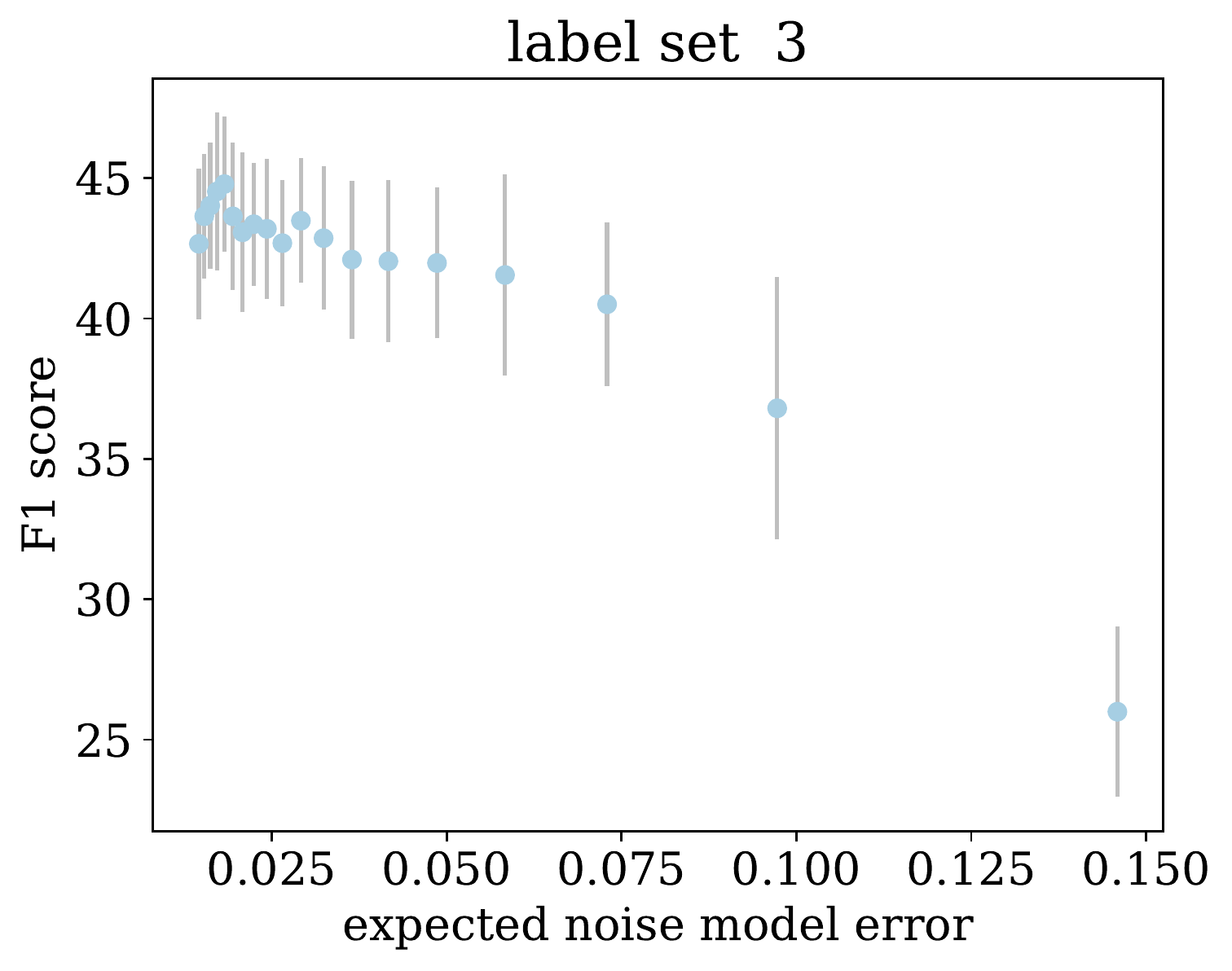}
    \includegraphics[height=3.35cm]{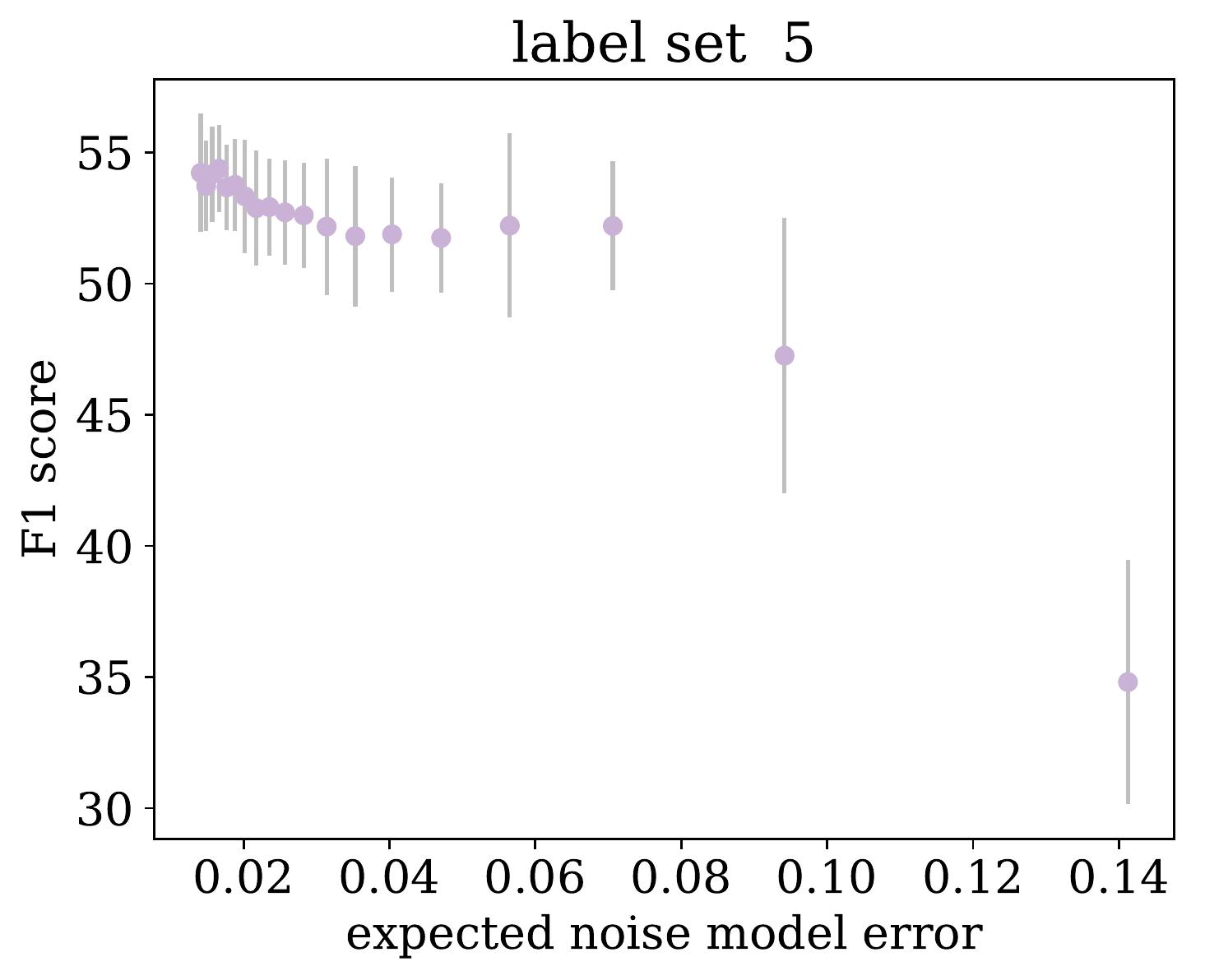}
    \includegraphics[height=3.35cm]{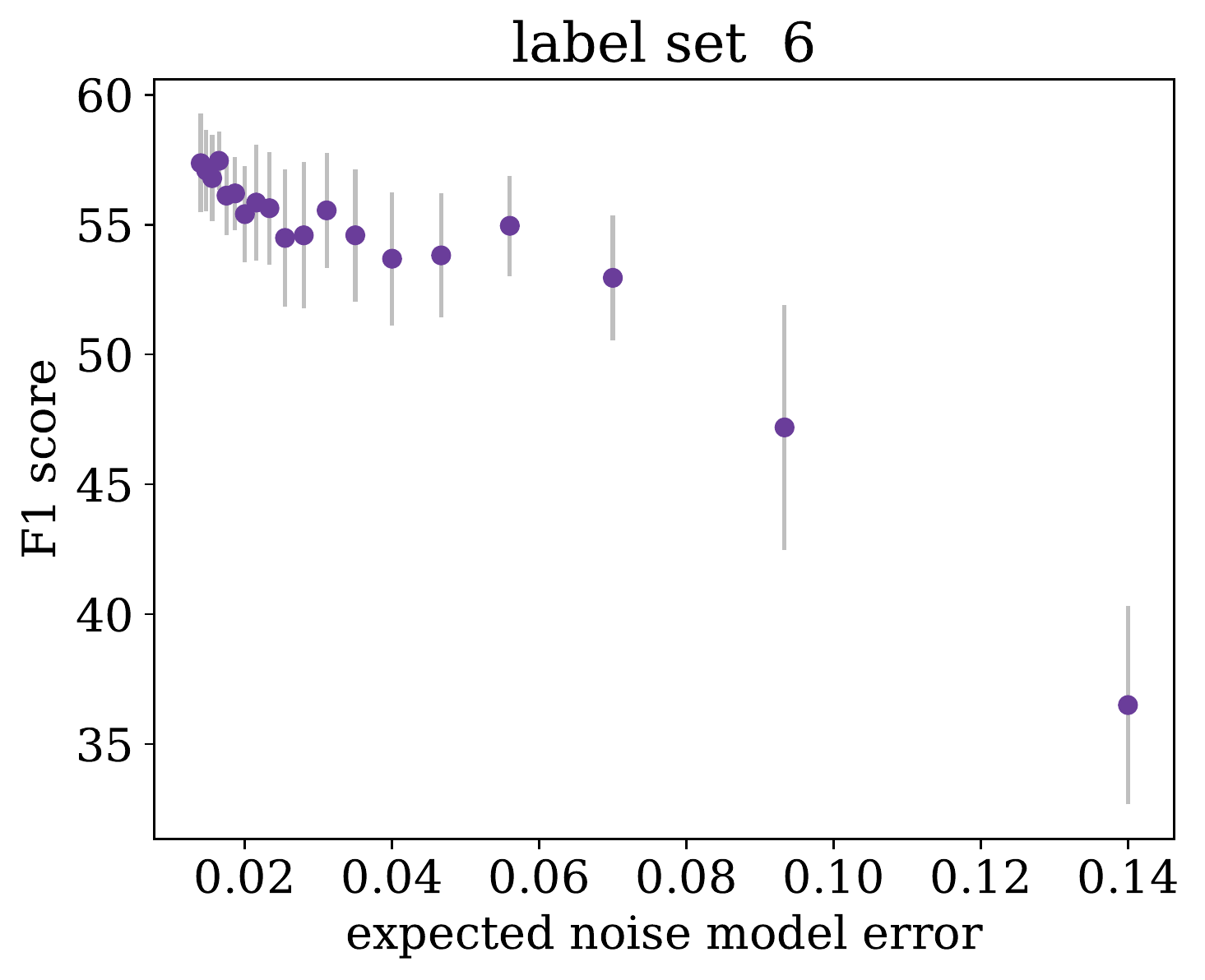}
    \caption{Relationship between the theoretically expected noise model error and the mean test performance of the base model for Clothing1M and NoisyNER label set 2, 3, 5 and 6 \textbf{with increasing $\mathbf{|D_C|}$} for the base model and varying for the noise model estimation and with \textbf{Fixed Sampling}. Each point corresponds to one sample size $n_i$. Grey error bars show the empirical standard deviation. }
\end{figure*}

\begin{figure*}
    \centering
    \includegraphics[height=3.35cm]{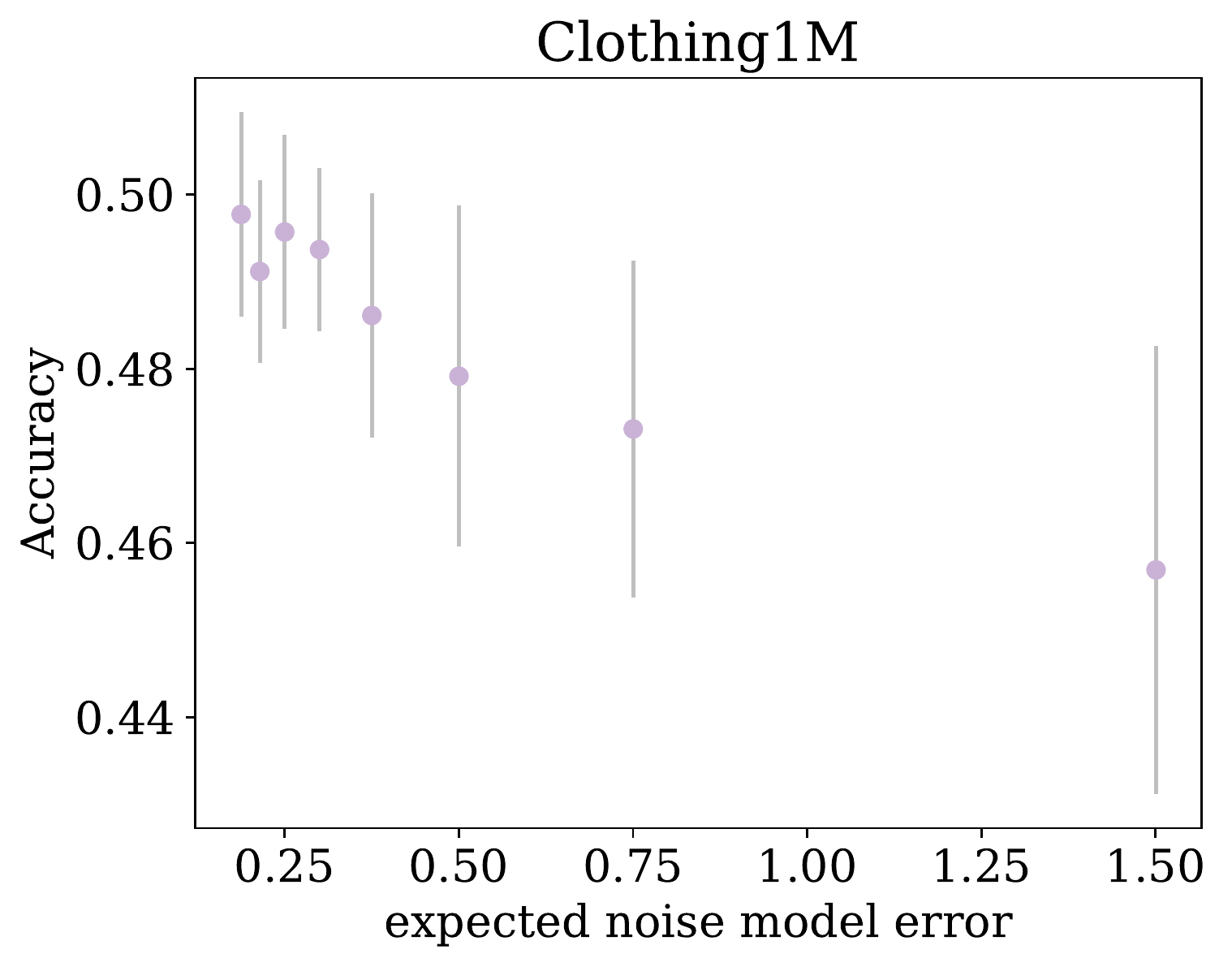}
    \includegraphics[height=3.35cm]{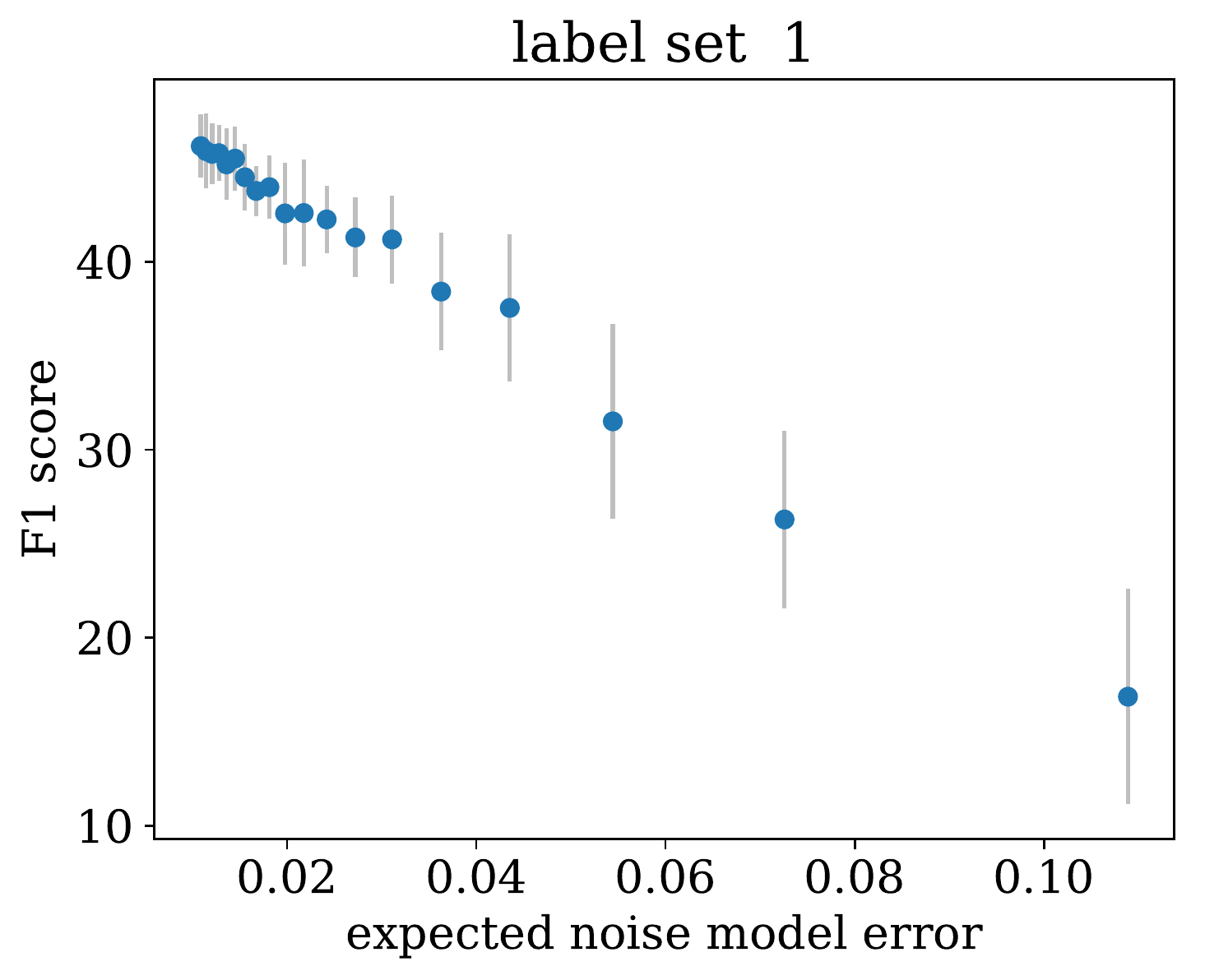}
    \includegraphics[height=3.35cm]{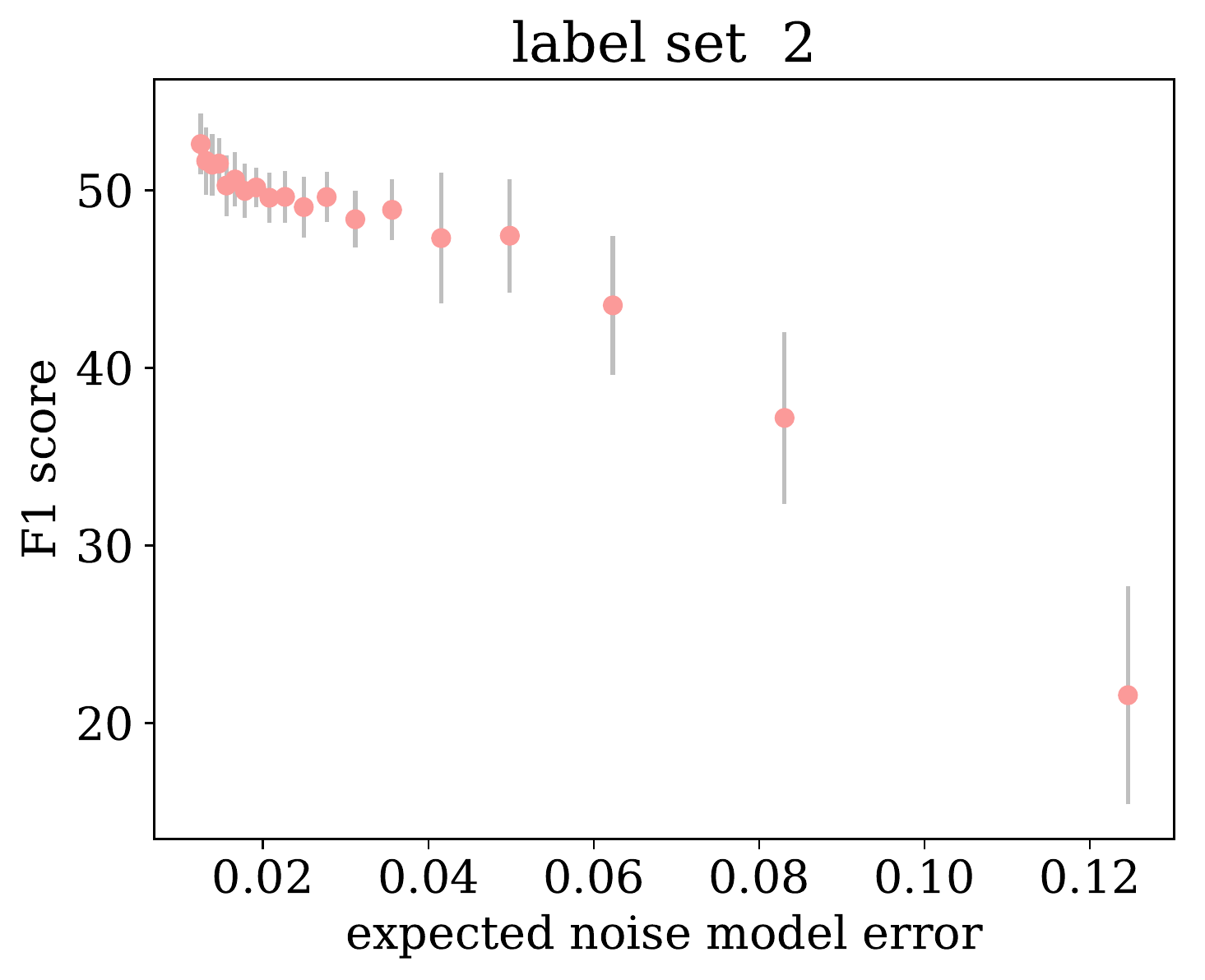}
    \includegraphics[height=3.35cm]{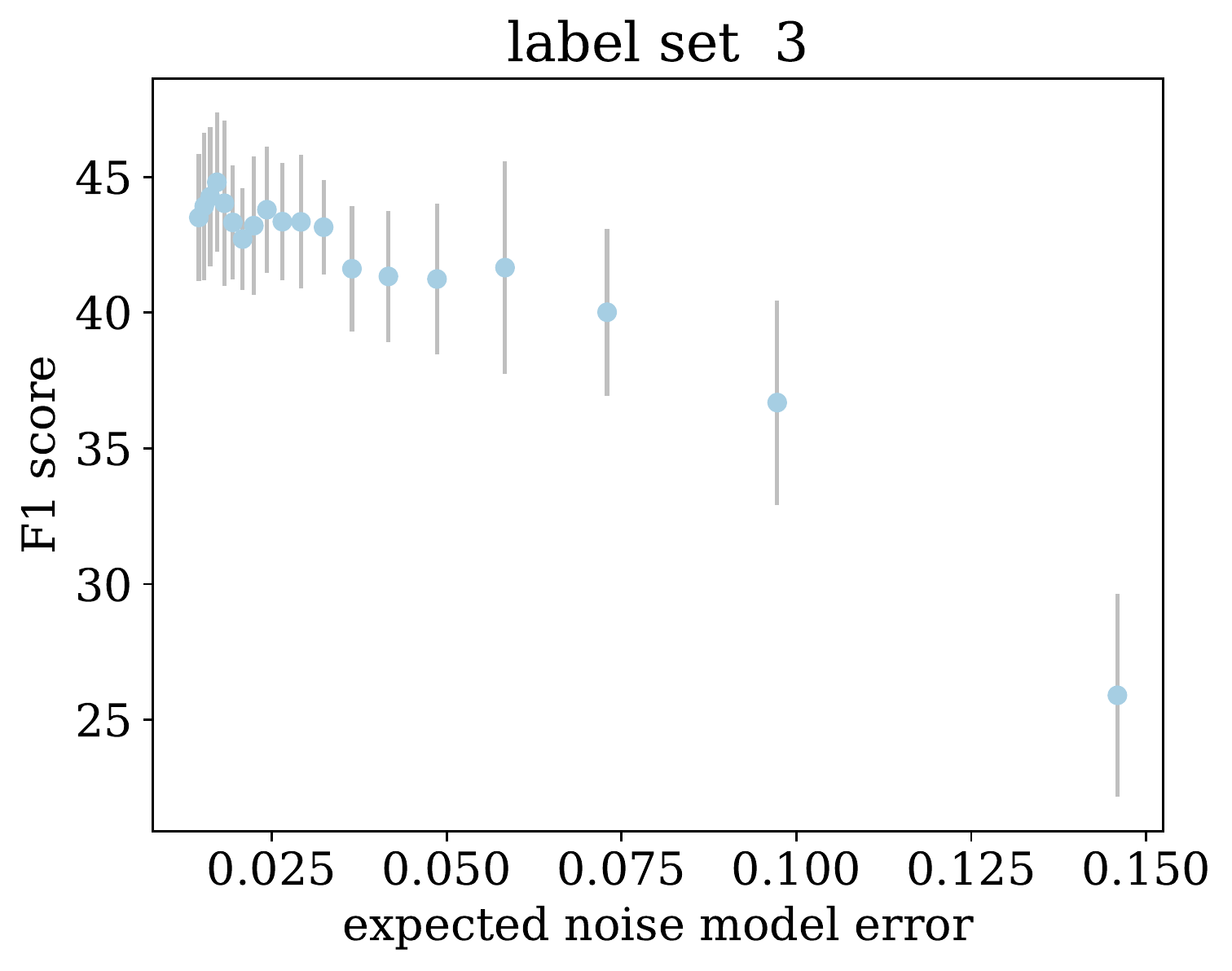}
    \includegraphics[height=3.35cm]{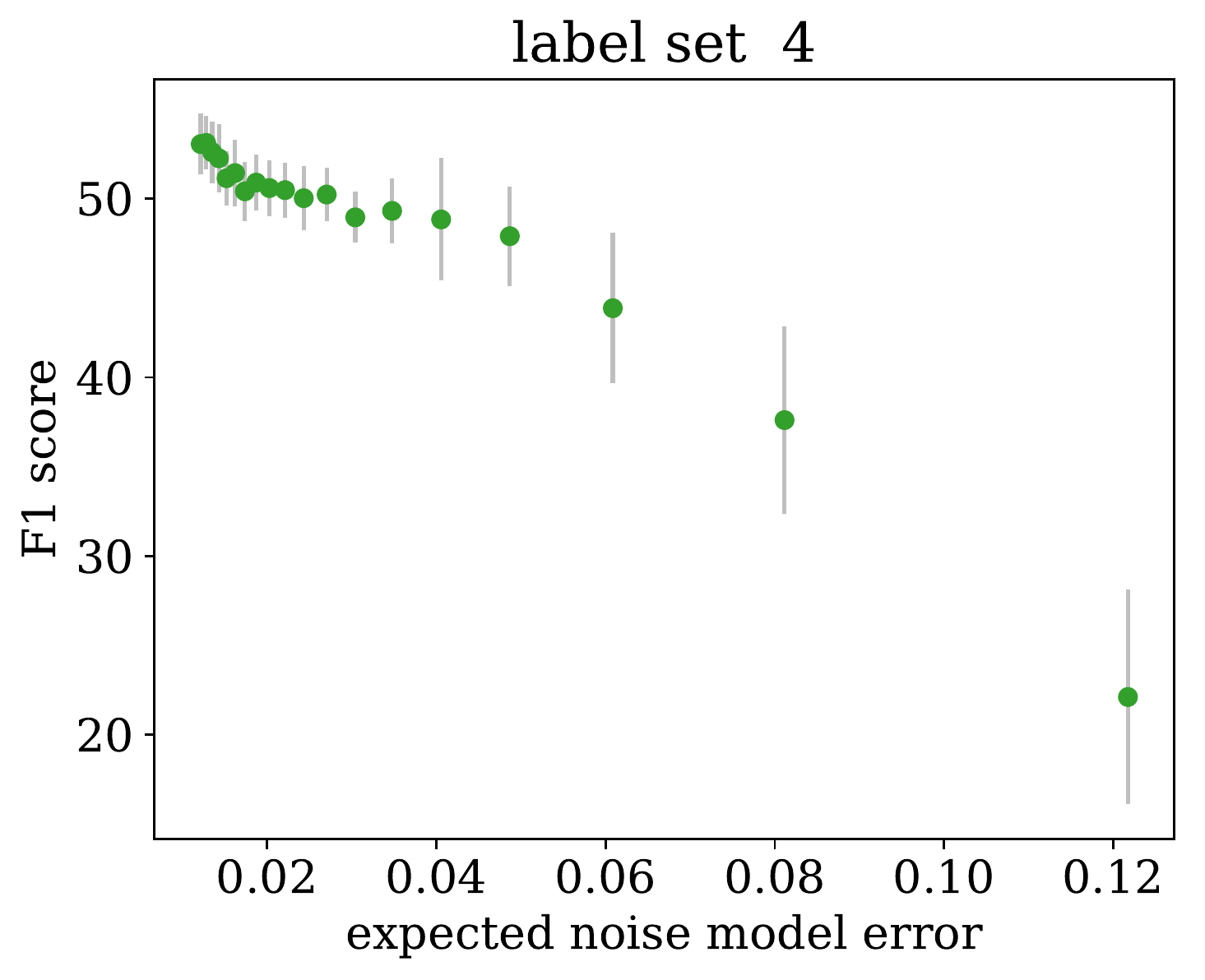}
    \includegraphics[height=3.35cm]{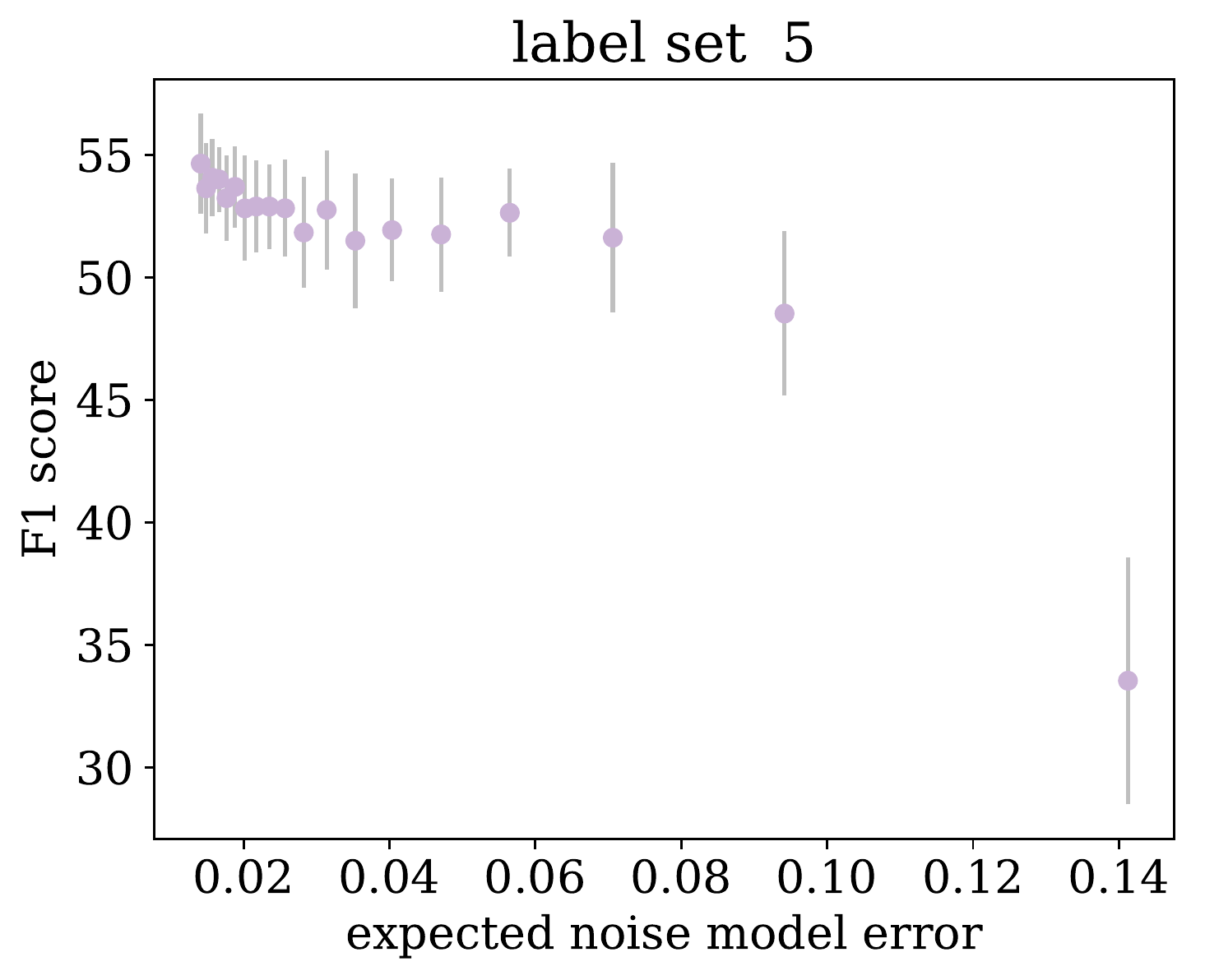}
    \includegraphics[height=3.35cm]{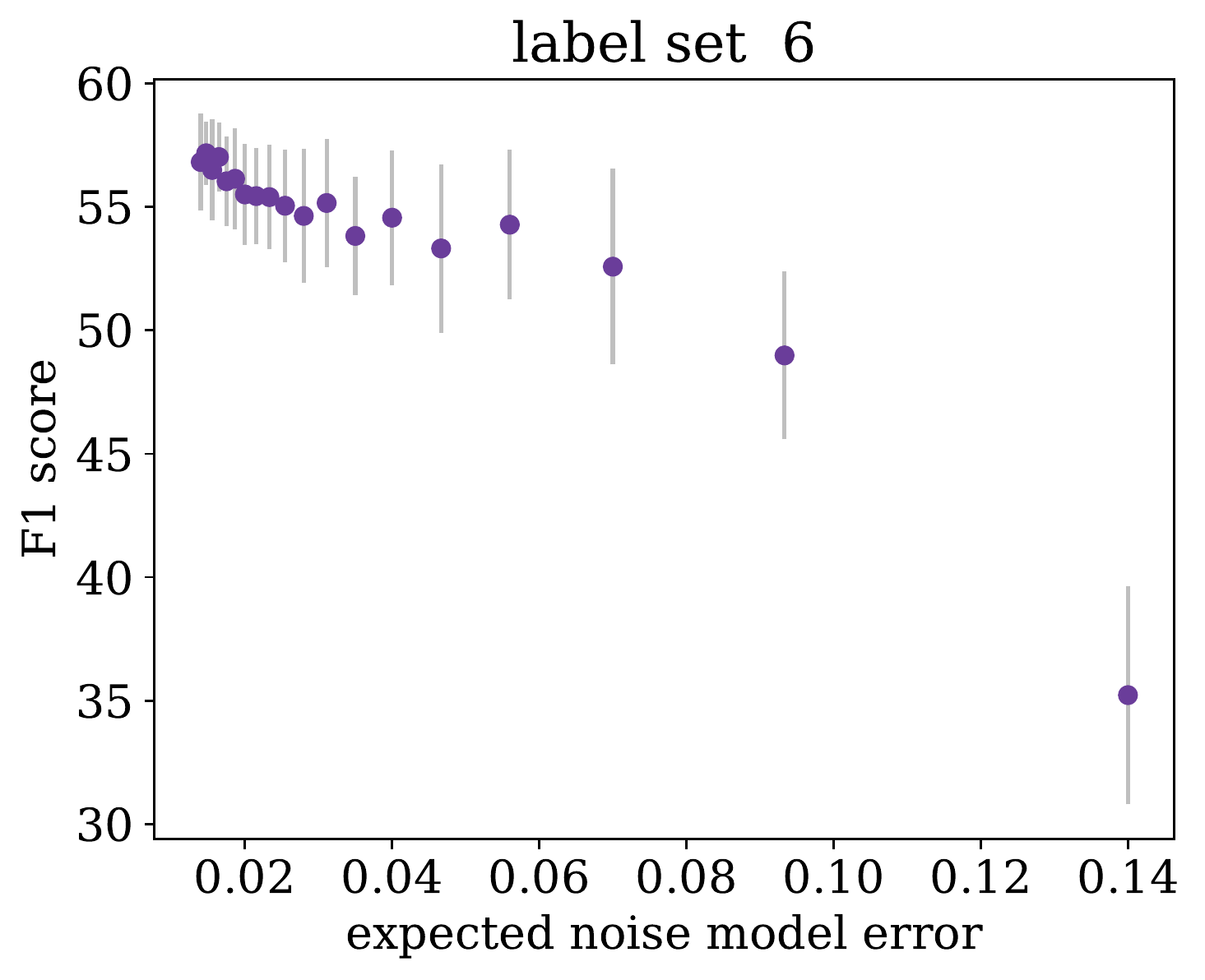}
    \includegraphics[height=3.35cm]{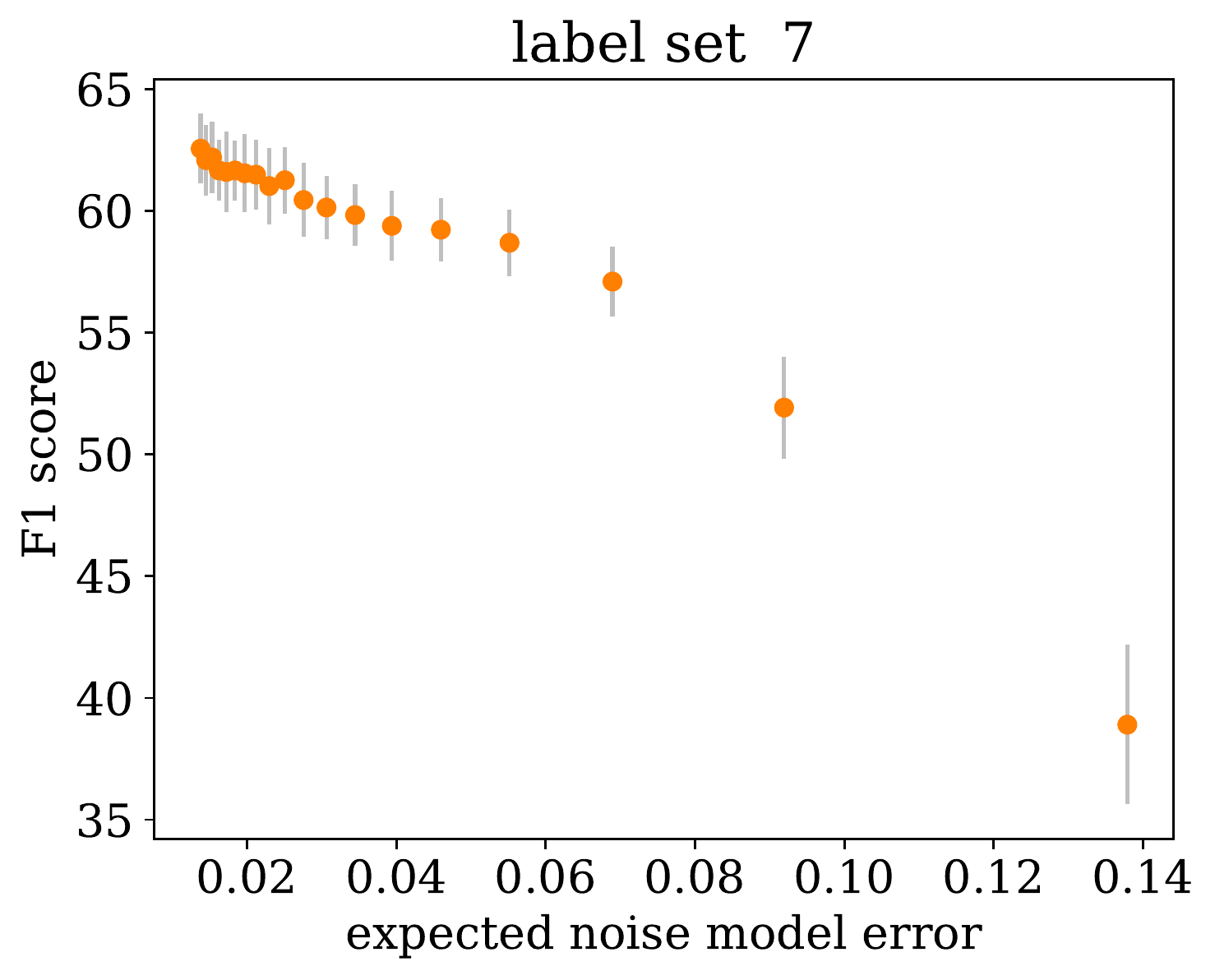}
    \caption{Relationship between the theoretically expected noise model error and the mean test performance of the base model for Clothing1M and NoisyNER \textbf{with $\mathbf{|D_C|}$ fixed} for the base model and varying for the noise model estimation and with \textbf{Fixed Sampling}. Each point corresponds to one sample size $n_i$. Grey error bars show the empirical standard deviation. }
\end{figure*}

\begin{figure*}
    \centering
    \includegraphics[height=3.5cm]{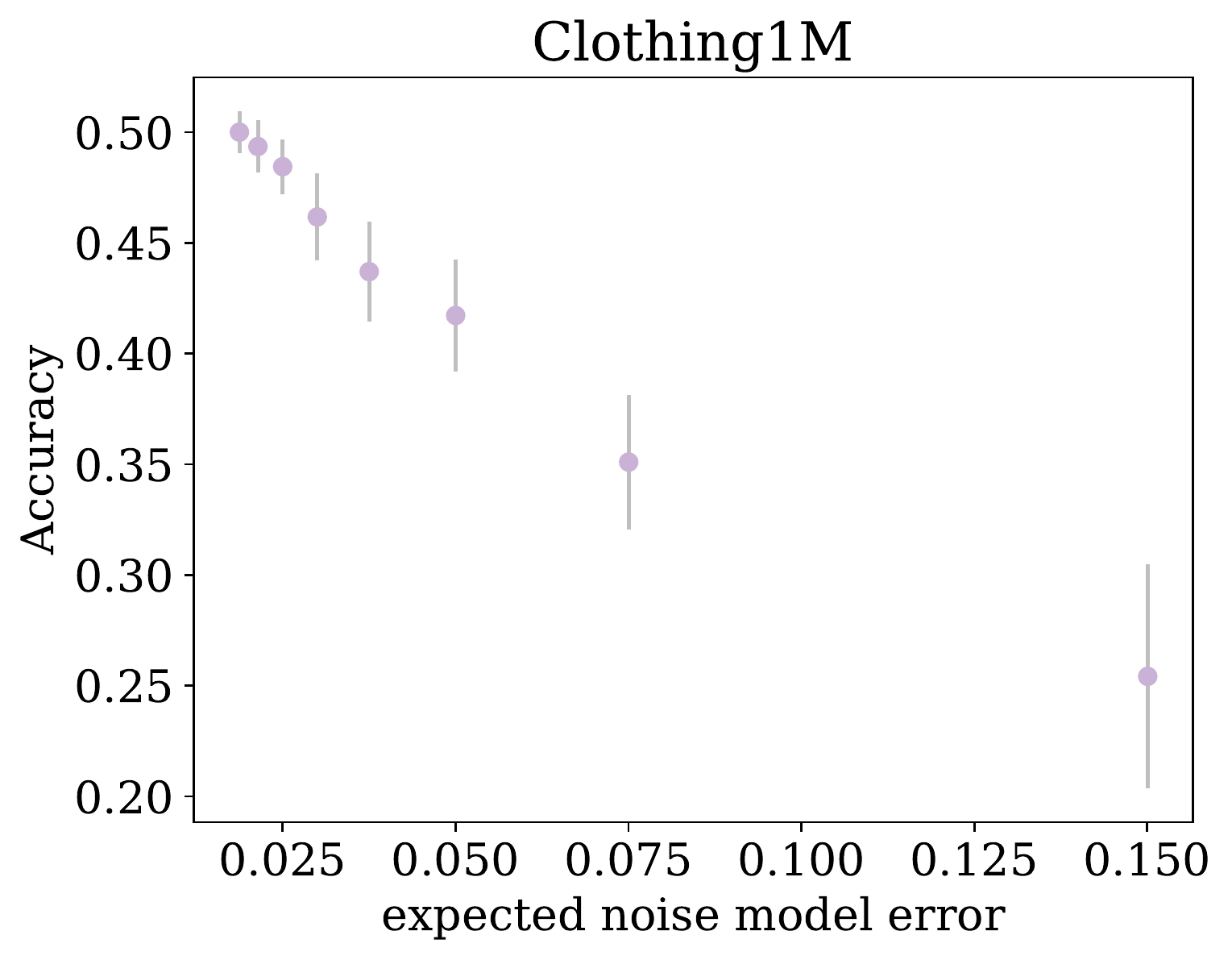}
    \includegraphics[height=3.5cm]{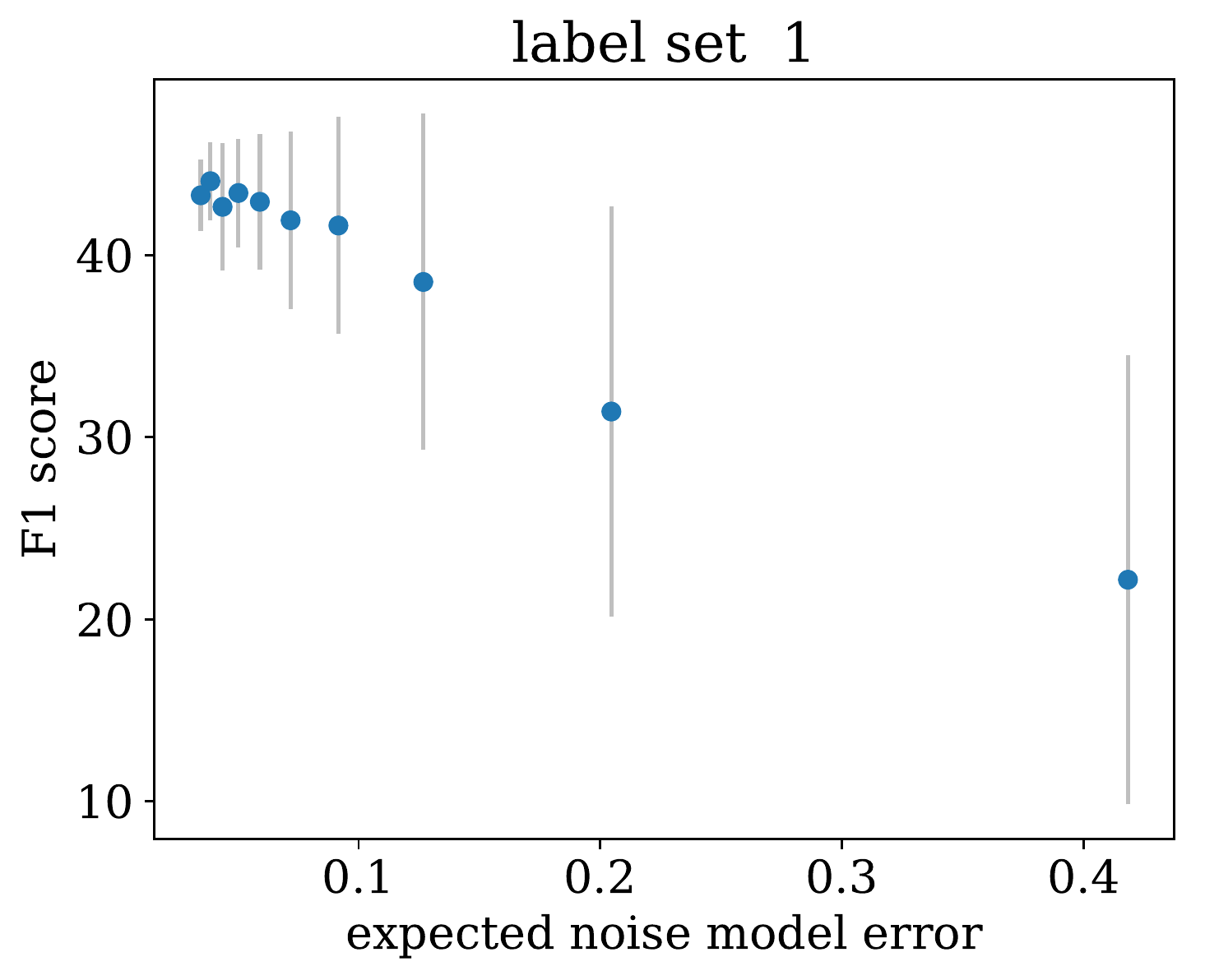}
    \includegraphics[height=3.5cm]{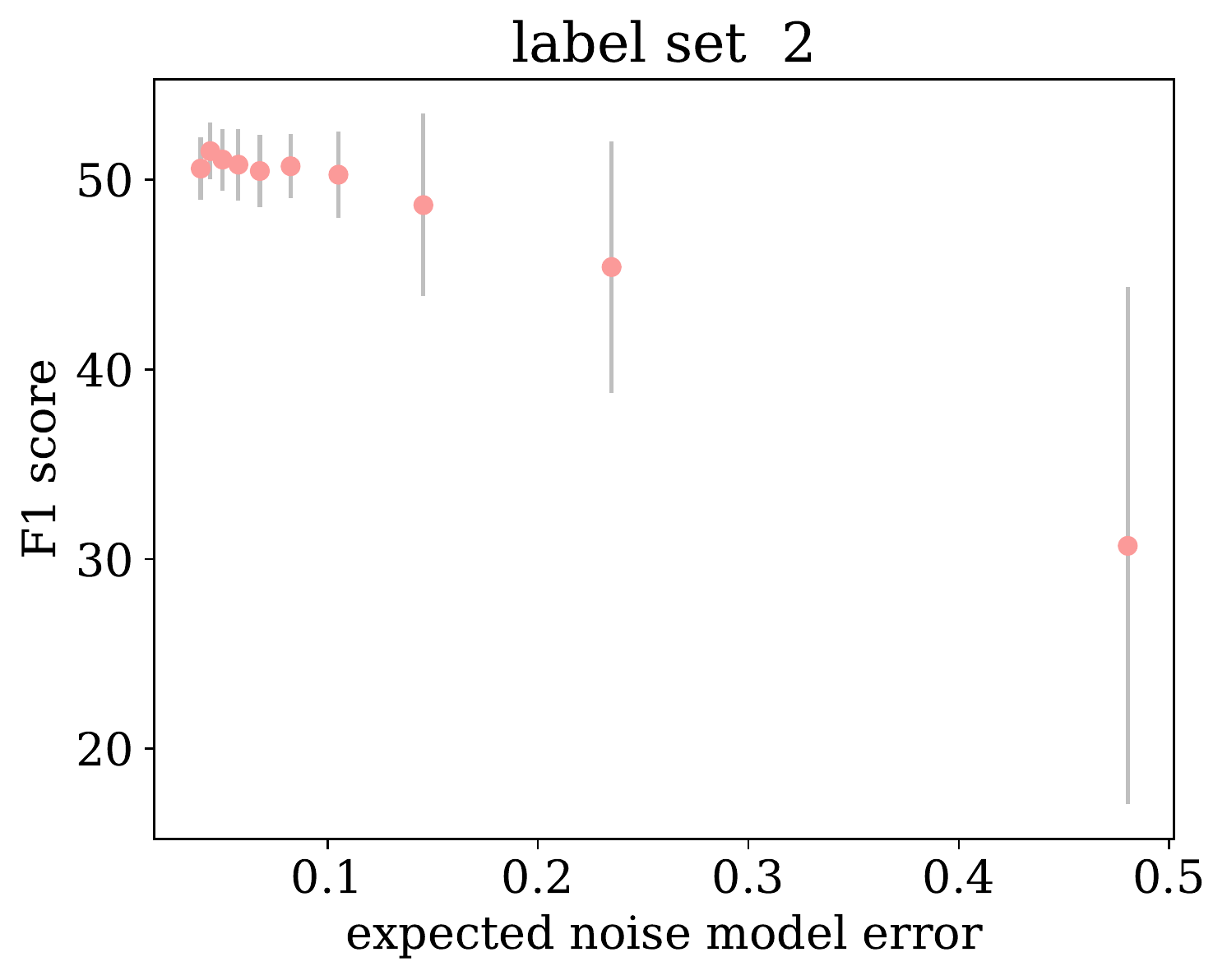}
    \includegraphics[height=3.5cm]{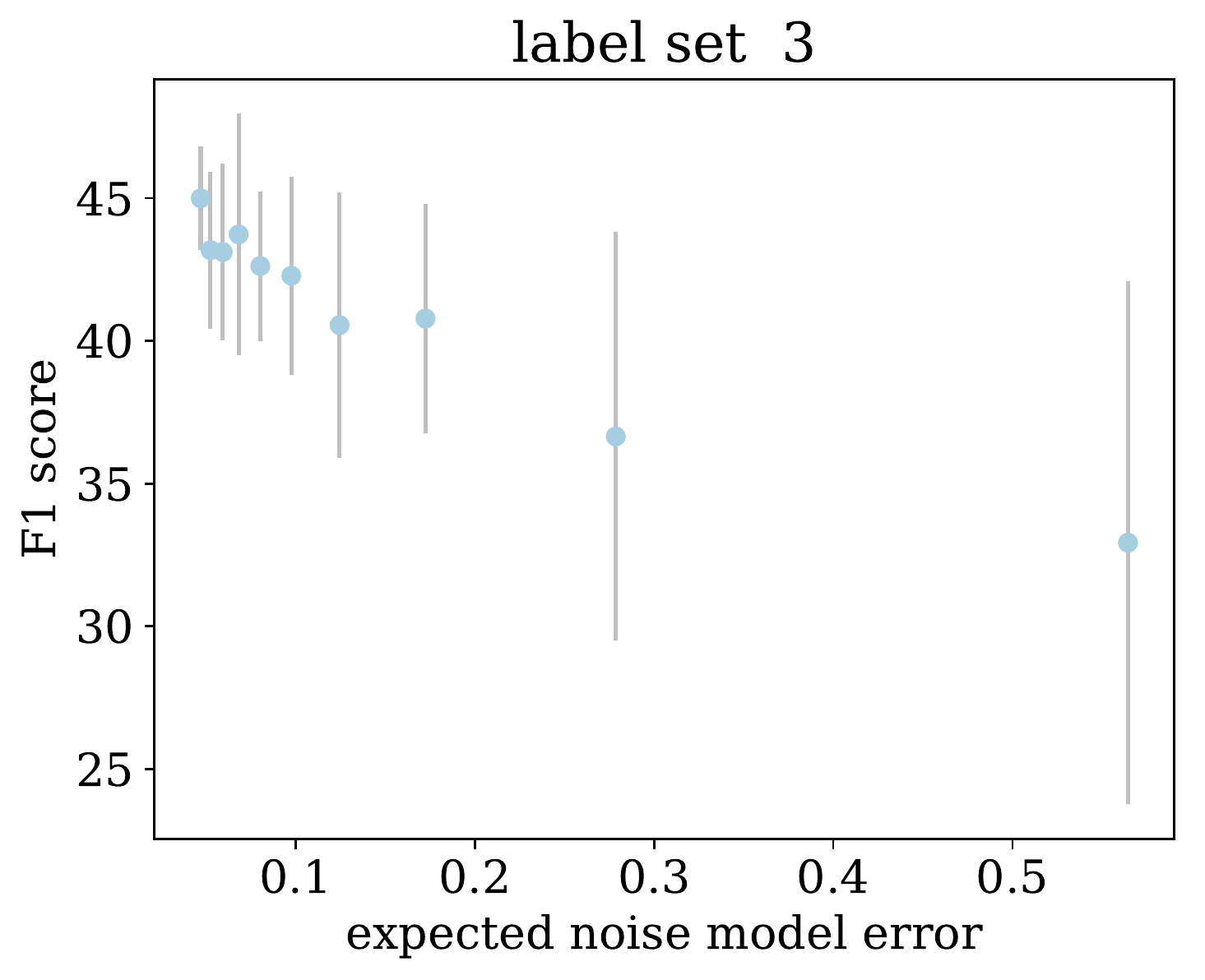}
    \includegraphics[height=3.5cm]{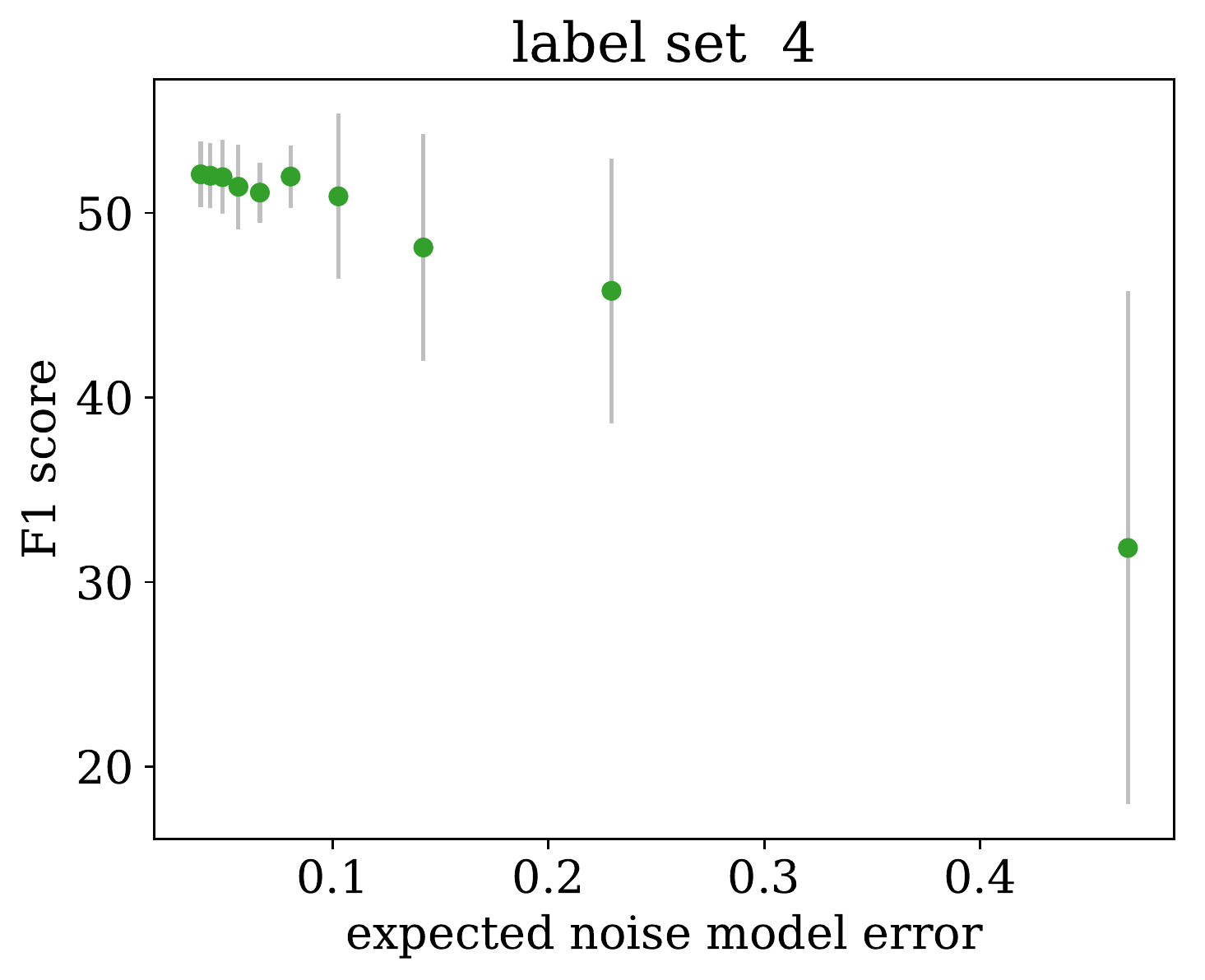}
    \includegraphics[height=3.5cm]{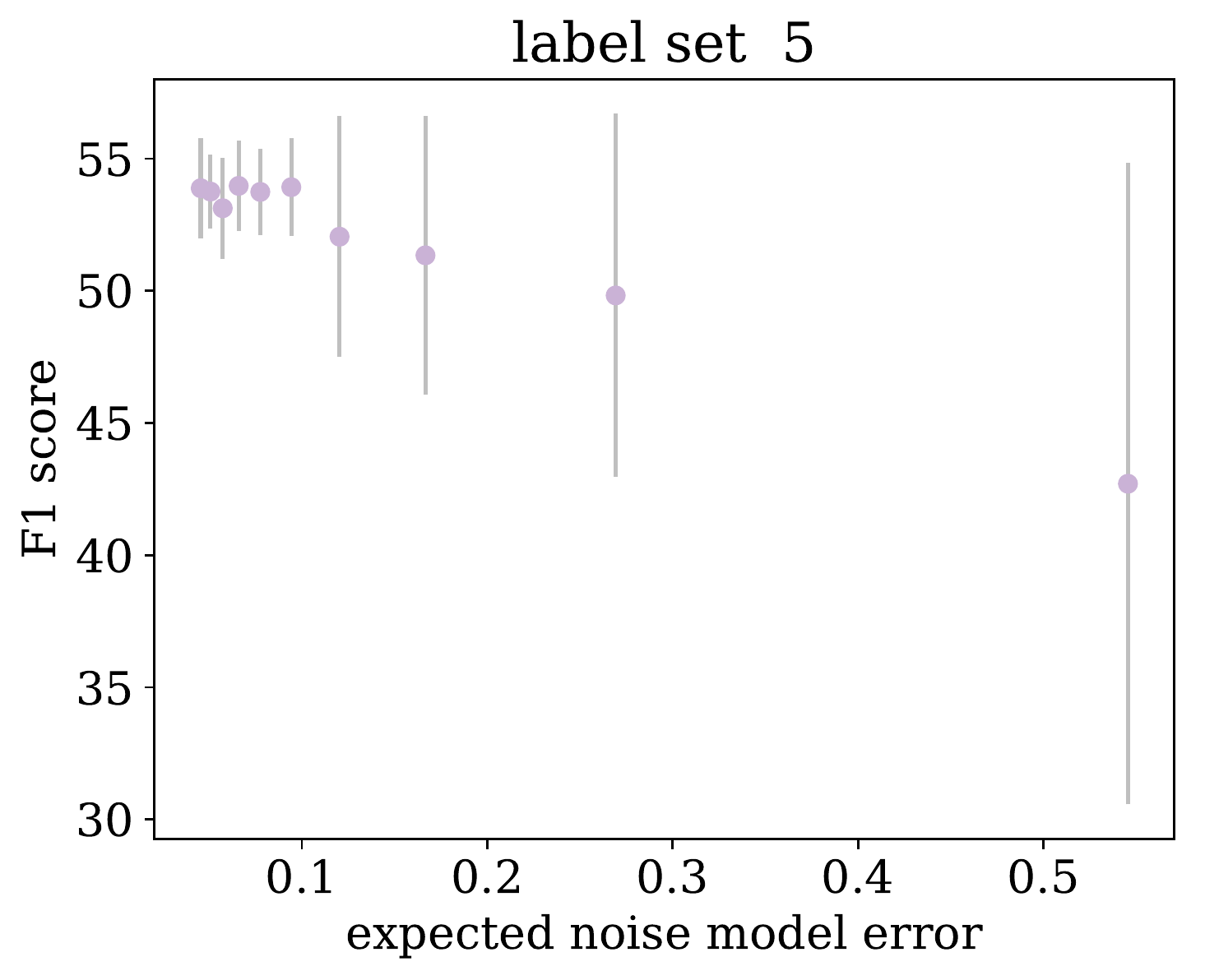}
    \includegraphics[height=3.5cm]{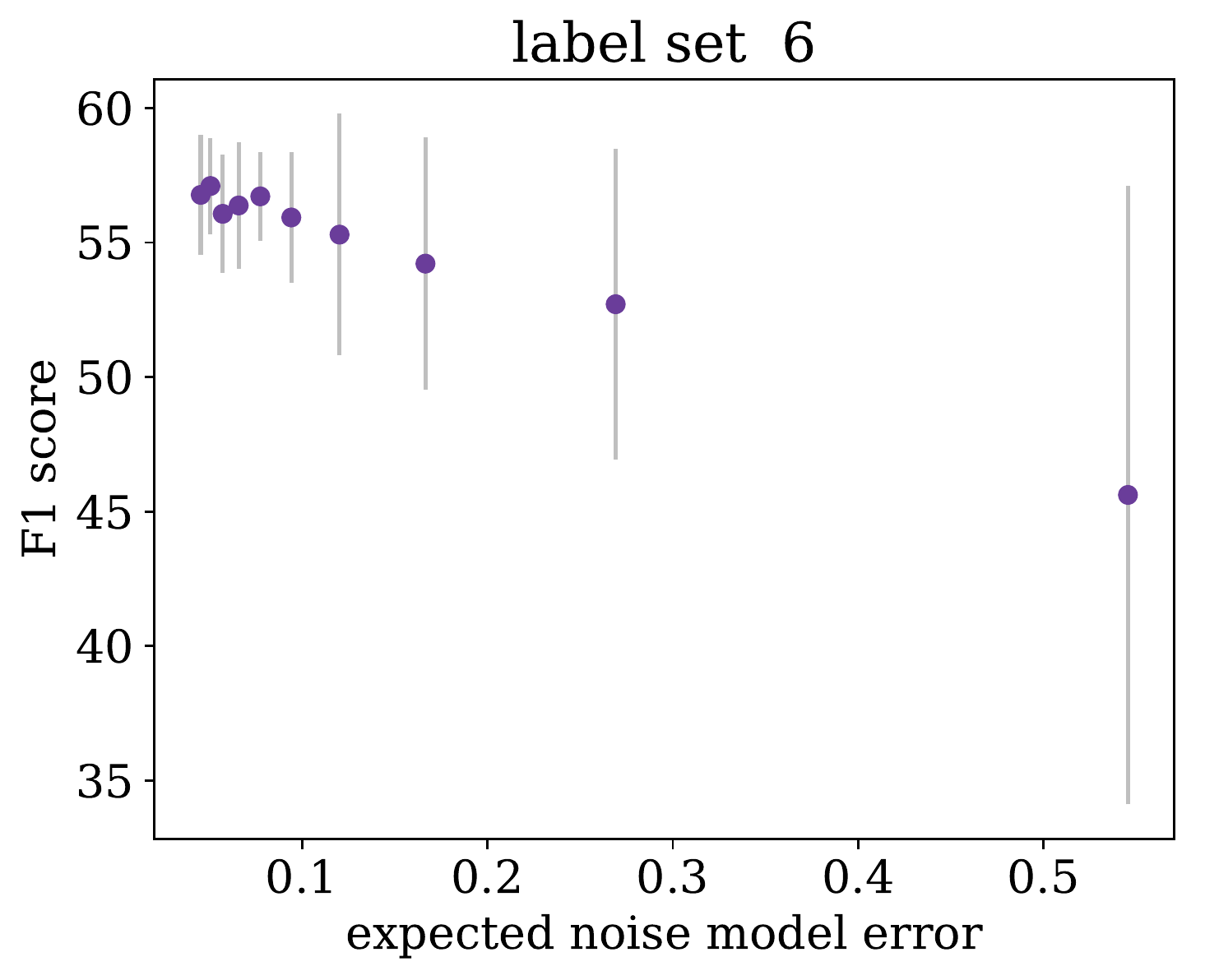}
    \includegraphics[height=3.5cm]{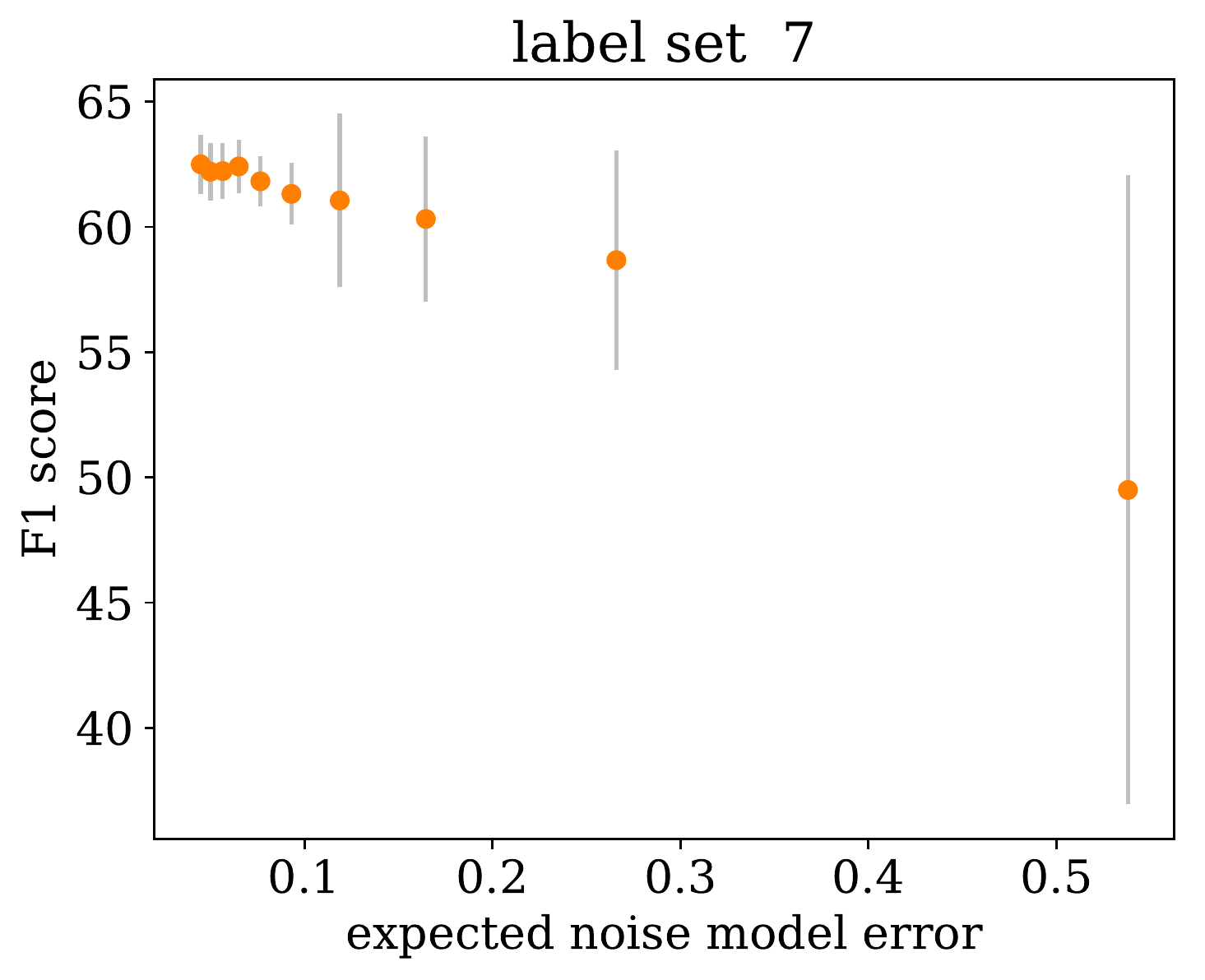}
    \caption{Relationship between the theoretically expected noise model error and the test performance (Accuracy/F1 score) of the base model for Clothing1M and NoisyNER  \textbf{with increasing $\mathbf{|D_C|}$} and with \textbf{Variable Sampling}. Each point corresponds to one sample size $n$.  Grey error bars show the empirical standard deviation.}
\end{figure*}

\begin{figure*}
    \centering
    \includegraphics[height=3.5cm]{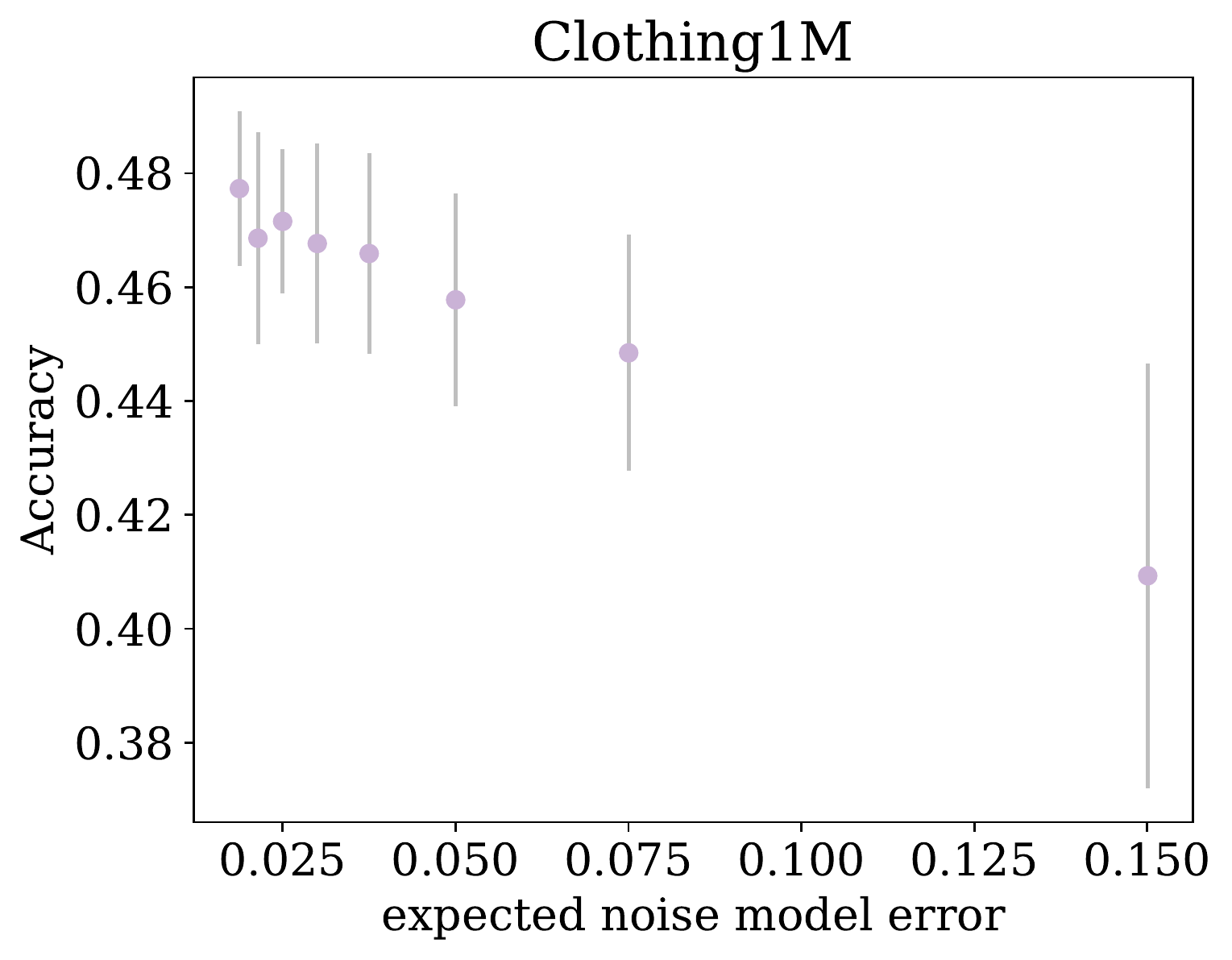}
    \includegraphics[height=3.5cm]{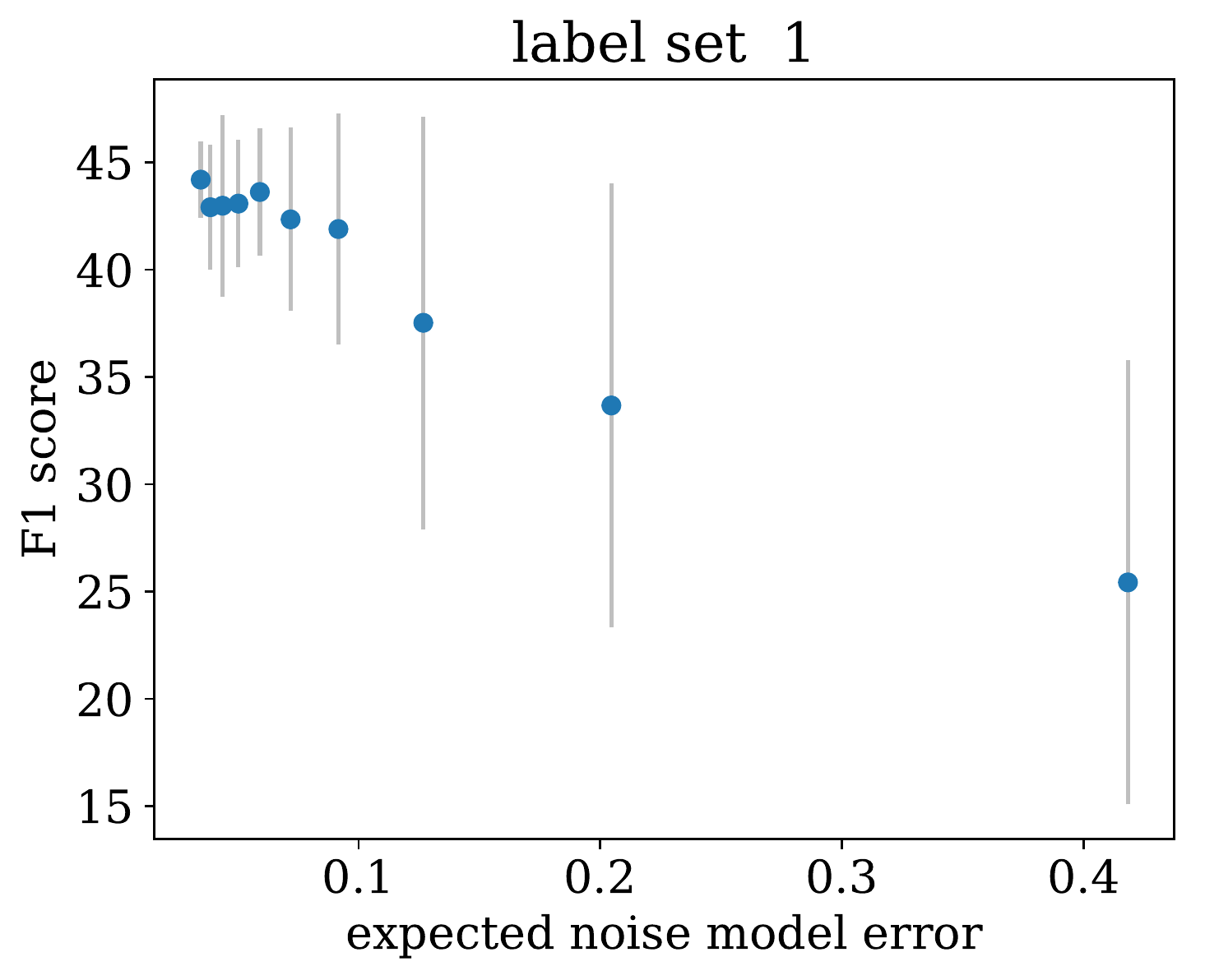}
    \includegraphics[height=3.5cm]{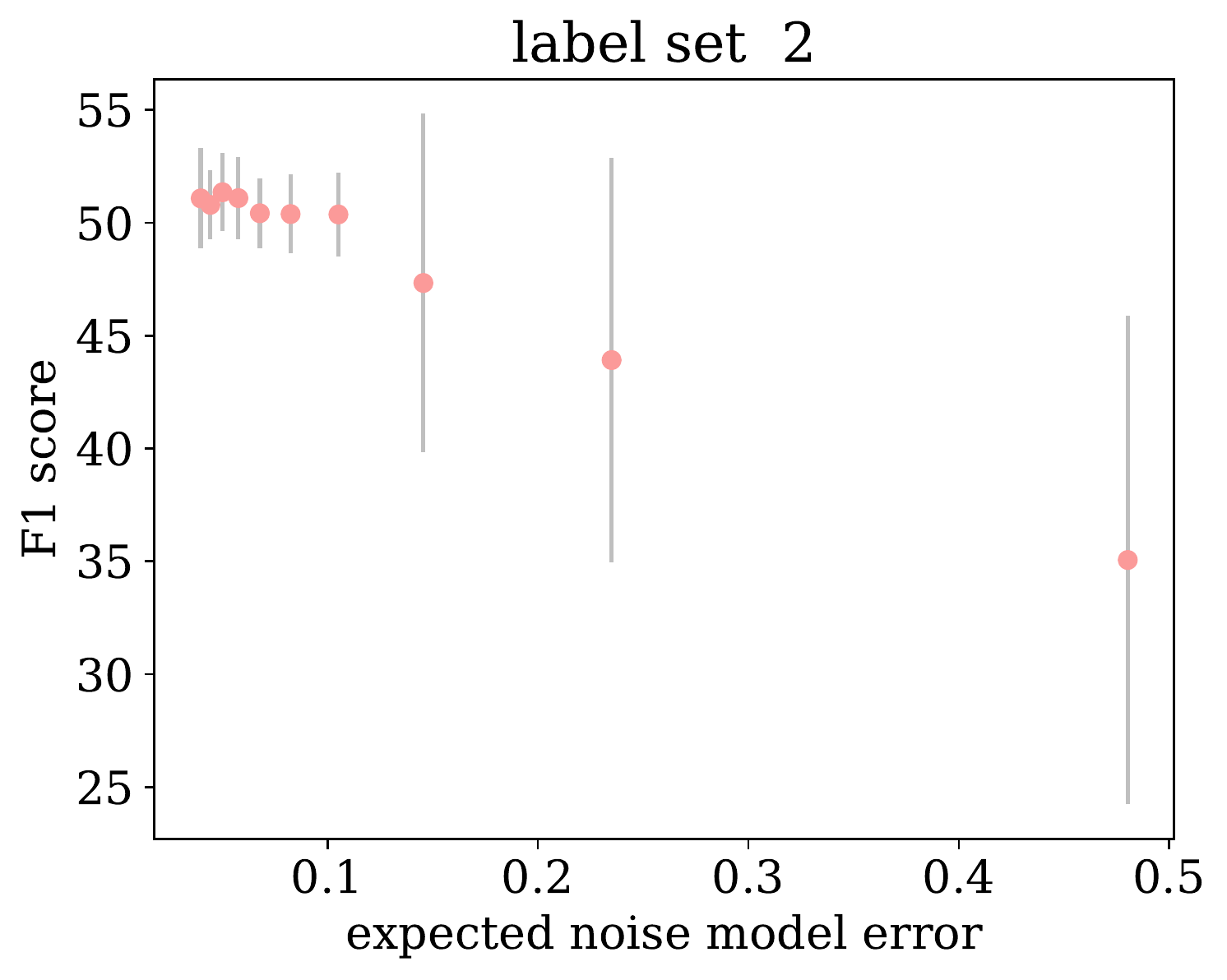}
    \includegraphics[height=3.5cm]{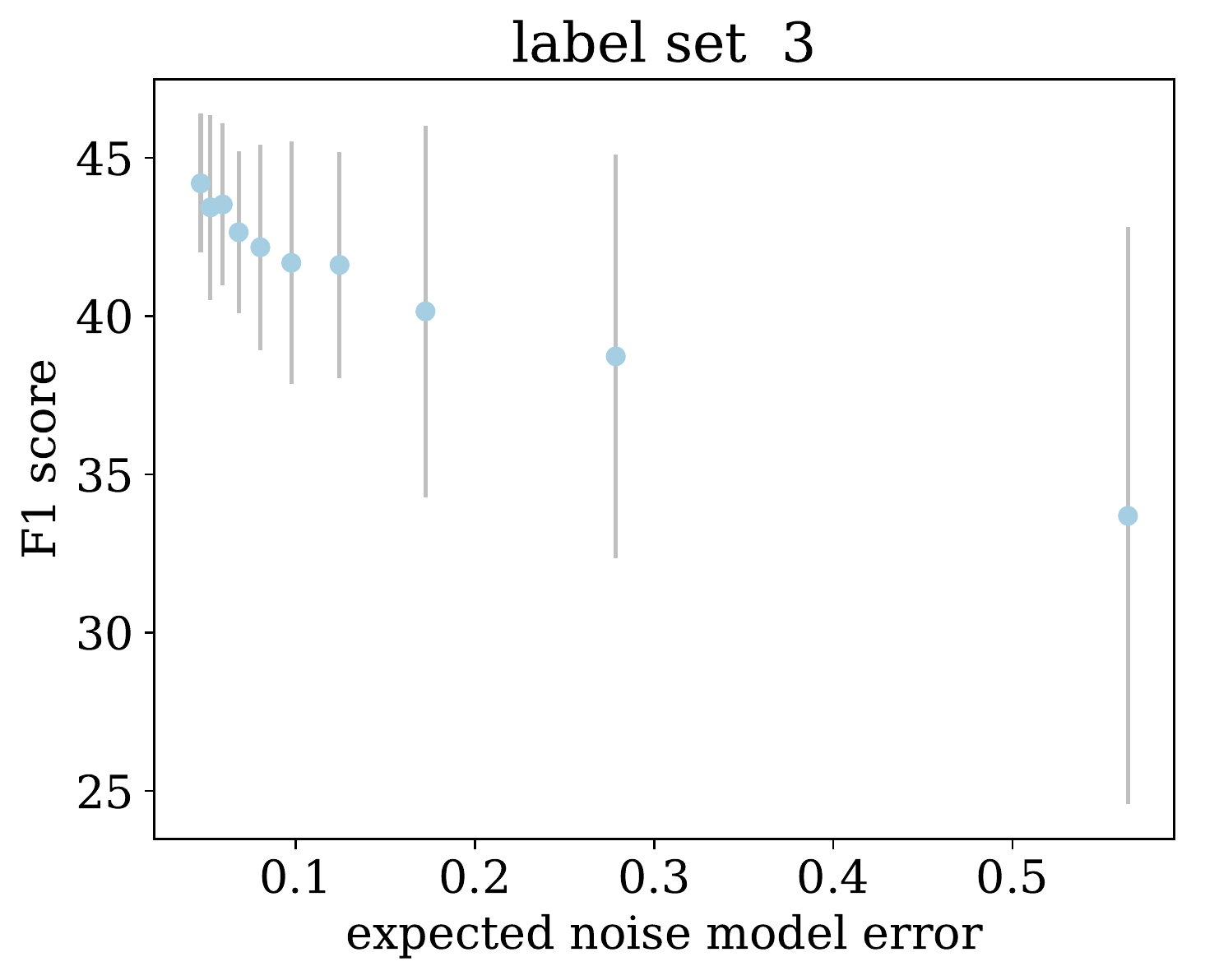}
    \includegraphics[height=3.5cm]{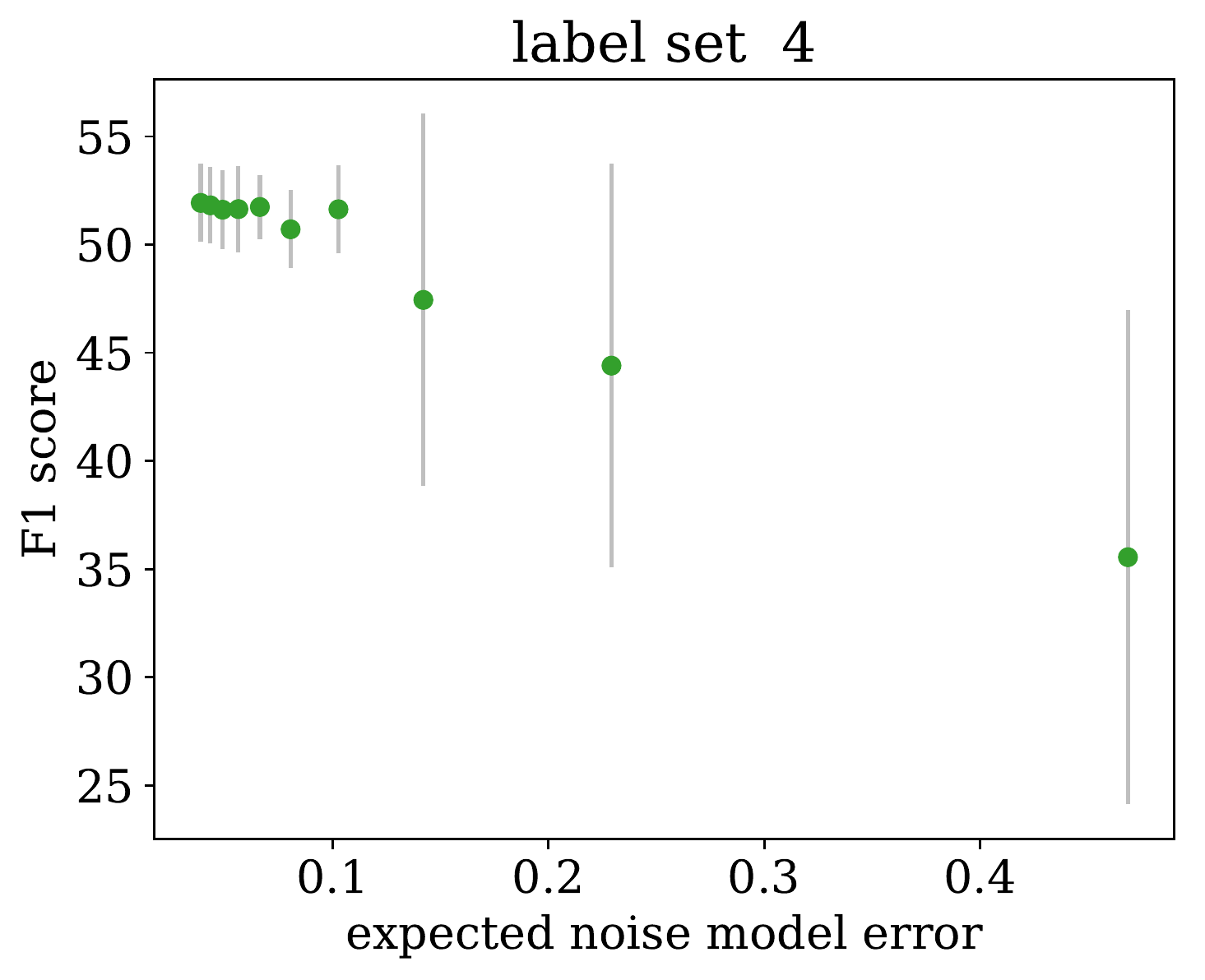}
    \includegraphics[height=3.5cm]{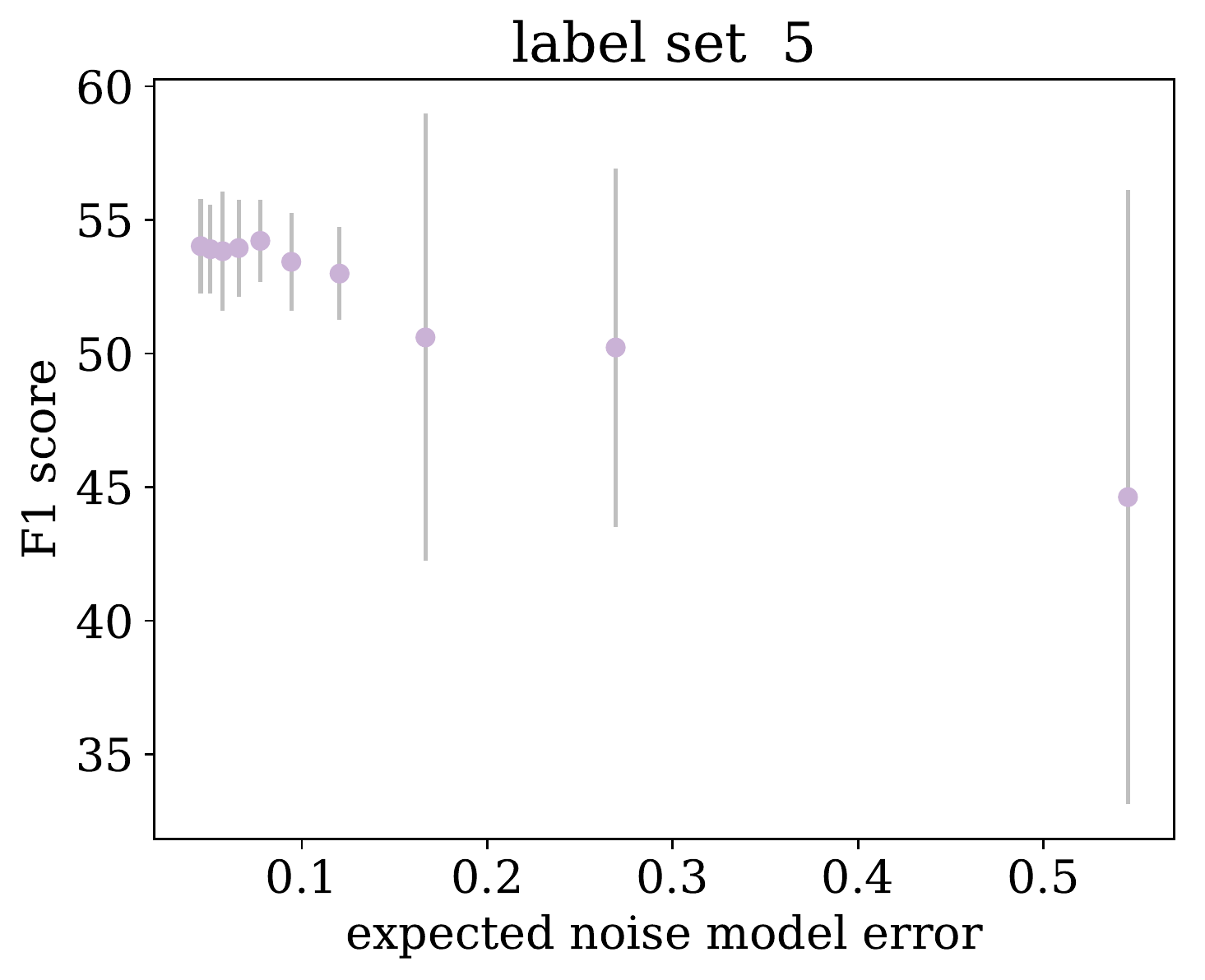}
    \includegraphics[height=3.5cm]{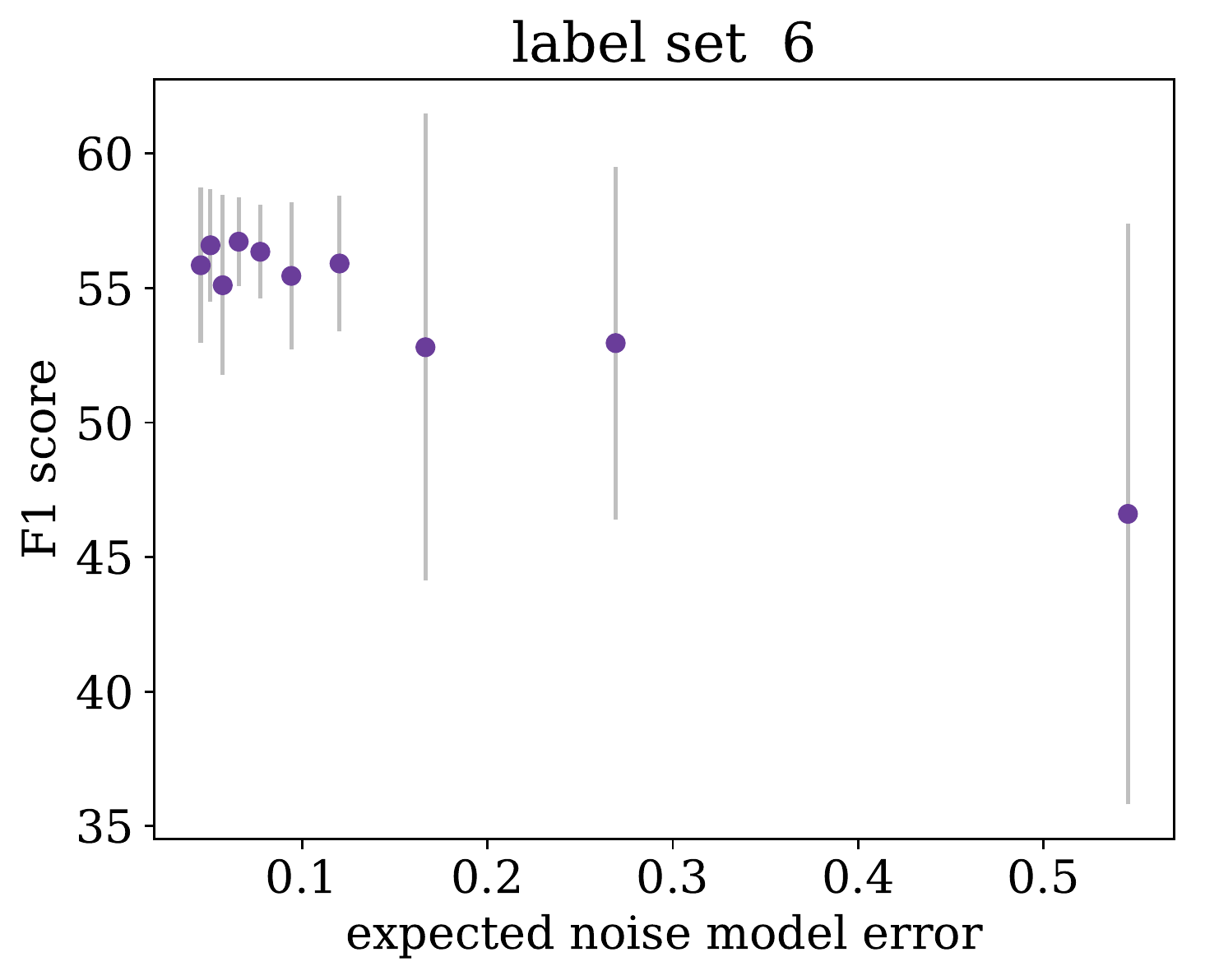}
    \includegraphics[height=3.5cm]{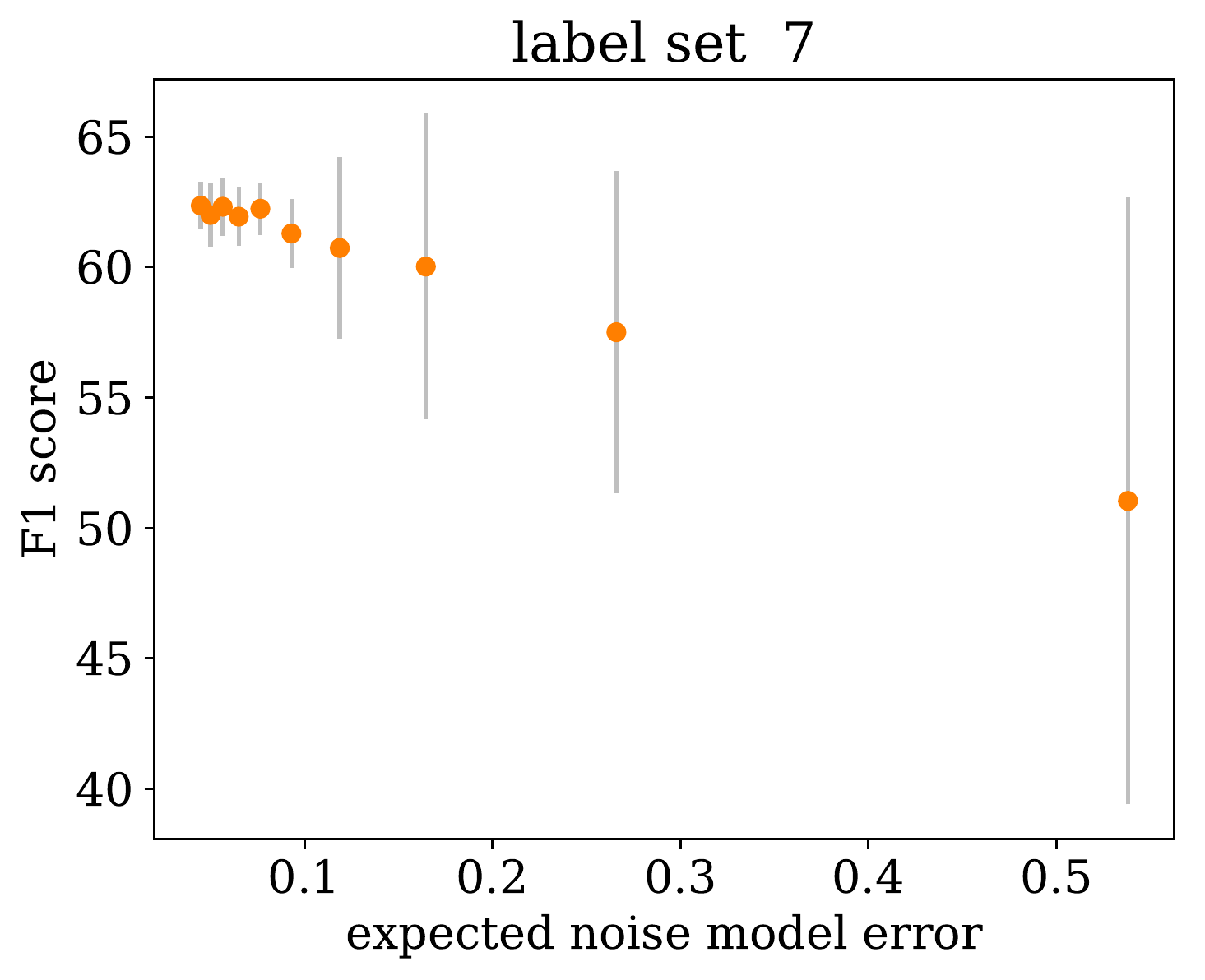}
    \caption{Relationship between the theoretically expected noise model error and the mean test performance of the base model for Clothing1M and NoisyNER \textbf{with $\mathbf{|D_C|}$ fixed} and with \textbf{Variable Sampling}. Each point corresponds to one sample size $n$. Grey error bars show the empirical standard deviation.}
\end{figure*}

\begin{figure}
    \centering
    \includegraphics[height=3.35cm]{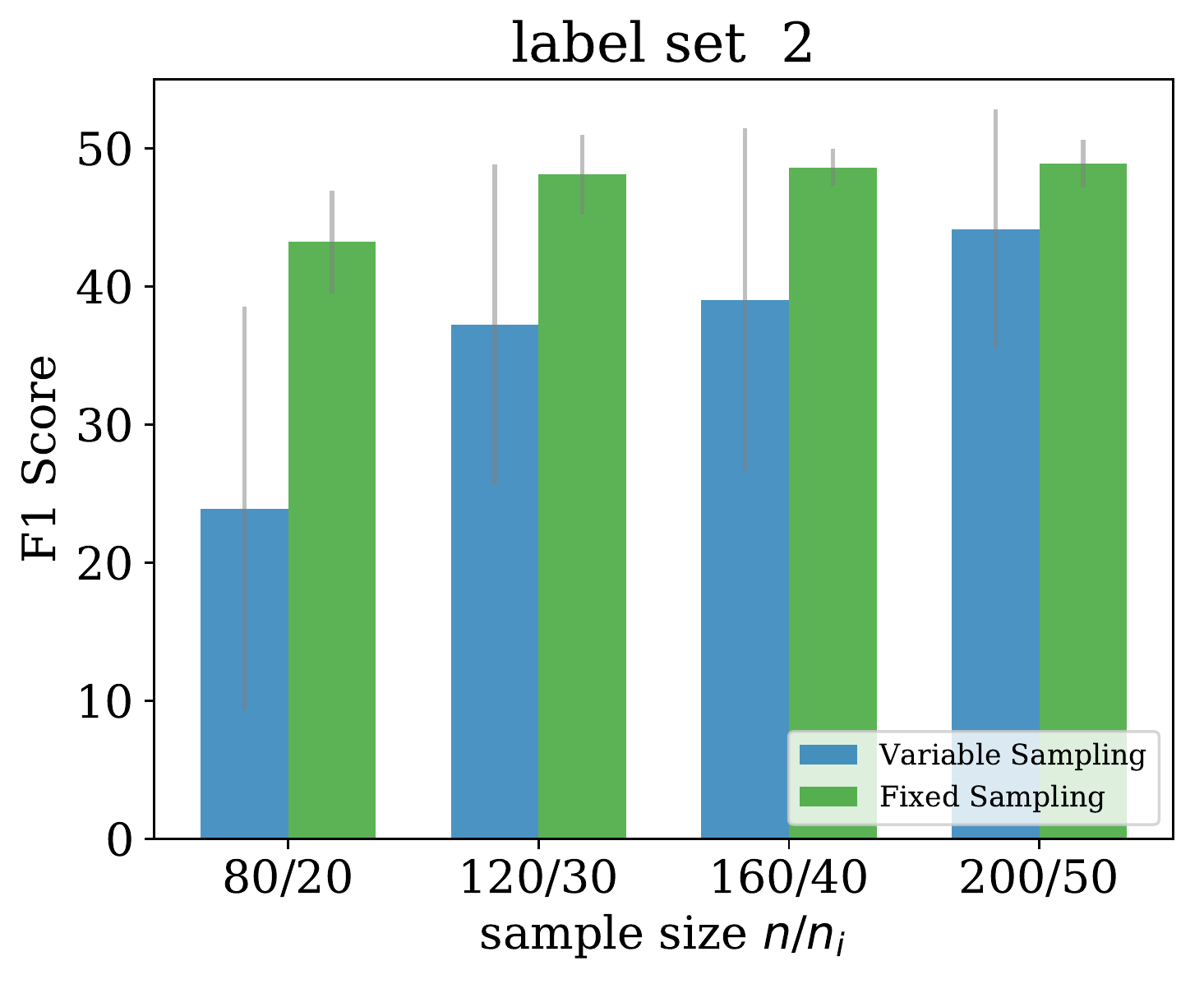}
    \includegraphics[height=3.35cm]{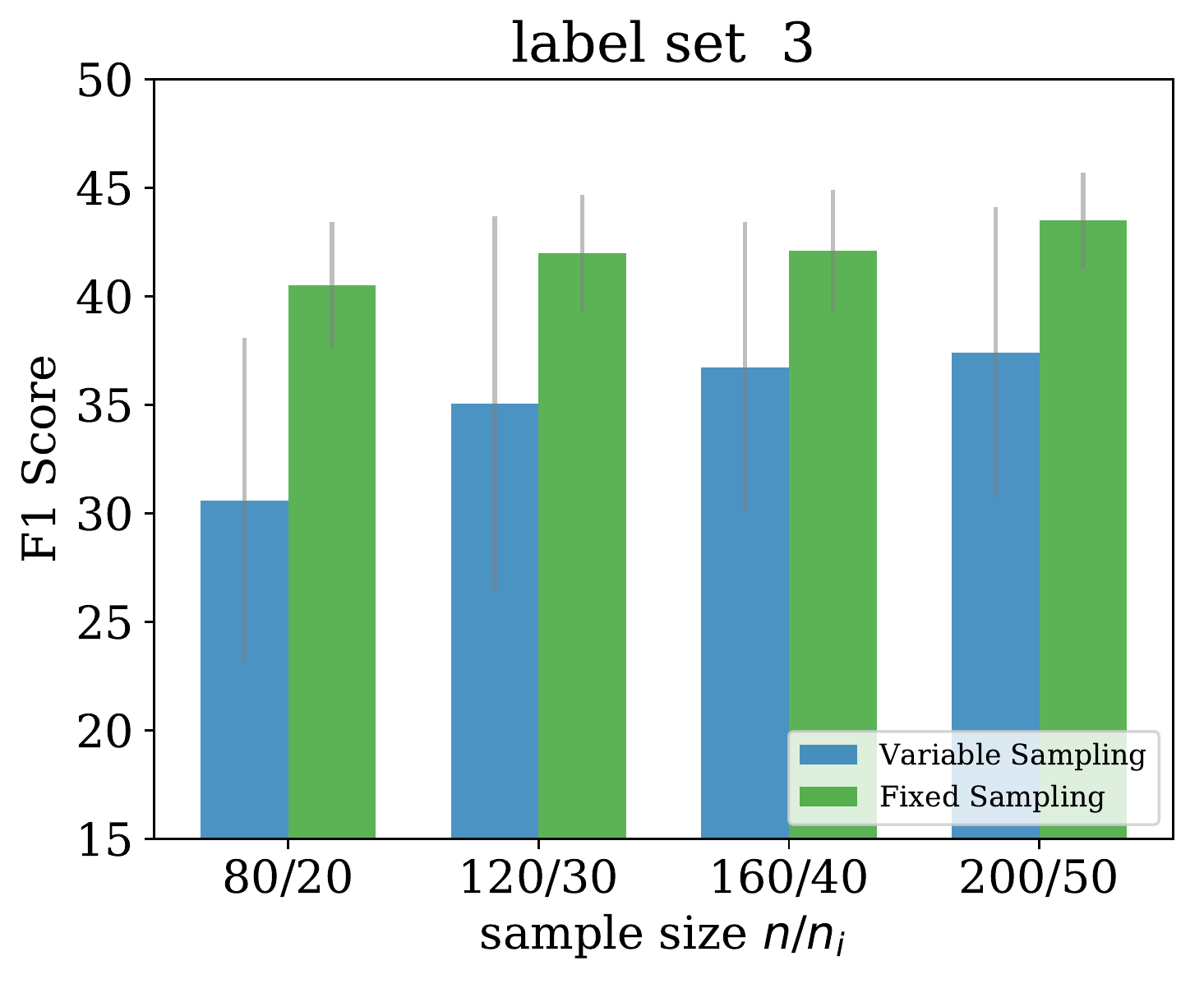}
    \includegraphics[height=3.35cm]{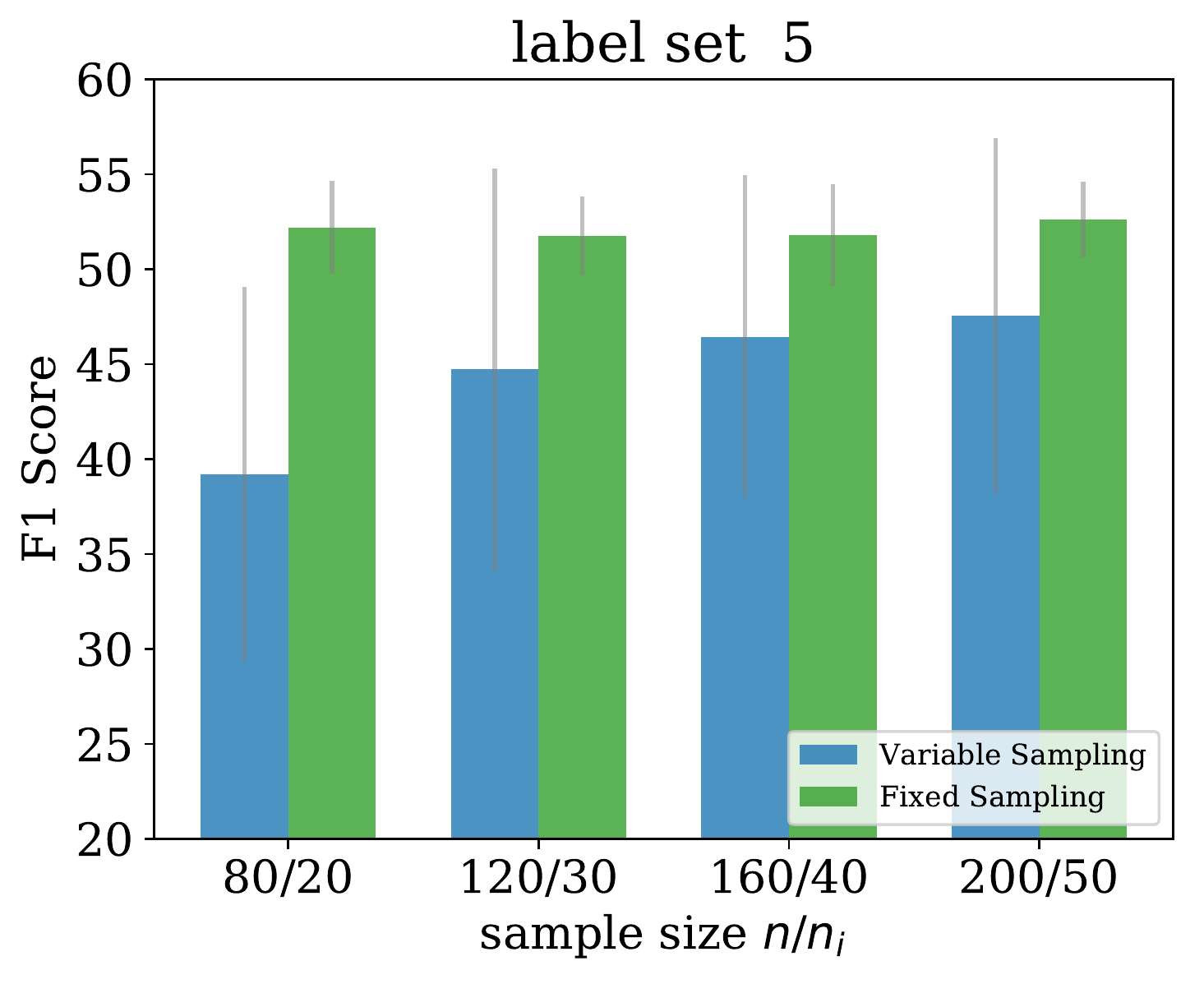}
    \includegraphics[height=3.35cm]{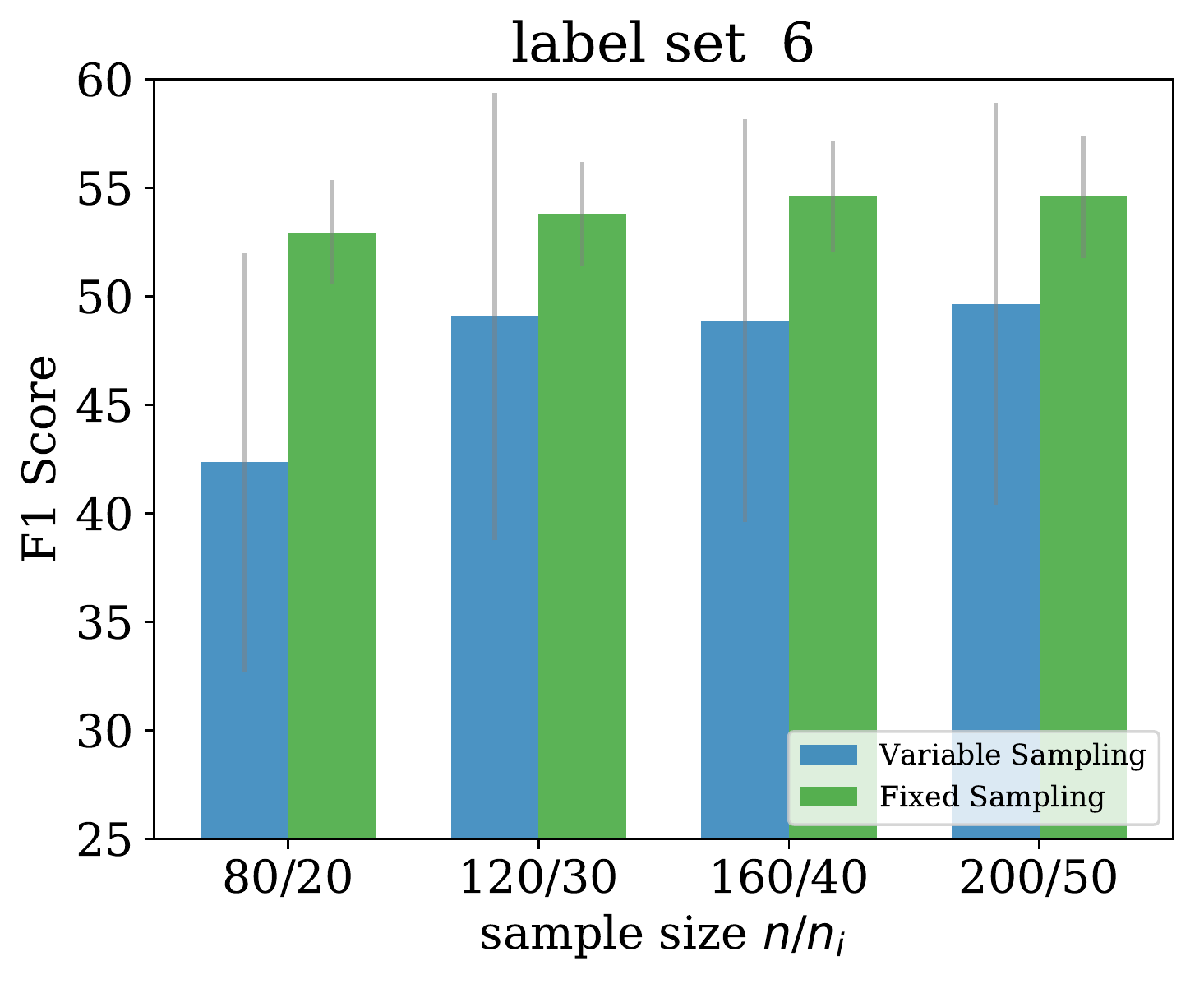}
    \caption{Comparing Variable and Fixed Sampling by mean test performance (F1 score) of the base model on NoisyNER label set 2, 3, 5 and 6 \textbf{with increasing $\mathbf{|D_C|}$} for the base model and varying for the noise model estimation. Error bars show the empirical standard deviation.}
\end{figure}

\begin{figure}
    \centering
    \includegraphics[height=3.35cm]{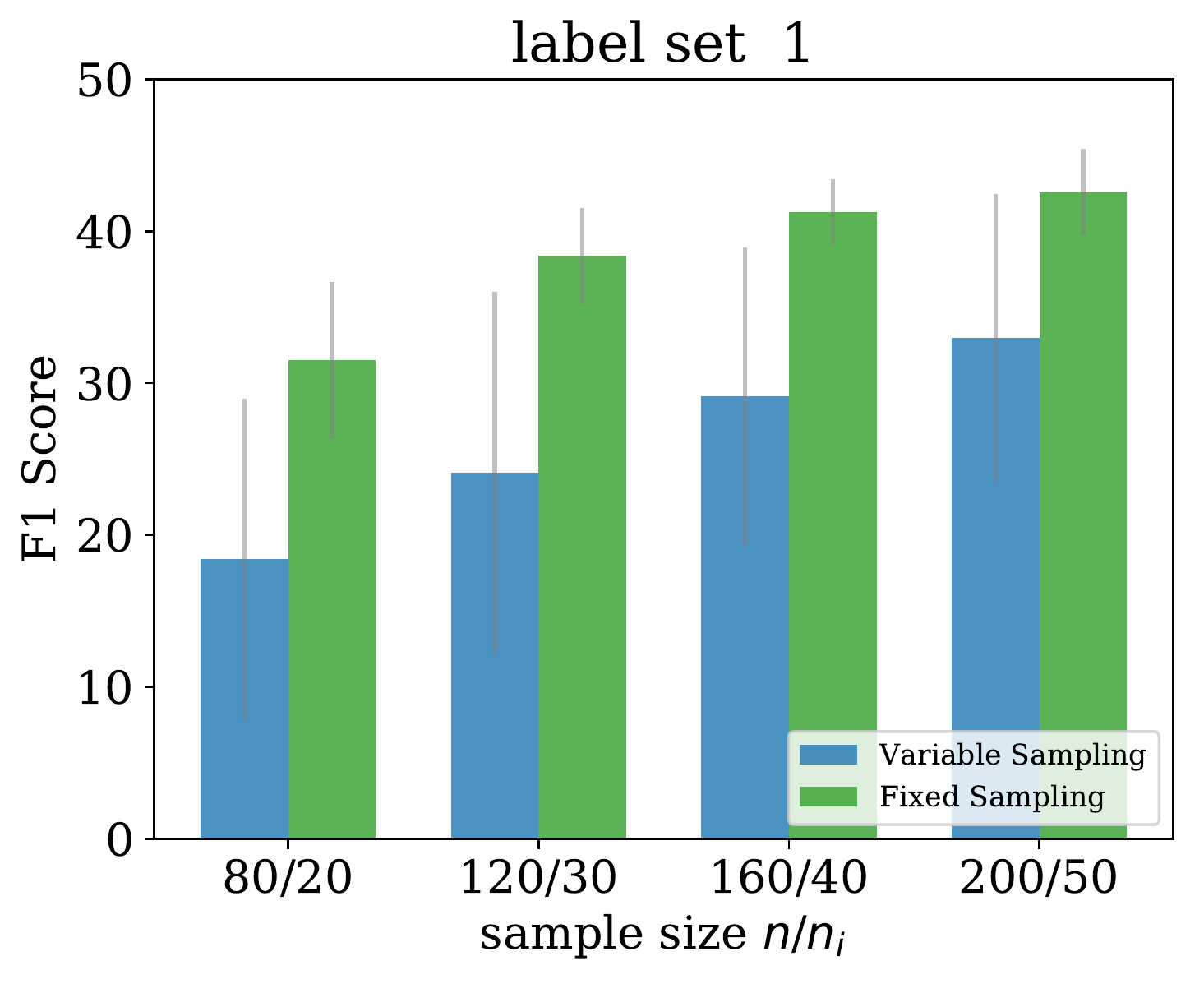}
    \includegraphics[height=3.35cm]{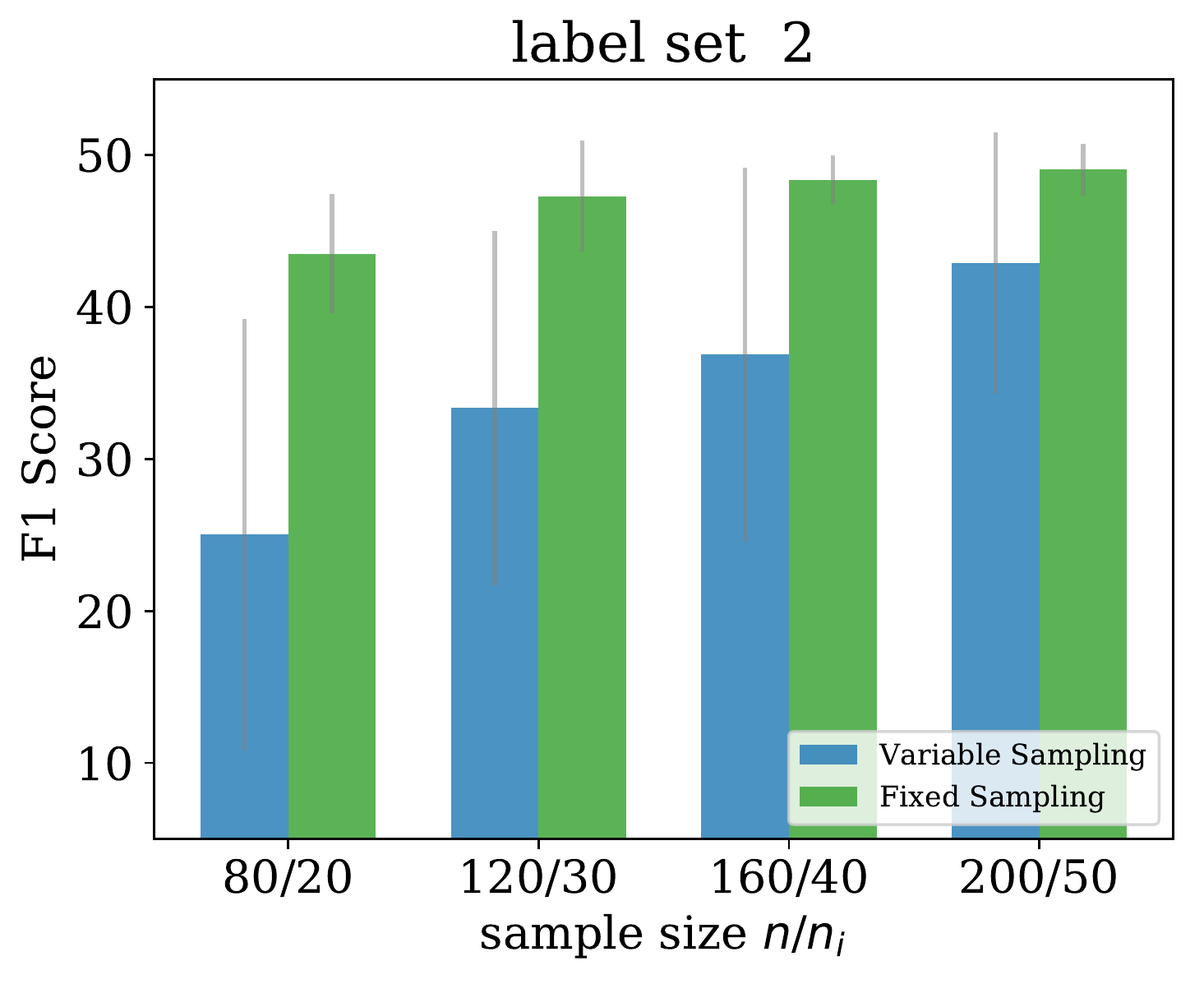}
    \includegraphics[height=3.35cm]{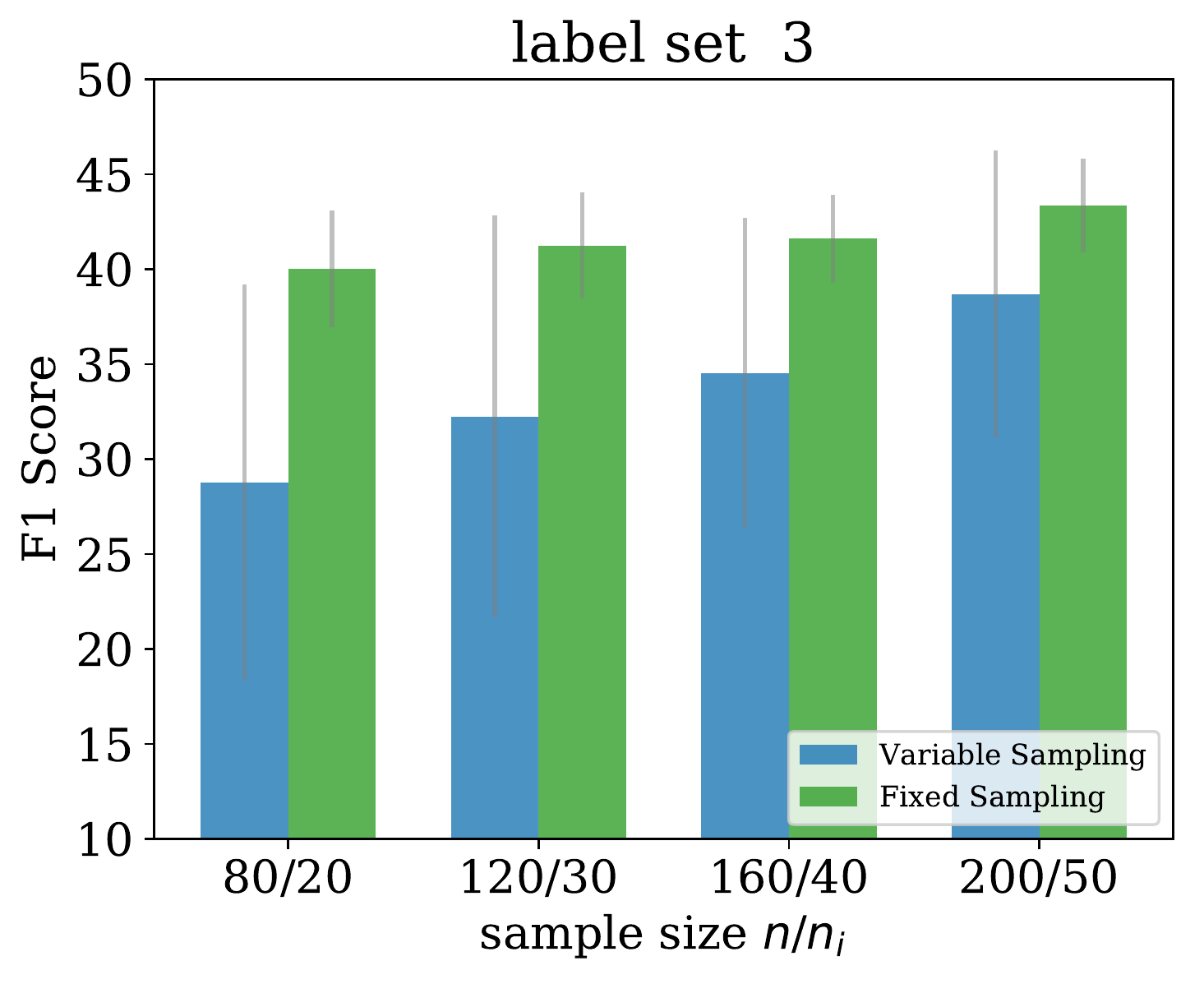}
    \includegraphics[height=3.35cm]{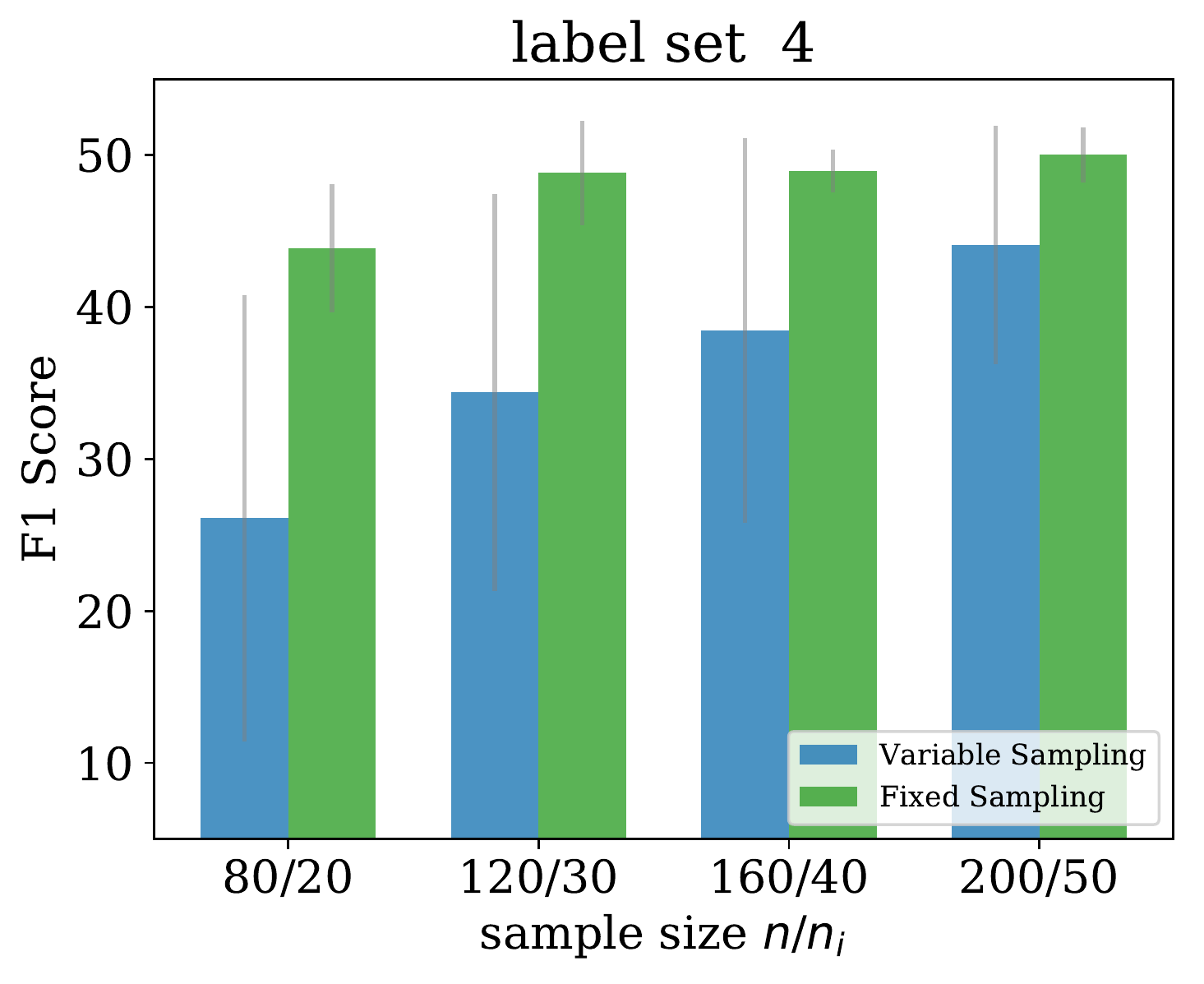}
    \includegraphics[height=3.35cm]{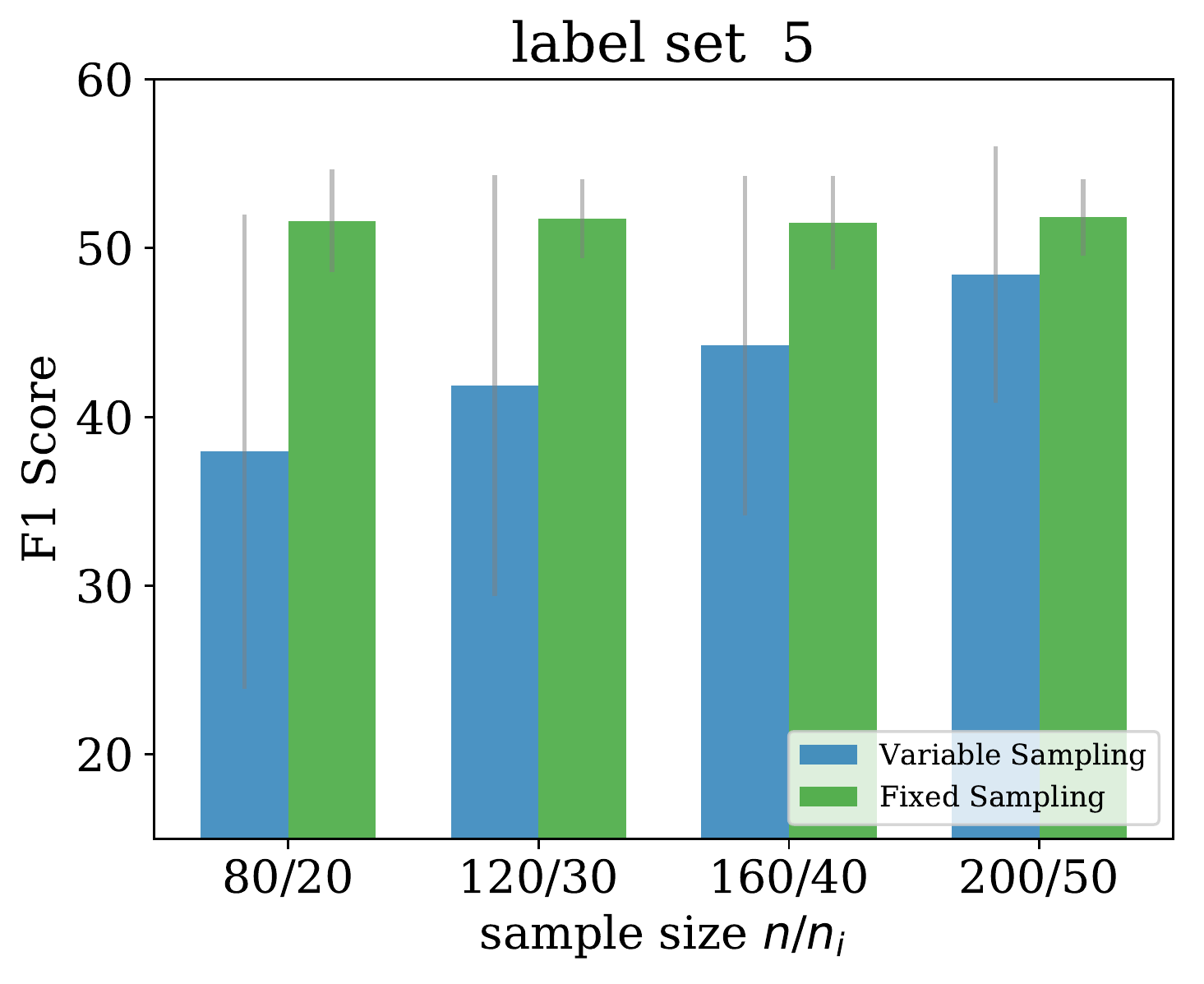}
    \includegraphics[height=3.35cm]{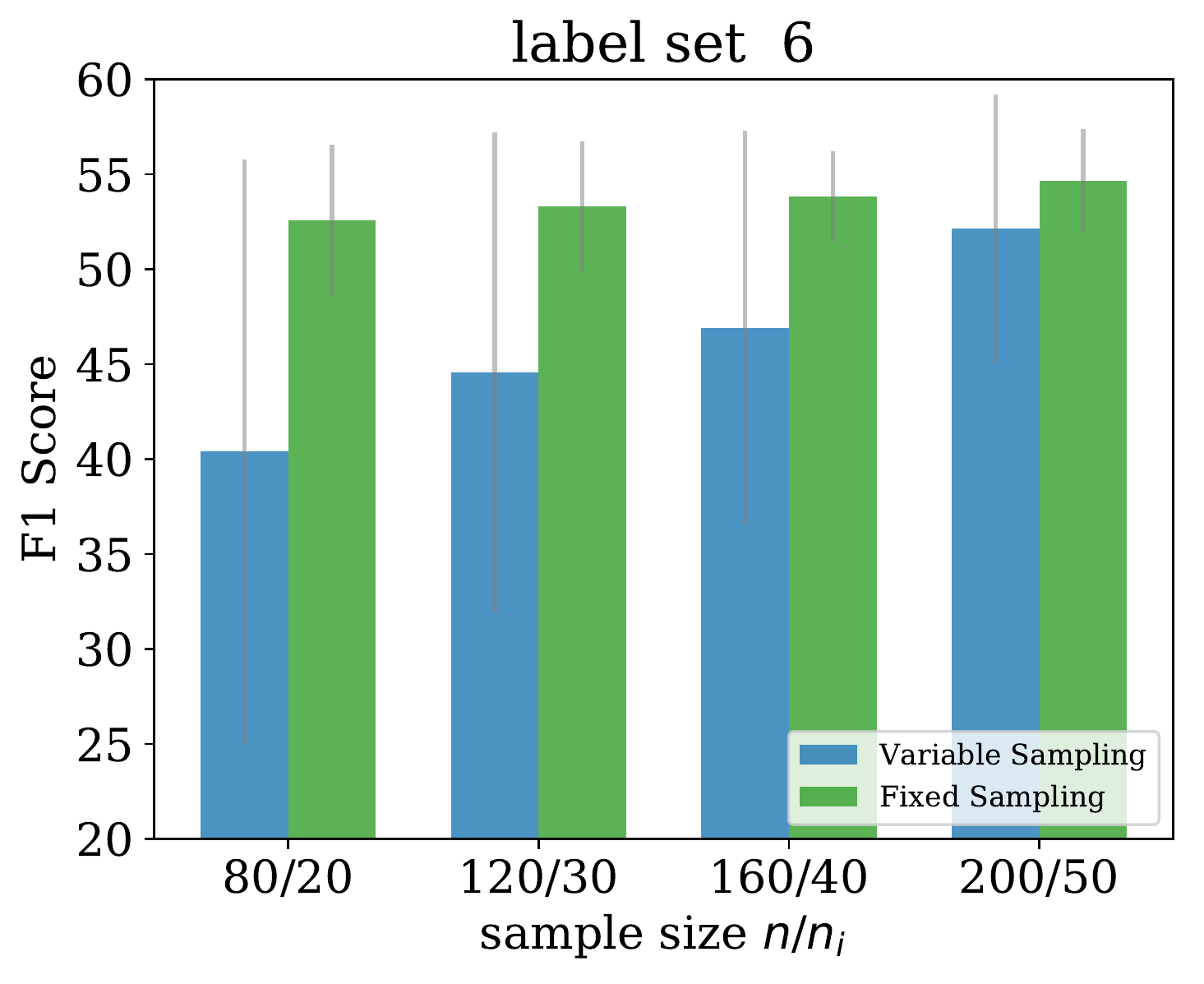}
    \includegraphics[height=3.35cm]{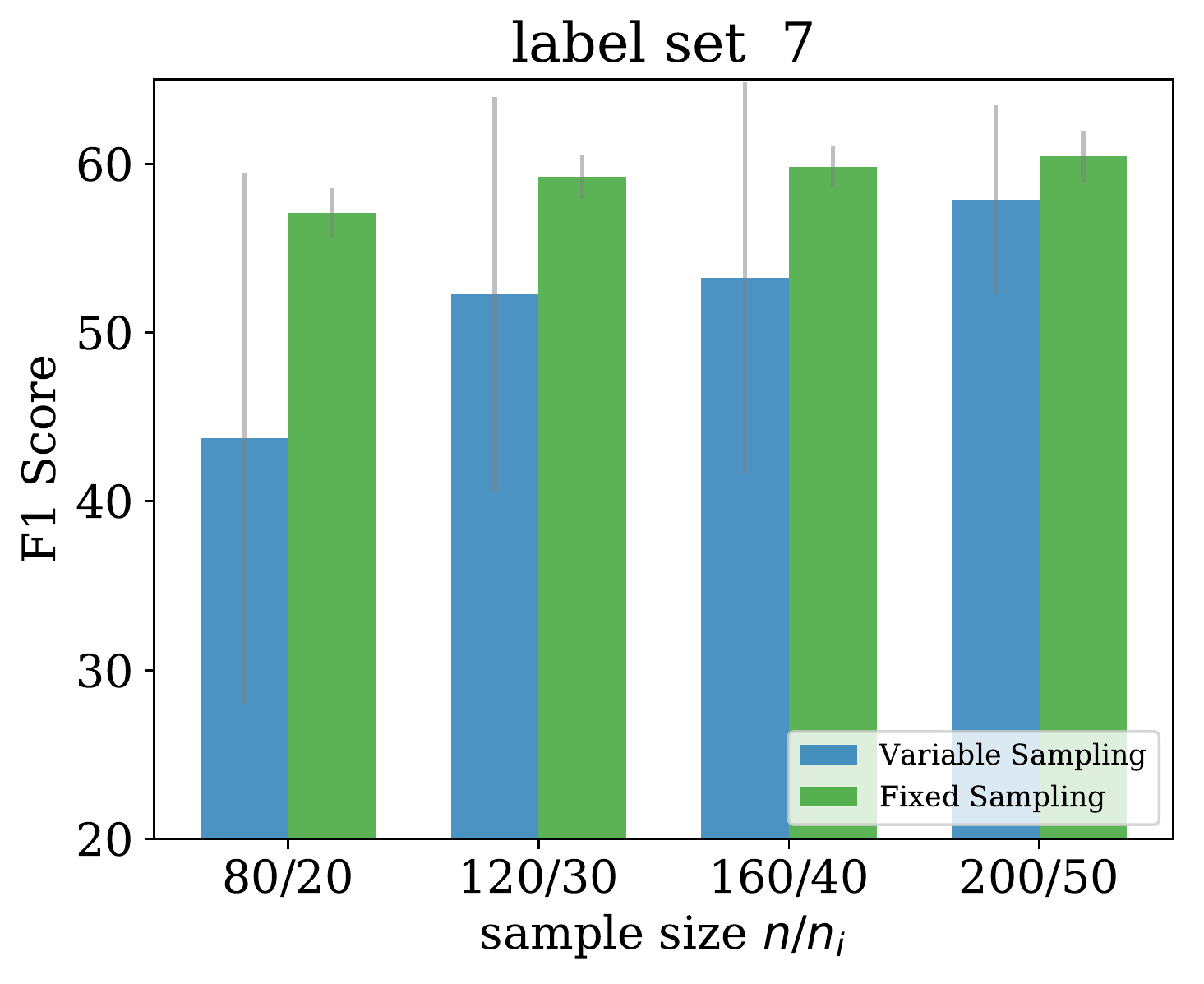}
    \caption{Comparing Variable and Fixed Sampling by mean test performance (F1 score) of the base model on NoisyNER \textbf{with $\mathbf{|D_C|}$ fixed} for the base model and varying for the noise model estimation. Error bars show the empirical standard deviation.}
\end{figure}

\end{document}